\PassOptionsToPackage{dvipsnames,table,pdftex}{xcolor}
\documentclass{article} 
\usepackage{iclr2025_conference,times}
\iclrfinalcopy
\usepackage{wrapfig}
\usepackage[utf8]{inputenc} 
\usepackage[T1]{fontenc}    
\usepackage{hyperref}       
\usepackage{url}            
\usepackage{booktabs}       
\usepackage{amsfonts}       
\usepackage{nicefrac}       
\usepackage{microtype}      

\usepackage{footmisc}
\usepackage{enumitem}
\usepackage{adjustbox}
\usepackage{pgfplots}
\usepackage{pgfplotstable}
\definecolor{zerothcolorbar}{HTML}{3c4e4b}
\definecolor{firstcolorbar}{HTML}{54bebe}
\definecolor{secondcolorbar}{HTML}{76c8c8}
\definecolor{thirdcolorbar}{HTML}{98d1d1}
\definecolor{fourthcolorbar}{HTML}{badbdb}
\definecolor{fifthcolorbar}{HTML}{dedad2}
\definecolor{sixthcolorbar}{HTML}{e4bcad}
\definecolor{seventhcolorbar}{HTML}{df979e}
\definecolor{eighthcolorbar}{HTML}{d7658b}
\definecolor{ninthcolorbar}{HTML}{c80064}

\definecolor{color1}{HTML}{1f77b4}
\definecolor{color2}{HTML}{ff7f0e}
\definecolor{color3}{HTML}{2ca02c}
\definecolor{color4}{HTML}{e4bcad}
\definecolor{color5}{HTML}{76c8c8}
\definecolor{color6}{HTML}{c80064}
\definecolor{color7}{HTML}{3c4e4b}
\definecolor{color8}{HTML}{7f7f7f}
\definecolor{color9}{HTML}{bcbd22}
\definecolor{color10}{HTML}{17becf}
\definecolor{color11}{HTML}{aec7e8}
\definecolor{color12}{HTML}{ffbb78}
\definecolor{color13}{HTML}{98df8a}
\definecolor{color14}{HTML}{ff9896}
\definecolor{color15}{HTML}{c5b0d5}
\definecolor{color16}{HTML}{c49c94}
\definecolor{color17}{HTML}{f7b6d2}
\definecolor{color18}{HTML}{c7c7c7}
\definecolor{color19}{HTML}{dbdb8d}
\definecolor{color20}{HTML}{9edae5}
\definecolor{color21}{HTML}{ad494a}

\usepackage{multirow}
\usepackage{makecell}
\usepackage{soul}
\usepackage{bbding}

\usepackage{graphicx}
\usepackage{booktabs}
\usepackage{array}
\usepackage{adjustbox}
\usepackage{tkz-kiviat} 
\usepackage{subcaption}
\usetikzlibrary{arrows}

\newsavebox{\measuredSize}

\usepackage{amsmath,amsfonts,bm}

\def\eqref#1{equation~\ref{#1}}

\def\1{\bm{1}}

\def\vt{{\bm{t}}}

\def\vz{{\bm{z}}}

\DeclareMathAlphabet{\mathsfit}{\encodingdefault}{\sfdefault}{m}{sl}
\SetMathAlphabet{\mathsfit}{bold}{\encodingdefault}{\sfdefault}{bx}{n}

\def\sB{{\mathbb{B}}}

\def\sR{{\mathbb{R}}}

\makeatother

\newcommand{\xfill}[2][.7ex]{{%
  \dimen0=#2\advance\dimen0 by #1
  \mbox{}\leaders\hrule height \dimen0 depth -#1\hfill%
}}

\newcommand{\revised}[1]{{#1}}

\newcommand{\Caltech}{{\small \textsc{Caltech}}}
\newcommand{\Cifarten}{{\small \textsc{CIFAR10}}}
\newcommand{\Cifaronehundred}{{\small \textsc{CIFAR100}}}
\newcommand{\Clevr}{{\small \textsc{CLEVR}}}
\newcommand{\Country}{{\small \textsc{Country}}}
\newcommand{\Cub}{{\small \textsc{CUB}}}
\newcommand{\Dtd}{{\small \textsc{DTD}}}
\newcommand{\Eurosat}{{\small \textsc{Eurosat}}}
\newcommand{\Gtsrb}{{\small \textsc{GTSRB}}}
\newcommand{\Mnist}{{\small \textsc{MNIST}}}
\newcommand{\FGVC}{{\small \textsc{FGVC}}}
\newcommand{\Pcam}{{\small \textsc{PCam}}}
\newcommand{\Renderedsst}{{\small  \textsc{SST}}}
\newcommand{\Resisc}{{\small \textsc{Resisc}}}
\newcommand{\STL}{{\small \textsc{STL}}}
\newcommand{\SUN}{{\small \textsc{SUN}}}
\newcommand{\Flowers}{{\small \textsc{Flowers}}}
\newcommand{\Pets}{{\small \textsc{Pets}}}
\newcommand{\Cars}{{\small \textsc{Cars}}}
\newcommand{\Food}{{\small \textsc{Food}}}
\newcommand{\Imagenet}{{\small \textsc{ImgNet-1k}}}

\newcommand{\GAC}{\textsc{GAC}}
\newcommand{\NNC}{\textsc{NLC}}
\newcommand{\MoE}{\textsc{MoE}}
\newcommand{\SuperLearner}{\textsc{SL}}
\newcommand{\LogitAvg}{\textsc{Log-Avg}}
\newcommand{\Best}{\textsc{Single-Best}}
\newcommand{\VoteOne}{\textsc{Vote T-1}}
\newcommand{\VoteThree}{\textsc{Vote T-3}}
\newcommand{\Confidence}{\textsc{Conf}}
\newcommand{\CalibratedConfidence}{\textsc{C-Conf}}
\newcommand{\CLogitAvg}{\textsc{C-Log-Avg}}

\newcommand\ColorBox[1]{\textcolor{#1}{\rule{2ex}{2ex}}}

\newcommand{\Oracle}{\textsc{Oracle}}

\newcommand{\upg}[1]{{\color{ForestGreen}#1}}
\newcommand{\downr}[1]{{\color{BrickRed}#1}} 
\newcommand{\samey}[1]{{\color{Dandelion}$-$#1}} 

\newcommand{\ie}{\textit{i.e.}}        

\title{Synergy and Diversity in CLIP: Enhancing Performance Through Adaptive Backbone Ensembling}

\author{%
Cristian Rodriguez-Opazo$^{1}$ \quad Ehsan Abbasnejad$^{1}$ \quad Damien Teney$^{2,1}$ \\
 \textbf{Hamed Damirchi}$^1$ \quad \textbf{Edison Marrese-Taylor}$^3$ \quad \textbf{Anton van den Hengel}$^1$ \\
$^1$Australian Institute for Machine Learning \quad University of Adelaide \\
$^2$Idiap Research Institute, Switzerland \quad $^3$The University of Tokyo\\
\texttt{\{cristian.rodriguezopazo,ehsan.abbasnejad\}@adelaide.edu.au}\\
}

\begin{document}

\maketitle

\begin{abstract}
Contrastive Language-Image Pretraining (CLIP) stands out as a prominent method for image representation learning. Various architectures, from vision transformers~(ViTs) to convolutional networks (ResNets) have been trained with CLIP to serve as general solutions to diverse vision tasks.
This paper explores the differences across various CLIP-trained vision backbones.
Despite using the same data and training objective, we find that these architectures have notably different representations,
different classification performances across datasets, and different robustness properties to certain types of image perturbations.
Our findings indicate a remarkable possible synergy across backbones
by leveraging their respective strengths.
In principle, classification accuracy could be improved by over 40 percent with an informed selection of the optimal backbone per test example. 
Using this insight, we develop a straightforward yet powerful approach to adaptively ensemble multiple backbones.
The approach uses as few as one labeled example per class
to tune the adaptive combination of backbones.
On a large collection of datasets, the method achieves a remarkable increase in accuracy of up to 39.1\% over the best single backbone, well beyond traditional ensembles.
\end{abstract}

\section{Introduction}
\label{sec:intro}

Large pre-trained models are transforming machine learning, computer vision \citep{radford2021learning, he2022masked, kirillov2023segment, liu2023improved}, and natural language processing~\citep{devlin2018bert,brown2020language,Touvron2023LLaMAOA}. 
These models are typically trained with self-supervised objectives, eschewing the need for manual annotations and enabling the use of very large datasets~\citep{jia2021scaling,schuhmann2022laion}.

Contrastive Language--Image Pre-training (CLIP)~\citep{radford2021learning} is one such approach that enables various downstream applications involving vision and language.
CLIP learns to align text and image representations
when training a pair of vision and language encoders, which can be implemented with various architectures.
These encoders are then used in downstream applications to obtain representations of images and text, whose similarity can be simply evaluated with a dot product. This enables 
e.g.\ zero-shot classification \citep{Zhai_2022_CVPR,li2022elevater,jia2021scaling} and cross-modal retrieval \citep{li2020unicoder,li2020oscar,yu2022coca}.
\begin{figure}[t!]
    \centering
    \includegraphics[width=\textwidth]{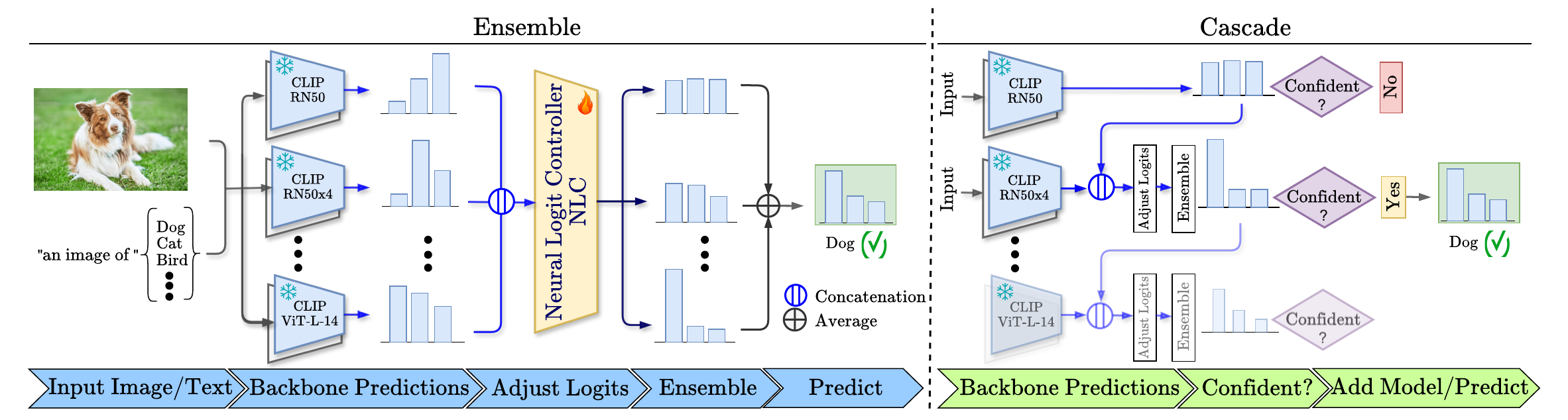}
    \caption{We propose a method 
    to improve CLIP's effectiveness for image classification
    by combining the strengths of different backbones.
    \textbf{(Left)}~For a given test image, the logits from different backbones are combined with a temperature scaling that weights their contribution to the final prediction.
    The scaling is implemented in the Neural Logit Controller (\NNC{}, a small MLP) that is learned from as little as one labeled example per class.
    \textbf{(Right)}~To reduce the computational load, our method can be combined with the Cascade framework~\citep{wang2022wisdom}.}
    \label{fig:teaser}
    \vspace{-2em}
\end{figure}

Despite extensive prior work on CLIP, there is still a gap in comparing the representations from different vision backbones trained with this paradigm. Existing work has compared backbones for their generalization capabilities~\citep{goldblum2023battle, li2022elevater, zhang2021tip, gao2021clip}, finding that larger architectures generally perform better.
This paper contributes to this area with an empirical study that compares CLIP-trained backbones.
We present new observations on how performance and robustness (e.g.\ invariance to image transformations)
vary substantially across architectures.
Within the same backbone family (e.g.\ different ViTs), different models present different patterns of performance across datasets (Figure~\ref{wrap-fig:oracle_kiviat}).
The relation with model scale is 
also more complicated than suggested in prior work (Figure~\ref{fig:venn_diagram}).

\begin{wrapfigure}[38]{RH}[0pt]{0.48\textwidth}
    \vspace{-1em}
    \centering
    \begin{adjustbox}{width=0.42\textwidth}
        \input{plots/improvement_vs_diversity}
    \end{adjustbox}
    \caption{\textbf{(Y axis)} Relative improvement of \NNC{} over best backbone vs. \textbf{(X axis)} predictions diversity $^1$.
    \NNC{} always improves over the best backbone.
    Moreover, the clear correlation shows that higher diversity in predictions tends to result in greater improvements with \NNC{}.
    }
    \label{wrap-fig:improvement_vs_diversity}
    \vspace{-0.5em}
    \begin{center}
    \includegraphics[width=0.42\textwidth]{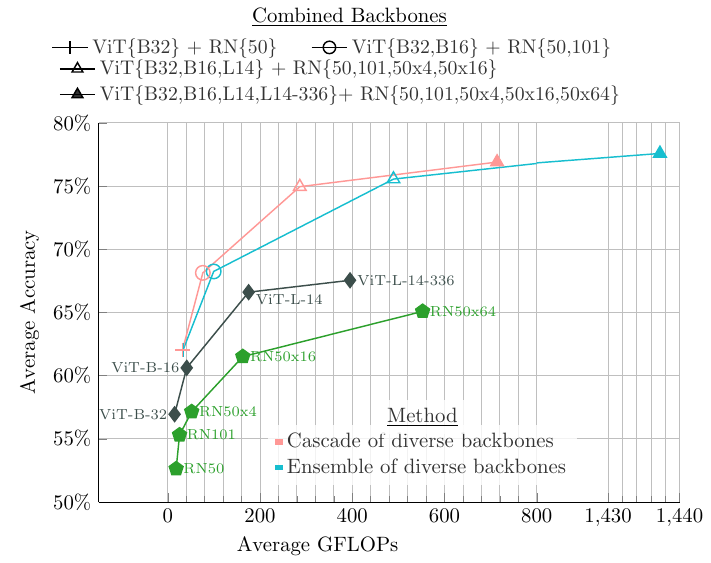}
    \end{center}
    \caption{Average accuracy across 21 datasets for zero-shot ResNets and ViTs backbones, \NNC{} Ensemble, and Cascade using 2 to 9 backbones, plotted against the Average GFLOPs. Demonstrates that ensembles can surpass the best zero-shot backbone with fewer GFLOPs.}
    \label{wrap-fig:accuracy_vs_flops}
\end{wrapfigure}
Our empirical findings suggest that CLIP's effectiveness for 
image classification could be improved by combining the strengths of different backbones. 
We thus present an ensemble method that combines 
backbones adaptively to each test 
example, using a temperature scaling that weights each backbone's contribution to the final prediction condition on the input image (see Figure~\ref{fig:teaser}).


A naive combination of backbones with traditional ensembling techniques \citep{lakshminarayanan2017simple, dietterich2000ensemble} 
(i.e.\ averaging the outputs)
does not consistently enhance generalization performance. 
These approaches focus on prediction agreements rather than leveraging diversity.
In contrast, our method uses a temperature scaling
based on the idea of calibrating each backbone's confidence.
This intuition is further supported by the
``modality gap''
and variation in performance observed in prior work
when adjusting CLIP's logits~\citep{clip-understanding}.
We compare our approach to more advanced ensembles, such as SuperLearner \cite{superlearner2018}, 
which also combines the logits of multiple models by learning temperature scaling. 
However, unlike our method, SuperLearner does not adaptively adjust these temperatures based on the input features\revised{, implying that SuperLearner cannot exploit the diversity of predictions (Tab. \ref{tab:zs-baselines}.})
\footnotetext[1]{The diversity of predictions is measured as:
$1\,-\,\frac{
    \texttt{\#\,samples correctly predicted by \emph{all} backbones}
}{
    \texttt{\#\,samples correctly predicted by \emph{any} backbone}
}.$\label{diversity:eq}}

Our proposed method, named the Neural Logit Controller (\NNC{}), uses a few labeled examples (as little as one per class) to tune the combination of backbones. Hence, we benchmark it against the state-of-the-art methods in a few-shot setting~\citep{zhang2021tip,radford2021learning,zhou2022learning,gao2021clip}.
\NNC{} shows remarkable performance and consistently outperforms other approaches.
Furthermore, we show that \NNC{} complements existing methods like Tip-Adapter~\citep{zhang2021tip}. Combining it with \NNC{} in its original experimental framework yields improvements in accuracy of over 15\%.

Finally, in-depth experiments reveal a clear correlation between
\NNC{}'s improvement over the best backbone and the diversity of correct predictions in each dataset (see Figure~\ref{wrap-fig:improvement_vs_diversity}).
It achieves over 120\% of relative improvement when the diversity of predictions is close to one, and never degrades the performance of the best backbone. In terms of efficiency, \NNC{} outperforms the best backbone by combining only the top-four most efficient backbones, while using approximately 300 fewer GFLOPs (see Figure~\ref{wrap-fig:accuracy_vs_flops}).
We also show how to use the Cascade framework~\revised{\citep{wang2022wisdom,varshney-baral-2022-model}}
with \NNC{} to enhance performance while maintaining computational requirements within the bounds of the original backbones.

Our contributions are summarized as follows.
\begin{itemize}[topsep=-2pt,itemsep=1pt,labelindent=2pt,labelsep=*,leftmargin=12pt]
    \item We perform extensive experiments across 21 datasets that reveal diverse predictions across CLIP backbones
    and distinct robustness properties to image transformations.

    \item We evaluate the complementarity of backbones
    by measuring the potential improvement in classification accuracy
    (up to 43.5\%)
    of an optimal oracle selection of backbone per test example.
    \item To leverage this complementarity, we propose an adaptive ensembling method (\NNC{})
    that uses temperature scaling and requires a single labeled example per class.
    It improves accuracy over the best backbone by 9.1\% on average, surpassing previous ensembling frameworks.

    \item We demonstrate significant advantages in computational efficiency. Combining the top four most efficient backbones enables \NNC{} to outperform the best backbone with $\sim$300 fewer GFLOPs. Integrating \NNC{} with the Cascade approach~\citep{wang2022wisdom} maintains computational efficiency.

    \item We demonstrate that
    \NNC{} consistently outperforms state-of-the-art few-shot methods.
    Moreover, integrating \NNC{} with Tip-Adapter~\citep{zhang2021tip}
    gives the latter a performance boost of over 15\%.

\end{itemize}


\section{Investigating Differences Across Vision Backbones}
\label{sec:prelim}

\paragraph{CLIP for zero-shot image classification.}
\begin{wrapfigure}[20]{r}[0pt]{0pt}
    \centering
    \vspace{-4em}
    \begin{adjustbox}{width=0.47\textwidth, keepaspectratio}
    \begin{tikzpicture}[scale=0.5]
    \tkzKiviatDiagram[scale=1.,
            label distance=2000,
            radial  = 100,
            gap     = 1,  
            lattice = 10]{\scriptsize{\textsc{Caltech101}},
                    \small{\textsc{Cars}},
                    \small{\textsc{Cifar10}},
                    \small{\textsc{Cifar100}},
                    \small{\textsc{CLEVR}},
                    \small{\textsc{Country211}},
                    \small{\textsc{Cub}},
                    \small{\textsc{DTD}},
                    \small{\textsc{Eurosat}},
                    \small{\textsc{FGVC}},
                    \small{\textsc{Flowers}},
                    \small{\textsc{Food}}, 
                    \small{\textsc{GTSRB}},
                    \small{\textsc{ImgNet-1k}},
                    \small{\textsc{MNIST}},
                    \small{\textsc{PCAM}},
                    \small{\textsc{Pets}},
                    \small{\textsc{SST2}},
                    \small{\textsc{Resisc45}},
                    \small{\textsc{STL10}},
                    \small{\textsc{SUN397}}
                    }
    \tkzKiviatLine[ultra thick,dashed,color=color11,mark=none,fill=color1!40,opacity=.0](7.79,5.42,7.15,4.03,2.16,1.54,4.66,4.12,2.82,1.71,6.61,7.79,3.51,5.98,4.91,6.39,8.58,5.66,4.4,9.42,5.86)
    \tkzKiviatLine[ultra thick,dashed,color=color12,mark=none,fill=color2!40,opacity=.0](8.18,6.11,8.08,4.76,2.44,1.69,4.96,4.37,2.64,1.86,6.52,8.19,3.75,6.23,4.6,5.81,8.69,6.39,5.45,9.68,5.72)
    \tkzKiviatLine[ultra thick,dashed,color=color13,mark=none,fill=color3!40,opacity=.0](8.2,6.69,7.94,4.51,2.05,2.04,5.42,4.86,2.74,2.11,6.99,8.55,3.62,6.62,5.33,5.67,8.89,6.72,5.43,9.66,5.97)
    \tkzKiviatLine[ultra thick,dashed,color=color14,mark=none,fill=color4!40,opacity=.0](8.51,7.32,8.13,5.21,1.96,2.44,5.79,5.34,4.12,2.75,7.2,8.97,3.98,7.07,6.27,6.25,9.01,6.75,5.94,9.78,6.36)
    \tkzKiviatLine[ultra thick,dashed,color=color15,mark=none,fill=color5!40,opacity=.0](8.39,7.59,8.51,5.99,2.27,2.98,6.27,5.31,4.8,3.1,7.61,9.11,4.79,7.39,7.89,5.39,9.36,7.1,6.39,9.83,6.6)
    \tkzKiviatLine[ultra thick,dashed,color=color16,mark=none,fill=color6!40,opacity=.0](8.37,5.97,8.98,6.42,2.32,1.72,5.3,4.4,3.74,1.97,6.65,8.26,3.26,6.33,3.81,6.23,8.75,5.86,5.42,9.71,6.13)
    \tkzKiviatLine[ultra thin,dashed,color=color17,mark=none,fill=color7!40,opacity=.0](8.66,6.46,9.08,6.69,2.12,2.28,5.53,4.51,4.41,2.44,7.14,8.79,4.33,6.83,5.95,5.07,8.91,6.05,5.83,9.82,6.37)
    \tkzKiviatLine[ultra thick,dashed,color=color18,mark=none,fill=color8!40,opacity=.0](8.59,7.78,9.56,7.58,1.94,3.19,6.21,5.53,4.72,3.17,7.91,9.23,5.06,7.55,7.33,5.2,9.36,6.89,6.46,9.94,6.67)
    \tkzKiviatLine[ultra thick,dashed,color=color19,mark=none,fill=color9!40,opacity=.0](8.61,7.94,9.5,7.44,2,3.45,6.3,5.64,4.5,3.32,7.83,9.31,5.24,7.66,7.52,6.07,9.38,7.06,6.37,9.94,6.77)
    
    \tkzKiviatLine[ultra thick,color=color1,mark=none,fill=color10!40,opacity=.0](9.02,9.13,9.65,7.68,5.62,4.12,8.25,7.35,8.19,5.23,8.6,9.61,6.52,8.49,9.02,8.7,9.81,9.53,8.23,9.93,8.27)    
    \tkzKiviatLine[ultra thick,color=color2,mark=none,fill=color11!40,opacity=.0](9.19,8.99,9.85,8.71,3.44,4.4,7.89,6.72,7.2,5.19,8.57,9.66,6.7,8.46,8.56,9.54,9.75,9.23,7.87,9.97,8.06)    
    \tkzKiviatLine[ultra thick,color=color3,mark=none,fill=color3!40,opacity=0.](9.37,9.54,9.93,9.1,6.04,5.2,8.81,7.82,9.15,6.63,8.89,9.8,7.79,8.9,9.51,9.91,9.91,9.86,8.87,9.98,8.7)
    
    \tkzKiviatLine[ultra thick,color=color21,mark=none,fill=color3!40,opacity=0.](9.11,8.41,9.66,7.95,5.44,3.61,7.36,6.32,8.71,3.76,8.21,9.4,7.05,7.88,9.18,8.49,9.54,7.73,7.83,9.95,7.32)
    \tkzKiviatGrad[prefix=,unity=10](0)  

\node[anchor=south east,xshift=-40pt,yshift=-55pt] at (current bounding box.south east) 
{
\begin{tabular}{llllllll}
    \ColorBox{color11} & RN50 & 
    \ColorBox{color12} & RN50x4 &
    \ColorBox{color13} & RN50x16 &
    \ColorBox{color14} & RN50x64 \\
    \ColorBox{color15} & RN101 &
    \ColorBox{color16} & ViT-B-32 &
    \ColorBox{color17} & ViT-B-16 &
    \ColorBox{color18} & ViT-L-14 \\
    \ColorBox{color19} & ViT-L-14-336 &
    \ColorBox{color1} & \textsc{Oracle} RN &
    \ColorBox{color2} & \textsc{Oracle} ViT &
    \ColorBox{color3} & \textsc{Oracle} All \\
    \ColorBox{color21} & \NNC & \\
    \end{tabular}
    };
\end{tikzpicture}
    \end{adjustbox}
    \caption{Zero-shot classification accuracy of various CLIP models on 21 datasets, and of the upper-bound ``\Oracle'' combination of ResNets (RN), ViTs, and all backbones.}
    \label{wrap-fig:oracle_kiviat}
\end{wrapfigure}
CLIP \citep{radford2021learning}
is a general technique to train a pair of vision and text encoders
$\phi_l$ and $\phi_v$ on paired image/text data.
The method uses a self-supervised objective to align the encoded representations of matching images and text descriptions.
Any image and text can then be mapped to a unified semantic space.
This enables
zero-shot image classification by computing a compatibility score
between an image and a class description (``prompt'') as $\operatorname{score}(x,l) = \phi_v(x)^\top \phi_l(l)$.
We follow previous work \citep{menon2022visual} for the generating of prompts
e.g.\ \texttt{an image of \{\texttt{label}\}, which is a \{\texttt{concept}\}}. Subsequently, we calculate the compatibility score between the image feature $\phi_v(x)$ and the prompt representations $\phi_l(y)$, where $y$ represents a label within the class set. The prompt with the highest similarity is then chosen as the label for the given image. 

\subsection{Complementarity of Different Backbones}
The vision encoder or ``backbone'' trained with CLIP can be implemented using various architectures.
We want to identify whether different backbones have distinctive behaviours.
We analyze the output of 9 of them
in a zero-shot classification setting on 21 standard benchmark datasets (see Section~\ref{subsec:db_bkbones} for details). 
We propose two approaches to assess the differences
in their zero-shot predictions
and their possible complementary behaviour.
First, we define an \Oracle{} prediction that combines
the output of the 9 models on a per-image basis, using a correct prediction if any of the models is correct, any other otherwise.
This simulates a scenario where an informed choice of the optimal backbone could be made for each test image to obtain the highest classification accuracy. Second, we use overlap diagrams to visualize the patterns of (in)correct predictions across backbones
and their different strengths on different subsets of images.

\begin{figure}[t]
    \centering
    {\textsc{ImageNet-1K}}
    \includegraphics[width=0.9\textwidth]{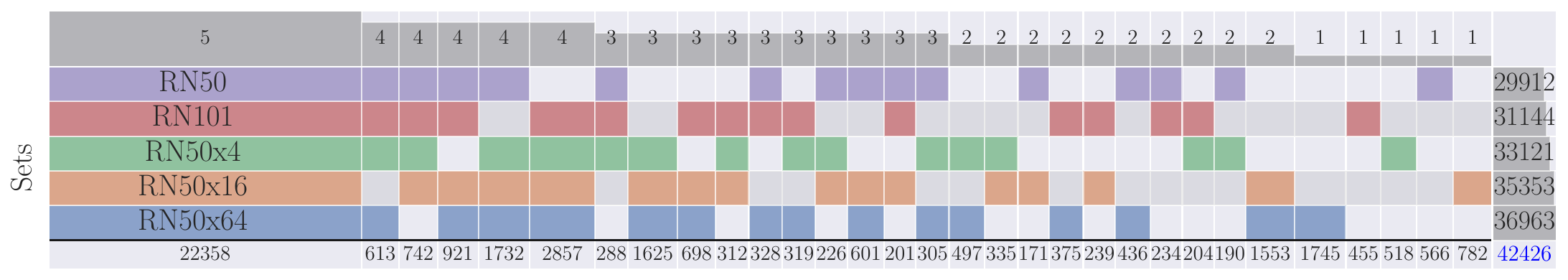}
    \includegraphics[width=0.9\textwidth]{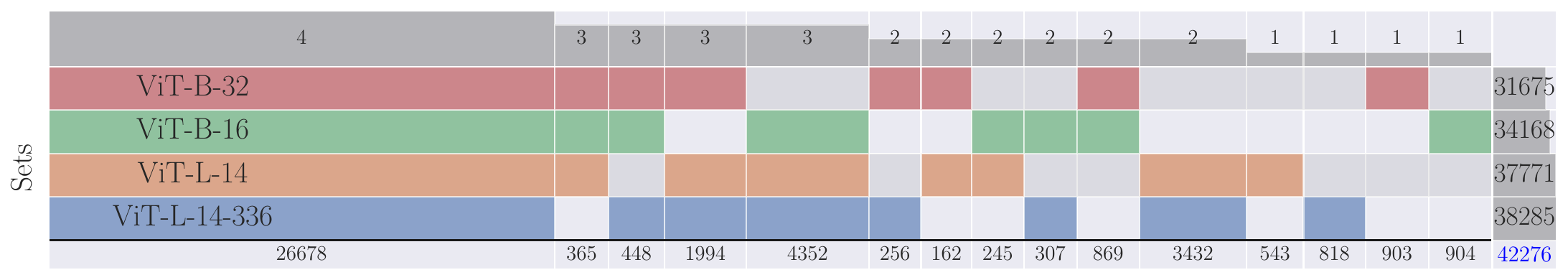}
    \includegraphics[width=0.9\textwidth]{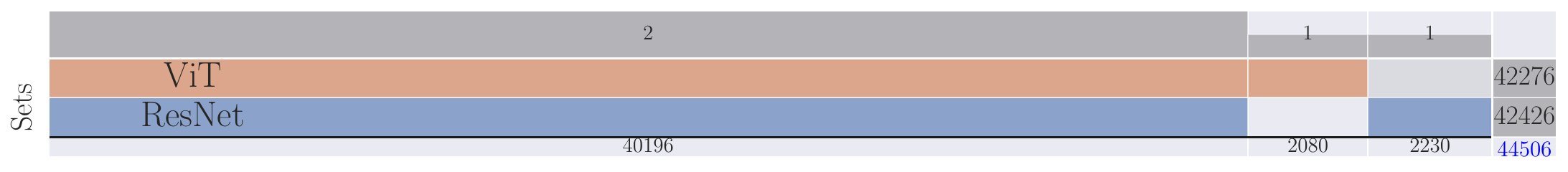}
    \caption{
    Linear Venn diagrams showing the overlap of test images from ImageNet-1k correctly classified by different backbones (rows).
    Each column represents a subset of images correctly classified by a specific group of backbones (group size in column header).
    Row/column sums indicate the number of correct predictions per backbone/subset.
    We observe that \textbf{(1)}~different backbones agree on a large part of the data.
    \textbf{(2)}~They also make additional correct predictions on different subsets.
    \textbf{(3)}~Accuracy usually grows with architectures size, but even within a same family (ViTs, ResNets), different models show different patterns of (in)correct predictions.}
    \label{fig:venn_diagram}
    \vspace{-2.5em}
\end{figure}

Figure \ref{wrap-fig:oracle_kiviat} and Table~\ref{tab:zs-all-table}
summarize the performance of the different backbones.
Note that they are pre-trained using the exact same dataset and objectives, as provided in the OpenClip project~\citep{ilharco_gabriel_2021_5143773}.
Our results show that backbones from different families (ViTs vs. ResNets)
show significant differences while controlling for model size (number of parameters).
For example, ViT-B-32 outperforms ResNet-50x4 in the \Cifarten{} and \Cifaronehundred{} benchmarks. However, this trend is not consistent across all datasets, e.g.\ on \Imagenet{}. Additionally, the results indicate that backbones with more parameters do not always outperform smaller ones.

We then assess the \Oracle{} predictions using backbones from the same family
(\Oracle{} RN for ResNets, \Oracle{} ViT for ViTs, and \Oracle{} All for a combination of all backbones).
The \Oracle{} All obviously outperforms the best individual
backbone in every case.
For instance, on \Eurosat, \Clevr, \Cub, \Dtd, \Cifaronehundred{}, and \Cars{}, the \Oracle{} All shows improvements of 43.5\%, 36.0\%, 25.1\%, 21.8\%, 16.6\% and 16.0\%, respectively.
\textbf{The large magnitude of these improvements highlights the complementarity of backbones.
In other words, the correctly-classified data is not simply growing
as one considers larger and better models.
Instead, different backbones perform well on different subsets of the data.}
\revised{This insight is the key motivation for the ensemble method we propose in Section~\ref{sec:method}. We also found that similar findings have been made in domains other than ours \cite{roth2024fantastic,zhong-etal-2021-larger,rame2022recycling}}.

We visualize in Figure~\ref{fig:venn_diagram}
the overlap between different subsets of \Imagenet{} correctly predicted by different backbones.
We observe comparable accuracy between the largest ViT (ViT-L-14-336, correct predictions) and ResNet (RN50x64) with $38,285$ vs. $36,963$ correct predictions.
However, there are significant differences within each family: for example, among ResNets, only $22,358$ images are correctly classified by all ResNets models, out of the $42,426$ ones correctly classified by \emph{any} ResNet.$^1$

We further observe that every model has unique strengths by examining the performance of the oracle models.
Specifically, \Oracle{} RN, \Oracle{} ViT, and \Oracle{} All increase the number of correct predictions to $42,426$, $42,276$, and $44,506$ out of $50,000$. Each backbone exhibits correct predictions unique to itself. In the \Oracle{} RN, RN50, RN101, RN50x4, RN50x16, and RN50x64 contribute $566$, $455$, $518$, $782$, and $1,745$ exclusive predictions, respectively. 
In the \Oracle{} ViT, B-32, B-16, L-14, and L-14-336 present $903$, $904$, $543$, and $818$ unique predictions, respectively. Finally, when all backbones are used, ResNet contributes $2,230$ exclusive predictions, and ViT contributes $2,080$. It is important to note that correct predictions of any given model are not simply a subset of those by the best model in its family.

\begin{wrapfigure}[38]{RH}[0pt]{0.4\textwidth}
    \vspace{-1.5em}
    \centering
    \includegraphics[width=0.38\textwidth]{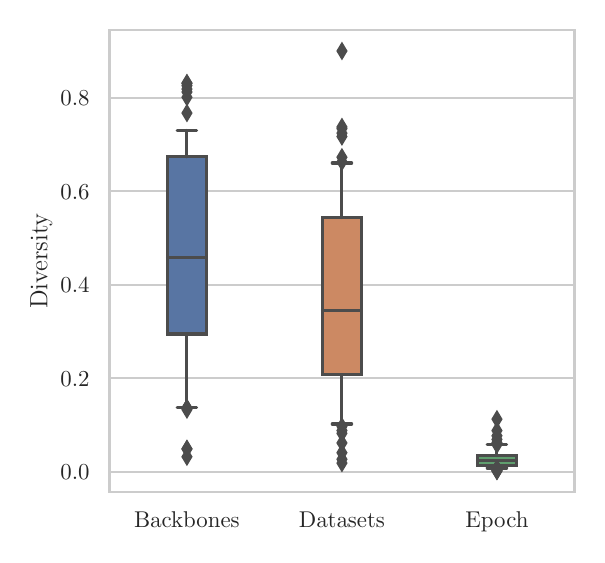}
    \caption{Diversity of predictions of (1) different \textbf{Backbones} with same pre-trained dataset, (2) same backbone with different pre-trained \textbf{Datasets}, and (3) same backbone and same pre-trained dataset in two different \textbf{Epoch}s. Results show complementarity is higher when we combine different backbones.}
    \label{wrap-fig:diversity_oracle}

    \vspace{0em}
    \centering
    \begin{adjustbox}{width=0.38\textwidth, keepaspectratio}
    \begin{tikzpicture}[scale=0.35]
    \tkzKiviatDiagram[scale=1.3,
            label distance=0cm,
            radial  = 9,
            gap     = 1,  
            lattice = 9]{ RESIZE,FLIP,GRAYSCALE, COLOR JITTER,GAUSSIAN BLUR} 
    \tkzKiviatLine[ultra thick,color=color11,mark=none,fill=color1!20,opacity=0.](3.81,-0.12,6.36,2.71,5.12)
    \tkzKiviatLine[ultra thick,color=color12,mark=*,fill=color2!20,opacity=.0](3.47,0.36,7.37,3,3.72)
    \tkzKiviatLine[ultra thick,color=color13,mark=*,fill=color3!20,opacity=.0](1.90,0.36,5.98,2.19,1.96)
    \tkzKiviatLine[ultra thick,color=color14,mark=none,fill=color4!20,opacity=.0](1.25,0.55,7.01,2.93,2.06)
    \tkzKiviatLine[ultra thick,color=color15,mark=none,fill=color5!20,opacity=.0](4.05,0.67,7.52,2.64,4.49)
    \tkzKiviatLine[ultra thick,color=color16,mark=none,fill=color6!20,opacity=.0](4.41,0.22,8.94,3.52,3.17)
    \tkzKiviatLine[ultra thick,color=color17,mark=none,fill=color7!20,opacity=.0](2.43,0.12,8.70,2.53,3.83)
    \tkzKiviatLine[ultra thick,color=color18,mark=none,fill=color8!20,opacity=.0](2.82,0.80,7.73,1.67,3.21)
    \tkzKiviatLine[ultra thick,color=color19,mark=none,fill=color9!20,opacity=.0](2.41,1.01,6.99,1.73,1.80)
    \tkzKiviatGrad[prefix=,unity=-1](1)  

\node[anchor=south east,xshift=-40pt,yshift=150pt] at (current bounding box.south east) 
{
\begin{tabular}{ll}
    \ColorBox{color11} & RN50 \\
    \ColorBox{color12} & RN50x4 \\
    \ColorBox{color13} & RN50x16 \\
    \ColorBox{color14} & RN50x64 \\
    \ColorBox{color15} & RN101 
    \end{tabular}
    };
    
\node[anchor=south east,xshift=-22pt,yshift=80pt] at (current bounding box.south east) 
{
\begin{tabular}{ll}
    \ColorBox{color16} & ViT-B-32 \\
    \ColorBox{color17} & ViT-B-16 \\
    \ColorBox{color18} & ViT-L-14 \\
    \ColorBox{color19} & ViT-L-14-336 
    \end{tabular}
    };
\end{tikzpicture}
    \end{adjustbox}
    \caption{Impact on test classification accuracy of image transformations. Smaller values (closer to center) mean higher robustness, suggesting models are particularly robust to specific transformations.}
    \label{wrap-fig:augmentation}
\end{wrapfigure}
\subsection{Other sources of diversity}
To better understand the effects of backbone complementarity, and contrast these effects against the role of variables that are known to drive performance in ensembles such random parameter initialization and learning via stochastic gradient descent, we propose to evaluate the diversity of correct predictions when combining multiple backbones, which as seen in our experiments above (Figure \ref{wrap-fig:improvement_vs_diversity}) strongly correlates with ensemble performance gains.


Concretely, we isolate the effects of three variables on prediction diversity: (1) backbone architecture, (2) pretraining dataset, and (3) number of training steps.
For (1), we evaluate the complementarity of ViT-L-14, ViT-B-32, and ViT-B-16 backbones using the same dataset.
For (2), we analyze each ViT's performance using different pretraining datasets (LAION-400M vs. OpenCLIP).
For (3), we consider ViTs pretrained on LAION-400M using the checkpoints at two different epochs (31 and 32).

Since, to the best of our knowledge, there are no publicly released CLIP models that differ only on their initialization, studies on this variable would, in principle, require us to train CLIP backbones from scratch, which is extremely compute-intensive. 

Our diversity metric considers the aggregate of test examples correctly classified and it is the ratio between the instances correctly classified by all backbones versus the instances correctly classified by any backbone~\footref{diversity:eq}.
It can be thought of as (1 - IoU), where IoU is the "intersection over union" of $n$ sets of correctly predicted samples, where $n$ is the number of methods considered.
If every sample in this aggregate is correctly classified by all backbones (i.e. they all agree while being correct), \textit{diversity=0}. Otherwise, if every sample can only be correctly classified by a single backbone, \textit{diversity=1}. This allows us to evaluate the complementarity of models.

Figure \ref{wrap-fig:diversity_oracle} (Tab. \ref{tab:diversity_pretrained_dataset}) demonstrates that combining multiple backbones achieves higher average diversity than using the same backbone across different datasets or nearby solutions in the loss landscape. This indicates that training stochasticity and pretraining dataset are not the primary factors. Instead, the most meaningful complementarity comes from diverse backbones, as they excel on different data subsets, as confirmed by the \Oracle{} results in Table \ref{tab:oracle_pretrained_dataset}.

\subsection{Robustness to Image Transformations}
To explore qualitative differences in behaviour of different backbones,
we selected a subset of each benchmark's test set and applied specific image transformations (\texttt{resize}, \texttt{flip}, \texttt{grayscale}, \texttt{color jitter}, and \texttt{Gaussian blur}).
We then evaluate the zero-shot classification performance of each backbone on these corrupted images.
Figure \ref{wrap-fig:augmentation} shows that ResNet 50 and ViT-B-16 are the most resilient 
to \texttt{flip}, RN50x64 to \texttt{resize}, and RN50x16 to \texttt{grayscale}.
ViT-L-14 and L-14-336 are the most robust to \texttt{color jitter} and \texttt{Gaussian blur}.
These findings suggest that the backbones' ability to deal with out-of-distribution images varies widely. Certain models prove particularly robust to specific transformations.

\section{Proposed Ensembling Method}
\label{sec:method}

The previous section showed that different backbones have different strengths
and perform well on different types of data.
We now use this insight to improve classification performance by combining multiple CLIP backbones.

\paragraph{Combining models by temperature scaling.}
Our approach is technically inspired by prior work on network calibration, which we apply to an ensemble of models. Specifically, our technique can be seen as a variation of \textbf{Platt scaling}~\citep{platt1999probabilistic}.
This classical method uses the logits of a model as features for a logistic regression, which is trained on the validation set to produce calibrated probabilities. More precisely, given logit scores $\vz_i$ for an example $i$, Platt scaling learns two scalars $a$ and $b$ and produces calibrated probabilities as $\hat q_i = \sigma(a \vz_i + b)$. The parameters $a$ and $b$ can be optimized using the negative log likelihood (NLL) loss over the validation set, with the model weights being frozen during this process.


A common use of Platt scaling is known as \textbf{temperature scaling} \citep{guoCalibrationModernNeural2017}.
In this approach, a single scalar parameter $t > 0$ is used for all classes of a given model. The new, calibrated confidence prediction is given by $\hat q_i = \max_k{\;\text{softmax}(\vz_i/t)^{(k)}}$, where $t$ is called the temperature, $\vz_i$ are the logits for example $i$ returned by the uncalibrated model.
Usually, $t$ is optimized with respect to the NLL on the validation set aiming to reduce the overconfidence of the model on its predictions and to produce more reliable predictions, but because the parameter $t$ does not change the maximum of the softmax function, the class prediction $\hat{y}_i$ remains unchanged, meaning that the performance of a given model remains the same. 

\paragraph{Learning temperature coefficients.}
In our method, we aim to jointly optimize a set of temperature parameters $t_b$ with $b \in [1, \ldots, B]$ for a set of $\sB$ backbones.
The aim is to combine their predictions by adjusting each backbone's confidence depending on the confidence of the others, and the input example.
We learn the temperatures $t_b$ that weigh the logit $\vz_i^b$ for a backbone $b$ and example $x$ using the cross-entropy loss\revised{, and then we combine the logits via a weighted sum}.

Concretely,
we train a one-layer MLP (our Neural Logit Controller) to predict the set of temperatures that best calibrate the backbone mixture.
The MLP takes as input the concatenated representations obtained by passing the images through the encoder $\phi_v$ of each backbone $b \in \sB$.
As depicted in Figure~\ref{fig:teaser},
The MLP directly produces a vector of temperatures $\vt \in \sR^B$ and is trained on a holdout set for the training set of each target dataset using the cross-entropy loss between final predictions and ground truth labels. 

\section{Experiments}
\label{sec:experiments}

\paragraph{Backbones.}
We consider the original selection of backbones described by \citet{radford2021learning}.
This includes the ResNets~\citep{He_2016_CVPR} RN50, RN50x4, RN50x16, RN50x64 and RN101, and the ViTs~\citep{dosovitskiy2020vit} B-16, B-32, L-14 and L-14-336.
All models are obtained through the open-source project OpenCLIP~\citep{ilharco_gabriel_2021_5143773}.

\paragraph{Datasets.}
\label{subsec:db_bkbones}
We use a selection of 21 popular image classification datasets: \Caltech101{} \citep{li_andreeto_ranzato_perona_2022}, \Cars{} \citep{krause2013collecting}, \Cifarten{} \citep{krizhevsky2009learning}, \Cifaronehundred{} \citep{krizhevsky2009learning}, \Clevr{} \citep{johnson2017clevr}, \Cub{} \citep{WahCUB_200_2011}, \Dtd{} \citep{cimpoi14describing}, \Eurosat{} \citep{helber2018introducing}, \FGVC{} \citep{maji13fine-grained}, \Flowers{} \citep{nilsback2008automated}, \Food{} \citep{bossard2014food}, \Gtsrb{} \citep{Houben-IJCNN-2013}, \Imagenet{} \citep{deng2009imagenet} \Mnist{} \citep{deng2012mnist}, \Pcam{} \citep{Veeling2018-qh}, \Pets{} \citep{parkhi2012cats}, Rendered\Renderedsst2 \citep{socher-etal-2013-recursive}, \Resisc45 \citep{Cheng_2017}, \STL10 \citep{coates2011analysis} and \SUN397 \citep{Xiao:2010}.
They include diverse images of nature, animals, places, medical scans, satellite images, and man-made objects.
The evaluation metric is simply the accuracy of each dataset's test split. 
The aggregated performance is the average accuracy across all datasets weighted equally.

\paragraph{Baselines.}
Our experiments consider several ensembling baselines. Non-parametric approaches:
\begin{itemize}[topsep=-2pt,itemsep=1pt,labelindent=2pt,labelsep=*,leftmargin=12pt]
\item \textbf{Logit averaging} (\LogitAvg) simply takes the average of the logits (scores) of the models.

\item \textbf{Voting} (\VoteOne{})
uses a majority vote. Each model casts a vote for its max-scored class.
The final prediction is the class with the highest number of votes.
Compared to \LogitAvg, this does not use the \emph{soft} score values and the uncertainty reflected therein.
\end{itemize}
Parametric approaches:
\begin{itemize}[topsep=-2pt,itemsep=1pt,labelindent=2pt,labelsep=*,leftmargin=12pt]
\item \textbf{Calibrated logit averaging} (\CLogitAvg)
first calibrates each model independently using temperature scaling.
They are then combined by simply averaging their logits.
\item \textbf{Super Learner} (\SuperLearner) \citep{superlearner2018} a flexible ensemble learning framework that optimally combines predictions from multiple base learners. In contrast to our method, SuperLearner learns temperature scaling factors without considering the input. Thus, finding the best scaling for the total validation set.

\item \textbf{Mixture of experts}  (\MoE{}) is a popular method to improve performance by combining specialized submodels of experts~\citep{NEURIPS2022_91edff07, lepikhin2020gshard, rau2019moe, Zoph2022STMoEDS}, each one of which focuses on a specific input region. A gating network determines which expert is relevant for a given input.
We use the Sparse MoE implementation~\citep{Zoph2022STMoEDS} by \citet{rau2019moe}, where the MoE layer's input is the concatenation of vision features $x_b$ from $\phi_b$. We use standard hyperparameters: $9$ experts trained for classification with cross-entropy loss with Adam \citep{kingma2014adam} and a learning rate of $2\mathrm{e}{-5}$ for 300 epochs.
\end{itemize}\vspace{1pt}

See Appendix~\ref{app:baselines} a discussion of additional baselines.
The appendix also contains additional results on
the combination of backbones and how they perform
under distribution shifts on the \Imagenet{} dataset (Section~\ref{app:robustness}). 
Finally, Section~\ref{app:alpha_values} presents
dataset-specific results including
overlap diagrams, an analysis of possible model combinations using \NNC{} and our \Oracle{}, as well as the dataset-specific learned temperature values. 

\pgfplotstableread[row sep=\\,col sep=&]{
dataset  & Conf & Vote T-1  & Vote T-3  & Log-Avg  & C-Log-Avg  & C-Conf  & GAC  & NLC & Single-Best & SL \\
Caltech101  & 83.2 & 86.3 & 86.4 & 86.9 & 86.6 & 88.9 & 87.4 & 91.1 & 86.6 & 87.6 \\
Cars  & 67.6 & 79.7 & 81.2 & 80.9 & 81.2 & 79.7 & 81.9 & 84.1 &  79.4 & 82.3 \\
Cifar10  & 82.5 & 94.3 & 94.9 & 94.9 & 94.9 & 88.9 & 95.4 & 96.7 & 95.6 & 96.1 \\
Cifar100  & 54.4 & 74.6 & 77 & 75 & 77.3 & 74.1 & 78.2 & 79.6 & 75.8 & 78.2 \\
CLEVR  & 21.6 & 22.7 & 22.5 & 23.6 & 23.8 & 14.5 & 24.3 & 54.4 & 24.4 & 29.0 \\
Country211  & 24.4 & 30.7 & 32.7 & 31.3 & 33.1 & 33.2 & 34.8 & 36.1 & 34.5 & 36.0\\
Cub  & 57.1 & 68.1 & 69.3 & 69.8 & 69.9 & 65.7 & 70 & 73.6 & 63.0 & 70.8 \\
DTD  & 50.2 & 58 & 59.4 & 57.7 & 58.8 & 58.8 & 58.5 & 63.2 & 56.4 & 59.6 \\
EUROSAT  & 35.7 & 48.2 & 50.7 & 49.4 & 51.6 & 53.4 & 51.9 & 87.1 & 48.0 & 53.5\\
FGVC  & 23.4 & 34.1 & 35 & 35.3 & 36.3 & 32.7 & 36.7 & 38.2 & 33.2 & 36.8 \\
Flowers  & 70.7 & 78.4 & 78.9 & 77.9 & 78.9 & 78.9 & 79 & 82.1 & 79.1 & 80.4\\ 
Food  & 85.4 & 92.5 & 93 & 92.9 & 92.9 & 92.5 & 93.8 & 94.3 & 93.1 & 94.2\\ 
GTSRB  & 40.8 & 53.9 & 55.9 & 54.8 & 48.1 & 44.3 & 54.2 & 70.5 & 52.4 & 54.7\\
Imgnet1k  & 65.2 & 76.1 & 76.8 & 76.6 & 76.8 & 75.9 & 78 & 78.8 & 76.6 & 78.3 \\
MNIST  & 66.5 & 70 & 74 & 74.8 & 60.9 & 57.2 & 84.7 & 93.9 & 78.9 & 84.1\\
PCAM  & 58.8 & 62.7 &  62.7  & 62.3 & 63.3 & 51.2 & 61.3 & 84.9 & 63.9 & 71.0\\
Pets  & 88.5 & 93.9 & 94.1 & 94.2 & 94.3 & 93.3 & 94.8 & 95.4 & 93.8 & 95.2\\
SST2  & 59.2 & 71.7 &  71.7 & 72.6 & 73.3 & 56.7 & 72.6 & 77.3 & 71.0 & 73.3\\
Resisc45  & 54.2 & 67.3 & 68.1 & 68.3 & 68.1 & 64.5 & 68.8 & 78.3 & 64.6 & 73.4\\
STL10  & 98.3 & 98.8 & 98.8 & 98.8 & 98.9 & 98 & 98.8 & 99.5 & 99.4 & 99.5\\
SUN397  & 63.1 & 70 & 70.4 & 70.9 & 70.4 & 67.7 & 71.6 & 74.2 & 67.7 & 71.9\\
}\mydata
\begin{figure}[t]
\begin{minipage}{0.8\textwidth}
    \centering
    \begin{tikzpicture}
      \begin{axis}[
        width=1.03\textwidth,
        height=3.2cm,
        ybar=0pt,
        bar width=2.1pt,
        ymin=0,
        enlarge x limits={abs=7pt},
        legend style={draw=none,at={(0.55,1.3)}, 
                    text width=1.3cm,
                    anchor=north,
                    legend columns=6,
                    fill=none,
                nodes={scale=0.7, transform shape},
                },
        ylabel={\scriptsize{Accuracy}},
        ylabel style={yshift=1cm},
        ymajorgrids=true,
        ytick={0,25,50,75,100},
        yticklabel={$\pgfmathprintnumber{\tick}$\%},
        ylabel near ticks,
        xlabel near ticks,
        axis x line*=bottom,
        symbolic x coords={Caltech101,Cars,Cifar10, Cifar100, CLEVR, Country211, Cub, DTD,EUROSAT,FGVC,Flowers,Food,GTSRB,Imgnet1k,MNIST,PCAM,Pets,SST2,Resisc45,STL10,SUN397}, 
        cycle list={zerothcolorbar, 
                    thirdcolorbar, 
                    firstcolorbar,
                    sixthcolorbar, 
                    eighthcolorbar, 
                    ninthcolorbar
                    },
        xtick=data,
        x tick label style={rotate=90, anchor=east, align=right,text width=1.2cm,xshift=0.2cm},
        xticklabel={\textsc{\tick}},
        font=\tiny,
        minor x tick num=1,
        xminorgrids,
        minor tick length=0,
        major x tick style = transparent,
      ]
      \pgfplotsinvokeforeach{Single-Best,
      Vote T-1,
      Log-Avg,
      C-Log-Avg,
      SL,
      NLC}{
        \addplot+[draw=none, fill, area legend,] table [col sep=&,y=#1]{\mydata};
        \addlegendentry{\tiny{\textsc{#1}}}
        }
    \end{axis}
    \end{tikzpicture}
\end{minipage}
\hfill
\begin{minipage}{0.19\textwidth}
    \centering
    \resizebox{\textwidth}{!}{%
    \begin{tabular}{rccc}
        \textbf{} &
          \rotatebox[origin=l]{90}{Mean $\Delta$} &
          \rotatebox[origin=l]{90}{Min $\Delta$} &
          \rotatebox[origin=l]{90}{Max $\Delta$} \\
        \cmidrule(lr){2-4} 
        \VoteOne              & \downr{-0.3} & \downr{-9.0}  & \upg{5.1} \\
        \LogitAvg             & \upg{0.5}    & \downr{-4.1}  & \upg{6.7} \\
        \cmidrule(lr){2-4} 
        \GAC                  & \upg{1.9}    & \downr{-2.6}  & \upg{6.9} \\
        \SuperLearner         & \upg{3.3}    & \upg{0.1}     & \upg{8.8} \\
        \cmidrule(lr){2-4} 
        \NNC{} &
          \upg{9.1} &
          \upg{0.1} &
          \upg{39.1}
    \end{tabular}%
    }
\end{minipage}
\captionof{figure}{Comparison of the top-performing single backbone (\Best{}) with \NNC{} and other ensemble strategies on top of zero-shot CLIP backbones. In the table, Mean, Max and Min $\Delta$ summarize the difference in performance across datasets with respect to the \Best{} backbone.}
\label{fig:barchar_nnc_zs}
\vspace{-1.5em}
\end{figure}
\subsection{Comparison of Backbone-Ensembling Methods}
\label{secMainResults}
In this section, we evaluate the proposed \NNC{} approach along with 
various non-parametric and parametric ensembling techniques.
The non-parametric ones are ``static'' while the
parametric ones use
the \textbf{training split} of each target dataset to adjust the ensemble adaptively.
The following results will show the importance of this adaptive setting to leverage the backbones' respective strengths, and the superiority of our method over baseline parametric techniques. 

Figure~\ref{fig:barchar_nnc_zs} and Table~\ref{tab:zs-baselines} show the performance of each technique.
Among \textbf{non-parametric baselines}, 
\LogitAvg{} fails to enhance overall performance beyond the best backbone.
One notable exception is the \Country{} dataset where \LogitAvg{} shows a substantial improvement over the best backbone.
Overall, though, simple score averaging does not seem to exploit the complementarity of the backbones, with an improvement in accuracy of only $+1.3$\%.
When the backbones are calibrated with \CLogitAvg{},
performance improves compared to the non-calibrated version, across all datasets except \Gtsrb{}, \Mnist{}, \Resisc45{}, and \SUN397{}.
\textbf{Vote T-1} is mostly ineffective and does not significantly improve over the best backbone.
Among \textbf{parametric methods}, the proposed \NNC{} shows
the most substantial improvement, up to $+39.1$\% on the \Eurosat{} dataset. On average, we obtain a commendable improvement of $+9.1$\% 
over the \Best{} backbone across all datasets.

\subsection{Investigating Benefits beyond Dataset-Specific Linear Classifiers}
We now design an additional experiment to investigate the benefits of the proposed \NNC{}.
A standard approach to adapt CLIP is to train a dataset-specific linear classifier on frozen visual features.
In this section, we first train such linear classifiers,
using each backbone's frozen features and each dataset's training split.
Whereas \NNC{} normally acts on raw CLIP scores,
we now train it to act on the logits from these classifiers instead.
Since the linear classifiers already are strong dataset-specific models, one might expect that only small additional improvements are possible.
Yet, the following results will show that the \NNC{} obtains further performance  gains thanks to the \emph{instance}-level adaptation.
In contrast to alternative approaches, it also never reduces the performance compared to the best single linear classifier.

\textbf{Setup.} We follow the setup of~\citet{li2022elevater}. We initialize the weights of the linear classifiers using language weights, which was shown to be more stable than a random initialization.
The output of each backbone $\phi_b$ is L2-normalized before being passed to each linear classifier, which is trained using 90\% of the target dataset's training split. The remaining 10\% are used to train \NNC{}. 


\pgfplotstableread[row sep=\\,col sep=&]{
dataset & Single-Best & Vote T-1 & Vote T-3 & Conf & Log-Avg & C-Conf & C-Log-Avg & GAC & MoE & NLC & SL \\
Caltech101 & 97.7 & 98.2 & 98.2 & 96.5 & 98 & 97.5 & 98.3 & 98.1 & 96.5 & 98.3 & 98.3\\
Cars & 90.6 & 91.3 & 91.4 & 86 & 91.5 & 91.3 & 91.9 & 92.3 & 90.5 & 91.8 & 91.9\\
Cifar10 & 98 & 97.2 & 97.2 & 91.5 & 98.4 & 96.4 & 97.2 & 98.2 & 97.6 & 98.3 & 98.4\\
Cifar100 & 85.5 & 87.1 & 87 & 73.7 & 89 & 85 & 87.2 & 88.6 & 87.1 & 88.8 & 88.8\\
CLEVR & 68.9 & 71 & 72 & 57.3 & 73.8 & 62.7 & 73.2 & 73.3 & 89.6 & 77.0 & 74.3\\
Country211 & 35.9 & 37.4 & 37.4 & 24.1 & 40.5 & 35.2 & 38.3 & 40.7 & 41.2 & 40.8 & 40.8\\
Cub & 86.3 & 86.7 & 86.5 & 80.7 & 87.3 & 87.3 & 87.5 & 88.3 & 84.8 & 87.2 & 87.7 \\
DTD & 79.1 & 81.3 & 80.9 & 73.5 & 82 & 80.1 & 81.3 & 82 & 72.7 & 79.4 & 81.0\\
EUROSAT & 97.1 & 97.6 & 97.7 & 96.5 & 98 & 96.4 & 97.7 & 51.9 & 96.9 & 97.8 & 98.1 \\
FGVC & 63.9 & 63.8 & 64 & 52.9 & 64.7 & 62.9 & 65.2 & 65.3 & 43 & 66.4 & 65.5\\
Flowers & 98.5 & 98.1 & 98.1 & 94.8 & 98.6 & 98.2 & 98.2 & 98.6 & 96.1 & 98.6 & 98.6\\
Food & 94.7 & 95 & 95 & 87.9 & 95.3 & 95 & 95.2 & 95.2 & 94.9 & 95.4 & 95.3\\
GTSRB & 92.6 & 93.3 & 93.4 & 87.9 & 94.3 & 91.9 & 93.4 & 94.3 & 93.4 & 94.4 & 94.3 \\
Imgnet1k & 84 & 82.6 & 83.4 & 80 & 84.6 & 84.6 & 83.3 & 84.8 & 93.4 & 84.7 & 84.8\\
MNIST & 98.8 & 99.2 & 99.2 & 97.8 & 99.3 & 98.9 & 99.2 & 99.2 & 98.8 & 99.3 & 99.3\\
PCAM & 82.4 & 80.4 & 80.4 & 79.6 & 80.8 & 71.9 & 80.6 & 82.6 & 82.3 & 83.4 & 83.6\\
Pets & 94.6 & 94.8 & 94.7 & 91.4 & 95.1 & 94.3 & 94.9 & 95.1 & 93.4 & 95 & 95\\
SST2 & 80.5 & 80.5 & 80.5 & 73.6 & 81.3 & 61.9 & 80.5 & 81.1 & 79.1 & 80.9 & 81.9\\
Resisc45 & 95.1 & 95.8 & 95.9 & 91.2 & 96.3 & 94.7 & 96 & 96.3 & 94.5 & 96.3 & 96.4\\
STL10 & 99.8 & 99.6 & 99.6 & 98.7 & 99.8 & 99.3 & 99.6 & 99.8 & 98.8 & 99.8 & 99.8\\
SUN397 & 82.5 & 85 & 85 & 79.4 & 85.5 & 83 & 85 & 71.6 & 85 & 85.9 & 85.9\\
}\mydatalp

\begin{figure}[t]
\begin{minipage}{0.8\textwidth}
    \begin{tikzpicture}
      \begin{axis}[
        width=1.04\textwidth,
        height=3cm,
        ybar=0pt,
        bar width=2.1pt,
        font=\tiny,
        ymin=0,
        enlarge x limits={abs=7pt},
        legend style={draw=none,at={(0.5,1.35)}, 
                    text width=1.3cm,
                    anchor=north,
                    legend columns=7},
        ylabel={\scriptsize{Accuracy}},
        ymajorgrids=true,
        ytick={0,25,50,75,100},
        ylabel style={yshift=0.3cm},
        yticklabel={\tiny$\pgfmathprintnumber{\tick}$\%},
        xlabel near ticks,
        axis x line*=bottom,
        symbolic x coords={Caltech101,Cars,Cifar10, Cifar100, CLEVR, Country211, Cub, DTD,EUROSAT,FGVC,Flowers,Food,GTSRB,Imgnet1k,MNIST,PCAM,Pets,SST2,Resisc45,STL10,SUN397}, 
        cycle list={zerothcolorbar, 
                    thirdcolorbar, 
                    firstcolorbar,
                    eighthcolorbar, 
                    color2,
                    ninthcolorbar},
        xtick=data,
        x tick label style={rotate=90, anchor=east, align=right,text width=1.2cm,xshift=0.2cm},
        xticklabel={\tiny{\textsc{\tick}}},
        minor x tick num=1,
        xminorgrids,
        minor tick length=0,
        major x tick style = transparent,
      ]
      \pgfplotsset{label style={yshift=-0.65cm, font=\tiny},}
      \pgfplotsset{every tick label/.append style={font=\tiny}}
      \pgfplotsinvokeforeach{Single-Best,
                            Vote T-1,
                            Log-Avg,
                            SL,
                            MoE,
                            NLC}{
        \addplot+[draw=none, fill, area legend,] table [col sep=&,y=#1]{\mydatalp};
        \addlegendentry{\tiny{\textsc{#1}}}
        }
    \end{axis}
    \end{tikzpicture}
\end{minipage}
\hfill
\begin{minipage}{0.19\textwidth}
    \centering
    \resizebox{\textwidth}{!}{%
    \begin{tabular}{rccc}
        \textbf{} &
          \rotatebox[origin=l]{90}{Mean $\Delta$} &
          \rotatebox[origin=l]{90}{Min $\Delta$} &
          \rotatebox[origin=l]{90}{Max $\Delta$} \\
        \cmidrule(lr){2-4} 
        \VoteOne              & \downr{-0.3} & \downr{-2.0}  & \upg{5.1} \\
        \LogitAvg             & \upg{1.3}    & \downr{-1.5}  & \upg{5.0} \\
        \cmidrule(lr){2-4} 
        \CLogitAvg &
          \upg{0.8} &
          \downr{-1.7} &
          \upg{4.4} \\
        \GAC                  & \upg{-1.4}    & \downr{-45.2}  & \upg{4.8} \\
        \MoE                  & \upg{0.2}    & \downr{-20.9}  & \upg{20.8} \\
        \SuperLearner & \upg{1.6} & 0.0 & \upg{5.4} \\
        \cmidrule(lr){2-4} 
        \NNC{} &
          \upg{1.6} &
          0.0 &
          \upg{8.1}
        
    \end{tabular}%
    }
\end{minipage}
\captionof{figure}{Comparison of the top-performing single backbone (\Best{}) with \NNC{} and ensemble strategies on top of CLIP backbones linear classifier, where Mean, Max and Min $\Delta$ summarize the difference in performance across datasets with respect to the \Best{}.}
\label{fig:nnc_lp}
\vspace{-1.5em}
\end{figure}
\begin{wraptable}[9]{RH}[0pt]{0.6\textwidth}
\caption{Accuracy (\%) in a few-shot setting} 
\label{tab:comparison}
\centering      
\resizebox{0.6\textwidth}{!}{
        \begin{tabular}{lccccc}
        \toprule
         Few-shot  & 1 & 2 & 4 & 8 & 16 \\
         \midrule
        Linear-probe CLIP \cite{radford2021learning} & 22.2 & 31.9 & 41.2 & 49.5 & 56.1 \\
        CoOP~\cite{zhou2022learning} & 47.6 & 50.9 & 56.2 & 59.9 & 63.0 \\
        CLIP-Adapter~\cite{gao2021clip} & 61.2 & 61.5 & 61.8 & 62.7 & 63.6 \\
        Tip-Adapter~\cite{zhang2021tip} & 60.7 & 61.0 & 61.0 & 61.5 & 62.0 \\
        Tip-Adapter-F~\cite{zhang2021tip} & 61.3 & 61.7 & 62.5 & 64.0 & 65.5 \\
        \midrule
        \NNC & \textbf{78.2} & \textbf{78.1} & \textbf{78.2} & \textbf{78.3} & \textbf{78.4} \\
         \bottomrule
        \end{tabular}%
}
\end{wraptable}
\textbf{Results.}
Figure \ref{fig:nnc_lp} and Table~\ref{tab:lp-baseline} present the results combining linear classifiers with both non-parametric and parametric approaches. Similar to Section~\ref{secMainResults}, the proposed \NNC{} consistently improves performance across all datasets, with an average improvement of $+1.6$\% over the \Best{} linear classifier.
As expected, the improvement is reduced compared to the original \NNC{} since some of the gains are now realised by the dataset-specific linear classifier.
Yet Table \ref{tab:lp-table} clearly shows that additional gains can be made over the linear classifiers, i.e.\ that there remains an exploitable complementarity across backbones.
By examining the \Oracle{} performance in this setting, we can see that the potential benefits of combining multiple backbones remain.
Interestingly, the \MoE{} approach does not reach the performance level of the best backbone on several datasets. It shows a degradation sometimes down to $-20.9$\%. This suggests that MoEs may struggle to partition the input space effectively into distinct clusters, specific to certain experts, potentially limiting its functionality.

\subsection{Comparison with Few-Shot Adapter Methods}
We now examine the performance of the proposed \NNC{} in a few-shot setting.
We also show that it complements existing few-shot adapter methods
and improves their accuracy across various datasets.



We compare our method against Tip-Adapter, Tip-Adapter-F~\citep{zhang2021tip}, few-shot Linear-probe CLIP~\citep{radford2021learning}, CoOP~\citep{zhou2022learning}, and CLIP-Adapter~\citep{gao2021clip}. Linear-probe CLIP fine-tunes a linear classifier on a few-shot training set on the frozen CLIP backbones. CoOP~\citep{zhou2022learning} generates different prompt designs to make prompts learnable. CLIP-Adapter~\citep{gao2021clip} enhances few-shot classification by introducing a feature adapter on CLIP's visual and textual encoders. Tip-Adapter~\citep{zhang2021tip} achieves comparable performance to CLIP-Adapter without requiring training, using a key-value cache model from the few-shot training set. Tip-Adapter-F can further enhance performance by fine-tuning the cache. 
As shown in Table \ref{tab:comparison}, our \NNC{} shows an outstanding performance over compared methods and consistently surpasses other few-shot methods by learning how to combine multiple backbones, which is a complementary approach to any of these adapters. In the appendix,
Table~\ref{tab:few_shot_bbone} shows that current few-shot adapter methods also have performance differences with various backbones, and still, our \NNC{} surpasses the best backbone reported by these existing methods.

\input{plots/tipadapter_nnc}
\begin{wrapfigure}[16]{RH}[0pt]{0.35\textwidth}
    \includegraphics[width=0.35\textwidth]{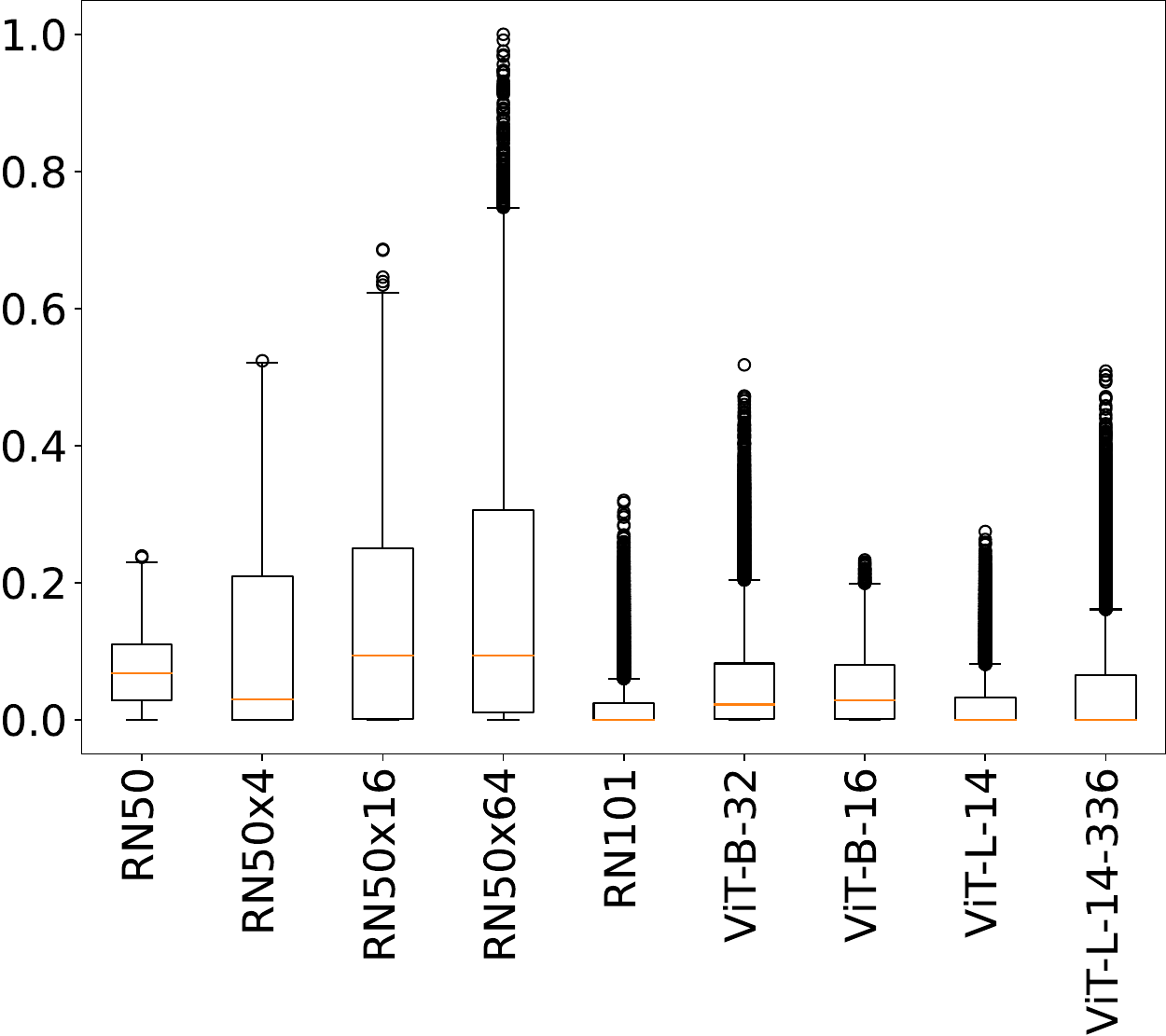}
    \caption{Distribution of max-normalized temperature values learned by our \NNC{} method on the \Clevr{} dataset.}
    \label{fig:alpha_box_plots_eurosat_mainpaper}
\end{wrapfigure}
Finally, we integrate Tip-Adapter~\citep{zhang2021tip} with our \NNC{} and conduct experiments on \Eurosat{}, \FGVC{}, \Dtd{}, and \Pets{}. Initially, we apply Tip-Adapter and Tip-Adapter-F independently on each backbone, following the protocol from~\citet{zhang2021tip} with 1, 2, 4, 8, and 16 shots. Subsequently, we employ our ensembling mechanism to fuse the adapted backbones. To train \NNC{}, we follow their setting and use the same validation split as used by Tip-Adapter to linearly combine their logits with CLIP.

See Figure~\ref{fig:tipadapter_nnc} and Tables \ref{tab:eurosat_tannc}--\ref{tab:pets_tannc} for a comparison of the performance of Tip-Adapter over different zero-shot CLIP backbones, and the combination of all these Tip-Adapter versions with \NNC. Across the four evaluated datasets, \NNC{} improves over each version of Tip-Adapter in the few-shot setting. 
Notably, \NNC{} improves the performance of the best Tip-Adapter backbone (L-14-336) of up to 10\% for \Eurosat{} using 16 shots. \revised{Moreover, when we compare \NNC{} and SL in the Tip-Adapter settting, \NNC{} shows better performance, Table \ref{tab:eurosat_tannc}--\ref{tab:pets_tannc}.}

\subsection{Examining the Learned Temperature Values}
We visualise the learned temperature values to understand how the \NNC{} method assigns weights to different backbones, revealing its strategy for leveraging the strengths of diverse architectures. 
Figure \ref{fig:alpha_box_plots_eurosat_mainpaper} shows the learned temperature values for \NNC{} on the \Clevr{} dataset (see Section \ref{app:alpha_values} for other datasets).
By examining the learned temperatures, we can gain insight into the contribution of each backbone. We note that the backbone with the largest overall weight is not consistently the deepest one, such as ResNet-101. 
The uniform distribution of learned temperatures across backbones indicates that \NNC{} effectively utilizes the strengths of multiple backbones, resulting in a more balanced and accurate labeling process, complementing our analysis in Figure \ref{wrap-fig:improvement_vs_diversity}.
These findings highlight that \NNC{} benefits from the diversity in model predictions, reinforcing the advantage of combining various backbones.


\section{Related Work}
\label{sec:related_work}

\textbf{Foundation models in computer vision.}
The trend in computer vision over the past decade has been to train 
larger and larger models
on increasingly diverse tasks
and increasingly larger datasets~\citep{gadre2023datacomp,schuhmann2022laion}.
Scale and generality have proved to be key enablers to the performance and robustness of these models
\citep{fang2022data,gan2022vision,mayilvahanan2023does,santurkar2022caption,tu2024closer}.
Various methods have been proposed to pre-train such models
\citep{he2022masked,radford2021learning,yu2022coca}.
CLIP~\citep{radford2021learning} is a notable one, specifically designed to exploit noisy image-text pairs scraped from the internet.
CLIP's significance lies in its scalability and its ability to generate meaningful alignments with prompts, facilitating zero-shot classification \citep{gao2021clip,zhang2021tip,li2022elevater}.
CLIP's popularity is due to its scalability and ability to handle diverse text prompts that enable a variety of downstream applications~\citep{gao2021clip,zhang2021tip,li2022elevater}. 
Various versions of CLIP have been trained on different datasets and vision backbones such as ResNets~\citep{He_2016_CVPR} and ViTs~\citep{dosovitskiy2020vit}.

\textbf{Comparing vision backbones}.
\citet{goldblum2023battle} recently conducted a comprehensive comparison of pretrained backbones across various tasks including image classification, object detection, segmentation, and image retrieval.
In comparison, this paper presents complementary observations
of the strengths of different backbones on different datasets and types of data.
Moreover, we leverage these findings with a novel ensembling approach this complementarity of different backbones.
Other works investigating differences across vision architectures include \citep{angarano2022back,pinto2022impartial,pinto2021vision,wang2022can} and others specific to CNNs~\citep{abello2021dissecting,hermann2020origins} and ViTs~\citep{naseer2021intriguing}.
\revised{Finally, we find that there are works analysing the diversity of models for knowledge distillation \cite{roth2024fantastic}, complementarity on models trained don different subset of data \cite{rame2022recycling} and \cite{zhong-etal-2021-larger} exploring the diversity on LLMs.}

\textbf{Model ensembling.}
Combining multiple machine learning models is a classical approach for improving predictive performance.
Simply averaging the outputs of several models is a simple, effective technique
\citep{bauer1999empirical,breiman1996bagging,dietterich2000ensemble,lakshminarayanan2017simple} 
with studies dating back more than three decades ago.
The approach has also been applied to deep neural networks (deep ensembles \citep{lakshminarayanan2017simple}), which has
shown benefits in various domains including higher accuracy under distribution shift~\citep{ovadia2019can,teney2018tips}.
The diversity of the combined models (in terms of uncorrelated errors) has also been shown to be critical to these improvements~\citep{hao2024benefits,wortsman2022model}. 
The closest to our work is the SuperLearner framework \citep{superlearner2018}, 
which similarly investigates multiple classical ensembles and introduces a new ensemble that learns temperature scaling. 
However, our study not only evaluates the performance of various classical ensembles but also explores the complementarity of CLIP backbones across 21 datasets. 
Furthermore, unlike SuperLearner, our method adapts its temperature scaling factors based on the input, which plays a crucial role in ensembling CLIP backbones.

\section{Conclusions}
\label{sec:conclusion}

This paper presents an analysis of vision backbones in the CLIP framework, focusing on the task of image classification. Unlike prior studies that span various downstream tasks (e.g. \cite{goldblum2023battle})
our emphasis lies in identifying the unique strengths of various backbones.
Our experiments revealed a distinctive complementarity across architectures and an avenue for enhancing CLIP's performance by synergistic combination.
We proposed an ensemble approach that reweights the logits from each backbone condition on the input data to yield more accurate predictions in image classification across a variety of datasets.

\label{sec:limitations}
\textbf{Limitations.}
First, our evaluation focused mostly on image classification, even though CLIP can be used for a variety of downstream tasks. 
Although we test our method on out-of-distribution Sec. \ref{app:robustness} and also obtain the upper bound performance on Image-Text retrieval Sec. \ref{app:imagetext_retrieval}, more work needs to be done on how to combine the representations. 
Second, our approach relies on a late fusion of backbones by adaptively adjusting their logits.
\revised{Alternative approaches that leverage the same initial motivation use knowledge distillation, which requires training the student backbone again \cite{roth2024fantastic}}.
Third, although we employed the Cascade method~\citep{wang2022wisdom} to reduce the computations, our method still needs to pass a test image multiple times through vision encoders. This adds complexity and computational overhead, which can be important if a large number of backbones are considered.
A possible solution would involve fusing backbones at the training stage to require only a single forward pass at test time.

\textbf{Future work.}
We could emulate the behavior of a given backbone by learning their representations using the main branch's early layers. 
Other efficient methods could also be developed to combine predictions from multiple backbones to enhance the scalability of the idea.






\section*{Acknowledge}
This research is funded in part by the Australian Government through the Australian Research Council (Project DP240103278) and the Centre of Augmented Reasoning at the Australian Institute for Machine Learning, established by a grant from the Department of Education. We thank the NVIDIA corporation for donating one of the GPUs used for the experiments. Finally, we would like to thank Stephen Gould for his valuable feedback on the paper.

\bibliography{egbib}

\clearpage
\newpage

\appendix
\renewcommand\thefigure{\thesection.\arabic{figure}}    
\renewcommand\thetable{\thesection.\arabic{table}}

\section{Reproducibility}
\label{app:reprod}
The one-layer MLP is trained with ADAM optimizer with a learning rate of 2e-4. We use a weight decay of 0.01. It receives as input the concatenation of the features, the hidden layer has a width of 128 and its outcome is the temperature values for the used backbones. All the backbones used in the paper are pre-trained using the same dataset called \texttt{openai} and objectives from the OpenAI Fundation \url{https://github.com/mlfoundations/open_clip}.
In the Cascade~\cite{wang2022wisdom}, we use the probability as the confident measurement with a threshold of $0.9$. When the method is confident with a probability bigger than $0.9$ we produce a final prediction, otherwise, we add the next backbone.
All our experiments are done in one NViDIA GeForce RTX 4090 GPU, in a small server with 64 GB of RAM and 32 CPU cores.
\section{Oracle Performance}
\label{app:oracle_performance}
\setcounter{figure}{0}    
\setcounter{table}{0}  
\begin{table}[b]
\centering
\caption{Zero-shot accuracy performance comparison of CLIP backbones on 21 datasets, showcasing the number of parameters and GFLOPs per image for each backbone. Additionally, explore the performance of our empirical upper-bound prediction, denoted as `\Oracle', representing the ideal combination of ResNet (RN), ViT, and All backbones.
}
\label{tab:zs-all-table}
\resizebox{\textwidth}{!}{
\begin{tabular}{ccccccccccccccccccccccccc}
 & & \rotatebox[origin=l]{90}{\makecell{Num \\ Parameters}} & \rotatebox[origin=l]{90}{\makecell{GFLOPs/ \\ Image}} & \rotatebox[origin=l]{90}{\Caltech101} & \rotatebox[origin=l]{90}{\Cars} & \rotatebox[origin=l]{90}{\Cifarten} & \rotatebox[origin=l]{90}{\Cifaronehundred} & \rotatebox[origin=l]{90}{\Clevr} & \rotatebox[origin=l]{90}{\Country211} & \rotatebox[origin=l]{90}{\Cub} & \rotatebox[origin=l]{90}{\Dtd} & \rotatebox[origin=l]{90}{\Eurosat} & \rotatebox[origin=l]{90}{\FGVC} & \rotatebox[origin=l]{90}{\Flowers} & \rotatebox[origin=l]{90}{\Food} & \rotatebox[origin=l]{90}{\Gtsrb} & \rotatebox[origin=l]{90}{\Imagenet} & \rotatebox[origin=l]{90}{\Mnist} & \rotatebox[origin=l]{90}{\Pcam} & \rotatebox[origin=l]{90}{\Pets} & \rotatebox[origin=l]{90}{\Renderedsst2} & \rotatebox[origin=l]{90}{\Resisc45} & \rotatebox[origin=l]{90}{\STL10} & \rotatebox[origin=l]{90}{\SUN397} \\
\addlinespace[-0.2cm]\cmidrule(lr){1-2}\cmidrule(lr){3-3} \cmidrule(lr){4-4} \cmidrule(lr){5-25}\multirow{5}{*}{\rotatebox[origin=c]{90}{CLIP-ResNet}}
 & 50 & 102.0M & 18.2 &77.9 & 54.2 & 71.5 & 40.3 & 21.6 & 15.4 & 46.6 & 41.2 & 28.2 & 17.1 & 66.1 & 77.9 & 35.1 & 59.8 & 49.1 & 63.9 & 85.8 & 56.6 & 44.0 & 94.2 & 58.6 \\
 & 101 & 119.7M & 25.5 &81.8 & 61.1 & 80.8 & 47.6 & 24.4 & 16.9 & 49.6 & 43.7 & 26.4 & 18.6 & 65.2 & 81.9 & 37.5 & 62.3 & 46.0 & 58.1 & 86.9 & 63.9 & 54.5 & 96.8 & 57.2 \\
 & 50x4 & 178.3M & 51.8 &82.0 & 66.9 & 79.4 & 45.1 & 20.5 & 20.4 & 54.2 & 48.6 & 27.4 & 21.1 & 69.9 & 85.5 & 36.2 & 66.2 & 53.3 & 56.7 & 88.9 & 67.2 & 54.3 & 96.6 & 59.7 \\
 & 50x16 & 291.9M & 162.7 &85.1 & 73.2 & 81.3 & 52.1 & 19.6 & 24.4 & 57.9 & 53.4 & 41.2 & 27.5 & 72.0 & 89.7 & 39.8 & 70.7 & 62.7 & 62.5 & 90.1 & 67.5 & 59.4 & 97.8 & 63.6 \\
 & 50x64 & 623.3M & 552.7 &83.9 & 75.9 & 85.1 & 59.9 & 22.7 & 29.8 & 62.7 & 53.1 & 48.0 & 31.0 & 76.1 & 91.1 & 47.9 & 73.9 & 78.9 & 53.9 & 93.6 & 71.0 & 63.9 & 98.3 & 66.0 \\
\cmidrule(lr){1-2}\cmidrule(lr){3-3} \cmidrule(lr){4-4} \cmidrule(lr){5-25}\multirow{4}{*}{\rotatebox[origin=c]{90}{CLIP-ViT}}
 & B-32 & 151.3M & 14.9 &83.7 & 59.7 & 89.8 & 64.2 & 23.2 & 17.2 & 53.0 & 44.0 & 37.4 & 19.7 & 66.5 & 82.6 & 32.6 & 63.3 & 38.1 & 62.3 & 87.5 & 58.6 & 54.2 & 97.1 & 61.3 \\
 & B-16 & 149.6M & 41.1 &86.6 & 64.6 & 90.8 & 66.9 & 21.2 & 22.8 & 55.3 & 45.1 & 44.1 & 24.4 & 71.4 & 87.9 & 43.3 & 68.3 & 59.5 & 50.7 & 89.1 & 60.5 & 58.3 & 98.2 & 63.7 \\
 & L-14 & 427.6M & 175.3 &85.9 & 77.8 & 95.6 & 75.8 & 19.4 & 31.9 & 62.1 & 55.3 & 47.2 & 31.7 & 79.1 & 92.3 & 50.6 & 75.5 & 73.3 & 52.0 & 93.6 & 68.9 & 64.6 & 99.4 & 66.7 \\
 & L-14-336 & 429.9M & 395.2 &86.1 & 79.4 & 95.0 & 74.4 & 20.0 & 34.5 & 63.0 & 56.4 & 45.0 & 33.2 & 78.3 & 93.1 & 52.4 & 76.6 & 75.2 & 60.7 & 93.8 & 70.6 & 63.7 & 99.4 & 67.7 \\
\cmidrule(lr){1-2}\cmidrule(lr){3-3} \cmidrule(lr){4-4} \cmidrule(lr){5-25}\multirow{3}{*}{\rotatebox[origin=c]{90}{\Oracle}}
& RN & & & 90.2 & 91.3 & 96.5 & 76.8 & 56.2 & 41.2 & 82.5 & 73.5 & 81.9 & 52.3 & 86.0 & 96.1 & 65.2 & 84.9 & 90.2 & 87.0 & 98.1 & 95.3 & 82.3 & 99.3 & 82.7 \\
& ViT & & & 91.9 & 89.9 & 98.5 & 87.1 & 34.4 & 44.0 & 78.9 & 67.2 & 72.0 & 51.9 & 85.7 & 96.6 & 67.0 & 84.6 & 85.6 & 95.4 & 97.5 & 92.3 & 78.7 & 99.7 & 80.6 \\
& All & & & 93.7 & 95.4 & 99.3 & 91.0 & 60.4 & 52.0 & 88.1 & 78.2 & 91.5 & 66.3 & 88.9 & 98.0 & 77.9 & 89.0 & 95.1 & 99.1 & 99.1 & 98.6 & 88.7 & 99.8 & 87.0 \\
\bottomrule
\end{tabular}
}
\end{table}

Table \ref{tab:zs-all-table} presents the results depicted in Figure \ref{wrap-fig:oracle_kiviat} from the main paper. It includes the number of parameters and GFLOPs/Image for each backbone, the performance of zero-shot CLIP (OpenAI) on the selected datasets, and the performance of the proposed \Oracle{} method.

We observe that backbones with a similar number of parameters from different families can show significant differences in performance. For instance, ViT-B-32 outperforms ResNet-50x4 in the \Cifarten{} and \Cifaronehundred{} benchmarks. However, this trend is not consistent across all datasets, as illustrated by the results on \Imagenet{}. Additionally, a higher number of parameters does not always correlate with better performance across different backbone families.

Notably, the \Oracle{} method, which leverages all possible backbones, shows substantial improvements on several datasets: 43.5\% on \Eurosat{}, 36.0\% on \Clevr{}, 25.1\% on \Cub{}, 21.8\% on \Dtd{}, 16.6\% on \Cifaronehundred{}, and 16.0\% on \Cars{}. This clearly demonstrates the complementarity of each zero-shot CLIP backbone.
\begin{table}[t]
\caption{Impact of artificial alterations of the input image on each backbone in terms of delta accuracy concerning the original performance}
\centering
\resizebox{0.65\textwidth}{!}{%
\begin{tabular}{cccccccccc}
 \cmidrule(lr){2-6} \cmidrule(lr){7-10}
 & \multicolumn{5}{c}{RN} & \multicolumn{4}{c}{ViT}  \\
 & 50 & 50x4 & 50x16 & 50x64 & 101 & B-32 & B-16 & L-14 & L-14-336 \\
 \cmidrule(lr){1-1} \cmidrule(lr){2-6} \cmidrule(lr){7-10}
Resize & -3.81 & -3.47 & -1.90 & \textbf{-1.25} & -4.05 & -4.41 & -2.43 & -2.82 & -2.41 \\
Flip & \textbf{0.12} & -0.36 & -0.36 & -0.55 & -0.67 & -0.22 & -0.12 & -0.80 & -1.01 \\
Grayscale & -6.36 & -7.37 & \textbf{-5.98} & -7.01 & -7.52 & -8.94 & -8.70 & -7.73 & -6.99 \\
Color Jitter & -2.71 & -3.00 & -2.19 & -2.93 & -2.64 & -3.52 & -2.53 & \textbf{-1.67} & -1.73 \\
Gaussian Blur & -5.12 & -3.72 & -1.96 & -2.06 & -4.49 & -3.17 & -3.83 & -3.21 & \textbf{-1.80} \\ 
 \cmidrule(lr){1-1} \cmidrule(lr){2-6} \cmidrule(lr){7-10}
\end{tabular}%
}
\vspace{-0.5cm}
\label{tab:augmentation}
\end{table}
In Table \ref{tab:augmentation}, we present the results of the artificial image variations shown in Figure \ref{wrap-fig:augmentation}. In this experiment, we selected a subset of each test set from the benchmarks and applied specific artificial image variations -\ie, Resize, Flip, Grayscale, Color Jitter, and Gaussian Blur- to evaluate the zero-shot classification performance on these ``corrupted'' images for each backbone.

The results indicate that ResNet-50 and ViT-B-16 are the most resilient to the Flip transformation, RN50x64 shows the highest resilience to Resize, and RN50x16 performs best for Grayscale. ViT-L-14 and L-14-336 exhibit the most robustness against Color Jitter and Gaussian Blur transformations. These findings suggest that each backbone has a unique ability to handle specific types of image variations.

\section{Additional Baselines}
\label{app:baselines}
\setcounter{figure}{0}    
\setcounter{table}{0}    

To complete the experiments and establish a comprehensive baseline compared to other approaches, we introduce two non-parametric and one parametric more ensemble methods to our benchmarks:

\paragraph{\textbf{Confidence} (\Confidence):} This method utilizes the Shannon entropy to assess the confidence of each backbone in making a prediction. The backbone with the highest confidence for a given prediction is selected as the source of the final prediction.

\paragraph{\textbf{Voting} (\VoteOne{} and \VoteThree{})}: Voting combines different classifiers to predict the class label. In the \VoteOne{} approach, we consider the top-1 prediction from each backbone, and the final prediction is determined by the label with the highest number of votes. In cases where multiple labels receive the same number of votes, we select the one with the highest probability. Additionally, we explore the \VoteThree{} method, which seeks consensus among the backbones by considering the three most likely predictions from each backbone. These predictions are weighted based on their position within the top-3 list.

\textbf{Genetic Algorithm} (\GAC). To showcase the importance of adaptively adjusting the logits, we use a genetic algorithm to find the optimal temperatures for combining the backbones. These temperatures are fixed after training, ensuring they remain constant regardless of the input.

\paragraph{\textbf{Calibrated confidence} (\CalibratedConfidence)} For this approach, we first calibrate the probabilities of each backbone independently using temperature scaling. Then follow the procedure above and utilize the Shannon entropy of each backbone to select the one with the highest confidence
\begin{table}[b]
\centering
\caption{Our results on ensembling the zero-shot predictions of CLIP backbones, which we group intro non-parametric and parametric techniques, and also compare to the best-performing single backbone (\Best). 
Mean, Max and Min $\Delta$ summarize the difference in performance when we compare it against the \Best{} backbone across datasets.}

\label{tab:zs-baselines}
\resizebox{\textwidth}{!}{
\begin{tabular}{cccccccccccccccccccccccccc}
 & \rotatebox[origin=l]{90}{\Caltech101} & \rotatebox[origin=l]{90}{\Cars} & \rotatebox[origin=l]{90}{\Cifarten} & \rotatebox[origin=l]{90}{\Cifaronehundred} & \rotatebox[origin=l]{90}{\Clevr} & \rotatebox[origin=l]{90}{\Country211} & \rotatebox[origin=l]{90}{\Cub} & \rotatebox[origin=l]{90}{\Dtd} & \rotatebox[origin=l]{90}{\Eurosat} & \rotatebox[origin=l]{90}{\FGVC} & \rotatebox[origin=l]{90}{\Flowers} & \rotatebox[origin=l]{90}{\Food} & \rotatebox[origin=l]{90}{\Gtsrb} & \rotatebox[origin=l]{90}{\Imagenet} & \rotatebox[origin=l]{90}{\Mnist} & \rotatebox[origin=l]{90}{\Pcam} & \rotatebox[origin=l]{90}{\Pets} & \rotatebox[origin=l]{90}{\Renderedsst2} & \rotatebox[origin=l]{90}{\Resisc45} & \rotatebox[origin=l]{90}{\STL10} & \rotatebox[origin=l]{90}{\SUN397} &  \rotatebox[origin=l]{90}{Mean $\Delta$} &  \rotatebox[origin=l]{90}{Min $\Delta$}  &  \rotatebox[origin=l]{90}{Max $\Delta$} \\
\addlinespace[-0.2cm]\cmidrule(lr){1-1} \cmidrule(lr){2-22} \cmidrule(lr){23-25}
\Best{} & 86.6 & 79.4 & 95.6 & 75.8 & 24.4 & 34.5 & 63.0 & 56.4 & 48.0 & 33.2 & 79.1 & 93.1 & 52.4 & 76.6 & 78.9 & 63.9 & 93.8 & 71.0 & 64.6 & 99.4 & 67.7 \\
\cmidrule(lr){1-1} \cmidrule(lr){2-22} \cmidrule(lr){23-25}
\Confidence & 83.2  & 67.6  & 82.5  & 54.4  & 21.6  & 24.4  & 57.1  & 50.2  & 35.7  & 23.4  & 70.7  & 85.4  & 40.8  & 65.2  & 66.5  & 58.8  & 88.5  & 59.2  & 54.2  & 98.3  & 63.1  & \downr{-8.9} & \downr{-21.4} & \downr{-1.1} \\
\VoteOne & 86.3  & 79.7  & 94.3  & 74.6  & 22.7  & 30.7  & 68.1  & 58.0  & 48.2  & 34.1  & 78.4  & 92.5  & 53.9  & 76.1  & 70.0  & 62.7  & 93.9  & 71.7  & 67.3  & 98.8  & 70.0  & \downr{-0.3} & \downr{-9.0} & \upg{5.1} \\
\VoteThree & 86.4  & 81.2  & 94.9  & 77.0  & 22.5  & 32.7  & 69.3  & 59.4  & 50.7  & 35.0  & 78.9  & 93.0  & 55.9  & 76.8  & 74.0  & - & 94.1  & - & 68.1  & 98.8  & 70.4  &\upg{0.9} & \downr{-4.9} & \upg{6.3} \\
\LogitAvg & 86.9  & 80.9  & 94.9  & 75.0  & 23.6  & 31.3  & 69.8  & 57.7  & 49.4  & 35.3  & 77.9  & 92.9  & \underline{54.8}  & 76.6  & 74.8  & 62.3  & 94.2  & 72.6  & 68.3  & 98.8  & 70.9  & \upg{0.5} & \downr{-4.1} & \upg{6.7} \\
\cmidrule(lr){1-1} \cmidrule(lr){2-22} \cmidrule(lr){23-25}
\CLogitAvg & 86.6  & 81.2  & 94.9  & 77.3  & 23.8  & 33.1  & 69.9  & 58.8  & 51.6  & 36.3  & 78.9  & 92.9  & 48.1  & 76.8  & 60.9  & 63.3  & 94.3  & \underline{73.3}  & 68.1  & 98.9  & 70.4  & \upg{0.1} & \downr{-18.0} & \upg{6.9} \\
\CalibratedConfidence & \underline{88.9}  & 79.7  & 88.9  & 74.1  & 14.5  & 33.2  & 65.7  & 58.8  & 53.4  & 32.7  & 78.9  & 92.5  & 44.3  & 75.9  & 57.2  & 51.2  & 93.3  & 56.7  & 64.5  & 98.0  & 67.7  & \downr{-3.2} & \downr{-21.7} & \upg{5.4} \\
\GAC & 87.4  & 81.9  & 95.4  & 78.2  & 24.3  & 34.8  & 70.0 & 58.5  & 51.9  & 36.7  & 79.0  & 93.8 & 54.2  & 78.0  & \underline{84.7}  & 61.3  & 94.8  & 72.6  & 68.8 & 98.8  & 71.6  & \upg{1.9} & \downr{-2.6} & \upg{6.9} \\
\SuperLearner & 87.6 & \underline{82.3} & \underline{96.1} & \underline{78.2} & \underline{29.0} & \underline{36.0} & \underline{70.8} & \underline{59.6} & \underline{53.5} & \underline{36.8} & \underline{80.4} & \underline{94.2} & \underline{54.7} & \underline{78.3} & 84.1 & \underline{71.0} & \underline{95.2} & \underline{73.3} & \underline{73.4} & \textbf{99.5} & \underline{71.9} &\upg{3.3}	&\upg{0.1}	&\upg{8.8} \\
\cmidrule(lr){1-1} \cmidrule(lr){2-22} \cmidrule(lr){23-25}
Ours & \textbf{91.1} & \textbf{84.1}	&\textbf{96.7}	&\textbf{79.6}	&\textbf{54.4}	&\textbf{36.1}	&\textbf{73.6} &\textbf{63.2}	&\textbf{87.1}	&\textbf{38.2}	&\textbf{82.1}	&\textbf{94.3}	&\textbf{70.5}	&\textbf{78.8}&	\textbf{93.9}	&\textbf{84.9}	&\textbf{95.4} &\textbf{77.3}	&\textbf{78.3}	&\textbf{99.5}	&\textbf{74.2}	&\upg{9.3}	&\upg{0.1}	&\upg{39.1} \\

\cmidrule(lr){1-1} \cmidrule(lr){2-22} \cmidrule(lr){23-25}
\end{tabular}
}
\end{table}
Table \ref{tab:zs-baselines} presents the baselines' results in the combination of zero-shot CLIP backbones. 
In non-parametric baselines, consistently with the presented in the main paper, leveraging the confidence of each backbone in its predictions consistently fails to enhance the overall performance beyond that of the best backbone. 
The lack of calibrated probabilities in the \Confidence{} contributes to overconfidence in some backbones, resulting in performance degradation when combined.
Calibrating the confidence (\CalibratedConfidence{}) leads to improvements, although the performance still falls short of matching the best backbone, except for \Dtd{}, \Caltech101{}, \Eurosat{} and \Cars{}. 
This trend persists in the \LogitAvg{} approach, where averaging performance across backbones does not effectively exploit their complementarity between prediction, yielding an average delta accuracy of 0.1\%. Notably, the \LogitAvg{} approach shows substantial improvement for the \Cub{} dataset compared to the best backbone. When the backbones are calibrated \CLogitAvg, the \LogitAvg{} approach enhances its performances across all datasets except \Gtsrb, \Mnist, \Resisc45, and \SUN397 when compared to the non-calibrated version. Intriguingly, conventional ensemble techniques such as \textbf{Vote T-1} and \textbf{Vote T-3} prove ineffective in providing a significant boost to prediction accuracy beyond that of the best backbone.

In employing parametric methods, we utilize the entire training set of the target datasets. Our proposed approach \NNC{} demonstrates a noteworthy capability to enhance the performance of the best backbone, achieving a substantial improvement of up to 39.1\% in the case of the \Eurosat{} dataset. On average, our method exhibits a commendable improvement of 9.3\% over the \Best{} backbone's performance across the evaluated datasets.

\section{\NNC{} Using Linear classifiers of CLIP Backbones} 
\setcounter{figure}{0}    
\setcounter{table}{0}   
\begin{table}[t]
\centering
\caption{Linear classifier accuracy across multiple datasets and employing different backbones, showcasing the performance improvement achieved by the linear classifier compared to the zero-shot version. Also, the combination of \Oracle{} of linear classifiers. The final three columns present the delta statistics with respect to the zero-shot version of each backbone. In the case of \Oracle{}, the delta with respect to the best linear probing.}
\resizebox{\textwidth}{!}{
\begin{tabular}{cccccccccccccccccccccccccc}
 & & \rotatebox[origin=l]{90}{\Caltech101} & \rotatebox[origin=l]{90}{\Cars} & \rotatebox[origin=l]{90}{\Cifarten} & \rotatebox[origin=l]{90}{\Cifaronehundred} & \rotatebox[origin=l]{90}{\Clevr} & \rotatebox[origin=l]{90}{\Country211} & \rotatebox[origin=l]{90}{\Cub} & \rotatebox[origin=l]{90}{\Dtd} & \rotatebox[origin=l]{90}{\Eurosat} & \rotatebox[origin=l]{90}{\FGVC} & \rotatebox[origin=l]{90}{\Flowers} & \rotatebox[origin=l]{90}{\Food} & \rotatebox[origin=l]{90}{\Gtsrb} & \rotatebox[origin=l]{90}{\Imagenet} & \rotatebox[origin=l]{90}{\Mnist} & \rotatebox[origin=l]{90}{\Pcam} & \rotatebox[origin=l]{90}{\Pets} & \rotatebox[origin=l]{90}{\Renderedsst2} & \rotatebox[origin=l]{90}{\Resisc45} & \rotatebox[origin=l]{90}{\STL10} & \rotatebox[origin=l]{90}{\SUN397} &  \rotatebox[origin=l]{90}{Mean $\Delta$} &  \rotatebox[origin=l]{90}{Min $\Delta$}  &  \rotatebox[origin=l]{90}{Max $\Delta$} \\
\addlinespace[-0.2cm]\cmidrule(lr){1-2} \cmidrule(lr){3-23}  \cmidrule(lr){24-26} \multirow{5}{*}{\rotatebox[origin=c]{90}{ResNet}}
& 50 & 93.8 & 77.5 & 88.5 & 68.8 & 61.3 & 17.9 & 65.6 & 69.0 & 95.0 & 39.9 & 90.9 & 81.7 & 84.5 & 70.3 & 97.4 & 78.3 & 85.5 & 71.3 & 89.0 & 97.1 & 74.9 & \upg{23.5} & \downr{-0.3} & \upg{66.8} \\
& 50x4 & 96.3 & 85.2 & 90.3 & 71.7 & 53.8 & 22.0 & 76.0 & 73.9 & 94.9 & 49.6 & 95.9 & 89.7 & 86.1 & 76.2 & 97.7 & 82.2 & 91.6 & 73.6 & 91.0 & 98.1 & 78.4 & \upg{22.6} & \upg{1.5} & \upg{67.5} \\
& 50x16 & 97.1 & 88.0 & 91.5 & 73.9 & 61.0 & 26.4 & 81.5 & 76.7 & 95.5 & 55.7 & 96.2 & 91.8 & 88.6 & 80.4 & 98.3 & 74.8 & 93.2 & 78.1 & 92.7 & 98.9 & 80.6 & \upg{20.5} & \upg{1.2} & \upg{54.3} \\
& 50x64 & 97.2 & 90.4 & 93.8 & 77.0 & 31.5 & 33.2 & 84.9 & 79.1 & 95.6 & 61.2 & 97.9 & 93.1 & 91.1 & 83.1 & 98.5 & 76.4 & 94.1 & 80.3 & 93.7 & 99.3 & 82.1 & \upg{17.5} & \upg{0.5} & \upg{47.7} \\
& 101 & 95.5 & 80.9 & 90.9 & 72.6 & 61.8 & 20.1 & 70.7 & 69.7 & 94.5 & 41.9 & 92.7 & 84.5 & 83.5 & 72.4 & 97.4 & 75.5 & 89.1 & 66.1 & 90.7 & 98.1 & 77.1 & \upg{22.1} & \upg{1.3} & \upg{68.1} \\

\cmidrule(lr){1-2} \cmidrule(lr){3-23}  \cmidrule(lr){24-26} \multirow{4}{*}{\rotatebox[origin=c]{90}{ViT}}
& B-32 & 95.8 & 76.8 & 94.8 & 77.6 & 59.1 & 18.8 & 70.9 & 71.8 & 95.1 & 41.3 & 92.6 & 83.2 & 85.6 & 73.0 & 98.0 & 77.6 & 87.7 & 69.9 & 90.1 & 98.5 & 74.6 & \upg{20.8} & \upg{0.2} & \upg{59.9} \\
& B-16 & 96.9 & 83.2 & 95.7 & 79.4 & 61.7 & 23.4 & 76.5 & 74.9 & 95.5 & 49.2 & 94.6 & 88.1 & 87.6 & 76.6 & 98.2 & 82.4 & 91.7 & 71.4 & 91.9 & 99.2 & 77.8 & \upg{20.1} & \upg{0.2} & \upg{51.4} \\
& L-14 & 97.6 & 89.1 & 98.0 & 85.5 & 68.9 & 32.2 & 83.3 & 78.3 & 97.1 & 60.9 & 98.4 & 92.4 & 92.6 & 82.2 & 98.8 & 75.5 & 94.6 & 79.4 & 94.6 & 99.8 & 81.4 & \upg{18.2} & \upg{0.1} & \upg{49.9} \\
& L-14-336 & 97.7 & 90.6 & 97.7 & 85.4 & 67.1 & 35.9 & 86.3 & 78.8 & 96.9 & 63.9 & 98.5 & 94.7 & 92.4 & 84.0 & 98.8 & 78.7 & 94.6 & 80.5 & 95.1 & 99.7 & 82.5 & \upg{18.2} & \upg{0.3} & \upg{51.9} \\

\cmidrule(lr){1-2} \cmidrule(lr){3-23}  \cmidrule(lr){24-26} \multirow{3}{*}{\rotatebox[origin=c]{90}{\Oracle}}
& RN & 99.1 & 96.1 & 98.2 & 91.4 & 91.9 & 48.2 & 93.0 & 89.5 & 99.0 & 76.1 & 99.1 & 97.6 & 96.3 & 88.9 & 99.8 & 92.6 & 97.6 & 94.9 & 98.5 & 99.8 & 92.4 & \upg{8.1} & \upg{0.5} & \upg{28.7} \\
& ViT & 99.0 & 95.7 & 99.5 & 94.1 & 91.4 & 49.4 & 92.8 & 88.6 & 99.2 & 78.1 & 99.3 & 97.6 & 96.9 & 89.1 & 99.7 & 92.7 & 97.7 & 92.6 & 98.5 & 99.9 & 91.5 & \upg{6.5} & \upg{0.1} & \upg{22.5} \\
& All & 99.4 & 97.6 & 99.7 & 96.7 & 97.2 & 59.5 & 95.6 & 92.7 & 99.6 & 84.6 & 99.6 & 98.8 & 98.2 & 91.8 & 99.9 & 96.0 & 98.7 & 96.9 & 99.5 & 100.0 & 95.0 & \upg{9.1} & \upg{0.2} & \upg{28.3} \\
\cmidrule(lr){1-2} \cmidrule(lr){3-23}  \cmidrule(lr){24-26} 
\end{tabular}
}
\label{tab:lp-table}
\end{table}
\begin{table}[bt]
\centering
\caption{Our results on combining the LinearProbe CLIP predictions with different backbones, which we group into non-parametric and parametric techniques. Mean, Max and Min $\Delta$ summarize the difference in performance across datasets.}
\label{tab:lp-baseline}
\resizebox{\textwidth}{!}{
\begin{tabular}{ccccccccccccccccccccccccc}
 & \rotatebox[origin=l]{90}{\Caltech101} & \rotatebox[origin=l]{90}{\Cars} & \rotatebox[origin=l]{90}{\Cifarten} & \rotatebox[origin=l]{90}{\Cifaronehundred} & \rotatebox[origin=l]{90}{\Clevr} & \rotatebox[origin=l]{90}{\Country211} & \rotatebox[origin=l]{90}{\Cub} & \rotatebox[origin=l]{90}{\Dtd} & \rotatebox[origin=l]{90}{\Eurosat} & \rotatebox[origin=l]{90}{\FGVC} & \rotatebox[origin=l]{90}{\Flowers} & \rotatebox[origin=l]{90}{\Food} & \rotatebox[origin=l]{90}{\Gtsrb} & \rotatebox[origin=l]{90}{\Imagenet} & \rotatebox[origin=l]{90}{\Mnist} & \rotatebox[origin=l]{90}{\Pcam} & \rotatebox[origin=l]{90}{\Pets} & \rotatebox[origin=l]{90}{\Renderedsst2} & \rotatebox[origin=l]{90}{\Resisc45} & \rotatebox[origin=l]{90}{\STL10} & \rotatebox[origin=l]{90}{\SUN397} & \rotatebox[origin=l]{90}{Mean $\Delta$} & \rotatebox[origin=l]{90}{Min $\Delta$} & \rotatebox[origin=l]{90}{Max $\Delta$}\\
\addlinespace[-0.2cm]\cmidrule(lr){1-1} \cmidrule(lr){2-22}   \cmidrule(lr){23-25}   
\Best & 97.7 & 90.6 & 98.0 & 85.5 & 68.9 & 35.9 & 86.3 & 79.1 & 97.1 & 63.9 & 98.5 & 94.7 & 92.6 & 84.0 & 98.8 & 82.4 & 94.6 & 80.5 & 95.1 & 99.8 & 82.5 & - & - & - \\
\cmidrule(lr){1-1} \cmidrule(lr){2-22}  \cmidrule(lr){23-25}  
\VoteOne & 98.2 & 91.3 & 97.2 & 87.1 & 71.0 & 37.4 & 86.7 & 81.3 & 97.6 & 63.8 & 98.1 & 95.0 & 93.3 & 82.6 & 99.2 & 80.4 & 94.8 & 80.5 & 95.8 & 99.6 & 85.0 & \upg{0.5} & \downr{-2.0} & \upg{2.5} \\
\VoteThree & 98.2 & 91.4 & 97.2 & 87.0 & 72.0 & 37.4 & 86.5 & 80.9 & 97.7 & 64.0 & 98.1 & 95.0 & 93.4 & 83.4 & 99.2 & 80.4 & 94.7 & 80.5 & 95.9 & 99.6 & 85.0 & \upg{0.5} & \downr{-2.0} & \upg{3.2} \\
\Confidence & 96.5 & 86.0 & 91.5 & 73.7 & 57.3 & 24.1 & 80.7 & 73.5 & 96.5 & 52.9 & 94.8 & 87.9 & 87.9 & 80.0 & 97.8 & 79.6 & 91.4 & 73.6 & 91.2 & 98.7 & 79.4 & \downr{-5.3} & \downr{-11.8} & \downr{-0.5} \\
\LogitAvg & 98.0 & 91.5 & 98.4 & 89.0 & 73.8 & 40.5 & 87.3 & 82.0 & 98.0 & 64.7 & 98.6 & 95.3 & 94.3 & 84.6 & 99.3 & 80.8 & 95.1 & 81.3 & 96.3 & 99.8 & 85.5 & \upg{1.3} & \downr{-1.5} & \upg{5.0} \\
\cmidrule(lr){1-1} \cmidrule(lr){2-22}   \cmidrule(lr){23-25}  
\CalibratedConfidence & 97.5 & 91.3 & 96.4 & 85.0 & 62.7 & 35.2 & 87.3 & 80.1 & 96.4 & 62.9 & 98.2 & 95.0 & 91.9 & 84.6 & 98.9 & 71.9 & 94.3 & 61.9 & 94.7 & 99.3 & 83.0 & \downr{-1.8} & \downr{-18.6} & \upg{1.0} \\
\CLogitAvg & 98.3 & 91.9 & 97.2 & 87.2 & 73.2 & 38.3 & 87.5 & 81.3 & 97.7 & 65.2 & 98.2 & 95.2 & 93.4 & 83.3 & 99.2 & 80.6 & 94.9 & 80.5 & 96.0 & 99.6 & 85.0 & \upg{0.8} & \downr{-1.7} & \upg{4.4} \\
\GAC & 98.1 & 92.3 & 98.2 & 88.6 & 73.3 & 40.7 & 88.3 & 82.0 & 51.9 & 65.3 & 98.6 & 95.2 & 94.3 & 84.8 & 99.2 & 82.6 & 95.1 & 81.1 & 96.3 & 99.8 & 71.6 & \downr{-1.4} & \downr{-45.2} & \upg{4.8} \\
\MoE & 96.5 & 90.5 & 97.6 & 87.1 & 89.6 & 41.2 & 84.8 & 72.7 & 96.9 & 43.0 & 96.1 & 94.9 & 93.4 & 93.4 & 98.8 & 82.3 & 93.4 & 79.1 & 94.5 & 98.8 & 85.0 & \upg{0.2} & \downr{-20.9} & \upg{20.8} \\
\SuperLearner & 98.3 & 91.9 & 98.4 & 88.8 & 74.3 & 40.8 & 87.7 & 81.0 & 98.1 & 65.5 & 98.7 & 95.3 & 94.3 & 84.8 & 99.3 & 83.6 & 95.0 & 81.9 & 96.4 & 99.8 & 	85.3 & \upg{1.6} & 0.0 & \upg{5.4} \\
\cmidrule(lr){1-1} \cmidrule(lr){2-22}  \cmidrule(lr){23-25}   
\NNC & 98.3 & 91.8 & 98.3 & 88.8 & 77.0 & 40.8 & 87.2 & 79.4 & 97.8 & 66.4 & 98.6 & 95.4 & 94.4 & 84.7 & 99.3 & 83.4 & 95.0 & 80.9 & 96.3 & 99.8 & 85.9 & \upg{1.6} & 0.0 & \upg{8.1} \\
\cmidrule(lr){1-1} \cmidrule(lr){2-22}  \cmidrule(lr){23-25}  
\end{tabular}
}
\end{table}
The results of adapting CLIP using a linear classifier are shown in Table \ref{tab:lp-table}. RN50 demonstrates the most significant improvement, averaging an impressive 23.5\% increase across all datasets, surpassing even ViT-B-32. However, RN50 still falls short of outperforming other backbones overall. ViT-L-14 and ViT-L-14-336 continue to be the top performers across all datasets, even after linear probing.

A notable observation is the superior performance of the \Oracle{} of linear probes compared to any individual backbone, highlighting the potential benefits of combining multiple backbones. Although the performance gap between the \Oracle{} and the best linear probe is consistently observed across datasets, it is less pronounced than in the zero-shot scenario, indicating a smaller margin for improvement.

\section{Robustness Under Distribution Shift of \Imagenet{}}
\label{app:robustness}
\setcounter{figure}{0}    
\setcounter{table}{0}   
\begin{table}[bt]
\caption{Detailed ImageNet robustness performance. IN is used to abbreviate for ImageNet}
\centering
\resizebox{0.5\textwidth}{!}{%
\begin{tabular}{cccccc}
 & IN & IN-V2 & IN-A & IN-R & IN-Sketch \\
 \cmidrule(lr){1-1} \cmidrule(lr){2-2} \cmidrule(lr){3-6}
Best Backbone & 76.6 & 70.3 & \textbf{77.6} & 89.0 & 60.9 \\
NNC (Train on IN) & \textbf{78.8} & \textbf{72.4} & 75.6 & \textbf{89.3} & \textbf{61.8} \\
 \cmidrule(lr){1-1} \cmidrule(lr){2-2} \cmidrule(lr){3-6}
\end{tabular}%
}
\vspace{-2em}
\label{tab:robustness}
\end{table}
To assess how well the combination of backbones performs under varying conditions, we test its robustness using natural distribution shifts in the ImageNet dataset. We evaluate its performance on four datasets representing different distribution shifts: ImageNet-V2 \cite{pmlr-v97-recht19a}, ImageNet Adversarial \cite{hendrycks2021nae}, ImageNet Rendition \cite{hendrycks2021many} and ImageNet Sketch \cite{wang2009learning}. Specifically, we employ a \NNC{} trained on the ZeroShot CLIP backbones using the original \Imagenet{} data \cite{deng2009imagenet}. This allows us to determine whether the learned combination of backbones can maintain its performance across these distribution shift datasets.
In Table \ref{tab:robustness}, we observe that the learned combination of backbones, denoted as \NNC{}, enhances performance in 3 out of 4 selected benchmarks, with improvements ranging from 0.3\% to 2.1\%. However, in the case of ImageNet Adversarial, the performance of the combination of backbones appears to suffer, possibly due to a more complex decision boundary.

\section{The Combination of \NNC{} with Tip-Adapter}
\setcounter{figure}{0}    
\setcounter{table}{0}   
\begin{table}[tb]
    \begin{minipage}{.49\linewidth}
      \caption{Classification accuracy of models under few-shot settings.}
        \label{tab:comparison_app}
      \centering
      
\resizebox{0.9\textwidth}{!}{
        \begin{tabular}{lccccc}
        \toprule
         Few-shot  & 1 & 2 & 4 & 8 & 16 \\
         \midrule
        Linear-probe CLIP \cite{radford2021learning} & 22.2 & 31.9 & 41.2 & 49.5 & 56.1 \\
        CoOP\cite{zhou2022learning} & 47.6 & 50.9 & 56.2 & 59.9 & 63.0 \\
        CLIP-Adapter\cite{gao2021clip} & 61.2 & 61.5 & 61.8 & 62.7 & 63.6 \\
        Tip-Adapter\cite{zhang2021tip} & 60.7 & 61.0 & 61.0 & 61.5 & 62.0 \\
        Tip-Adapter-F\cite{zhang2021tip} & 61.3 & 61.7 & 62.5 & 64.0 & 65.5 \\
        \midrule
        \NNC & \textbf{78.2} & \textbf{78.1} & \textbf{78.2} & \textbf{78.3} & \textbf{78.4} \\
         \bottomrule
        \end{tabular}%
        }
    \end{minipage}%
    \quad
    \begin{minipage}{.45\linewidth}
      \centering
        \caption{Classification accuracy of models on various vision backbones using 16-shots.}
        \label{tab:few_shot_bbone}
\resizebox{0.85\textwidth}{!}{
        \begin{tabular}{lcccccc}
        \toprule
          & \multicolumn{3}{c}{ResNet} & \multicolumn{2}{c}{ViT}\\
         Models & 50 & 101 & 50x16 & B-32 & B-16 &  \\
         \midrule
        Zero-shot CLIP \cite{radford2021learning}& 60.3 & 62.5 & 70.9 & 63.8 & 68.7 \\
        CoOP\cite{zhou2022learning} & 47.6 & 50.9 & 56.2 & 59.9 & 63.0 \\
        CoOP & 63.0 & 66.6 & - & 66.9 & 71.9 \\
        CLIP-Adapter\cite{gao2021clip} & 63.6 & 65.4 & - & 66.2 & 71.1 \\
        Tip-Adapter\cite{zhang2021tip} & 62.0 & 64.8 & 73.0 & 65.6 & 70.8 \\
        Tip-Adapter-F\cite{zhang2021tip} & 65.5 & 68.6 & 75.8 & 68.7 & 73.7 \\
        \midrule
        \NNC & \multicolumn{5}{c}{\textbf{78.4}}  \\
        \bottomrule
        \end{tabular}%
        }
    \end{minipage} 
\end{table}

We integrate the Tip-Adapter~\cite{zhang2021tip} with our \NNC{} ensembling mechanism and conduct experiments on \Eurosat{}, \FGVC{}, \Dtd{}, and \Pets{} datasets. Initially, we apply Tip-Adapter and Tip-Adapter-F independently on each backbone, following their protocol with 1, 2, 4, 8, and 16 shots. Subsequently, we employ our ensembling mechanism to fuse the adapted backbones. For training \NNC{}, we utilize the validation set used by Tip-Adapter to combine their logits with CLIP linearly.

In Table \ref{tab:comparison_app} and \ref{tab:few_shot_bbone}, we compare the performance of different few-shot adapters of CLIP on the ImageNet dataset. It shows that current few-shot adapter methods also have performance differences with various backbones; still, our \NNC{} surpassed the best backbone reported by their method.

Table \ref{tab:eurosat_tannc}, \ref{tab:fgvc_tannc}, \ref{tab:dtd_tannc}, and \ref{tab:pets_tannc} show the zero-shot performance of the CLIP used by Tip-Adapter. It also shows the Tip-Adapter and Tip-Adapter-F performance on each backbone and when we combine all the Tip-Adapter versions and backbones with \NNC{}. Across the four datasets used for this experiment, \NNC{} improve the performance of each version of Tip-Adapter. Notably, \NNC{} for \Eurosat{} obtained an improvement of up to 15\% with respect to the best Tip-Adapter backbone (L-14-336) using 1 shot.

\begin{table}[bt]
    \begin{minipage}[t]{.49\linewidth}
        \centering
        \caption{Tip-Adapter and Tip-Adapter-F fused with our \NNC{} and applied to \Eurosat{} dataset}  
        \label{tab:eurosat_tannc}
        \resizebox{\textwidth}{!}{
        \begin{tabular}{ccccccccc}
        \toprule
        \multicolumn{9}{c}{ZeroShot} \\
        \cmidrule(lr){1-5} \cmidrule(lr){6-9}
        \multicolumn{5}{c}{ResNet} & \multicolumn{4}{c}{ViT} \\
        \cmidrule(lr){1-5} \cmidrule(lr){6-9}
        50 & 50x4 & 50x16 & 50x64 & 101 & B-32 & B-16 & L-14 & L-14-336\\
        37.5 & 32.0 & 40.3 & 49.4 & 32.5 & 45.2 & 47.6 & 58.1 & 63.5\\
        \bottomrule
        \end{tabular}
        }
        \vspace{2em}
        
        \resizebox{\textwidth}{!}{
        \begin{tabular}{clccccc}
        \toprule
        \multicolumn{7}{c}{TiP-Adapter} \\
        \midrule
        & & \multicolumn{5}{c}{Shots} \\
        \multicolumn{2}{c}{Model} & 1 & 2 & 4 & 8 & 16 \\
        \midrule
        \multirow{5}{*}{\rotatebox[origin=l]{90}{ResNet}}
        & 50 & 55.5 & 60.8 & 68.1 & 66.3 & 70.3\\
        & 50x4 & 53.0 & 56.8 & 67.0 & 68.0 & 71.7\\
        & 50x16 & 54.5 & 61.0 & 62.7 & 65.1 & 71.4\\
        & 50x64 & 65.1 & 66.5 & 70.5 & 73.5 & 73.9\\
        & 101 & 47.9 & 49.7 & 60.6 & 64.7 & 67.6\\
        \midrule
        \multirow{4}{*}{\rotatebox[origin=l]{90}{ViT}}
        & B-32 & 54.1 & 61.7 & 68.2 & 70.6 & 69.8\\
        & B-16 & 66.9 & 68.8 & 75.1 & 72.9 & 78.1\\
        & L-14 & 73.0 & 71.1 & 78.4 & 79.2 & 82.4\\
        & L-14-336 & 73.5 & 72.9 & 75.7 & 79.2 & 80.4\\
        \midrule
        \multicolumn{2}{c}{\revised{With SL}} & \revised{77.8} & \revised{81.5} & \revised{85.6} & \revised{86.4} & \revised{85.4} \\
        \multicolumn{2}{c}{With \NNC} & \textbf{88.4} & \textbf{86.8} & \textbf{89.7} & \textbf{90.1} & \textbf{90.7}\\
        \bottomrule
        \end{tabular}
        }
        \vspace{2em}
        
        \resizebox{\textwidth}{!}{
        \begin{tabular}{clccccc}
        \toprule
        \multicolumn{7}{c}{TiP-Adapter-F} \\
        \midrule
        & & \multicolumn{5}{c}{Shots} \\
        \multicolumn{2}{c}{Model} & 1 & 2 & 4 & 8 & 16 \\
        \midrule
        \multirow{5}{*}{\rotatebox[origin=l]{90}{ResNet}}
        & 50 & 60.7 & 64.4 & 73.3 & 77.7 & 84.9\\
        & 50x4 & 59.5 & 61.9 & 76.2 & 81.9 & 84.9\\
        & 50x16 & 61.4 & 68.7 & 75.9 & 80.8 & 83.7\\
        & 50x64 & 71.2 & 69.7 & 78.7 & 81.4 & 86.6\\
        & 101 & 62.7 & 57.7 & 75.4 & 78.8 & 83.6\\
        \midrule
        \multirow{4}{*}{\rotatebox[origin=l]{90}{ViT}}
        & B-32 & 59.4 & 70.1 & 76.5 & 79.9 & 84.9\\
        & B-16 & 66.6 & 71.0 & 79.1 & 83.8 & 88.9\\
        & L-14 & 74.9 & 75.0 & 86.1 & 86.4 & 90.6\\
        & L-14-336 & 72.5 & 76.4 & 86.1 & 85.3 & 91.0\\
        \midrule
        \multicolumn{2}{c}{\revised{With SL}} & \revised{84.5} & \revised{86.7} & \revised{89.9} & \revised{89.0} & \revised{93.1} \\
        \multicolumn{2}{c}{With \NNC} & \textbf{89.7} & \textbf{88.3} & \textbf{91.4} & \textbf{91.1} & \textbf{93.8}\\
        \bottomrule
        \end{tabular}
        }
    \end{minipage}
    \begin{minipage}[t]{.49\linewidth}
        \caption{Tip-Adapter and Tip-Adapter-F fused with our \NNC{} and applied to \FGVC{} dataset}
        \label{tab:fgvc_tannc}
        \centering
        \resizebox{\textwidth}{!}{
        \begin{tabular}{ccccccccc}
            \toprule
            \multicolumn{9}{c}{ZeroShot} \\
            \cmidrule(lr){1-5} \cmidrule(lr){6-9}
            \multicolumn{5}{c}{ResNet} & \multicolumn{4}{c}{ViT} \\
            \cmidrule(lr){1-5} \cmidrule(lr){6-9}
            50 & 50x4 & 50x16 & 50x64 & 101 & B-32 & B-16 & L-14 & L-14-336\\
            17.2 & 21.4 & 27.0 & 30.2 & 18.1 & 19.3 & 24.8 & 32.6 & 33.4\\
            \bottomrule
        \end{tabular}
        }
        \vspace{2em}
        
        \resizebox{\textwidth}{!}{
        \begin{tabular}{clccccc}
            \toprule
            \multicolumn{7}{c}{TiP-Adapter} \\
            \midrule
            & & \multicolumn{5}{c}{Shots} \\
            \multicolumn{2}{c}{Model} & 1 & 2 & 4 & 8 & 16 \\
            \midrule
            \multirow{5}{*}{\rotatebox[origin=l]{90}{ResNet}}
            & 50 & 19.0 & 21.4 & 22.5 & 25.9 & 29.6\\
            & 50x4 & 24.1 & 27.8 & 29.3 & 31.5 & 37.8\\
            & 50x16 & 29.0 & 32.4 & 35.7 & 38.5 & 42.6\\
            & 50x64 & 35.0 & 38.5 & 41.0 & 44.8 & 49.9\\
            & 101 & 20.4 & 22.9 & 24.8 & 28.3 & 32.1\\
            \midrule
            \multirow{4}{*}{\rotatebox[origin=l]{90}{ViT}}
            & B-32 & 21.6 & 23.9 & 23.7 & 27.2 & 31.6\\
            & B-16 & 28.0 & 31.1 & 32.3 & 36.7 & 39.6\\
            & L-14 & 35.6 & 41.7 & 44.9 & 48.9 & 52.3\\
            & L-14-336 & 37.7 & 43.6 & 46.5 & 50.0 & 53.8\\
            \midrule
            \multicolumn{2}{c}{\revised{With SL}} & \revised{39.1} & \revised{44.2} & \revised{47.7} & \revised{51.2} & \revised{54.8} \\
            \multicolumn{2}{c}{With \NNC} & \textbf{40.9} & \textbf{44.5} & \textbf{47.9} & \textbf{51.5} & \textbf{55.0}\\
            \bottomrule
        \end{tabular}
        }
        \vspace{2em}
        
        \resizebox{\textwidth}{!}{
        \begin{tabular}{clccccc}
            \toprule
            \multicolumn{7}{c}{TiP-Adapter-F} \\
            \midrule
            & & \multicolumn{5}{c}{Shots} \\
            \multicolumn{2}{c}{Model} & 1 & 2 & 4 & 8 & 16 \\
            \midrule
            \multirow{5}{*}{\rotatebox[origin=l]{90}{ResNet}}
            & 50 & 20.5 & 22.9 & 26.5 & 30.4 & 35.5\\
            & 50x4 & 26.3 & 29.1 & 32.8 & 36.8 & 42.2\\
            & 50x16 & 31.2 & 36.9 & 38.3 & 43.6 & 49.4\\
            & 50x64 & 37.0 & 40.8 & 45.0 & 49.8 & 54.8\\
            & 101 & 21.7 & 24.0 & 26.9 & 32.0 & 38.0\\
            \midrule
            \multirow{4}{*}{\rotatebox[origin=l]{90}{ViT}}
            & B-32 & 22.7 & 25.3 & 27.5 & 32.9 & 36.9\\
            & B-16 & 30.2 & 34.1 & 36.1 & 40.9 & 44.6\\
            & L-14 & 38.6 & 44.1 & 48.5 & 51.9 & 57.4\\
            & L-14-336 & 40.9 & 45.2 & 49.6 & 52.7 & 59.0\\
            \midrule
            \multicolumn{2}{c}{\revised{With SL}} & \revised{46.7} & \revised{49.1} & \revised{53.4} & \revised{57.4} & \revised{62.4} \\
            \multicolumn{2}{c}{With \NNC} & \textbf{46.8} & \textbf{49.2} & \textbf{53.6} & \textbf{57.5} & \textbf{62.6}\\
            \bottomrule
        \end{tabular}
        }
    \end{minipage}
\end{table}

\begin{table}[bt]
    \begin{minipage}[t]{.49\linewidth}
        \centering
        \caption{Tip-Adapter and Tip-Adapter-F fused with our \NNC{} and applied to \Dtd{} dataset}
        \label{tab:dtd_tannc}
        \resizebox{\textwidth}{!}{
        \begin{tabular}{ccccccccc}
        \toprule
        \multicolumn{9}{c}{ZeroShot} \\
        \cmidrule(lr){1-5} \cmidrule(lr){6-9}
        \multicolumn{5}{c}{ResNet} & \multicolumn{4}{c}{ViT} \\
        \cmidrule(lr){1-5} \cmidrule(lr){6-9}
        50 & 50x4 & 50x16 & 50x64 & 101 & B-32 & B-16 & L-14 & L-14-336\\
        42.1 & 50.8 & 50.3 & 48.4 & 38.4 & 43.7 & 44.6 & 53.1 & 54.0\\
        \bottomrule
        \end{tabular}
        }
        \vspace{2em}
        
        \resizebox{\textwidth}{!}{
        \begin{tabular}{clccccc}
        \toprule
        \multicolumn{7}{c}{TiP-Adapter} \\
        \midrule
        & & \multicolumn{5}{c}{Shots} \\
        \multicolumn{2}{c}{Model} & 1 & 2 & 4 & 8 & 16 \\
        \midrule
        \multirow{5}{*}{\rotatebox[origin=l]{90}{ResNet}}
        & 50 & 46.3 & 49.9 & 54.3 & 58.5 & 61.3\\
        & 50x4 & 52.5 & 56.3 & 59.8 & 62.4 & 65.8\\
        & 50x16 & 56.9 & 59.5 & 61.5 & 65.3 & 67.6\\
        & 50x64 & 55.6 & 61.5 & 63.2 & 67.3 & 70.0\\
        & 101 & 48.2 & 51.5 & 56.6 & 60.2 & 63.6\\
        \midrule
        \multirow{4}{*}{\rotatebox[origin=l]{90}{ViT}}
        & B-32 & 49.3 & 51.9 & 55.3 & 59.4 & 63.2\\
        & B-16 & 51.5 & 52.6 & 58.6 & 62.8 & 67.0\\
        & L-14 & 59.6 & 61.6 & 64.8 & 69.7 & 71.3\\
        & L-14-336 & 59.3 & 61.8 & 64.5 & 69.5 & 72.6\\
        \midrule
        \multicolumn{2}{c}{\revised{With SL}} & \revised{58.7} & \revised{64.1} & \revised{65.5} & \revised{70.3} & \revised{71.5} \\
        \multicolumn{2}{c}{With \NNC} & \textbf{63.7} & \textbf{66.1} & \textbf{68.4} & \textbf{71.6} & \textbf{73.0}\\
        \bottomrule
        \end{tabular}
        }
        \vspace{2em}
        
        \resizebox{\textwidth}{!}{
        \begin{tabular}{clccccc}
        \toprule
        \multicolumn{7}{c}{TiP-Adapter-F} \\
        \midrule
        & & \multicolumn{5}{c}{Shots} \\
        \multicolumn{2}{c}{Model} & 1 & 2 & 4 & 8 & 16 \\
        \midrule
        \multirow{5}{*}{\rotatebox[origin=l]{90}{ResNet}}
        & 50 & 48.8 & 53.5 & 56.6 & 61.9 & 66.5\\
        & 50x4 & 53.7 & 57.2 & 61.6 & 65.7 & 70.6\\
        & 50x16 & 58.8 & 59.2 & 63.1 & 67.4 & 73.5\\
        & 50x64 & 59.0 & 61.8 & 66.6 & 69.9 & 76.0\\
        & 101 & 47.5 & 50.7 & 58.3 & 64.8 & 68.8\\
        \midrule
        \multirow{4}{*}{\rotatebox[origin=l]{90}{ViT}}
        & B-32 & 50.9 & 55.3 & 60.4 & 63.8 & 68.9\\
        & B-16 & 52.7 & 56.0 & 61.0 & 67.6 & 71.0\\
        & L-14 & 60.4 & 60.7 & 67.1 & 71.6 & 75.8\\
        & L-14-336 & 58.9 & 60.9 & 67.1 & 72.4 & 75.7\\
        \midrule
        \multicolumn{2}{c}{\revised{With SL}} & \revised{62.2} & \revised{67.3} & \revised{71.4} & \revised{74.8} & \revised{78.8} \\
        \multicolumn{2}{c}{With \NNC} & \textbf{67.2} & \textbf{68.7} & \textbf{73.3} & \textbf{75.4} & \textbf{79.1}\\
        \bottomrule
        \end{tabular}
        }
    \end{minipage}
    \begin{minipage}[t]{.49\linewidth}
        \centering
        \caption{Tip-Adapter and Tip-Adapter-F fused with our \NNC{} and applied to \Pets{} dataset}
        \label{tab:pets_tannc}
        \resizebox{\textwidth}{!}{
        \begin{tabular}{ccccccccc}
        \toprule
        \multicolumn{9}{c}{ZeroShot} \\
        \cmidrule(lr){1-5} \cmidrule(lr){6-9}
        \multicolumn{5}{c}{ResNet} & \multicolumn{4}{c}{ViT} \\
        \cmidrule(lr){1-5} \cmidrule(lr){6-9}
        50 & 50x4 & 50x16 & 50x64 & 101 & B-32 & B-16 & L-14 & L-14-336\\
        85.8 & 88.9 & 90.1 & 93.7 & 86.9 & 87.4 & 89.1 & 93.6 & 93.8\\
        \bottomrule
        \end{tabular}
        }
        \vspace{2em}
        
        \resizebox{\textwidth}{!}{
        \begin{tabular}{clccccc}
        \toprule
        \multicolumn{7}{c}{TiP-Adapter} \\
        \midrule
        & & \multicolumn{5}{c}{Shots} \\
        \multicolumn{2}{c}{Model} & 1 & 2 & 4 & 8 & 16 \\
        \midrule
        \multirow{5}{*}{\rotatebox[origin=l]{90}{ResNet}}
        & 50 & 86.2 & 87.2 & 86.0 & 86.7 & 88.8\\
        & 50x4 & 88.7 & 89.2 & 89.1 & 88.9 & 90.8\\
        & 50x16 & 90.5 & 90.8 & 91.4 & 91.7 & 90.9\\
        & 50x64 & 93.7 & 93.3 & 93.8 & 93.8 & 94.1\\
        & 101 & 86.8 & 86.6 & 86.9 & 87.3 & 88.1\\
        \midrule
        \multirow{4}{*}{\rotatebox[origin=l]{90}{ViT}}
        & B-32 & 87.1 & 88.0 & 87.6 & 87.9 & 88.1\\
        & B-16 & 89.3 & 90.6 & 90.7 & 90.7 & 91.7\\
        & L-14 & 93.8 & 93.5 & 94.1 & 94.2 & 94.5\\
        & L-14-336 & 93.8 & 93.8 & 94.0 & 93.7 & 94.0\\
        \midrule
        \multicolumn{2}{c}{\revised{With SL}} & \revised{93.8} & \revised{93.6} & \revised{94.3} & \revised{94.4} & \revised{94.5} \\
        \multicolumn{2}{c}{With \NNC} & \textbf{94.5} & \textbf{94.2} & \textbf{94.9} & \textbf{95.0} & \textbf{95.3}\\
        \bottomrule
        \end{tabular}
        }
        \vspace{2em}
        
        \resizebox{\textwidth}{!}{
        \begin{tabular}{clccccc}
        \toprule
        \multicolumn{7}{c}{TiP-Adapter-F} \\
        \midrule
        & & \multicolumn{5}{c}{Shots} \\
        \multicolumn{2}{c}{Model} & 1 & 2 & 4 & 8 & 16 \\
        \midrule
        \multirow{5}{*}{\rotatebox[origin=l]{90}{ResNet}}
        & 50 & 86.6 & 86.8 & 87.2 & 87.5 & 89.5\\
        & 50x4 & 89.7 & 89.5 & 90.7 & 90.7 & 91.8\\
        & 50x16 & 92.8 & 92.0 & 92.9 & 92.9 & 92.9\\
        & 50x64 & 94.2 & 94.1 & 94.2 & 94.1 & 94.3\\
        & 101 & 87.6 & 88.4 & 88.7 & 88.7 & 90.5\\
        \midrule
        \multirow{4}{*}{\rotatebox[origin=l]{90}{ViT}}
        & B-32 & 88.2 & 89.0 & 88.9 & 89.6 & 90.7\\
        & B-16 & 90.9 & 91.3 & 92.0 & 92.3 & 92.8\\
        & L-14 & 94.2 & 94.1 & 94.6 & 94.3 & 95.1\\
        & L-14-336 & 94.3 & 94.3 & 94.8 & 94.1 & 95.0\\
        \midrule
        \multicolumn{2}{c}{\revised{With SL}} & \revised{94.5} & \revised{94.6} & \revised{95.2} & \revised{95.2} & \revised{95.1} \\
        \multicolumn{2}{c}{With \NNC} & \textbf{95.2} & \textbf{95.0} & \textbf{95.3} & \textbf{95.2} & \textbf{95.5}\\
        \bottomrule
        \end{tabular}
        }
    \end{minipage}
\end{table}
\clearpage
\newpage

\section{Training \NNC{} Under Limited Samples}
\setcounter{figure}{0}    
\setcounter{table}{0}   
\begin{table}[bt]
\centering
\caption{Ablation of \NNC{} performance when changing the number of samples used to train. We use \NNC($n$) to denote \NNC{} with $n$ samples per class. \upg{} and \downr{} showcase the improvement in performance compared with \Best{} backbone in a zero-shot setting.}
\label{tab:NNC_limited_samples}
\resizebox{\textwidth}{!}{
\begin{tabular}{ccccccccccccccccccccccccc}
 & \rotatebox[origin=l]{90}{\Caltech101} & \rotatebox[origin=l]{90}{\Cars} & \rotatebox[origin=l]{90}{\Cifarten} & \rotatebox[origin=l]{90}{\Cifaronehundred} & \rotatebox[origin=l]{90}{\Clevr} & \rotatebox[origin=l]{90}{\Country211} & \rotatebox[origin=l]{90}{\Cub} & \rotatebox[origin=l]{90}{\Dtd} & \rotatebox[origin=l]{90}{\Eurosat} & \rotatebox[origin=l]{90}{\FGVC} & \rotatebox[origin=l]{90}{\Flowers} & \rotatebox[origin=l]{90}{\Food} & \rotatebox[origin=l]{90}{\Gtsrb} & \rotatebox[origin=l]{90}{\Imagenet} & \rotatebox[origin=l]{90}{\Mnist} & \rotatebox[origin=l]{90}{\Pcam} & \rotatebox[origin=l]{90}{\Pets} & \rotatebox[origin=l]{90}{\Renderedsst2} & \rotatebox[origin=l]{90}{\Resisc45} & \rotatebox[origin=l]{90}{\STL10} & \rotatebox[origin=l]{90}{\SUN397} & \rotatebox[origin=l]{90}{Mean $\Delta$}& \rotatebox[origin=l]{90}{Min $\Delta$}& \rotatebox[origin=l]{90}{Max $\Delta$} \\
\addlinespace[-0.2cm]\cmidrule(lr){1-1} \cmidrule(lr){2-22}  \cmidrule(lr){23-25}  
\Best & 86.6 & 79.4 & 95.6 & 75.8 & 24.4 & 34.5 & 63.0 & 56.4 & 48.0 & 33.2 & 79.1 & 93.1 & 52.4 & 76.6 & 78.9 & 63.9 & 93.8 & 71.0 & 64.6 & 99.4 & 67.7 & \\
\cmidrule(lr){1-1} \cmidrule(lr){2-22}  \cmidrule(lr){23-25}  
NNC(1) & 87.3 & 82.3 & 93.6 & 78.0 & 25.1 & 36.0 & 71.0 & 59.3 & 55.0 & 37.1 & 79.4 & 93.6 & 54.8 & 77.6 & 83.3 & 62.5 & 94.3 & 73.8 & 70.0 & 98.9 & 71.1 & \upg{2.2} & \downr{-2.0} & \upg{8.0} \\
NNC(2) & 87.3 & 81.4 & 93.0 & 78.2 & 25.1 & 33.2 & 70.0 & 59.9 & 52.3 & 37.2 & 79.5 & 93.5 & 55.3 & 77.4 & 84.4 & 62.5 & 94.5 & 73.1 & 70.2 & 98.9 & 71.1 & \upg{1.9} & \downr{-2.6} & \upg{7.0} \\
NNC(4) & 87.1 & 81.3 & 95.3 & 75.9 & 26.9 & 35.4 & 70.2 & 59.9 & 54.0 & 35.9 & 78.2 & 93.1 & 55.2 & 78.1 & 81.6 & 62.7 & 94.5 & 72.7 & 68.8 & 99.1 & 71.2 & \upg{1.9} & \downr{-1.3} & \upg{7.2} \\
NNC(8) & 87.1 & 81.6 & 95.6 & 76.2 & 26.9 & 36.1 & 70.3 & 58.1 & 54.0 & 37.0 & 80.3 & 93.1 & 54.9 & 78.3 & 82.5 & 67.0 & 94.2 & 72.4 & 68.9 & 99.5 & 71.2 & \upg{2.3} & 0.0 & \upg{7.3} \\
NNC(16) & 87.1 & 82.8 & 95.8 & 77.2 & 27.0 & 36.2 & 71.3 & 60.4 & 53.5 & 37.0 & 80.4 & 93.5 & 55.2 & 78.4 & 83.3 & 69.6 & 94.2 & 73.6 & 69.6 & 99.4 & 72.2 & \upg{2.9} & 0.0 & \upg{8.3} \\
NNC(32) & 87.3 & 82.9 & 96.0 & 78.2 & 27.2 & 36.1 & 71.2 & 60.4 & 51.6 & 37.2 & 80.4 & 94.2 & 55.1 & 78.1 & \revised{85.3} & 75.5 & 94.4 & 74.1 & 70.1 & 99.4 & 72.4 & \revised{3.3} & \revised{0.0} & 11.6 \\
\cmidrule(lr){1-1} \cmidrule(lr){2-22}  \cmidrule(lr){23-25} 

\end{tabular}
}
\end{table}
In our exploration of the effectiveness of the \NNC{} approach, we extend our analysis to a scenario where we limit the samples to combine the zero-shot CLIPs. This experiment allows us to assess the adaptability and performance of our proposed method under limited training data conditions. Table \ref{tab:NNC_limited_samples} presents the performance of \NNC{} by means of a limited number of samples. Although the performance of \NNC{} overall improves when it has more data available to combine the backbones, in most cases, just using one sample \NNC(1) per class is enough to improve its performance. Notably, there is a stop in performance degradation when we use more than 8 samples \NNC(8) in all the benchmarks.

\revised{Moreover, we run five different seeds to train NLC on the different few-shots settings. Results show that the standard deviation of our obtained performance is stable and low across all of our studied datasets, suggesting that our approach is not sensitive to the effects of different samples. It varies with the number of samples used with a mean standard deviation of 0.62 when we used 1 shot and 0.48 when we used 32 shots.}
\section{Learned Temperature Values}
\label{app:alpha_values}
\setcounter{figure}{0}    
\setcounter{table}{0}   
Figure \ref{fig:alpha1} and \ref{fig:alpha2} present the distribution of the temperature values using a box plot for the \NNC{} method, normalized by their maximum value. Notably, there is a dominance in the temperature values for the best backbones in each family, \ie, RN50x64 and ViT-L-14 or ViT-L-14-336. Particularly prominent in \STL10, \Pets, \Dtd, \FGVC and \Flowers, making these backbones especially relevant for \NNC. Interestingly, it is observed that the most weighted backbone within the ResNet family is not consistently ResNet-101, despite its deeper architecture. This observation is evident in datasets such as \Pets{}, \Cub{}, \FGVC{} and \SUN397, where the mean value of the temperature corresponding to ResNet-50 surpasses that of ResNet-101.

Furthermore, the distribution of the temperature values across backbones for datasets such as \Clevr, \Eurosat, and \Gtsrb{} is more uniform compared to other datasets. This suggests that the \NNC{} method is effectively leveraging the strengths of each backbone to arrive at accurate labels for each sample, resulting in a more balanced distribution of weights across different backbones, complementing our analysis on Fig. \ref{wrap-fig:improvement_vs_diversity}.
\begin{figure*}
\centering
\begin{subfigure}{0.32\textwidth}
    \caption{\Caltech}
    \includegraphics[width=\textwidth]{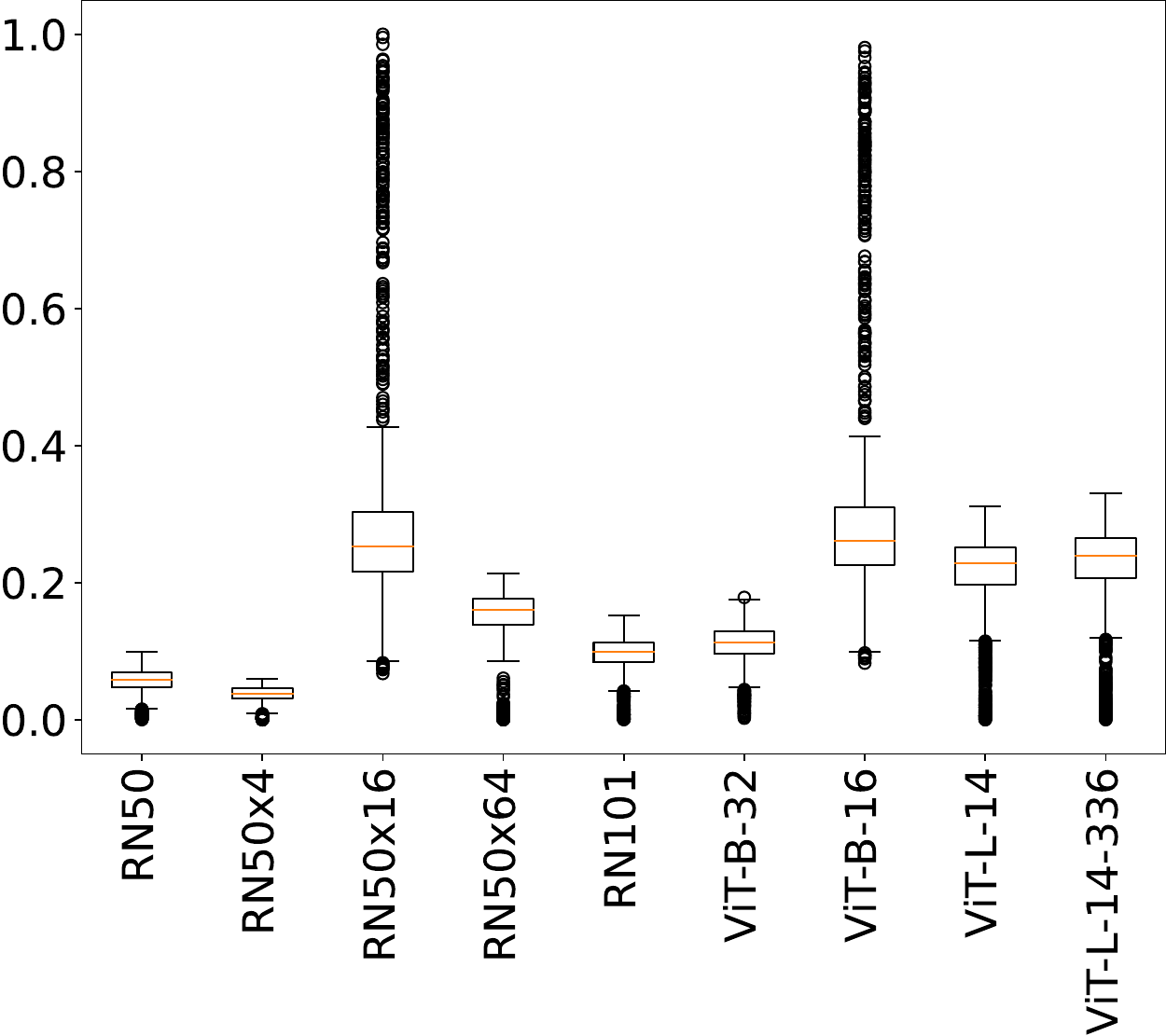}
    \label{fig:alpha_box_plots_caltech}
\end{subfigure}
\begin{subfigure}{0.32\textwidth}
    \caption{\Cars}
    \includegraphics[width=\textwidth]{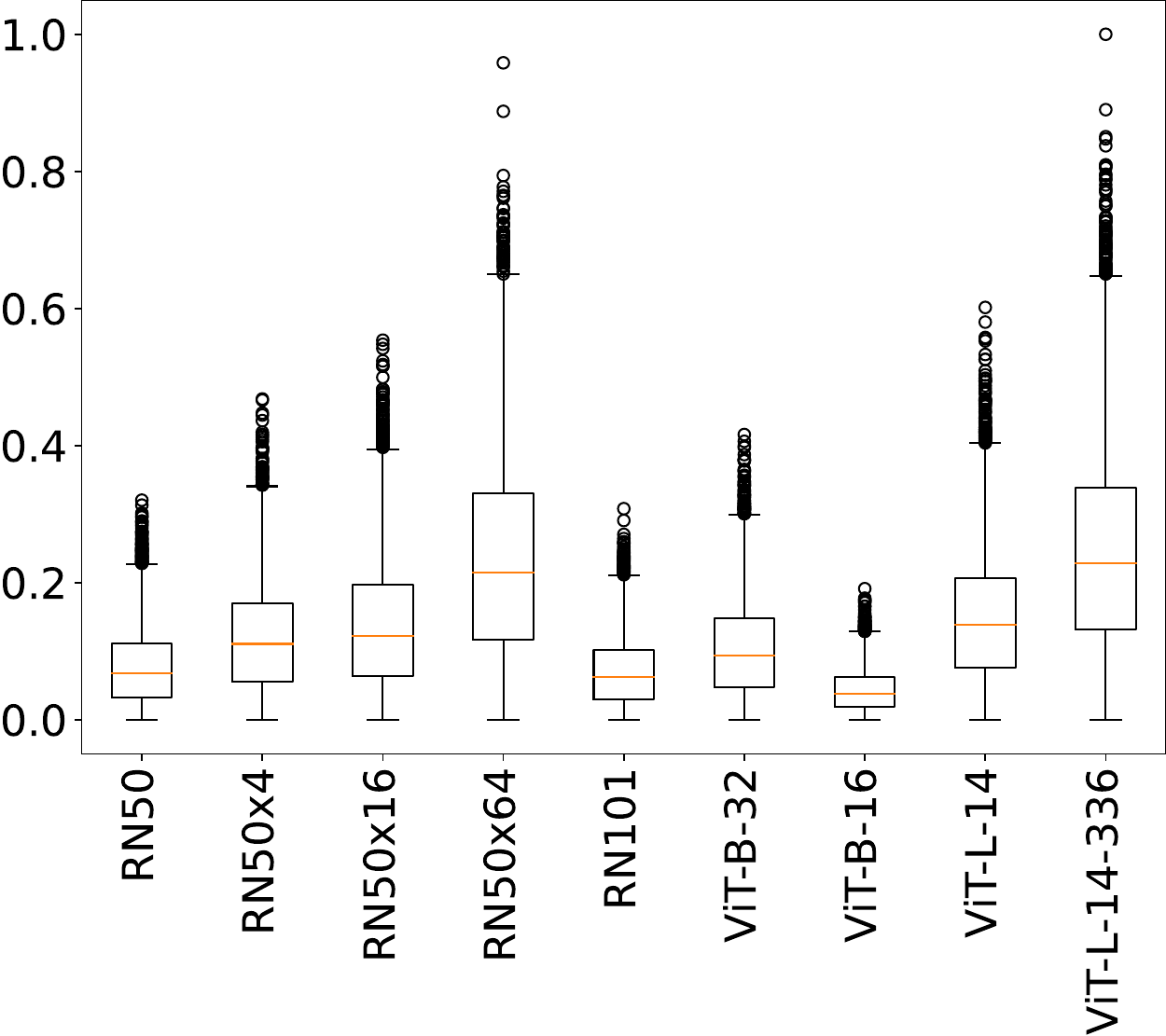}
    \label{fig:alpha_box_plots_cars}
\end{subfigure}
\begin{subfigure}{0.32\textwidth}
    \caption{\Cifarten}
    \includegraphics[width=\textwidth]{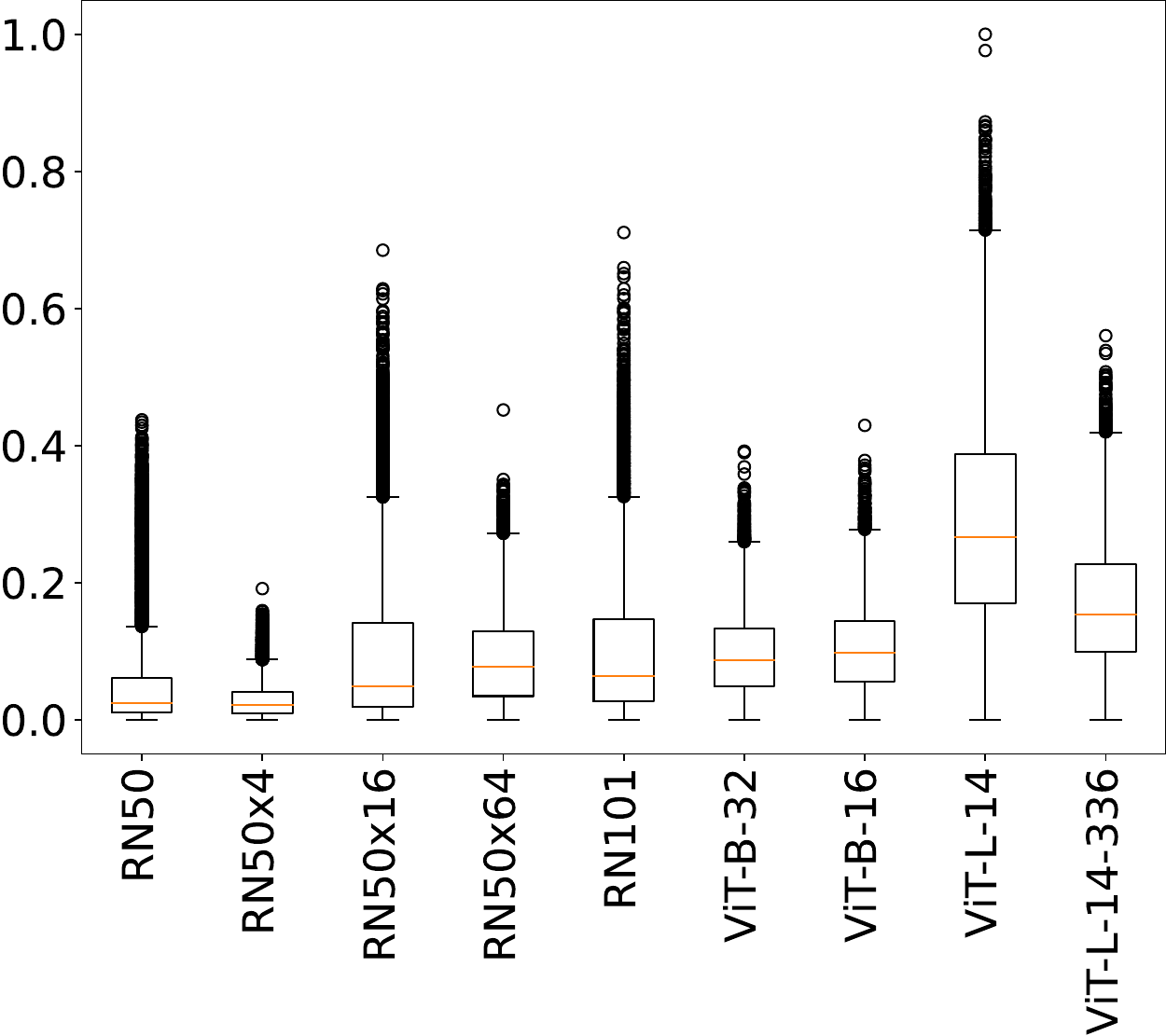}
    \label{fig:alpha_box_plots_cifar10}
\end{subfigure}
\begin{subfigure}{0.32\textwidth}
    \caption{\Cifaronehundred}
    \includegraphics[width=\textwidth]{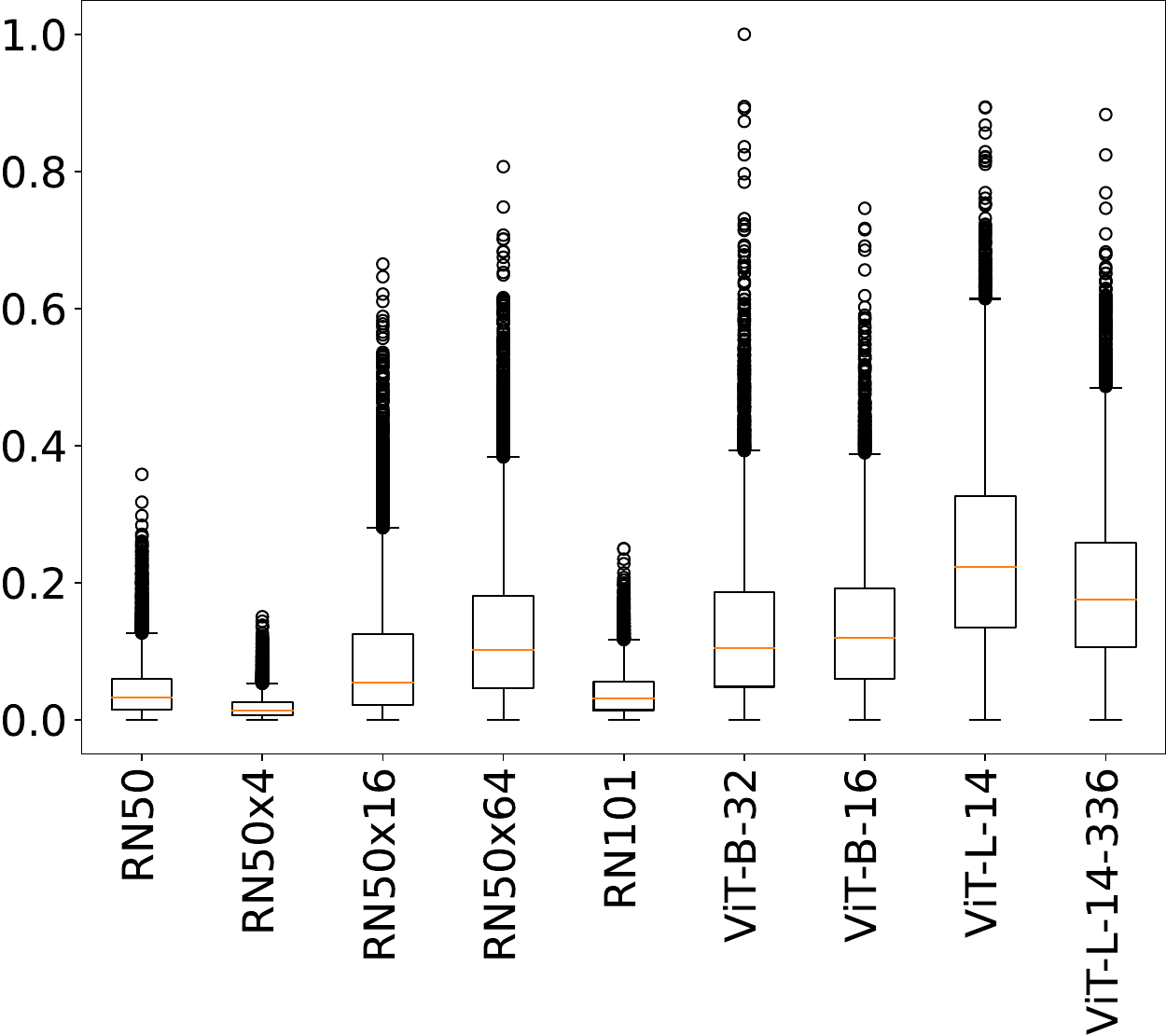}
    \label{fig:alpha_box_plots_cifar100}
\end{subfigure}
\begin{subfigure}{0.32\textwidth}
    \caption{\Clevr}
    \includegraphics[width=\textwidth]{figures/clevr_alphas.pdf}
    \label{fig:alpha_box_plots_clevr}
\end{subfigure}
\hfill
\begin{subfigure}{0.32\textwidth}
    \caption{\Country211}
    \includegraphics[width=\textwidth]{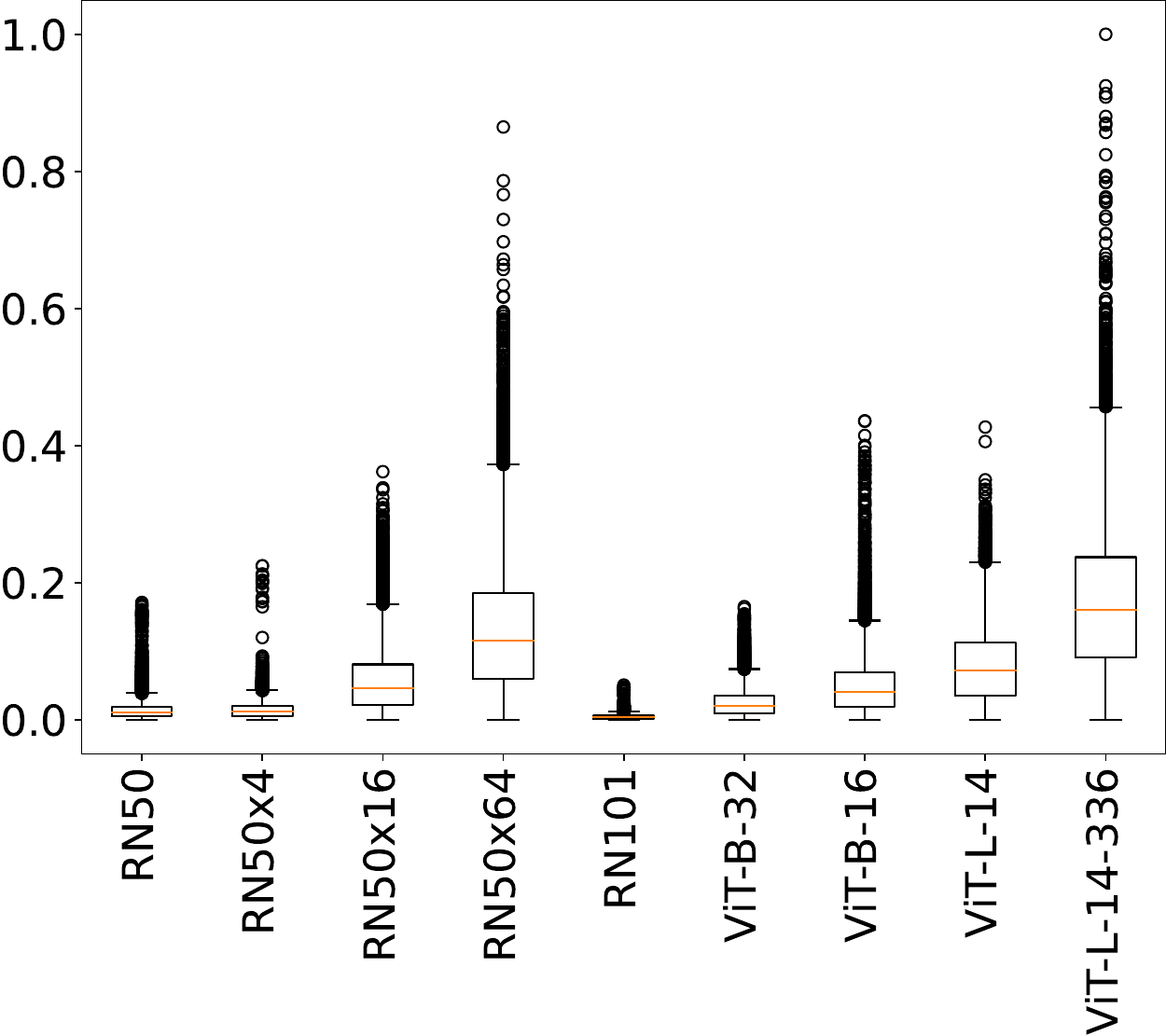}
    \label{fig:alpha_box_plots_country}
\end{subfigure}
\hfill
\begin{subfigure}{0.32\textwidth}
    \caption{\Cub}
    \includegraphics[width=\textwidth]{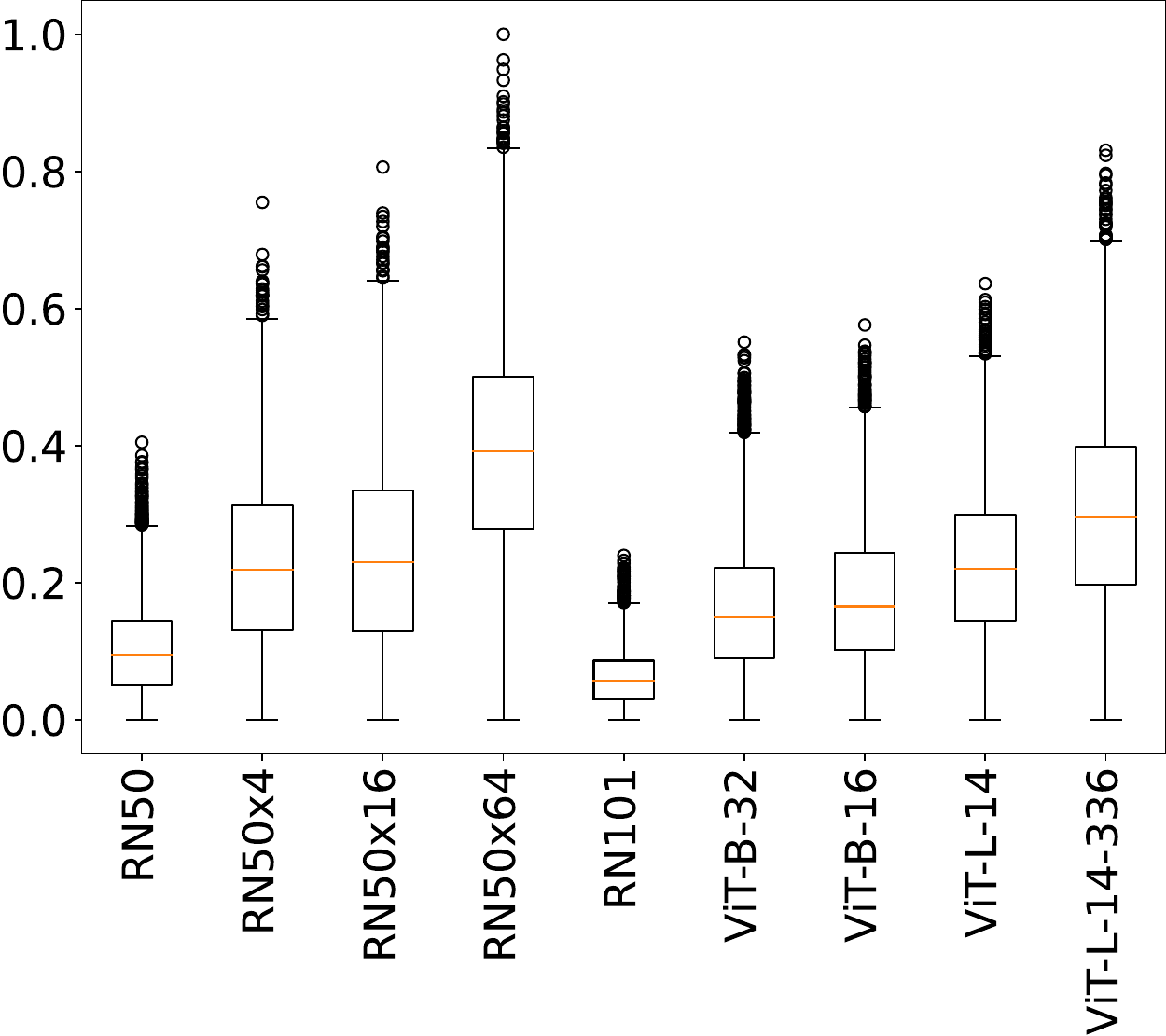}
    \label{fig:alpha_box_plots_cub}
\end{subfigure}
\hfill
\begin{subfigure}{0.32\textwidth}
    \caption{\Dtd}
    \includegraphics[width=\textwidth]{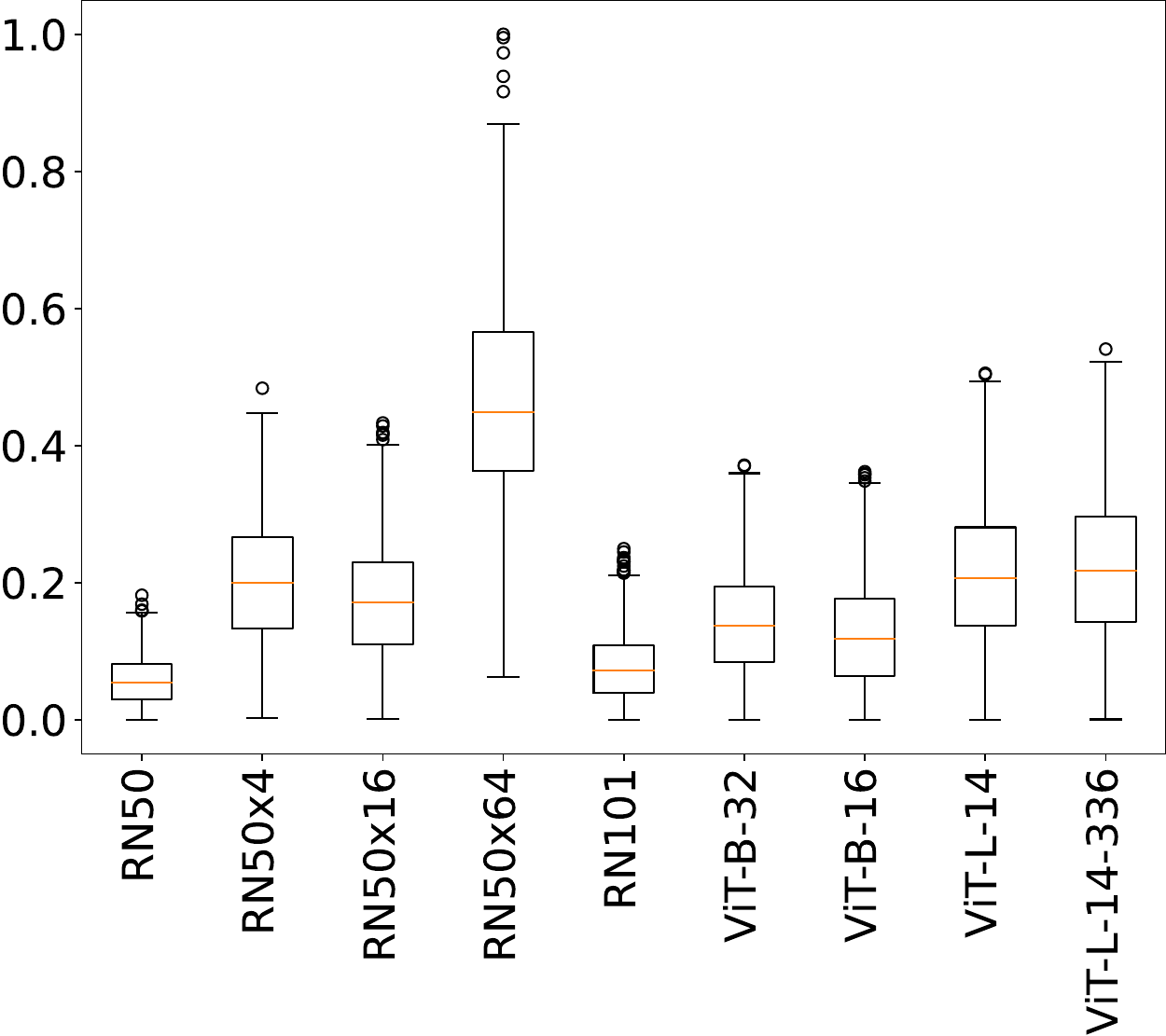}
    \label{fig:alpha_box_plots_dtd}
\end{subfigure}
\begin{subfigure}{0.32\textwidth}
    \caption{\Eurosat{}}
    \includegraphics[width=\textwidth]{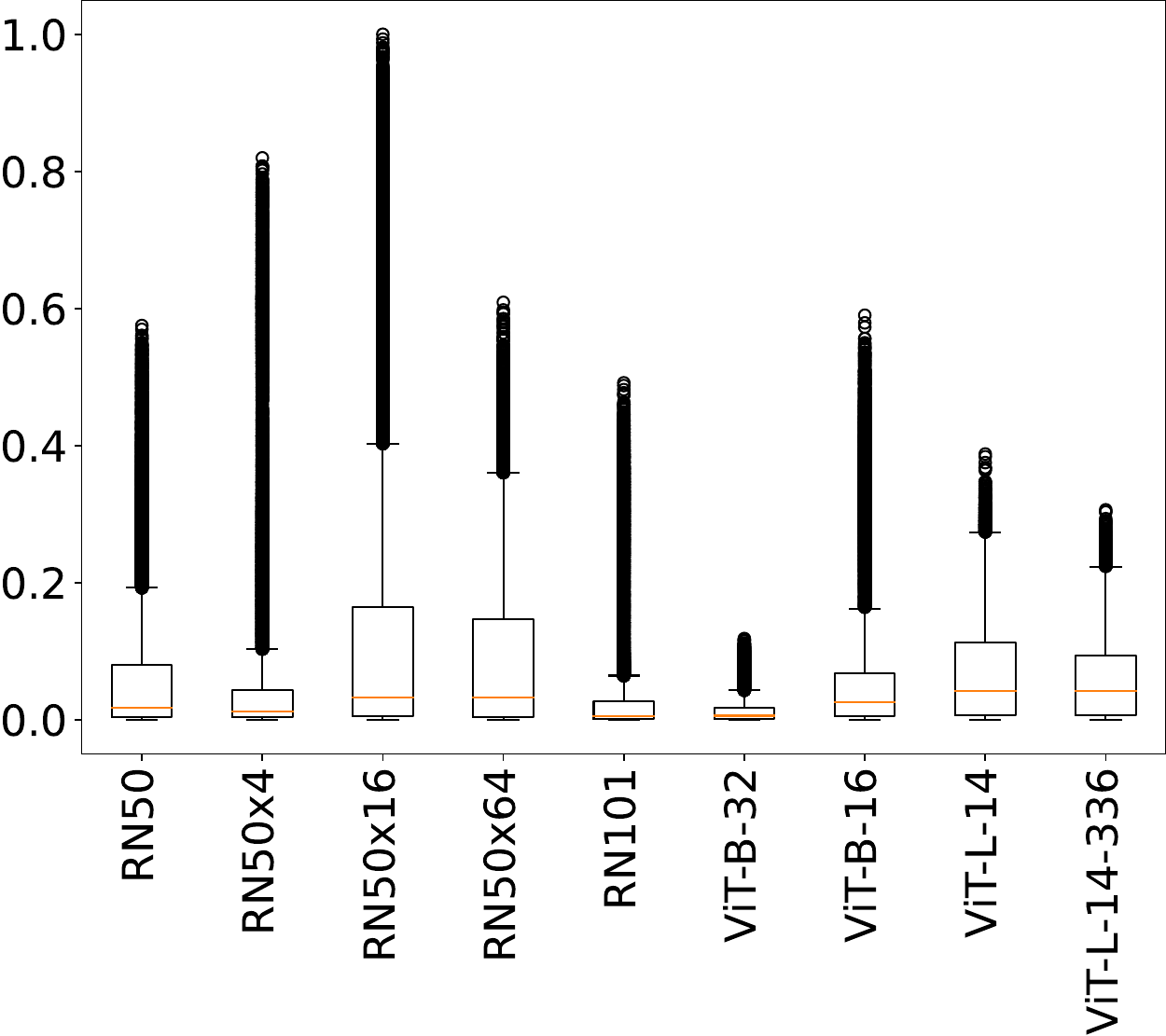}
    \label{fig:alpha_box_plots_eurosat}
\end{subfigure}
\hfill       
\begin{subfigure}{0.32\textwidth}
    \caption{\FGVC{}}
    \includegraphics[width=\textwidth]{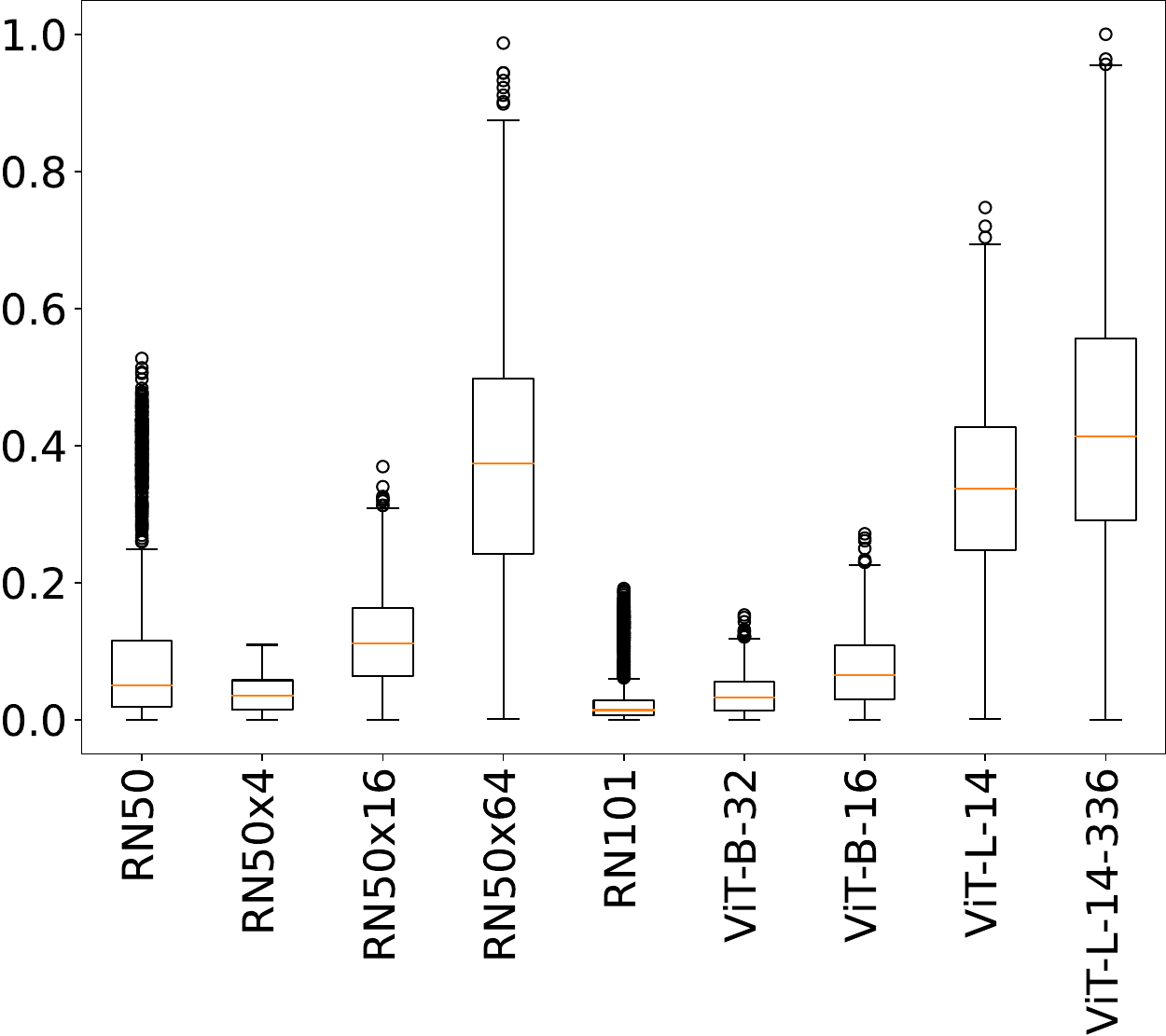}
    \label{fig:alpha_box_plots_fgvc}
\end{subfigure}
\begin{subfigure}{0.32\textwidth}
    \caption{\Flowers}
    \includegraphics[width=\textwidth]{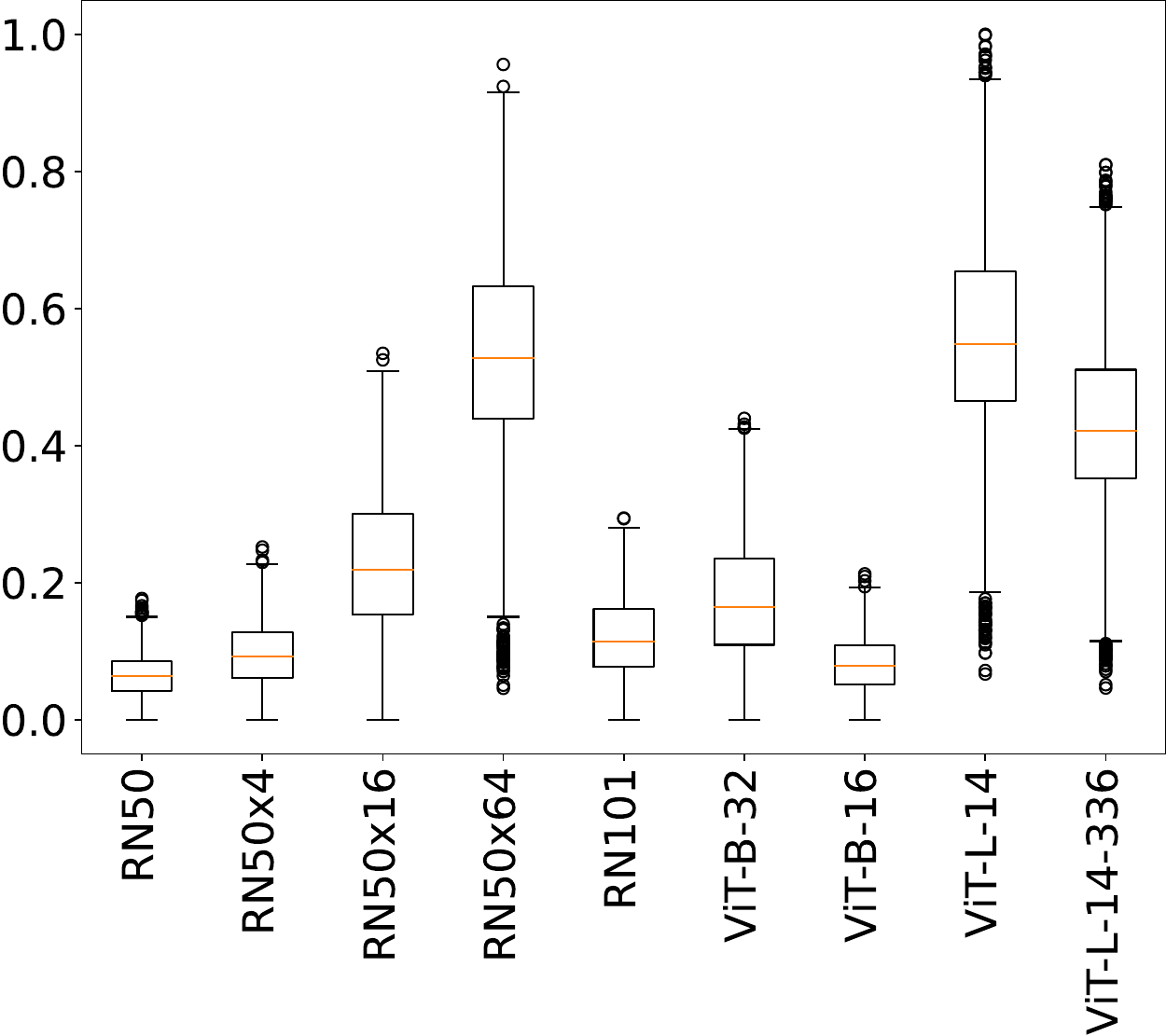}
    \label{fig:alpha_box_plots_flowers}
\end{subfigure}
\hfill
\begin{subfigure}{0.32\textwidth}
    \caption{\Food}
    \includegraphics[width=\textwidth]{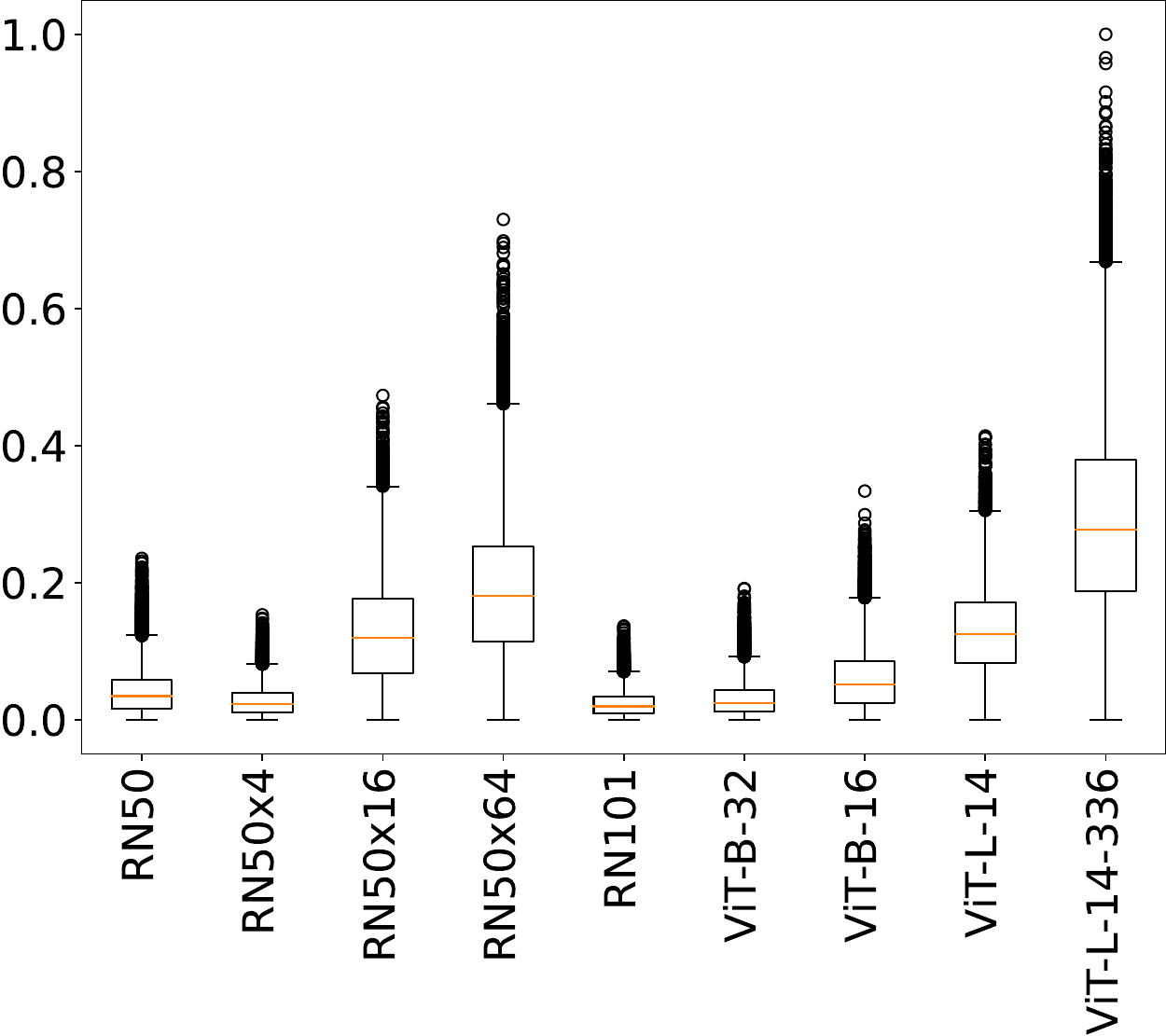}
    \label{fig:alpha_box_plots_food}
\end{subfigure}
\caption{Alpha values for each dataset using \NNC}
\label{fig:alpha1}
\end{figure*}

\begin{figure*}
    
\hfill
\begin{subfigure}{0.32\textwidth}
    \caption{\Gtsrb}
    \includegraphics[width=\textwidth]{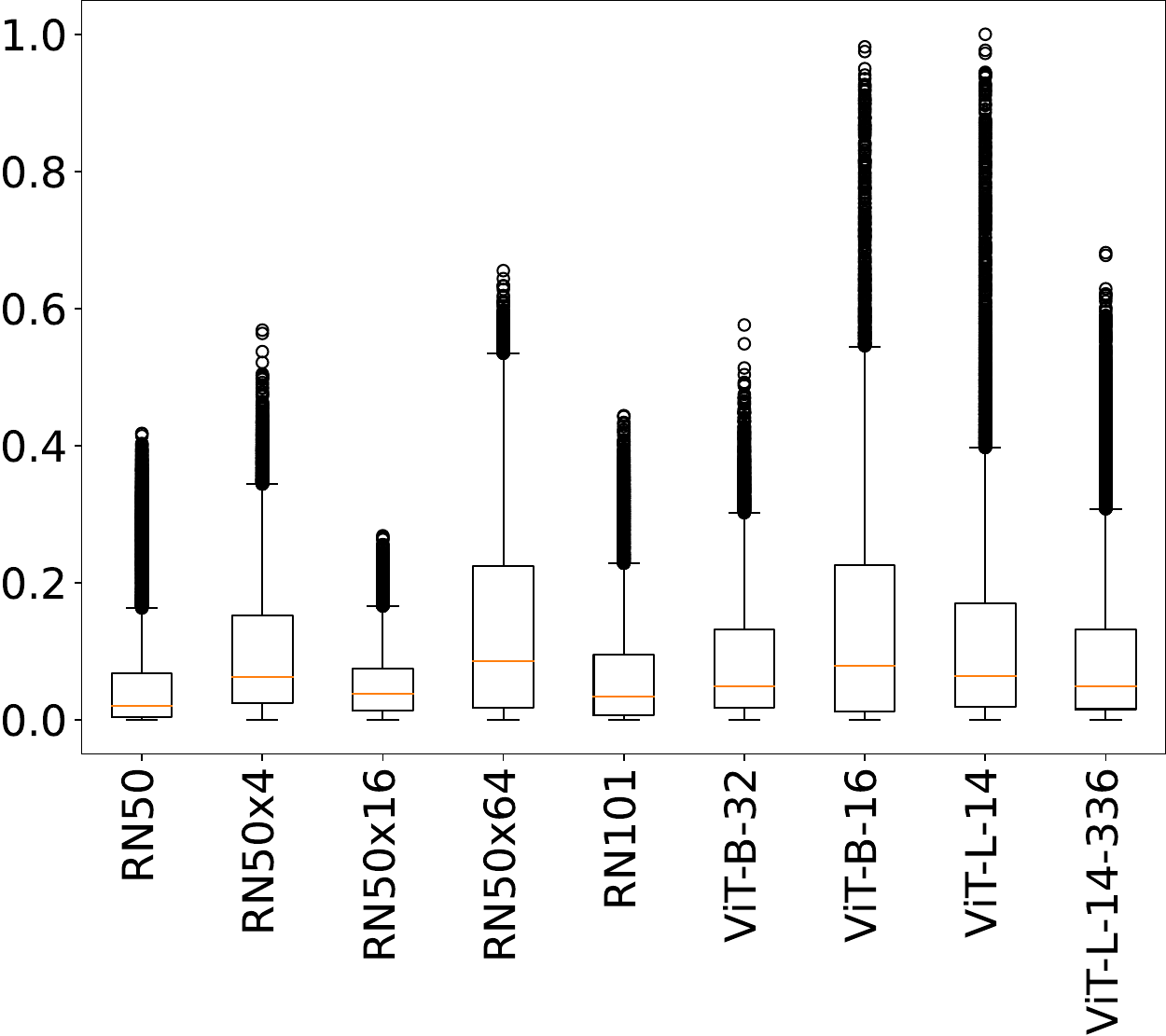}
    \label{fig:alpha_box_plots_gtsrb}
\end{subfigure}
\begin{subfigure}{0.32\textwidth}
    \caption{\Mnist}
    \includegraphics[width=\textwidth]{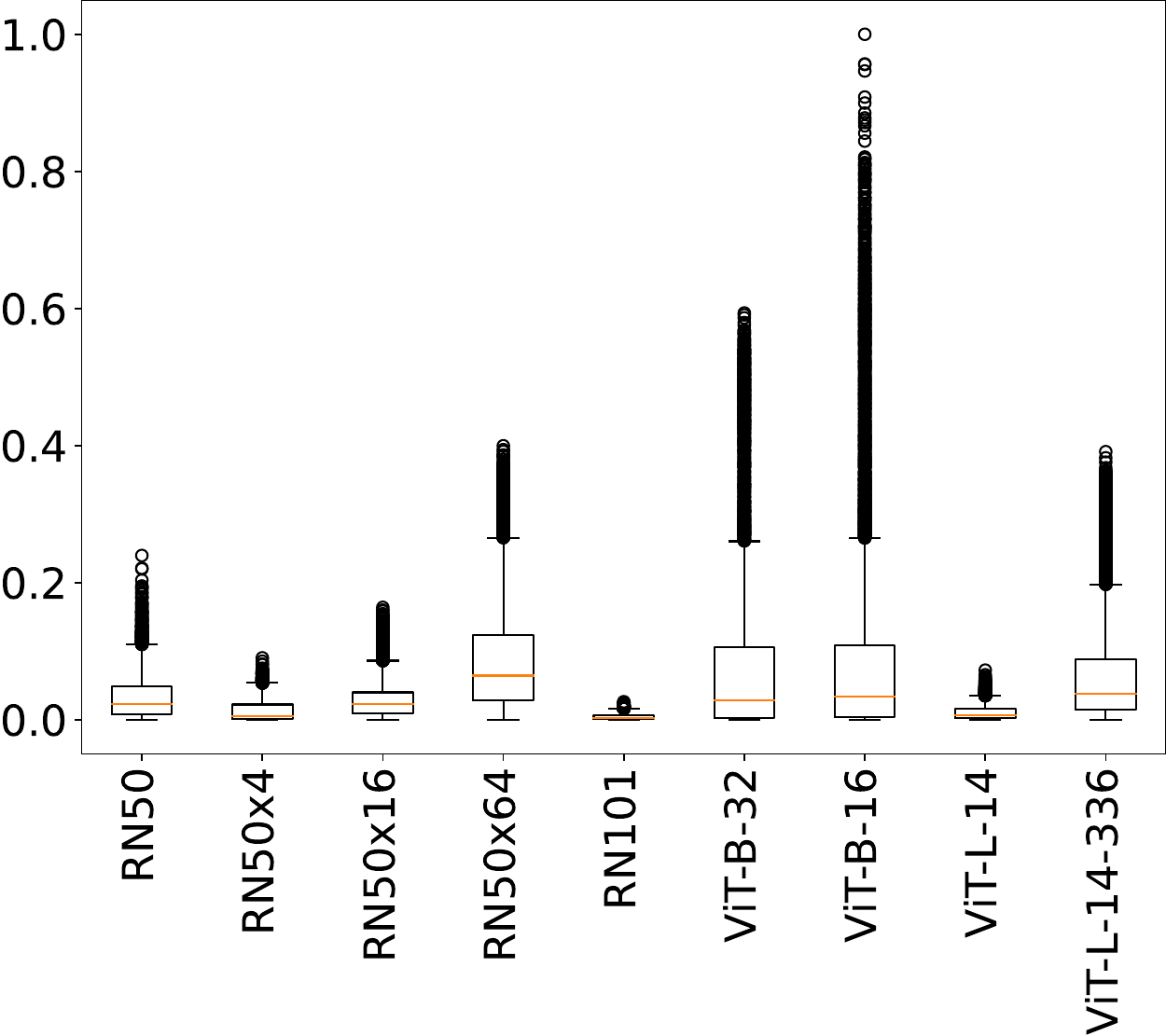}
    \label{fig:alpha_box_plots_mnist}
\end{subfigure}
\begin{subfigure}{0.32\textwidth}
    \caption{\Imagenet}
    \includegraphics[width=\textwidth]{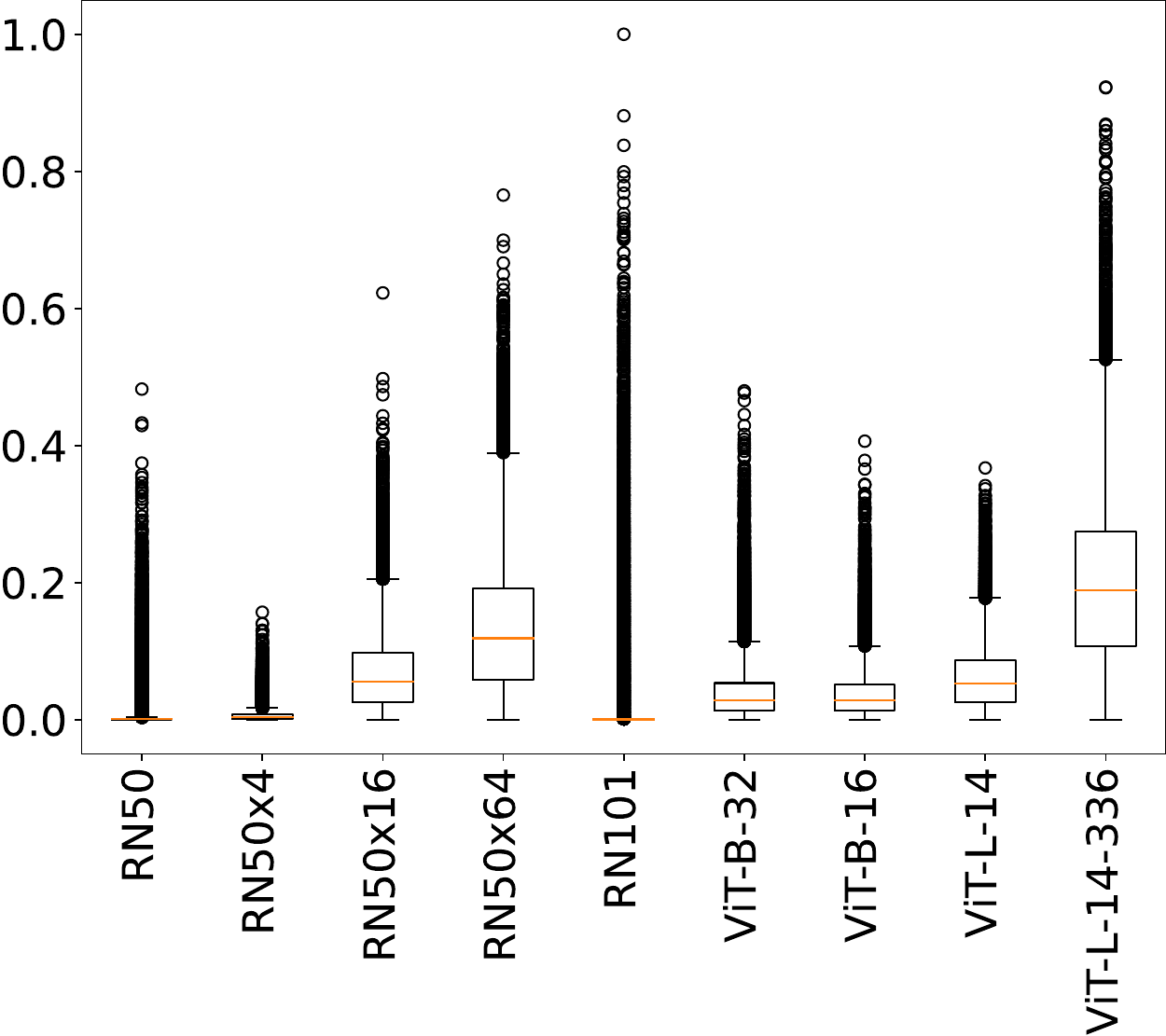}
    \label{fig:alpha_box_plots_inet}
\end{subfigure}
\begin{subfigure}{0.32\textwidth}
    \caption{\Pcam}
    \includegraphics[width=\textwidth]{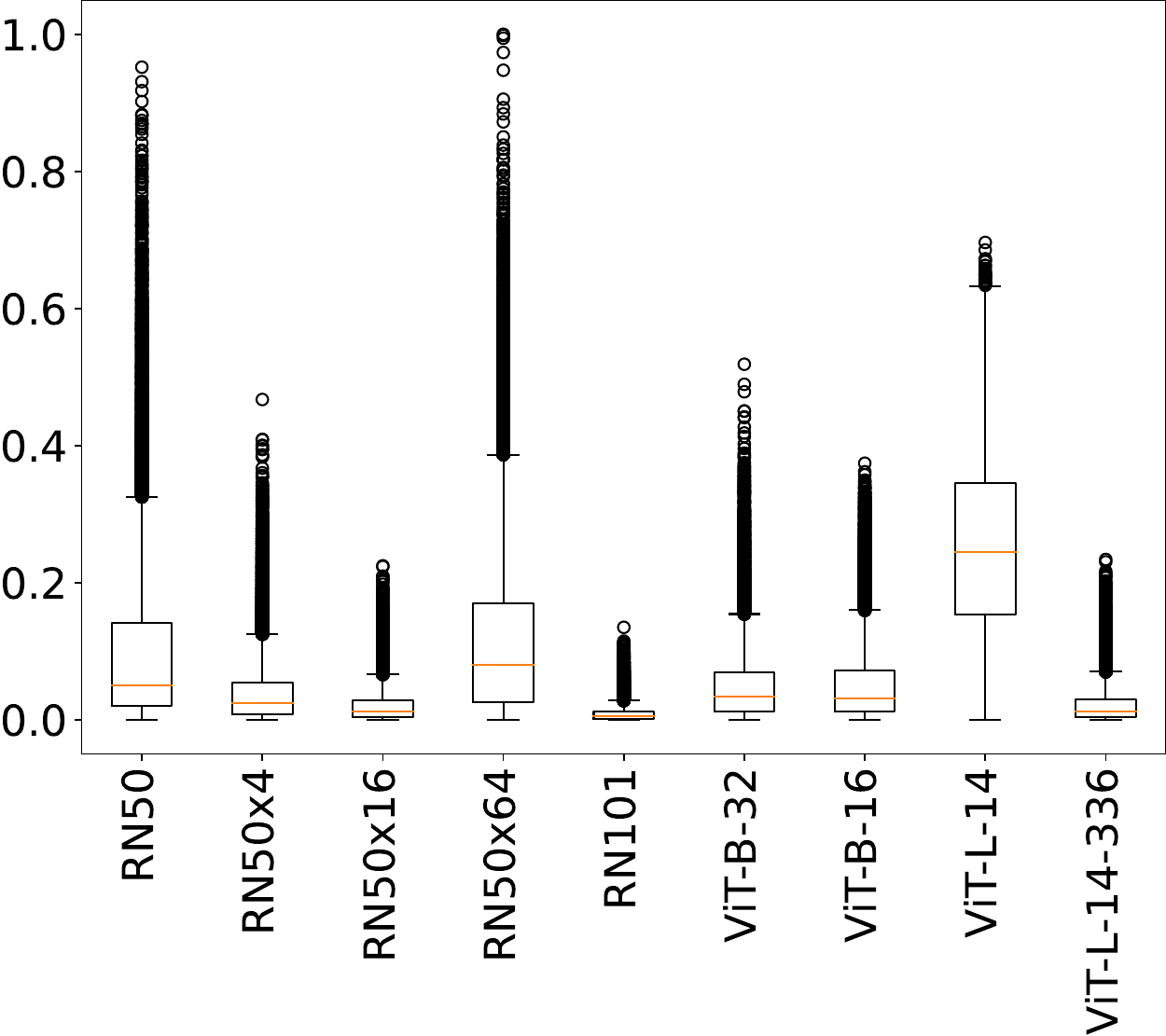}
    \label{fig:alpha_box_plots_pcam}
\end{subfigure}
\begin{subfigure}{0.32\textwidth}
    \caption{\Pets{}}
    \includegraphics[width=\textwidth]{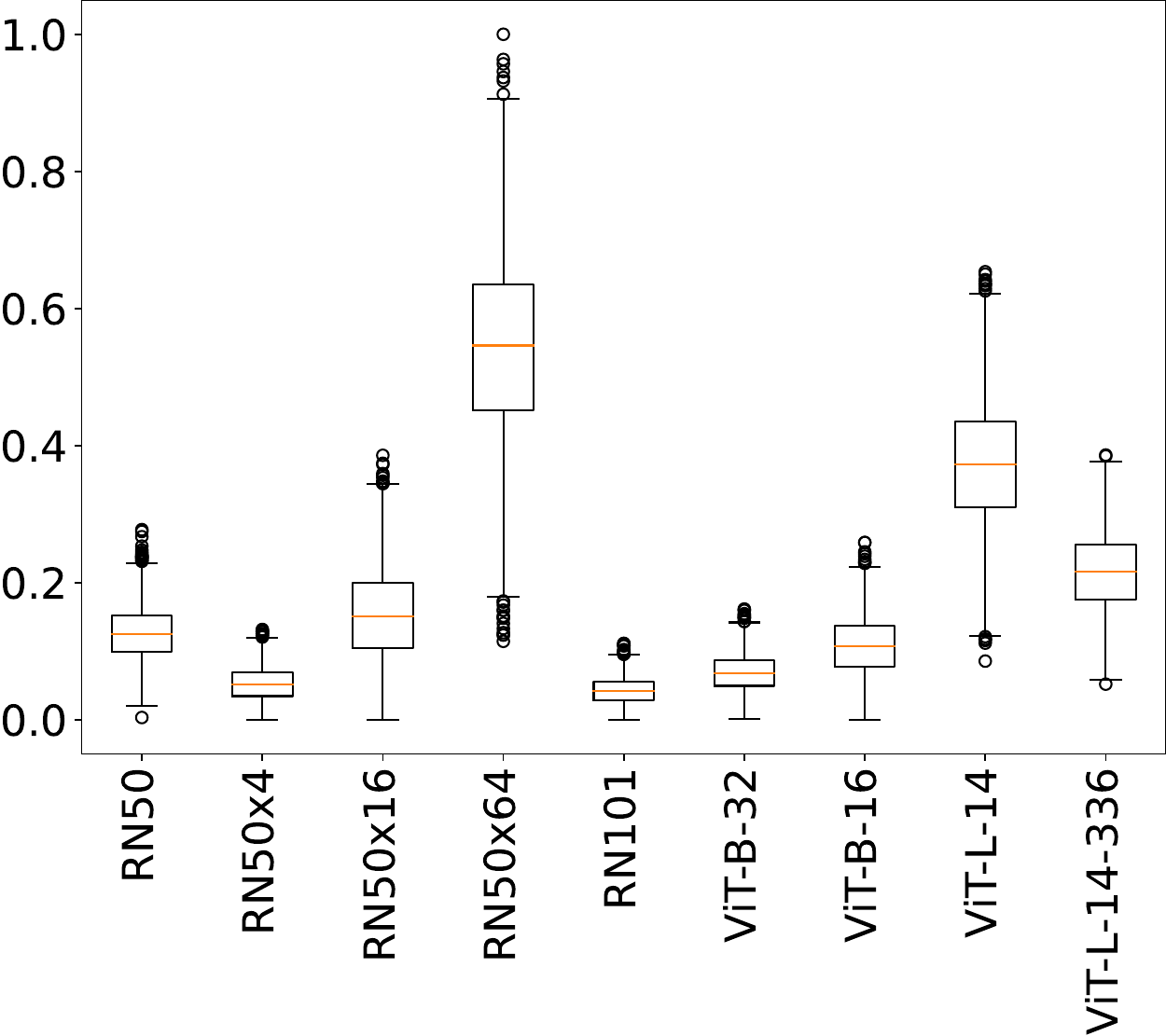}
    \label{fig:alpha_box_plots_pets}
\end{subfigure}
\begin{subfigure}{0.32\textwidth}
    \caption{\Renderedsst2}
    \includegraphics[width=\textwidth]{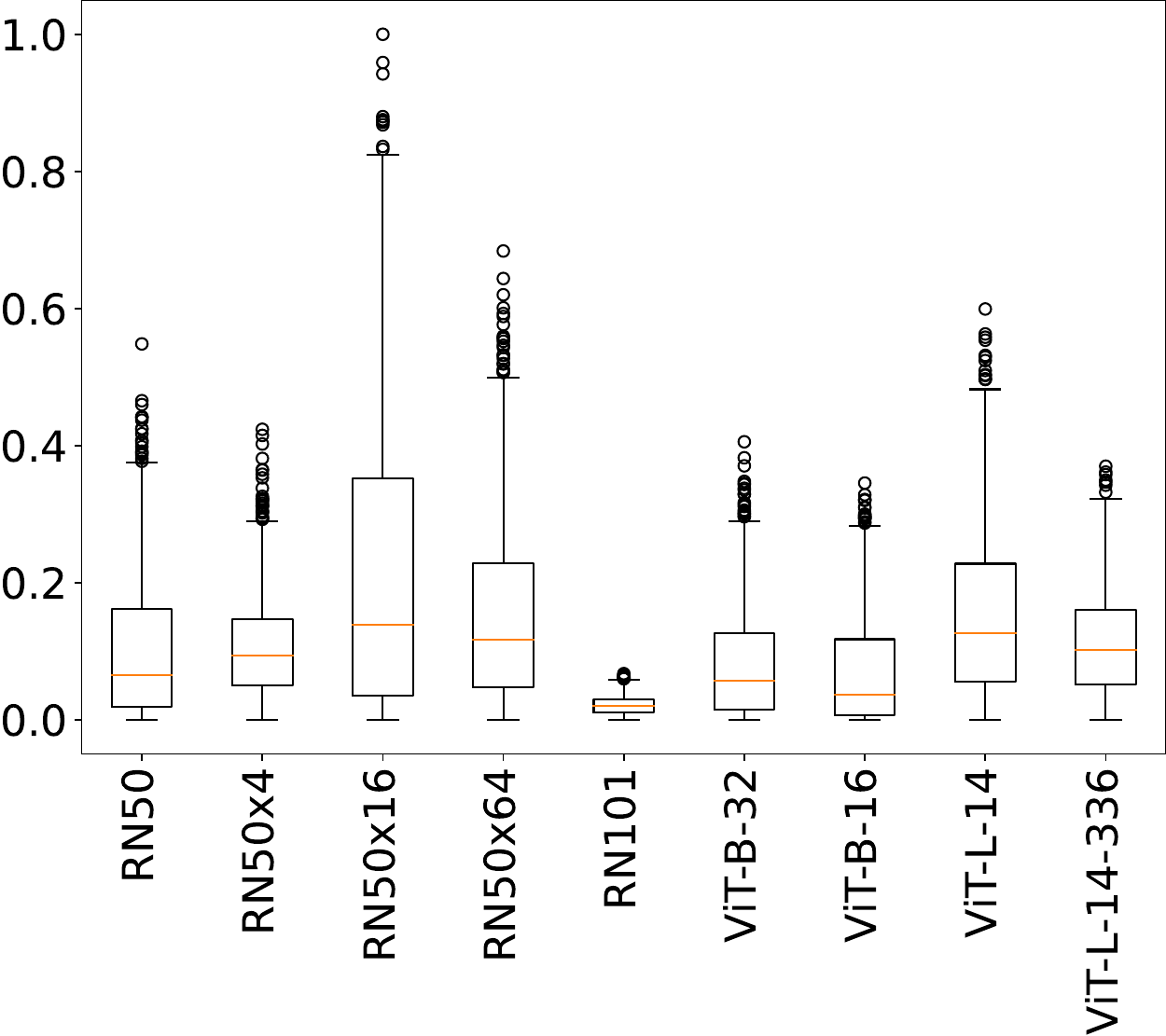}
    \label{fig:alpha_box_plots_renderedsst}
\end{subfigure}
\begin{subfigure}{0.32\textwidth}
    \caption{\Resisc45}
    \includegraphics[width=\textwidth]{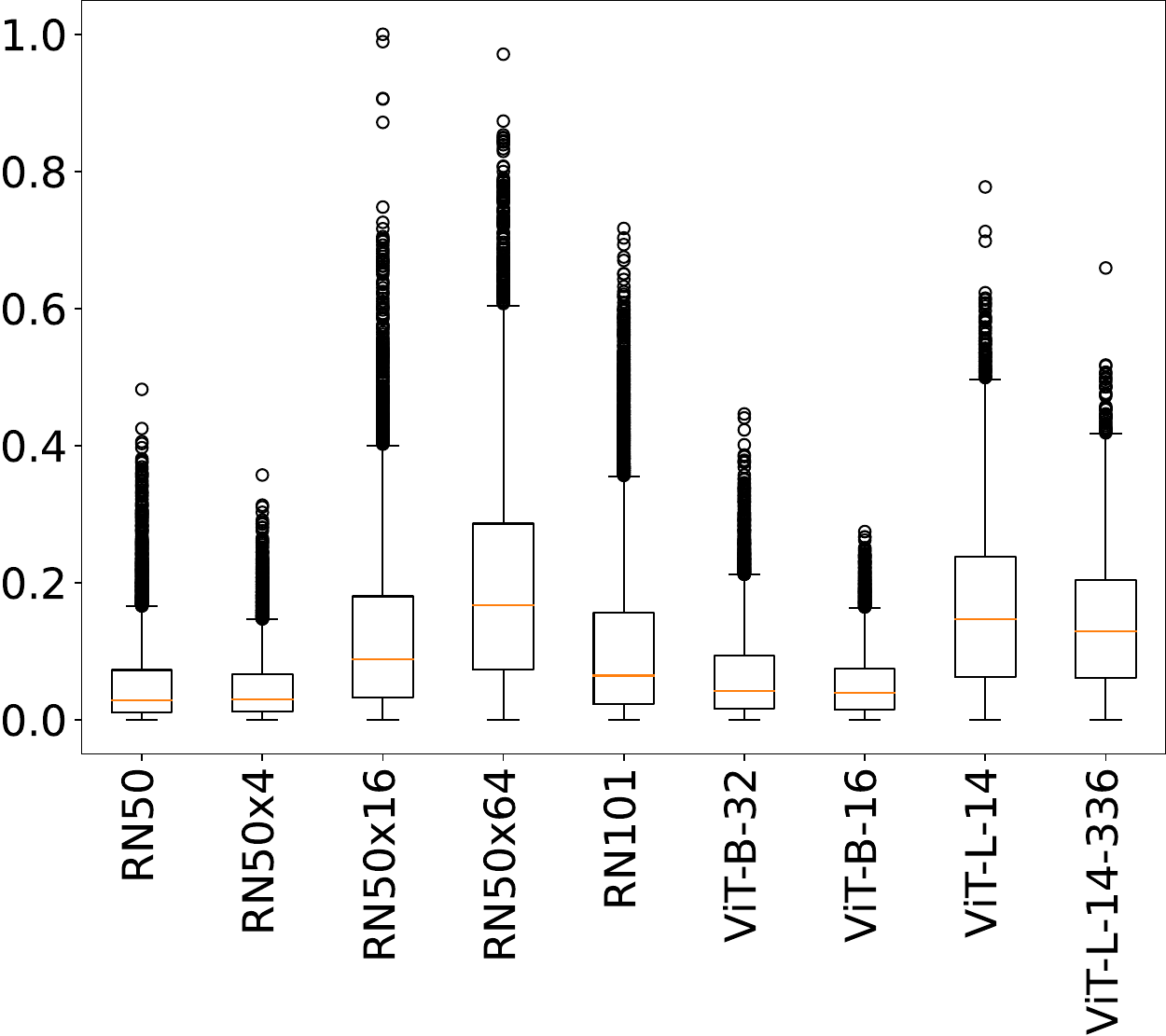}
    \label{fig:alpha_box_plots_resisc}
\end{subfigure}
\begin{subfigure}{0.32\textwidth}
    \caption{\STL10}
    \includegraphics[width=\textwidth]{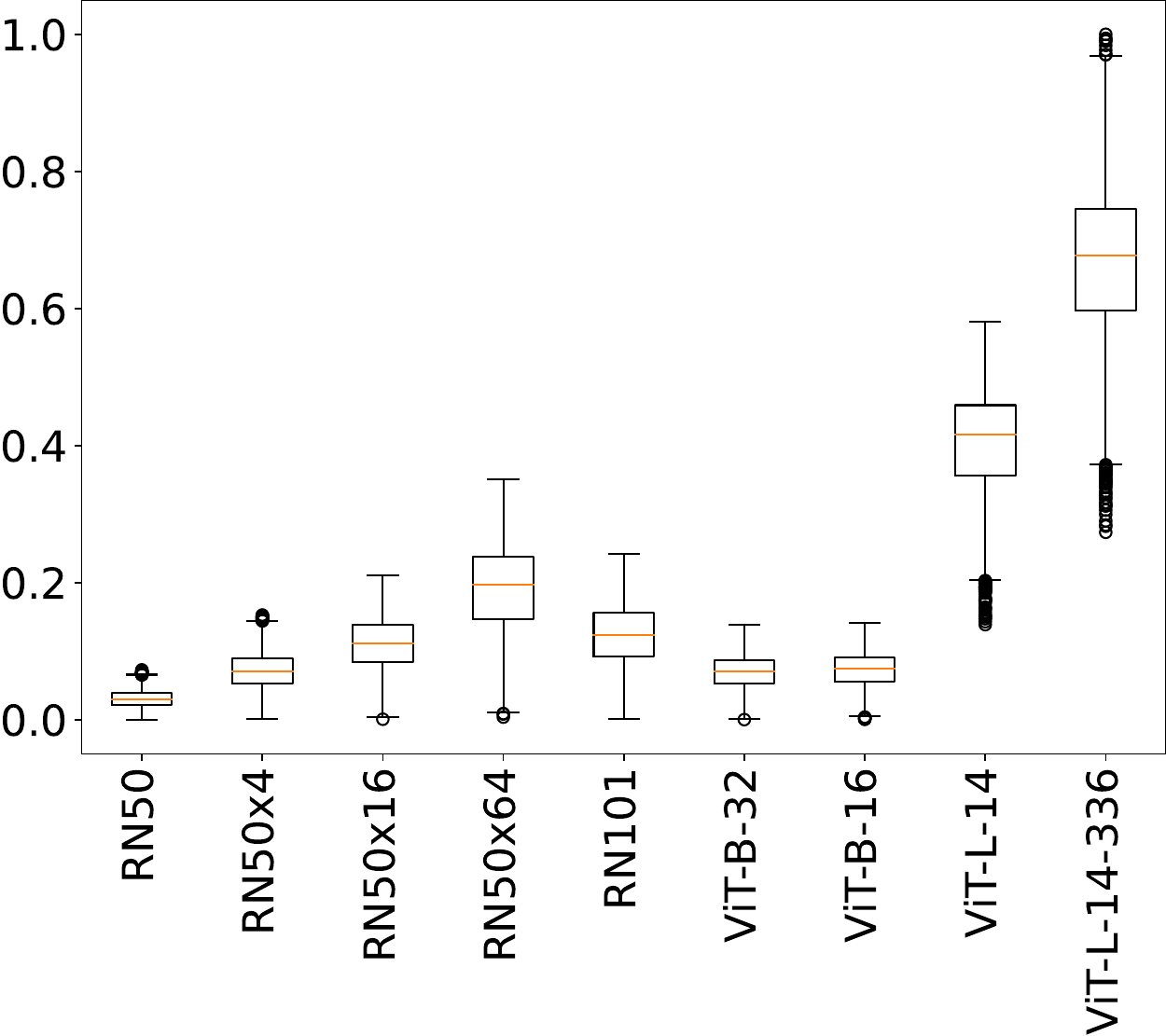}
    \label{fig:alpha_box_plots_stl}
\end{subfigure}
\begin{subfigure}{0.32\textwidth}
    \caption{\SUN397}
    \includegraphics[width=\textwidth]{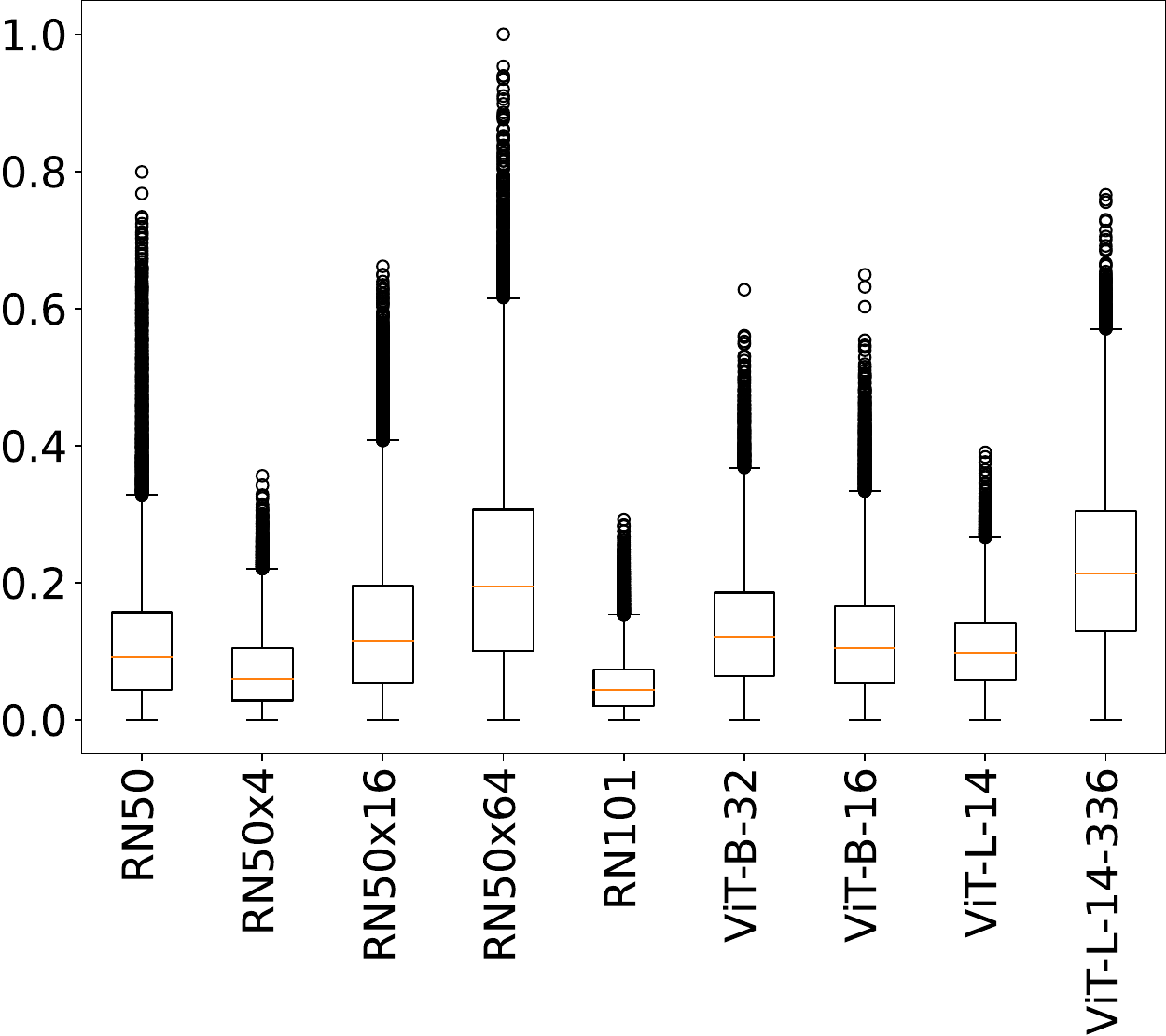}
    \label{fig:alpha_box_plots_sun}
\end{subfigure}
\caption{Alpha values for each dataset using \NNC}
\label{fig:alpha2}
\end{figure*}

\section{Diversity and Oracle Performance of Different Combinations}

Table \ref{tab:oracle_pretrained_dataset} shows the Oracle performance by combinaning different options of diversity. Table \ref{tab:diversity_pretrained_dataset} present the diversity of different source of complementarity.

\begin{table}[t]
\centering
\caption{\Oracle{} performance by combining: 1) ViT \textbf{backbones} pretrained with same datasets. 2) Same ViT backbone trained on different \textbf{datasets} 3) Same ViT backbone trained on same dataset but two different \textbf{epochs}.}
\resizebox{\textwidth}{!}{
\begin{tabular}{clcccccccccccccccccccccc}
               & & \rotatebox[origin=l]{90}{Average} & \rotatebox[origin=l]{90}{\Caltech101} & \rotatebox[origin=l]{90}{\Cars} & \rotatebox[origin=l]{90}{\Cifarten} & \rotatebox[origin=l]{90}{\Cifaronehundred} & \rotatebox[origin=l]{90}{\Clevr} & \rotatebox[origin=l]{90}{\Country211} & \rotatebox[origin=l]{90}{\Cub} & \rotatebox[origin=l]{90}{\Dtd} & \rotatebox[origin=l]{90}{\Eurosat} & \rotatebox[origin=l]{90}{\FGVC} & \rotatebox[origin=l]{90}{\Flowers} & \rotatebox[origin=l]{90}{\Food} & \rotatebox[origin=l]{90}{\Gtsrb} & \rotatebox[origin=l]{90}{\Imagenet} & \rotatebox[origin=l]{90}{\Mnist} & \rotatebox[origin=l]{90}{\Pcam} & \rotatebox[origin=l]{90}{\Pets} & \rotatebox[origin=l]{90}{\Renderedsst2} & \rotatebox[origin=l]{90}{\Resisc45} & \rotatebox[origin=l]{90}{\STL10} & \rotatebox[origin=l]{90}{\SUN397} \\ \addlinespace[-0.2cm]
\cmidrule(lr){2-2} \cmidrule(lr){3-3}   \cmidrule(lr){4-23}
               
\multirow{3}{*}{\rotatebox[origin=l]{90}{Backbone}} & OpenAI         & 78.2    & 91.6       & 87.4 & 98.2    & 85.9     & 33.6  & 39.1       & 76.6 & 65.3 & 70.0    & 46.6           & 85.0    & 95.8 & 64.9  & 82.9       & 83.0  & 93.8 & 97.0 & 90.6         & 76.5     & 99.6  & 79.1   \\
& LAION-400M\_e31 & 76.9    & 91.7       & 94.9 & 97.4    & 85.9     & 39.9  & 30.0       & 83.3 & 70.3 & 68.0    & 36.2           & 81.6    & 93.3 & 59.7  & 79.9       & 77.1  & 79.8 & 96.1 & 90.6         & 77.5     & 98.8  & 82.1   \\
& LAION-400M\_e32 & 76.8    & 91.6       & 94.9 & 97.4    & 85.9     & 39.6  & 30.1       & 83.3 & 70.5 & 67.5    & 36.8           & 81.5    & 93.4 & 59.8  & 80.0       & 76.6  & 80.2 & 96.1 & 90.3         & 77.3     & 98.9  & 82.1  \\
\cmidrule(lr){2-2} \cmidrule(lr){3-3}   \cmidrule(lr){4-23}

\multirow{3}{*}{\rotatebox[origin=l]{90}{Dataset}} & ViT-B-16       & 73.7    & 90.8       & 88.9 & 95.9    & 80.0     & 45.7  & 28.8       & 75.1 & 60.4 & 57.2    & 33.3           & 79.2    & 92.8 & 55.8  & 77.6       & 76.7  & 80.8 & 94.7 & 84.3         & 72.9     & 98.9  & 78.2   \\
& ViT-B-32       & 69.1    & 88.1       & 82.9 & 94.7    & 77.4     & 26.9  & 22.4       & 68.6 & 62.3 & 60.7    & 27.4           & 76.1    & 87.9 & 51.9  & 72.9       & 60.5  & 74.9 & 93.4 & 83.3         & 65.5     & 98.0  & 74.7   \\
& ViT-L-14       & 77.4    & 90.7       & 93.6 & 98.1    & 86.6     & 31.4  & 38.1       & 81.8 & 69.7 & 62.1    & 44.0           & 85.1    & 95.5 & 62.8  & 82.8       & 83.8  & 79.4 & 96.8 & 84.3         & 77.4     & 99.6  & 80.8   \\
\cmidrule(lr){2-2} \cmidrule(lr){3-3}   \cmidrule(lr){4-23}

\multirow{3}{*}{\rotatebox[origin=l]{90}{Epoch}} & ViT-B-16 & 62.9 & 85.7 & 84.2 & 92.0 & 72.1 & 29.2 & 18.7 & 65.8 & 51.8 & 45.4 & 18.3 & 70.0 & 85.1 & 44.4 & 67.8 & 58.0 & 59.8 & 89.7 & 55.6 & 60.4 & 97.1 & 70.2 \\
& ViT-B-32 & 57.8 & 83.5 & 75.3 & 88.8 & 68.6 & 15.6 & 13.9 & 57.5 & 53.1 & 45.1 & 15.5 & 66.8 & 76.4 & 41.4 & 60.9 & 45.3 & 56.2 & 85.9 & 52.7 & 52.3 & 95.0 & 65.0 \\
& ViT-L-14 & 67.2 & 88.1 & 90.0 & 94.9 & 78.1 & 24.6 & 23.6 & 74.0 & 60.5 & 50.9 & 25.9 & 76.0 & 89.6 & 50.5 & 73.3 & 71.4 & 50.7 & 92.2 & 57.5 & 67.6 & 98.1 & 73.1 \\
\cmidrule(lr){2-2} \cmidrule(lr){3-3}   \cmidrule(lr){4-23}
\end{tabular}
}
\label{tab:oracle_pretrained_dataset}
\centering
\caption{Diversity of combining: 1) ViT \textbf{backbones} pretrained with same datasets. 2) Same ViT backbone trained on different \textbf{datasets} 3) Same ViT backbone trained on same dataset but two different \textbf{epochs}.}
\resizebox{\textwidth}{!}{
\begin{tabular}{clcccccccccccccccccccccc}
 & & \rotatebox[origin=l]{90}{Average} & \rotatebox[origin=l]{90}{\Caltech101} & \rotatebox[origin=l]{90}{\Cars} & \rotatebox[origin=l]{90}{\Cifarten} & \rotatebox[origin=l]{90}{\Cifaronehundred} & \rotatebox[origin=l]{90}{\Clevr} & \rotatebox[origin=l]{90}{\Country211} & \rotatebox[origin=l]{90}{\Cub} & \rotatebox[origin=l]{90}{\Dtd} & \rotatebox[origin=l]{90}{\Eurosat} & \rotatebox[origin=l]{90}{\FGVC} & \rotatebox[origin=l]{90}{\Flowers} & \rotatebox[origin=l]{90}{\Food} & \rotatebox[origin=l]{90}{\Gtsrb} & \rotatebox[origin=l]{90}{\Imagenet} & \rotatebox[origin=l]{90}{\Mnist} & \rotatebox[origin=l]{90}{\Pcam} & \rotatebox[origin=l]{90}{\Pets} & \rotatebox[origin=l]{90}{\Renderedsst2} & \rotatebox[origin=l]{90}{\Resisc45} & \rotatebox[origin=l]{90}{\STL10} & \rotatebox[origin=l]{90}{\SUN397} \\ 
\addlinespace[-0.2cm]
\cmidrule(lr){2-2} \cmidrule(lr){3-3}   \cmidrule(lr){4-24}
\multirow{3}{*}{\rotatebox[origin=l]{90}{Backbone}} & OpenAI         & 47.2    & 13.3       & 47.5 & 14.3    & 40.8     & 70.4  & 71.8       & 51.3 & 50.8 & 76.7    & 81.2           & 32.4    & 19.3 & 68.1  & 34.8       & 66.7  & 80.1 & 15.9 & 65.8         & 46.5     & 3.2   & 39.7   \\
& LAION400M\_e31 & 43.7    & 13.6       & 29.6 & 13.7    & 33.6     & 69.0  & 72.7       & 46.0 & 45.3 & 62.7    & 83.0           & 29.6    & 24.5 & 50.8  & 34.9       & 53.6  & 68.3 & 16.7 & 82.5         & 47.7     & 4.9   & 34.6   \\
& LAION400M\_e32 & 43.6    & 13.6       & 29.3 & 13.6    & 33.5     & 68.6  & 72.9       & 45.7 & 45.7 & 62.1    & 83.1           & 29.9    & 24.4 & 50.9  & 35.0       & 53.6  & 67.9 & 17.1 & 81.8         & 47.5     & 4.9   & 34.5  \\
\cmidrule(lr){2-2} \cmidrule(lr){3-3}   \cmidrule(lr){4-24}
\\
\multirow{3}{*}{\rotatebox[origin=l]{90}{Dataset}}  & ViT-B-16       & 37.7    & 10.6       & 33.2 & 9.7     & 27.3     & 90.0  & 57.5       & 39.7 & 40.6 & 45.3    & 73.4           & 22.5    & 14.5 & 43.7  & 25.8       & 47.9  & 64.0 & 11.8 & 64.1         & 38.5     & 2.7   & 29.6   \\
& ViT-B-32       & 39.8    & 10.3       & 37.8 & 12.0    & 29.4     & 56.5  & 62.4       & 40.3 & 45.1 & 66.3    & 73.7           & 26.0    & 20.3 & 58.0  & 30.8       & 66.0  & 44.0 & 15.1 & 67.2         & 38.7     & 4.1   & 31.9   \\
& ViT-L-14       & 32.0    & 8.2        & 21.3 & 6.2     & 23.3     & 60.8  & 56.1       & 34.5 & 34.8 & 43.4    & 71.7           & 18.6    & 10.2 & 39.9  & 21.2       & 28.9  & 72.4 & 8.8  & 52.5         & 30.1     & 1.8   & 27.7   \\
\cmidrule(lr){2-2} \cmidrule(lr){3-3} \cmidrule(lr){4-24}
\multirow{3}{*}{\rotatebox[origin=l]{90}{Epoch}} & ViT-B-16 & 2.7 & 0.8 & 1.4 & 0.6 & 2.4 & 3.2 & 6.2 & 2.9 & 2.5 & 5.4 & 7.7  & 1.9 & 1.4 & 4.1 & 2.1 & 2.9 & 1.3 & 0.9 & 4.1 & 3.5 & 0.3 & 1.9 \\
& ViT-B-32 & 3.3 & 0.7 & 1.9 & 1.0 & 2.4 & 2.1 & 5.7 & 3.8 & 3.3 & 6.0 & 11.2 & 2.3 & 1.7 & 3.0 & 2.3 & 6.9 & 4.7 & 1.2 & 3.2 & 2.6 & 0.2 & 2.1 \\
& ViT-L-14 & 2.6 & 0.7 & 1.0 & 0.6 & 1.7 & 3.4 & 5.8 & 2.3 & 2.1 & 3.4 & 8.8  & 1.3 & 0.9 & 3.1 & 1.7 & 2.9 & 5.7 & 0.9 & 4.5 & 2.4 & 0.2 & 1.8
\\
\cmidrule(lr){2-2} \cmidrule(lr){3-3}   \cmidrule(lr){4-24}
\end{tabular}
}
\label{tab:diversity_pretrained_dataset}
\end{table}

\section{Cascading and Ensemble}
\begin{figure}[t!]
    \centering
    \includegraphics[width=\linewidth]{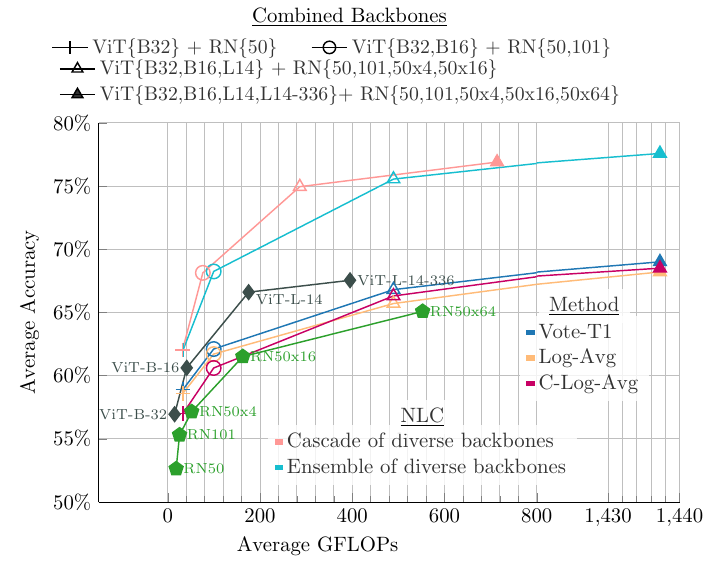}
    \caption{Average accuracy across 21 datasets for zero-shot ResNets and ViTs backbones, \LogitAvg{} Ensemble, \VoteOne{} Ensemble, \CLogitAvg{} Ensemble, \NNC{} Ensemble, and \NNC{} Cascade using 2 to 9 backbones, plotted against the Average GFLOPs. Demonstrates that our \NNC{} ensembles can surpass the best zero-shot backbone, and the \NNC{} cascade can also surpass the best zero-shot backbone with fewer GFLOPs. Moreover, our method surpasses the standard ensembling techniques.}
    \label{app:enter-label}
\end{figure}

Figure \ref{app:enter-label} shows the performance of multiple ensemble and cascading methods of 2 to 9 backbones, notice that \NNC{} obtains the best performance, and when we add cascade it maintains the computational requirements

\section{Image-Text Retrieval}

We present additional image and text retrieval experiments in Table \ref{tab:imgtextret} that hint at the potential benefits of combining multiple backbones of CLIP. 
We evaluated the complementarity of CLIP backbones on Flickr30k and MSCOCO. 
We measured the upper-bound improvement of an "oracle selection" of backbones as described in Section \ref{sec:prelim}. Table \ref{tab:imgtextret} shows patterns of complementarity across backbones comparable to those in the classification setting. On Flick30k, the upper bound shows possible improvements of 20 percentage points on both text and image retrieval. On MSCOCO, >25pp.

\label{app:imagetext_retrieval}
\begin{table}[hb!]
\caption{Zero-shot retrieval performance of CLIP backbones on the Flickr30k and MSCOCO benchmark and the performance of our empirical upper-bound, denoted as \Oracle{}, representing the ideal combination of ResNet (RN), ViT, and All backbones}
\label{tab:imgtextret}
\resizebox{\textwidth}{!}{
\begin{tabular}{clcccccccccccc}
\multicolumn{1}{l}{} &  & \multicolumn{6}{c}{Text Retrieval}                                                              & \multicolumn{6}{c}{Image Retrieval}                                                           \\
\multicolumn{1}{l}{} &  & \multicolumn{3}{c}{Flickr30k}                   & \multicolumn{3}{c}{MSCOCO}                    & \multicolumn{3}{c}{Flickr30k}                 & \multicolumn{3}{c}{MSCOCO}                    \\ \cmidrule(lr){3-5} \cmidrule(lr){6-8}  \cmidrule(lr){9-11} \cmidrule(lr){12-14} 
\multicolumn{1}{l}{} &  & R@1           & R@5            & R@10           & R@1           & R@5           & R@10          & R@1           & R@5           & R@10          & R@1           & R@5           & R@10          \\ \cmidrule(lr){1-2} \cmidrule(lr){3-5} \cmidrule(lr){6-8}  \cmidrule(lr){9-11} \cmidrule(lr){12-14} 
                                             & 50                       & 80.4          & 96.1           & 98.1           & 48.1          & 73.9          & 83.0          & 59.2          & 85.9          & 91.6          & 28.3          & 53.0          & 64.1          \\
                                             & 50x4                     & 84.8          & 96.6           & 98.9           & 52.1          & 76.6          & 84.5          & 64.6          & 87.3          & 92.7          & 32.5          & 56.7          & 67.1          \\
                                             & 50x16                    & 86.2          & 97.8           & 99.5           & 55.3          & 78.2          & 86.3          & 67.4          & 88.9          & 93.6          & 35.2          & 59.5          & 69.6          \\
                                             & 50x64                    & 88.3          & 98.4           & 99.6           & 57.8          & 80.6          & 87.7          & 71.4          & 91.6          & 95.5          & 34.7          & 60.0          & 70.0          \\
\multirow{-5}{*}{ResNet}                     & 101                      & 83.8          & 96.5           & 98.3           & 49.8          & 74.4          & 82.7          & 62.5          & 85.7          & 91.3          & 30.2          & 54.2          & 65.3          \\
\cmidrule(lr){1-2} \cmidrule(lr){3-5} \cmidrule(lr){6-8}  \cmidrule(lr){9-11} \cmidrule(lr){12-14} 
                                             & B-32                     & 82.2          & 95.1           & 97.5           & 50.0          & 75.0          & 83.2          & 61.4          & 86.0          & 91.7          & 30.4          & 54.8          & 66.1          \\
                                             & B-16                     & 85.1          & 97.3           & 98.9           & 51.8          & 76.8          & 84.3          & 65.2          & 87.8          & 92.8          & 32.7          & 57.8          & 68.3          \\
                                             & L-14                     & 87.0          & 98.4           & 99.8           & 56.1          & 79.6          & 86.9          & 67.8          & 89.8          & 94.3          & 35.3          & 59.6          & 70.1          \\
\multirow{-4}{*}{ViT}                        & L-14-336                 & 88.1          & 98.2           & 99.6           & 57.5          & 80.4          & 87.6          & 71.5          & 91.8          & 95.5          & 36.1          & 60.7          & 70.8          \\ \cmidrule(lr){1-2} \cmidrule(lr){3-5} \cmidrule(lr){6-8}  \cmidrule(lr){9-11} \cmidrule(lr){12-14} 
\multicolumn{2}{c}{\Oracle{} ResNet}                                       & \textbf{97.9} & \textbf{99.9}  & \textbf{100.0} & \textbf{77.7} & \textbf{93.0} & \textbf{96.3} & \textbf{86.6} & \textbf{96.8} & \textbf{98.4} & \textbf{54.7} & \textbf{76.9} & \textbf{84.9} \\
\multicolumn{2}{c}{\Oracle{} ViT}                                          & \textbf{96.3} & \textbf{99.5}  & \textbf{100.0} & \textbf{73.0} & \textbf{90.4} & \textbf{94.6} & \textbf{84.3} & \textbf{96.1} & \textbf{97.9} & \textbf{50.9} & \textbf{74.1} & \textbf{82.5} \\
\multicolumn{2}{c}{\Oracle{} All}                                              & \textbf{98.8} & \textbf{100.0} & \textbf{100.0} & \textbf{83.3} & \textbf{95.3} & \textbf{97.5} & \textbf{90.3} & \textbf{97.8} & \textbf{98.8} & \textbf{61.9} & \textbf{82.1} & \textbf{88.9} \\
\bottomrule
\end{tabular}
}
\end{table}

\clearpage
\newpage
\section{Overlap Diagrams For Other Datasets.}
\begin{figure}[b]
    \centering
    \underline{\textsc{Caltech101}}    
    \includegraphics[width=\textwidth]{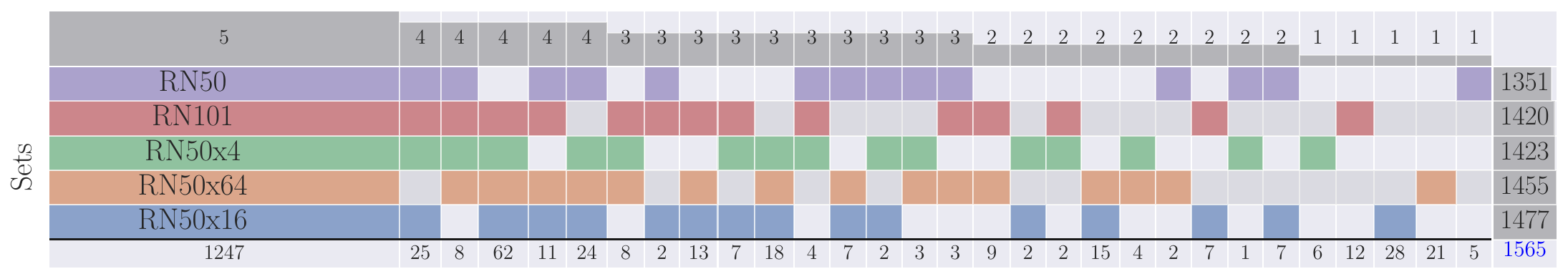}
    \includegraphics[width=\textwidth]{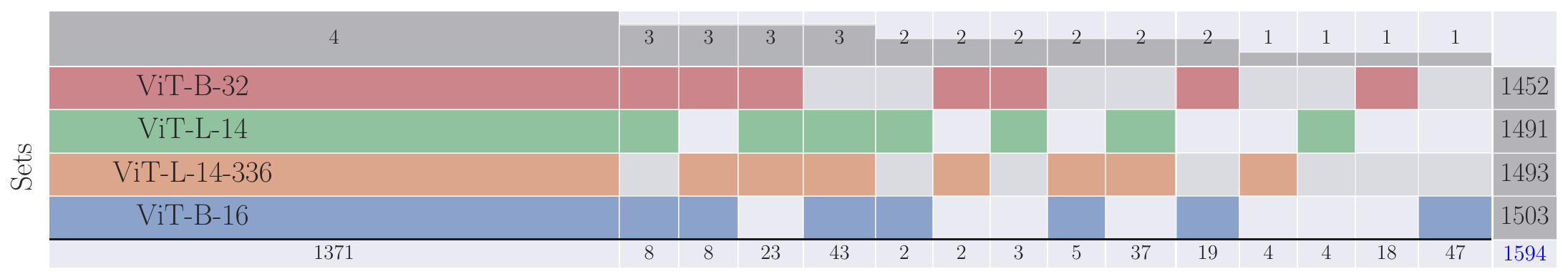}
    \includegraphics[width=\textwidth]{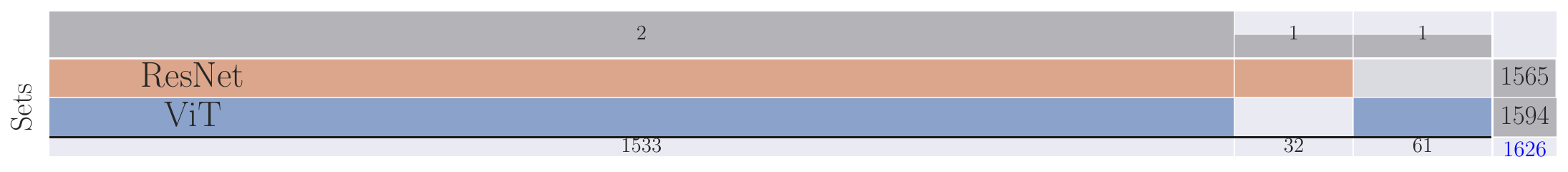}
    \caption{\textsc{Caltech101} Overlap diagrams with the correct prediction of each backbone. The Top part of the Overlap diagram shows the number of backbones that are predicting correctly a set of images. Each column represents a set of image instances that are predicted correctly by some group of backbones. Each row in the diagram shows in colour the backbone that correctly predicts a certain set of image instances, in grey when the backbone is not correctly predicting those instances. The bottom part of the Overlap diagram shows the number of images in a certain set. The right part is the total amount of correctly predicted images per backbone.}
    \label{fig:caltech101_venn_diagram}
\end{figure}

\begin{figure}[b]
    \centering
    \underline{\textsc{Cars}}    
    \includegraphics[width=\textwidth]{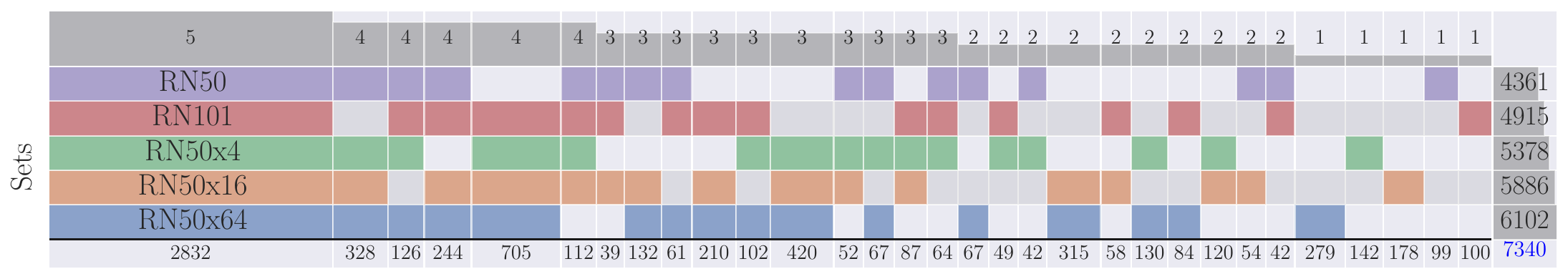}
    \includegraphics[width=\textwidth]{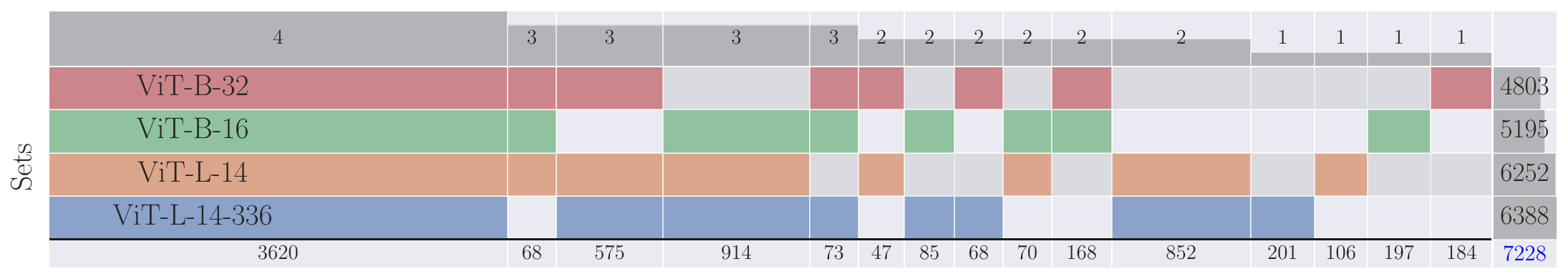}
    \includegraphics[width=\textwidth]{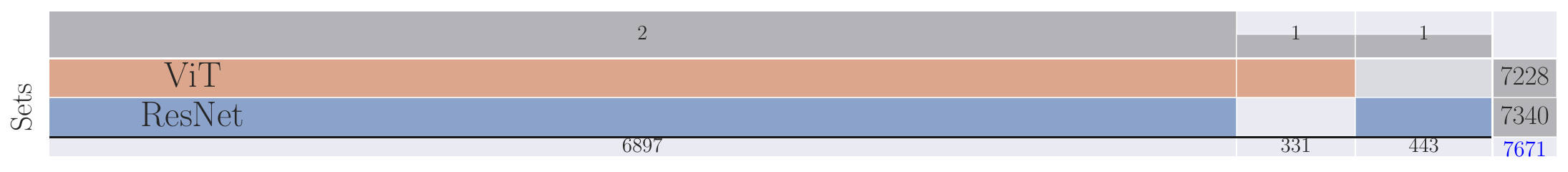}
    \caption{Overlap diagrams for the \Cars{} dataset with the correct prediction of each backbone. The Top part of the Overlap diagram shows the number of backbones that are predicting correctly a set of images. Each column represents a set of image instances that are predicted correctly by some group of backbones. Each row in the diagram shows in colour the backbone that correctly predicts a certain set of image instances, in grey when the backbone is not correctly predicting those instances. The bottom part of the Overlap diagram shows the number of images in a certain set. The right part is the total amount of correctly predicted images per backbone.}
    \label{fig:cars_diagram}
\end{figure}

\begin{figure}[bt]
    \centering
    \underline{\textsc{Cifar10}}    
    \includegraphics[width=\textwidth]{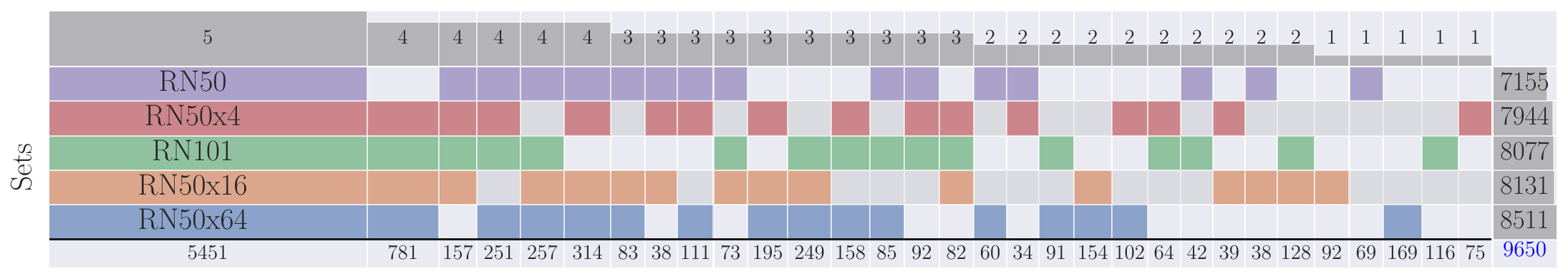}
    \includegraphics[width=\textwidth]{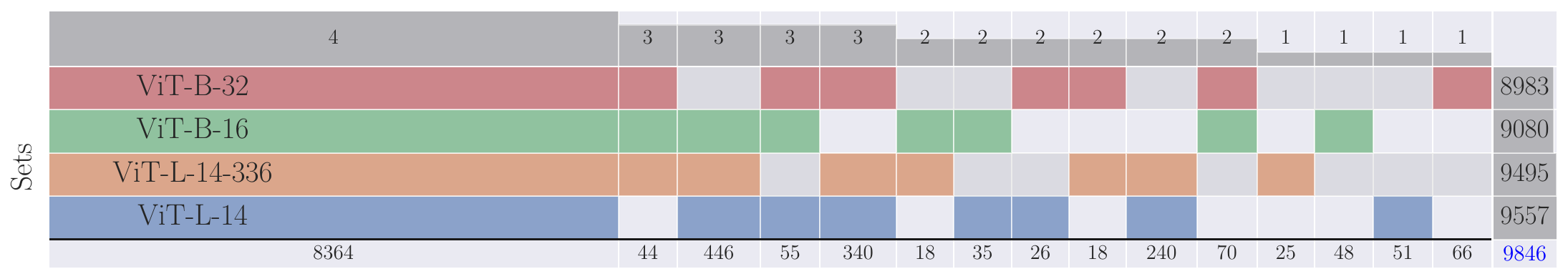}
    \includegraphics[width=\textwidth]{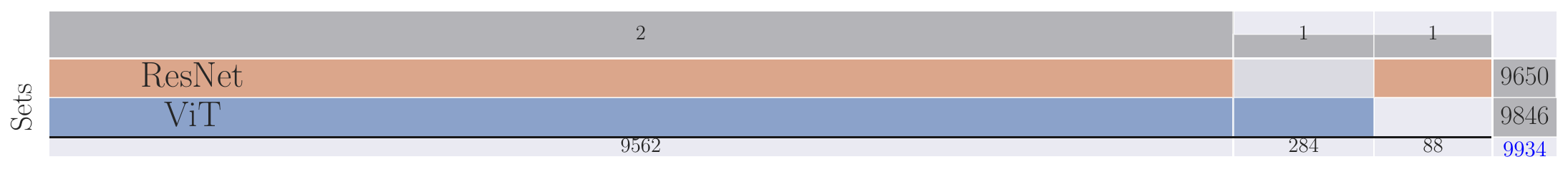}
    \caption{\textsc{Cifar10} Overlap diagrams with the correct prediction of each backbone. The Top part of the Overlap diagram shows the number of backbones that are predicting correctly a set of images. Each column represents a set of image instances that are predicted correctly by some group of backbones. Each row in the diagram shows in colour the backbone that correctly predicts a certain set of image instances, in grey when the backbone is not correctly predicting those instances. The bottom part of the Overlap diagram shows the number of images in a certain set. The right part is the total amount of correctly predicted images per backbone.}
    \label{fig:cifar10_diagram}
\end{figure}

\begin{figure}[bt]
    \centering
    \underline{\textsc{Cifar100}}    
    \includegraphics[width=\textwidth]{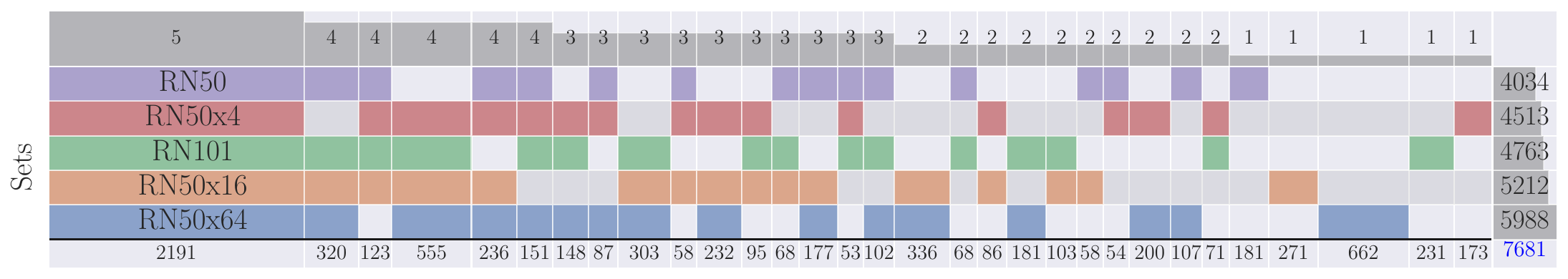}
    \includegraphics[width=\textwidth]{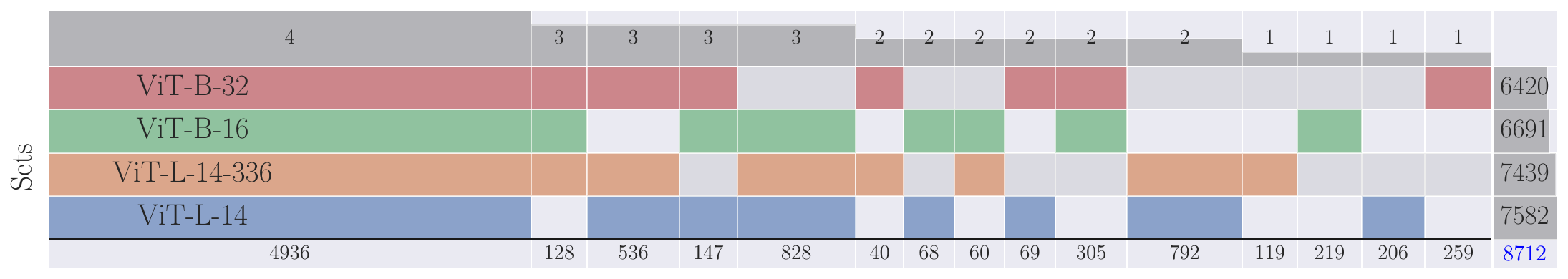}
    \includegraphics[width=\textwidth]{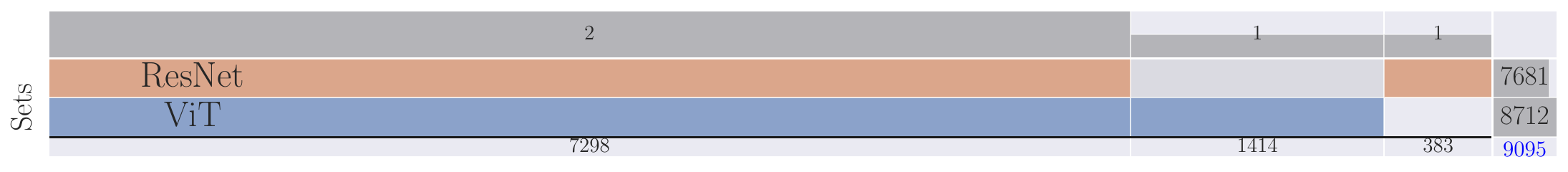}
    \caption{\textsc{Cifar100} Venn diagrams with the correct prediction of each backbone. The Top part of the Venn diagram shows the number of backbones that are predicting correctly a set of images. Each column represents a set of image instances that are predicted correctly by some group of backbones. Each row in the diagram shows in colour the backbone that correctly predicts a certain set of image instances, in grey when the backbone is not correctly predicting those instances. The bottom part of the Venn diagram shows the number of images in a certain set. The right part is the total amount of correctly predicted images per backbone.}
    \label{fig:cifar100_diagram}
\end{figure}

\begin{figure}[bt]
    \centering
    \underline{\textsc{Clever}}    
    \includegraphics[width=\textwidth]{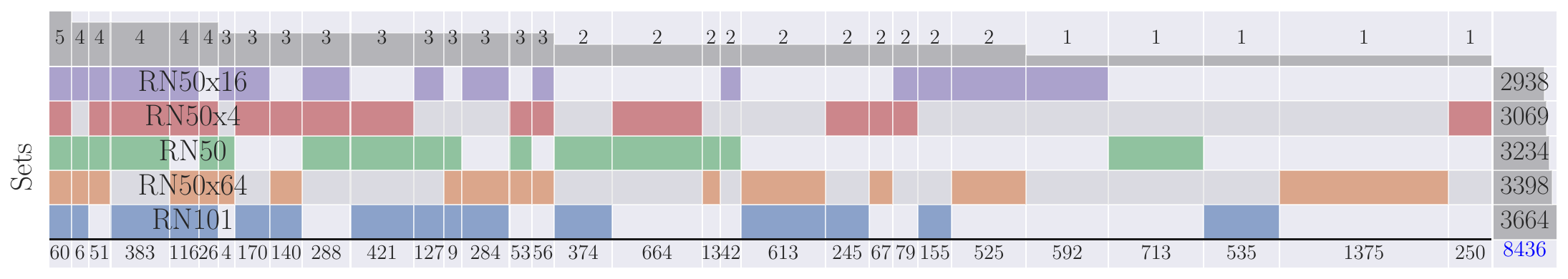}
    \includegraphics[width=\textwidth]{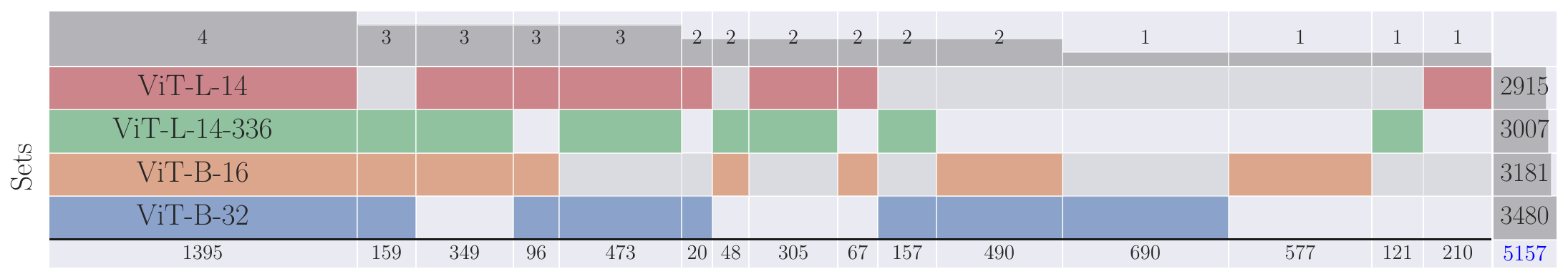}
    \includegraphics[width=\textwidth]{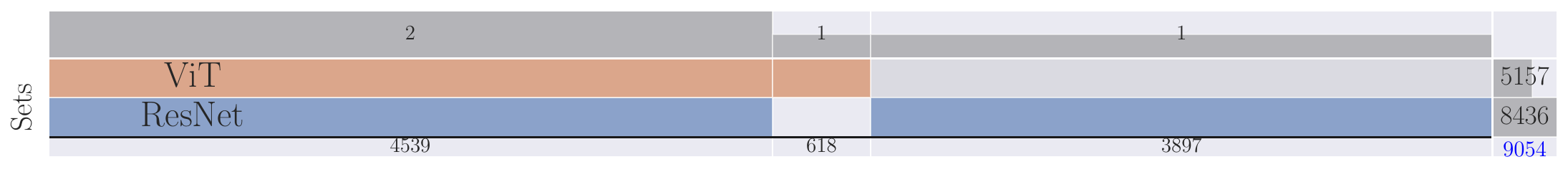}
    \caption{\textsc{Clever} Overlap diagrams with the correct prediction of each backbone. The Top part of the Overlap diagram shows the number of backbones that are predicting correctly a set of images. Each column represents a set of image instances that are predicted correctly by some group of backbones. Each row in the diagram shows in colour the backbone that correctly predicts a certain set of image instances, in grey when the backbone is not correctly predicting those instances. The bottom part of the Overlap diagram shows the number of images in a certain set. The right part is the total amount of correctly predicted images per backbone.}
    \label{fig:clever_diagram}
\end{figure}

\begin{figure}[bt]
    \centering
    \underline{\textsc{Country211}}    
    \includegraphics[width=\textwidth]{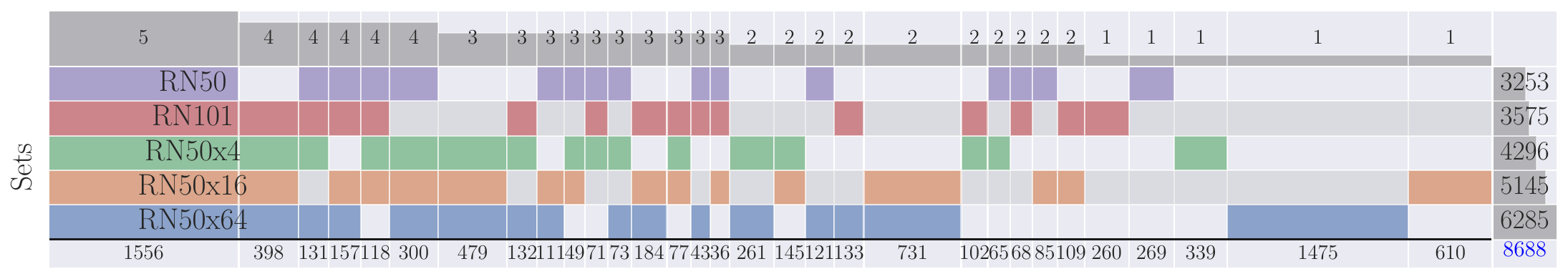}
    \includegraphics[width=\textwidth]{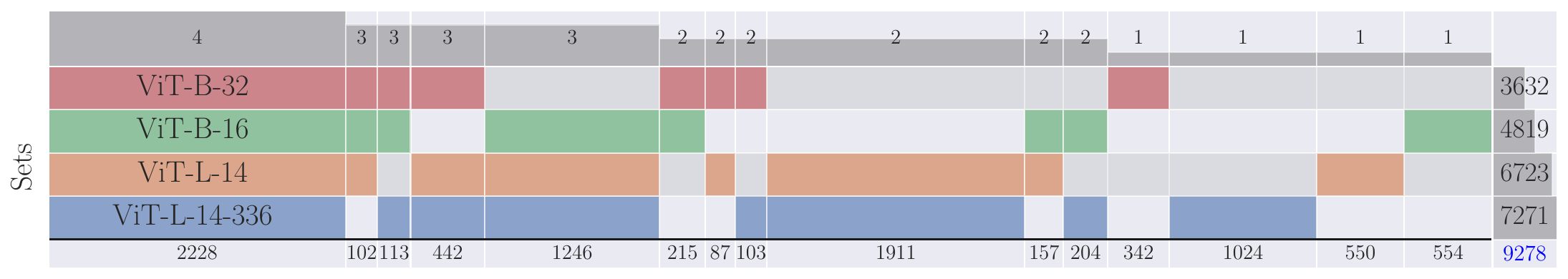}
    \includegraphics[width=\textwidth]{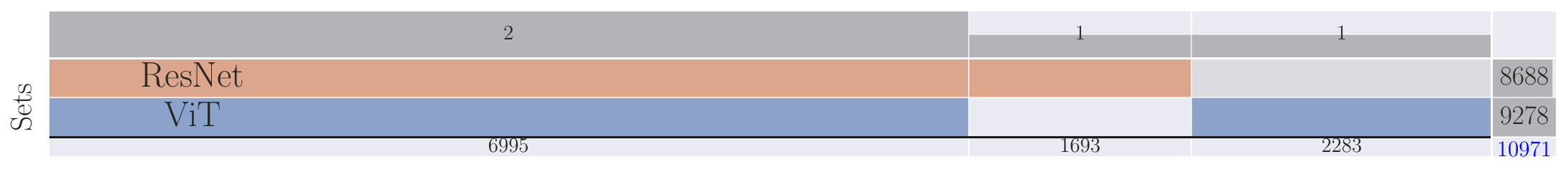}
    \caption{\textsc{Country211} Overlap diagrams with the correct prediction of each backbone. The Top part of the Overlap diagram shows the number of backbones that are predicting correctly a set of images. Each column represents a set of image instances that are predicted correctly by some group of backbones. Each row in the diagram shows in colour the backbone that correctly predicts a certain set of image instances, in grey when the backbone is not correctly predicting those instances. The bottom part of the Overlap diagram shows the number of images in a certain set. The right part is the total amount of correctly predicted images per backbone.}
    \label{fig:country211_diagram}
\end{figure}

\begin{figure}[bt]
    \centering
    \underline{\textsc{Cub}}    
    \includegraphics[width=\textwidth]{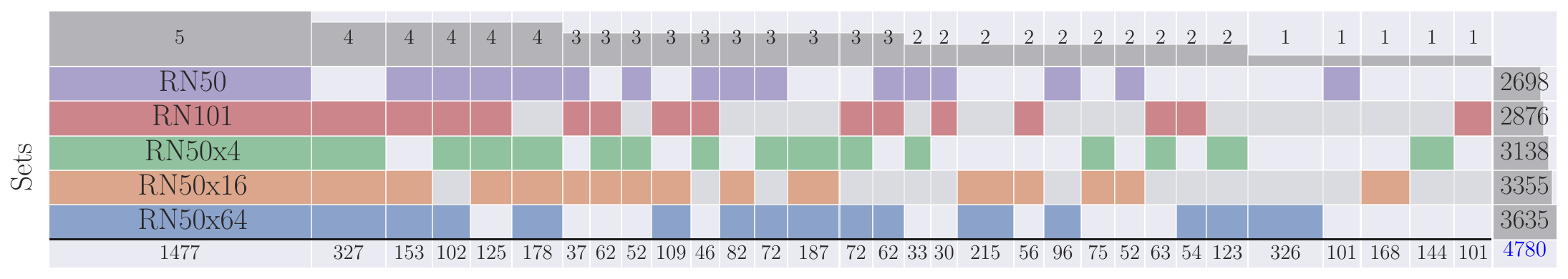}
    \includegraphics[width=\textwidth]{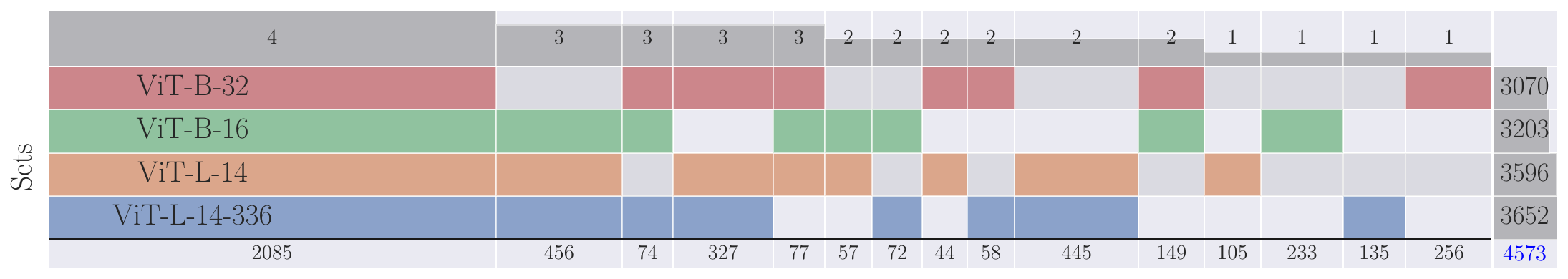}
    \includegraphics[width=\textwidth]{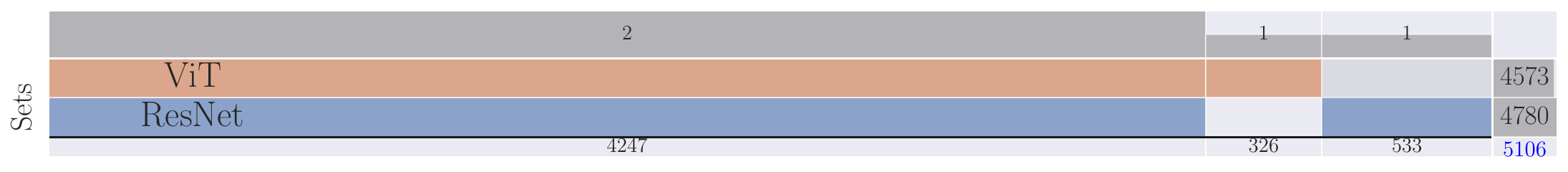}
    \caption{\textsc{Cub} Overlap diagrams with the correct prediction of each backbone. The Top part of the Overlap diagram shows the number of backbones that are predicting correctly a set of images. Each column represents a set of image instances that are predicted correctly by some group of backbones. Each row in the diagram shows in colour the backbone that correctly predicts a certain set of image instances, in grey when the backbone is not correctly predicting those instances. The bottom part of the Overlap diagram shows the number of images in a certain set. The right part is the total amount of correctly predicted images per backbone.}
    \label{fig:cub_diagram}
\end{figure}

\begin{figure}[bt]
    \centering
    \underline{\textsc{Dtd}}    
    \includegraphics[width=\textwidth]{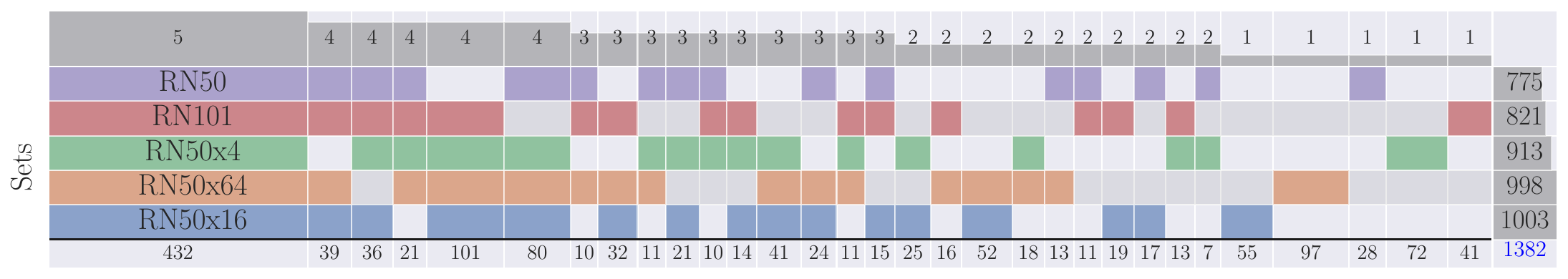}
    \includegraphics[width=\textwidth]{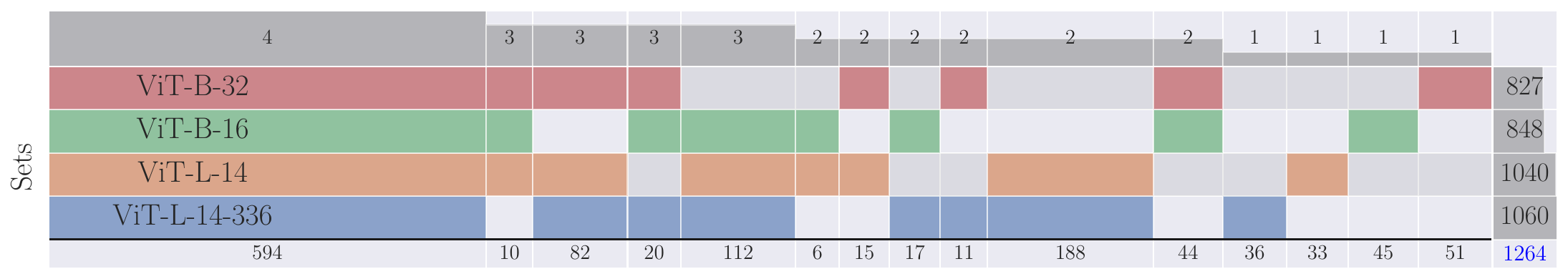}
    \includegraphics[width=\textwidth]{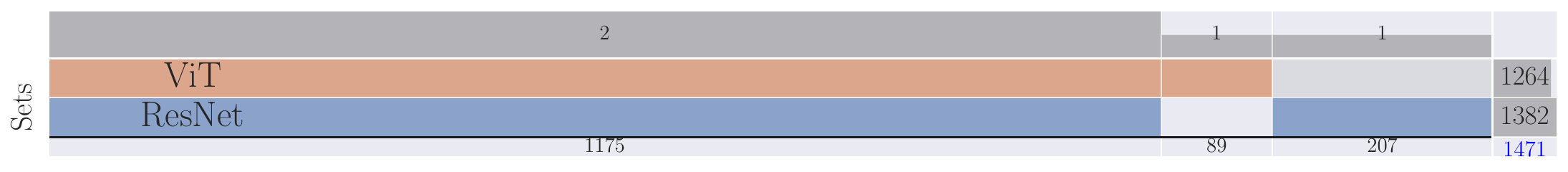}
    \caption{\textsc{Dtd} Overlap diagrams with the correct prediction of each backbone. The Top part of the Overlap diagram shows the number of backbones that are predicting correctly a set of images. Each column represents a set of image instances that are predicted correctly by some group of backbones. Each row in the diagram shows in colour the backbone that correctly predicts a certain set of image instances, in grey when the backbone is not correctly predicting those instances. The bottom part of the Overlap diagram shows the number of images in a certain set. The right part is the total amount of correctly predicted images per backbone.}
    \label{fig:dtd_diagram}
\end{figure}

\begin{figure}[bt]
    \centering
    \underline{\textsc{Eurosat}}    
    \includegraphics[width=\textwidth]{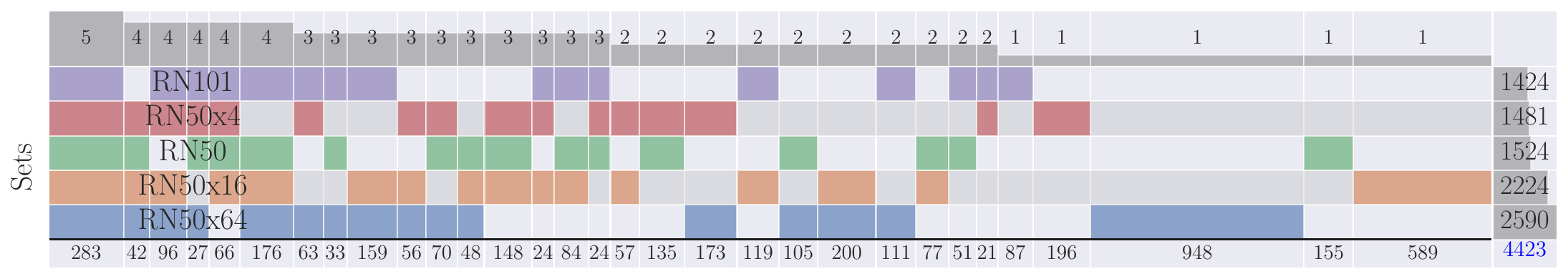}
    \includegraphics[width=\textwidth]{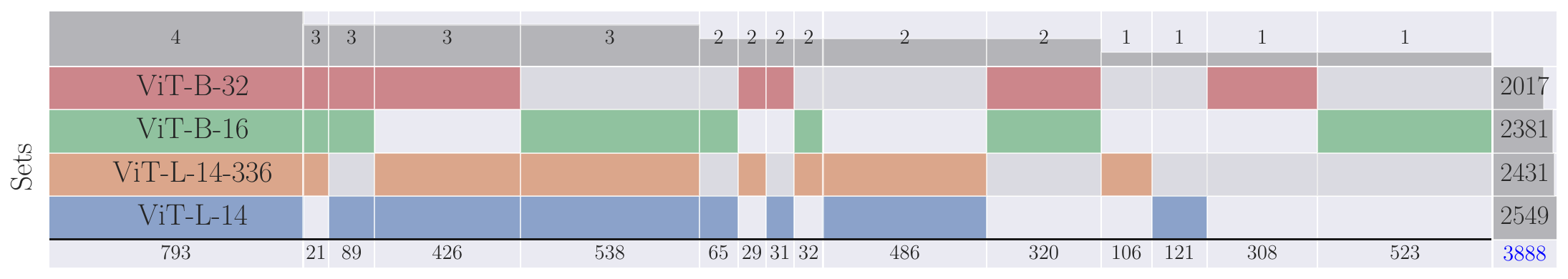}
    \includegraphics[width=\textwidth]{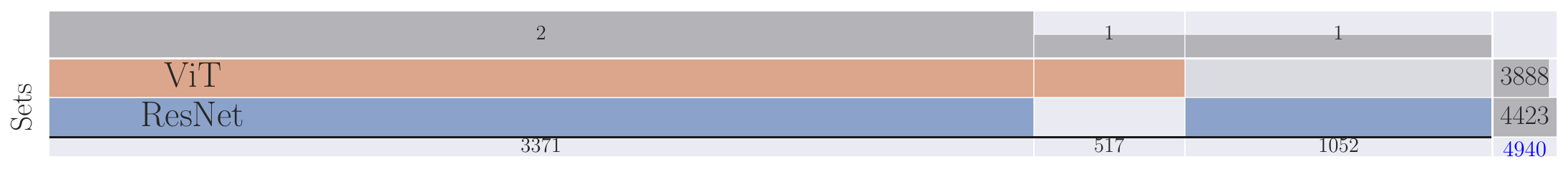}
    \caption{\textsc{Eurosat} Overlap diagrams with the correct prediction of each backbone. The Top part of the Overlap diagram shows the number of backbones that are predicting correctly a set of images. Each column represents a set of image instances that are predicted correctly by some group of backbones. Each row in the diagram shows in colour the backbone that correctly predicts a certain set of image instances, in grey when the backbone is not correctly predicting those instances. The bottom part of the Overlap diagram shows the number of images in a certain set. The right part is the total amount of correctly predicted images per backbone.}
    \label{fig:eurosat_diagram}
\end{figure}

\begin{figure}[bt]
    \centering
    \underline{\FGVC{}}    
    \includegraphics[width=\textwidth]{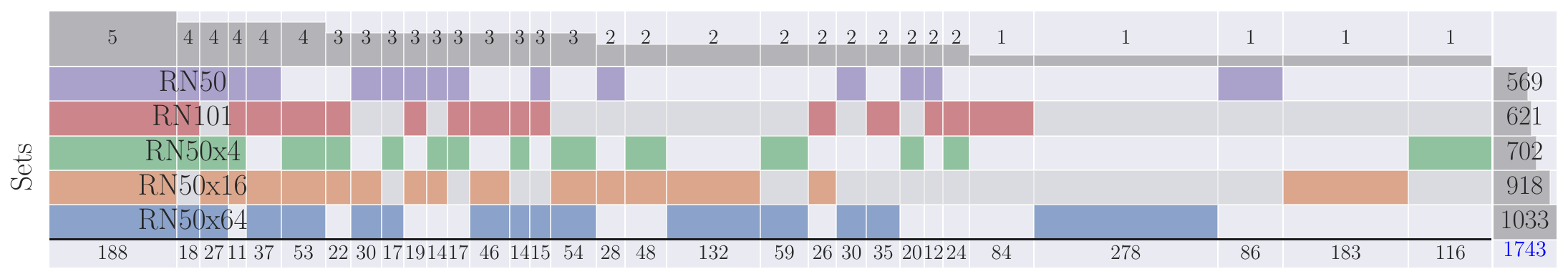}
    \includegraphics[width=\textwidth]{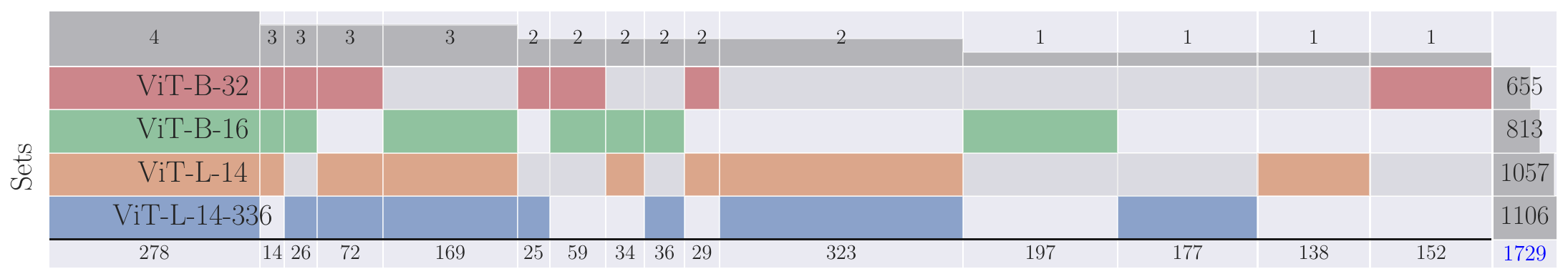}
    \includegraphics[width=\textwidth]{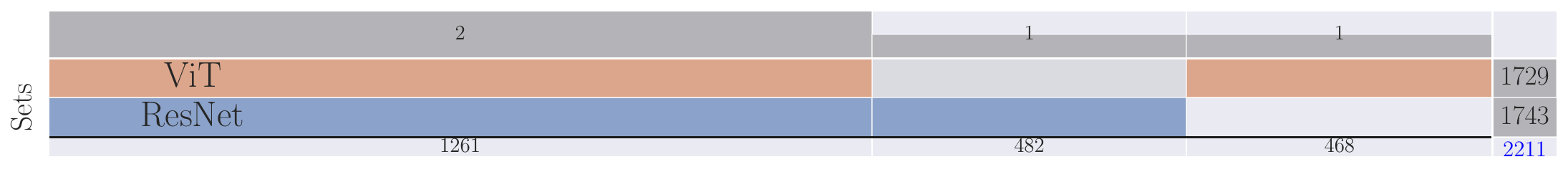}
    \caption{\FGVC{} Overlap diagrams with the correct prediction of each backbone. The Top part of the Overlap diagram shows the number of backbones that are predicting correctly a set of images. Each column represents a set of image instances that are predicted correctly by some group of backbones. Each row in the diagram shows in colour the backbone that correctly predicts a certain set of image instances, in grey when the backbone is not correctly predicting those instances. The bottom part of the Overlap diagram shows the number of images in a certain set. The right part is the total amount of correctly predicted images per backbone.}
    \label{fig:fgvc_diagram}
\end{figure}

\begin{figure}[bt]
    \centering
    \underline{\Flowers{}}    
    \includegraphics[width=\textwidth]{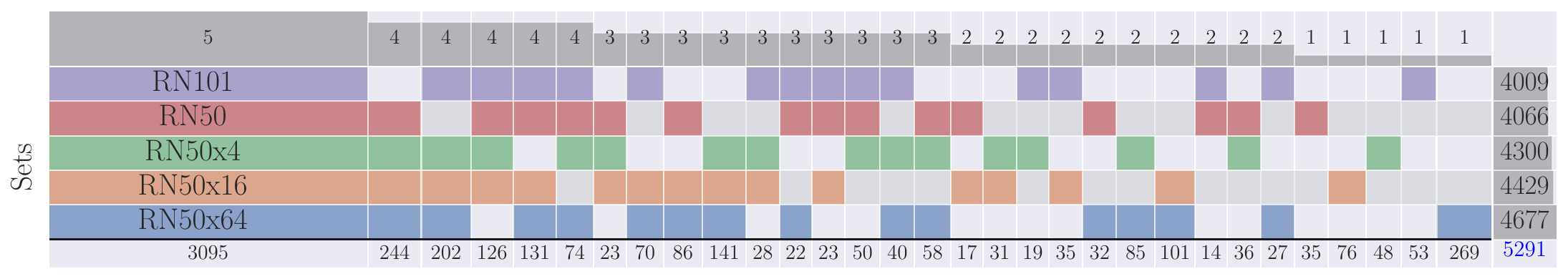}
    \includegraphics[width=\textwidth]{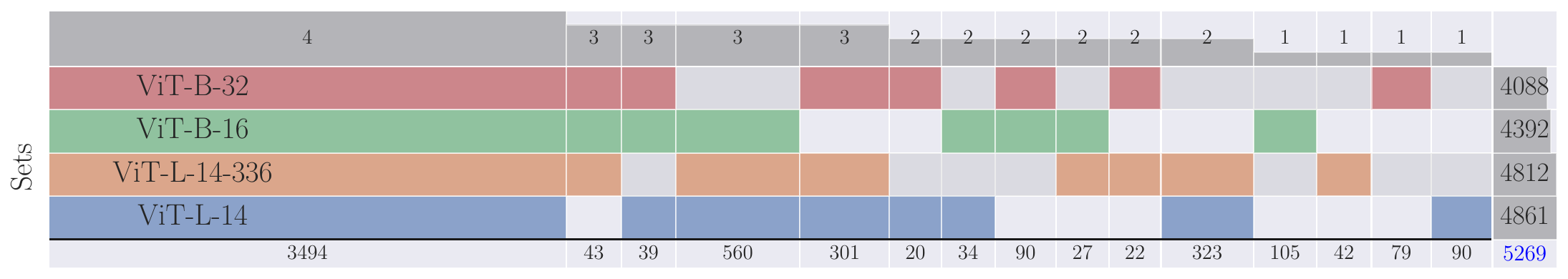}
    \includegraphics[width=\textwidth]{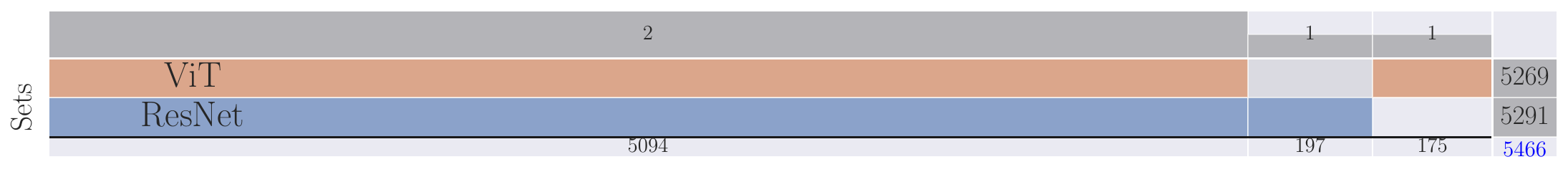}
    \caption{\Flowers{} Overlap diagrams with the correct prediction of each backbone. The Top part of the Overlap diagram shows the number of backbones that are predicting correctly a set of images. Each column represents a set of image instances that are predicted correctly by some group of backbones. Each row in the diagram shows in colour the backbone that correctly predicts a certain set of image instances, in grey when the backbone is not correctly predicting those instances. The bottom part of the Overlap diagram shows the number of images in a certain set. The right part is the total amount of correctly predicted images per backbone.}
    \label{fig:flowers_diagram}
\end{figure}

\begin{figure}[bt]
    \centering
    \underline{\Food{}}    
    \includegraphics[width=\textwidth]{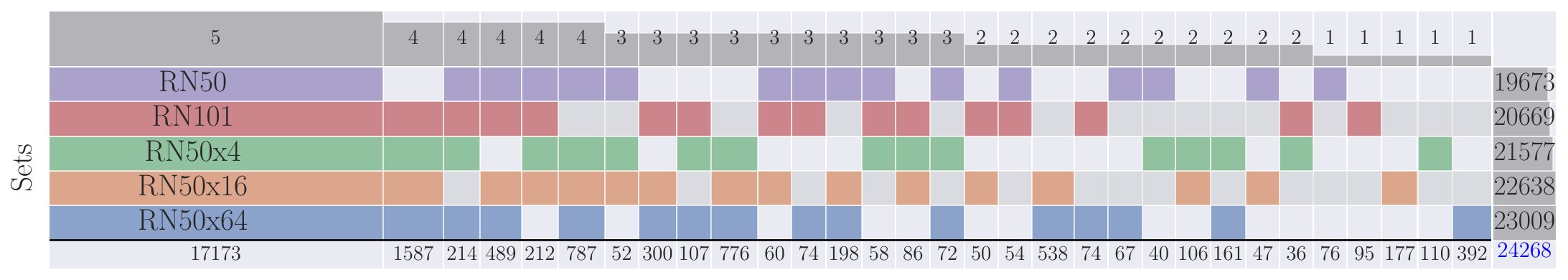}
    \includegraphics[width=\textwidth]{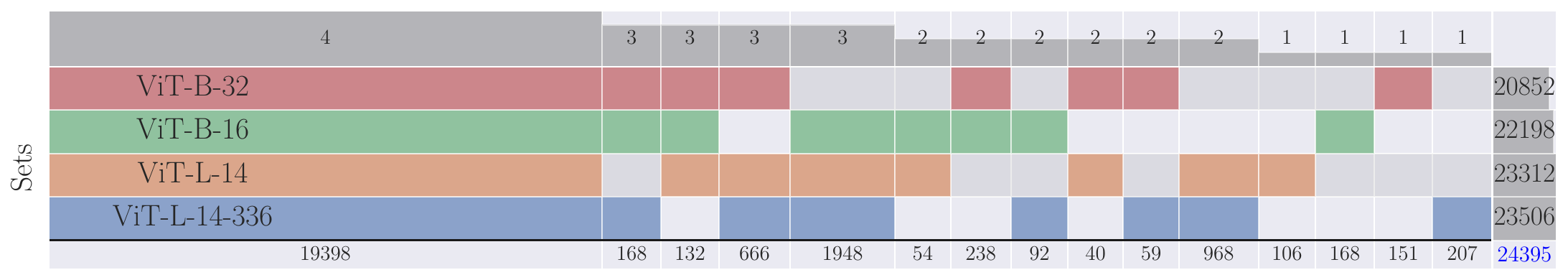}
    \includegraphics[width=\textwidth]{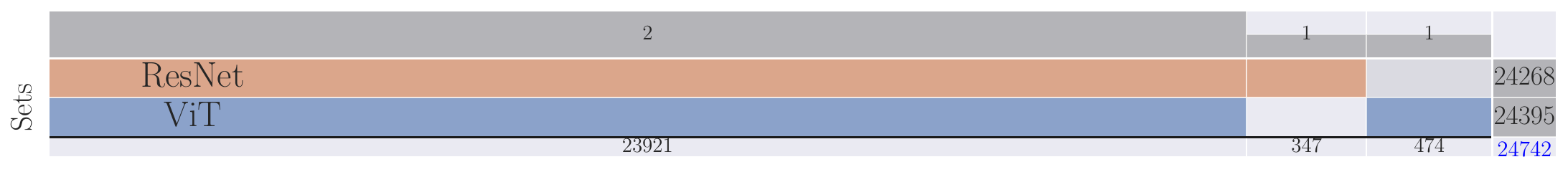}
    \caption{\Food{} Overlap diagrams with the correct prediction of each backbone. The Top part of the Overlap diagram shows the number of backbones that are predicting correctly a set of images. Each column represents a set of image instances that are predicted correctly by some group of backbones. Each row in the diagram shows in colour the backbone that correctly predicts a certain set of image instances, in grey when the backbone is not correctly predicting those instances. The bottom part of the Overlap diagram shows the number of images in a certain set. The right part is the total amount of correctly predicted images per backbone.}
    \label{fig:food_diagram}
\end{figure}

\begin{figure}[bt]
    \centering
    \underline{\Gtsrb{}}    
    \includegraphics[width=\textwidth]{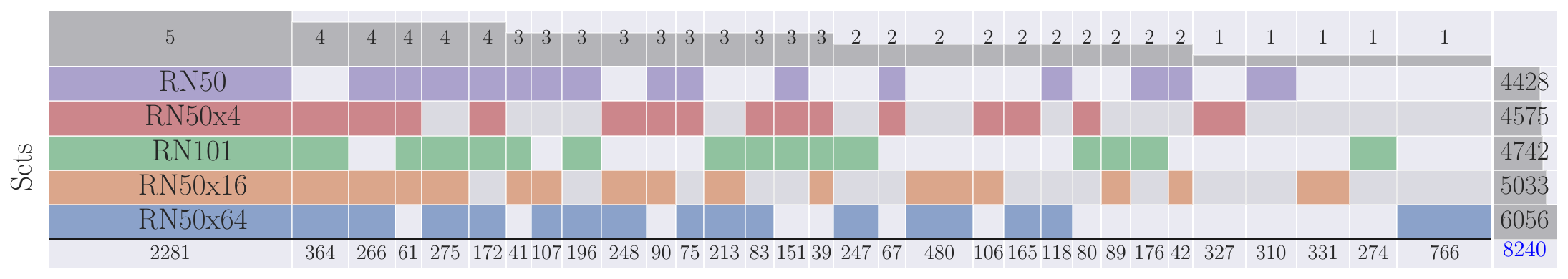}
    \includegraphics[width=\textwidth]{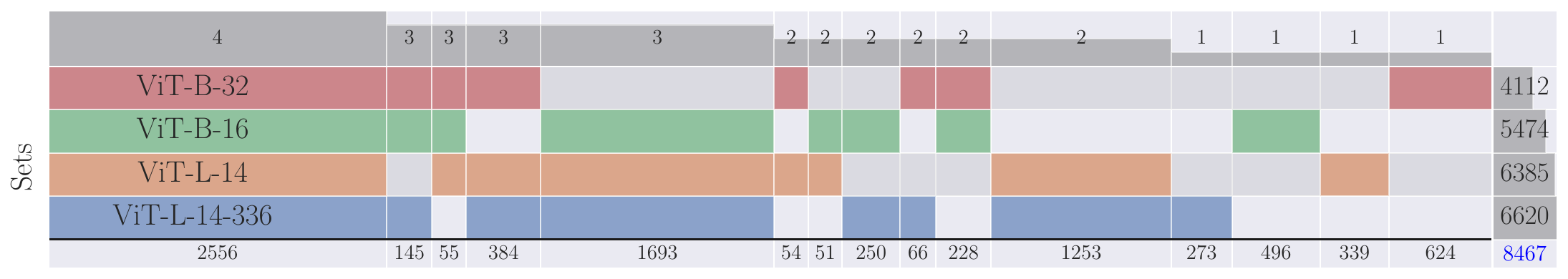}
    \includegraphics[width=\textwidth]{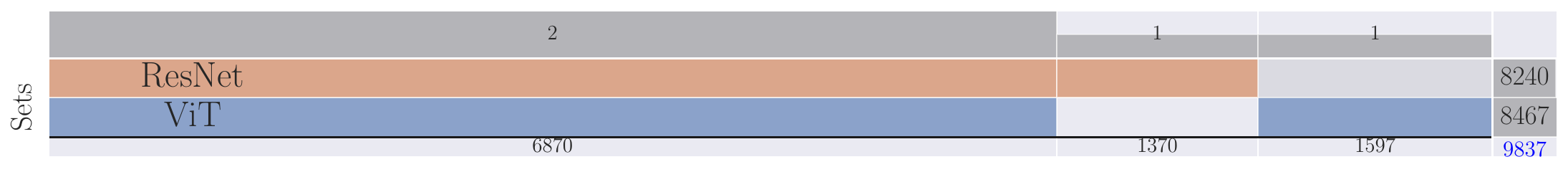}
    \caption{\Gtsrb{} Overlap diagrams with the correct prediction of each backbone. The Top part of the Overlap diagram shows the number of backbones that are predicting correctly a set of images. Each column represents a set of image instances that are predicted correctly by some group of backbones. Each row in the diagram shows in colour the backbone that correctly predicts a certain set of image instances, in grey when the backbone is not correctly predicting those instances. The bottom part of the Overlap diagram shows the number of images in a certain set. The right part is the total amount of correctly predicted images per backbone.}
    \label{fig:gtsrb_diagram}
\end{figure}

\begin{figure}[bt]
    \centering
    \underline{\Mnist{}}    
    \includegraphics[width=\textwidth]{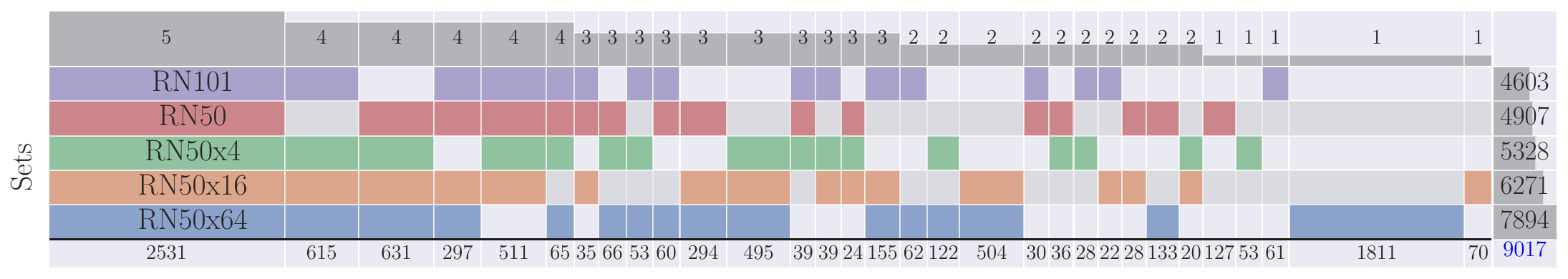}
    \includegraphics[width=\textwidth]{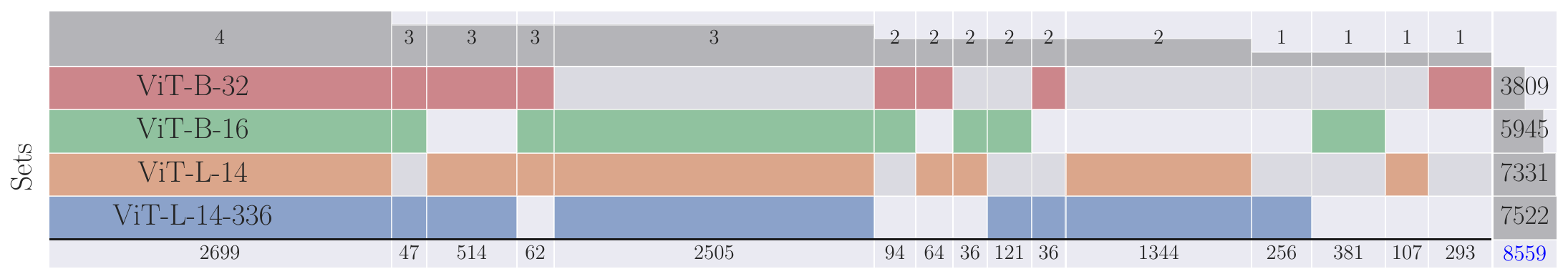}
    \includegraphics[width=\textwidth]{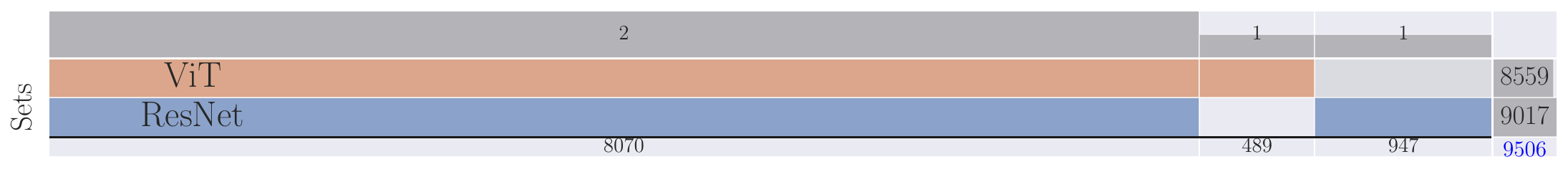}
    \caption{\Mnist{} Overlap diagrams with the correct prediction of each backbone. The Top part of the Overlap diagram shows the number of backbones that are predicting correctly a set of images. Each column represents a set of image instances that are predicted correctly by some group of backbones. Each row in the diagram shows in colour the backbone that correctly predicts a certain set of image instances, in grey when the backbone is not correctly predicting those instances. The bottom part of the Overlap diagram shows the number of images in a certain set. The right part is the total amount of correctly predicted images per backbone.}
    \label{fig:mnist_diagram}
\end{figure}

\begin{figure}[bt]
    \centering
    \underline{\Pcam{}}    
    \includegraphics[width=\textwidth]{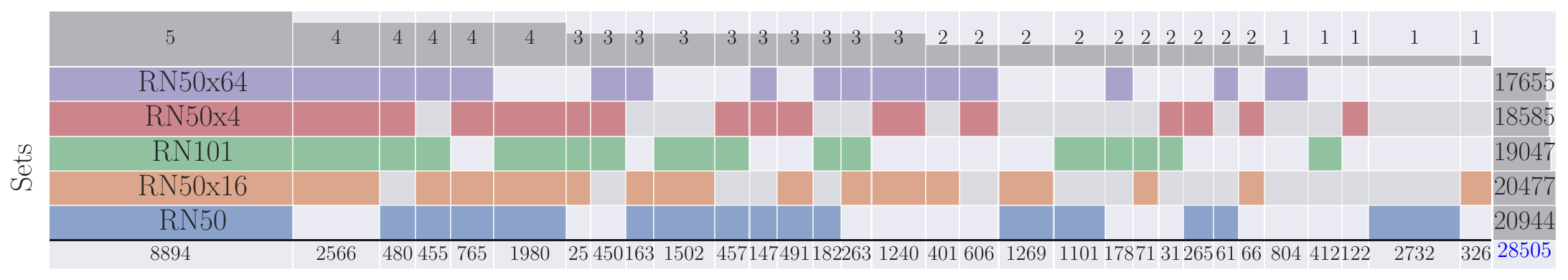}
    \includegraphics[width=\textwidth]{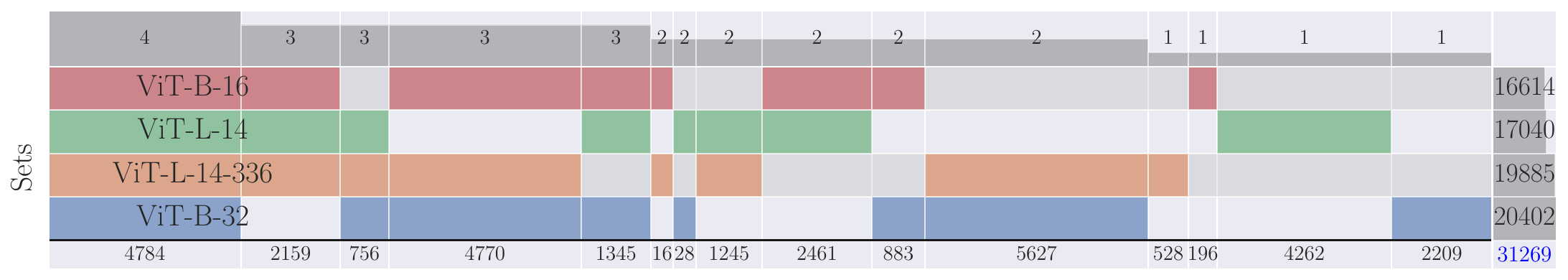}
    \includegraphics[width=\textwidth]{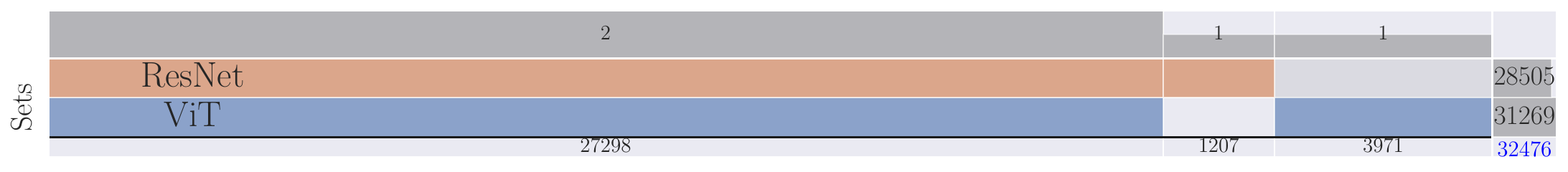}
    \caption{\Pcam{} Overlap diagrams with the correct prediction of each backbone. The Top part of the Overlap diagram shows the number of backbones that are predicting correctly a set of images. Each column represents a set of image instances that are predicted correctly by some group of backbones. Each row in the diagram shows in colour the backbone that correctly predicts a certain set of image instances, in grey when the backbone is not correctly predicting those instances. The bottom part of the Overlap diagram shows the number of images in a certain set. The right part is the total amount of correctly predicted images per backbone.}
    \label{fig:pcam_diagram}
\end{figure}

\begin{figure}[bt]
    \centering
    \underline{\Pets{}}    
    \includegraphics[width=\textwidth]{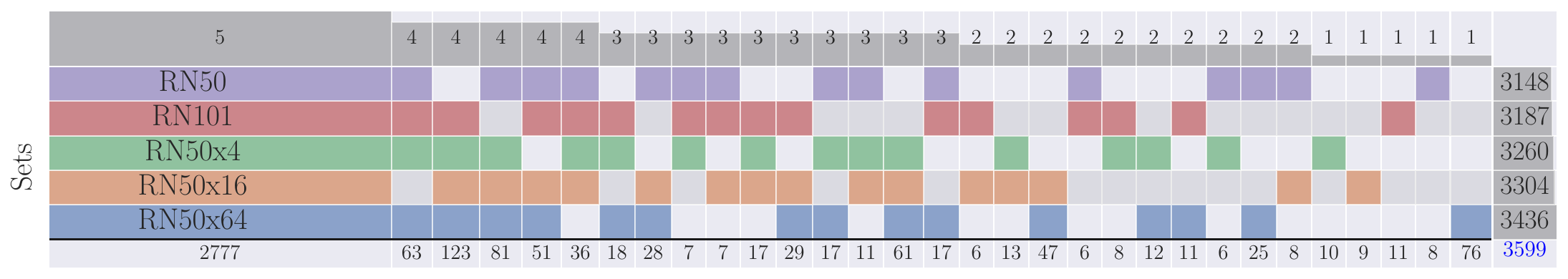}
    \includegraphics[width=\textwidth]{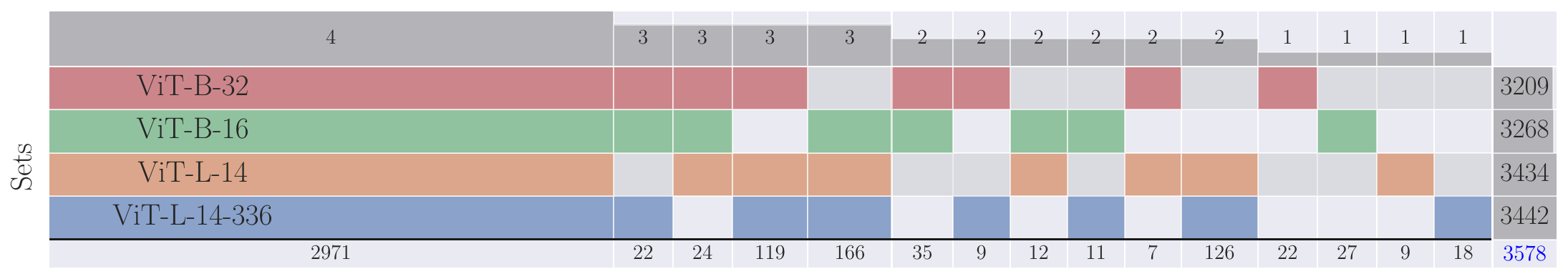}
    \includegraphics[width=\textwidth]{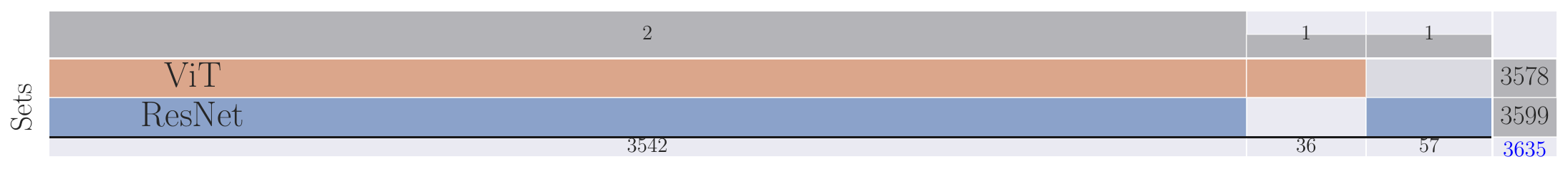}
    \caption{\Pets{} Overlap diagrams with the correct prediction of each backbone. The Top part of the Overlap diagram shows the number of backbones that are predicting correctly a set of images. Each column represents a set of image instances that are predicted correctly by some group of backbones. Each row in the diagram shows in colour the backbone that correctly predicts a certain set of image instances, in grey when the backbone is not correctly predicting those instances. The bottom part of the Overlap diagram shows the number of images in a certain set. The right part is the total amount of correctly predicted images per backbone.}
    \label{fig:pets_diagram}
\end{figure}

\begin{figure}[bt]
    \centering
    \underline{\textsc{Render}\Renderedsst{}2}    
    \includegraphics[width=\textwidth]{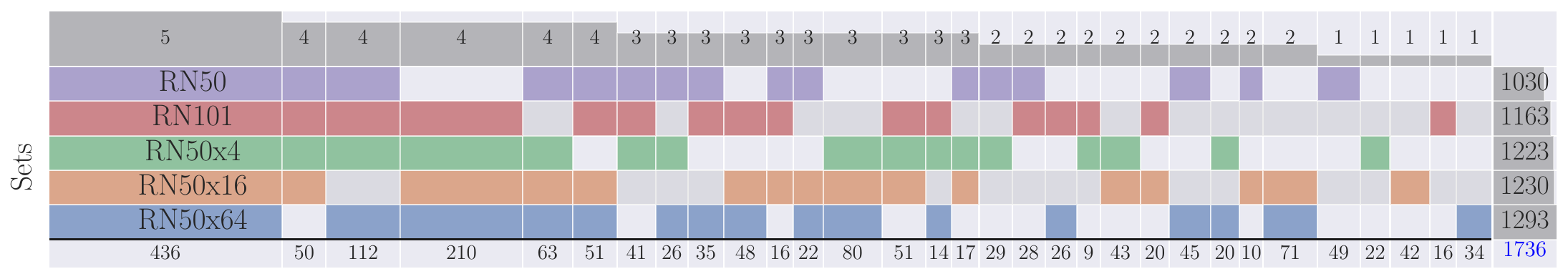}
    \includegraphics[width=\textwidth]{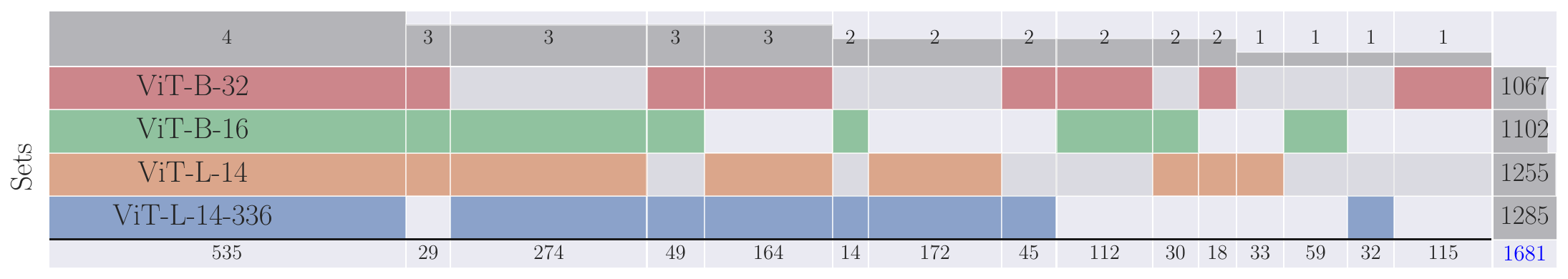}
    \includegraphics[width=\textwidth]{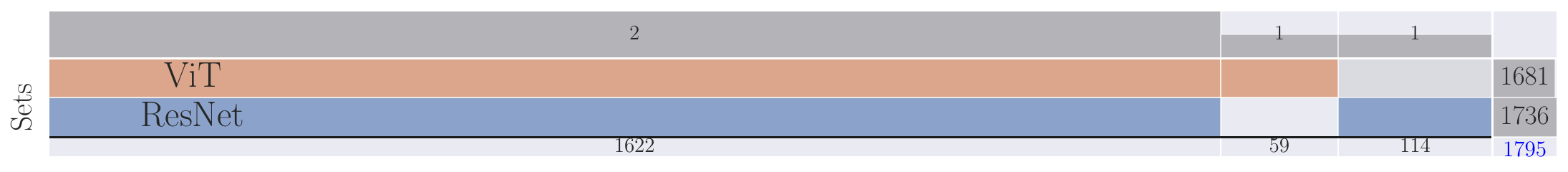}
    \caption{\textsc{Render}\Renderedsst{}2 Overlap diagrams with the correct prediction of each backbone. The Top part of the Overlap diagram shows the number of backbones that are predicting correctly a set of images. Each column represents a set of image instances that are predicted correctly by some group of backbones. Each row in the diagram shows in colour the backbone that correctly predicts a certain set of image instances, in grey when the backbone is not correctly predicting those instances. The bottom part of the Overlap diagram shows the number of images in a certain set. The right part is the total amount of correctly predicted images per backbone.}
    \label{fig:sst2_diagram}
\end{figure}

\begin{figure}[bt]
    \centering
    \underline{\Resisc{}45}    
    \includegraphics[width=\textwidth]{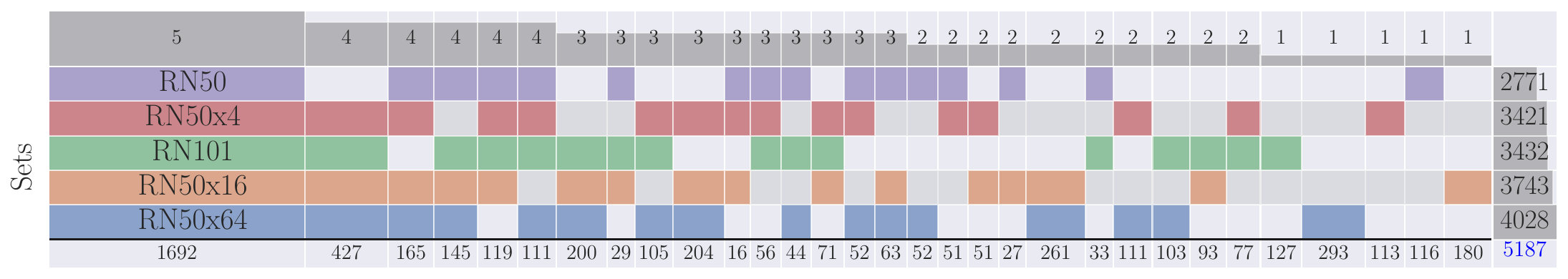}
    \includegraphics[width=\textwidth]{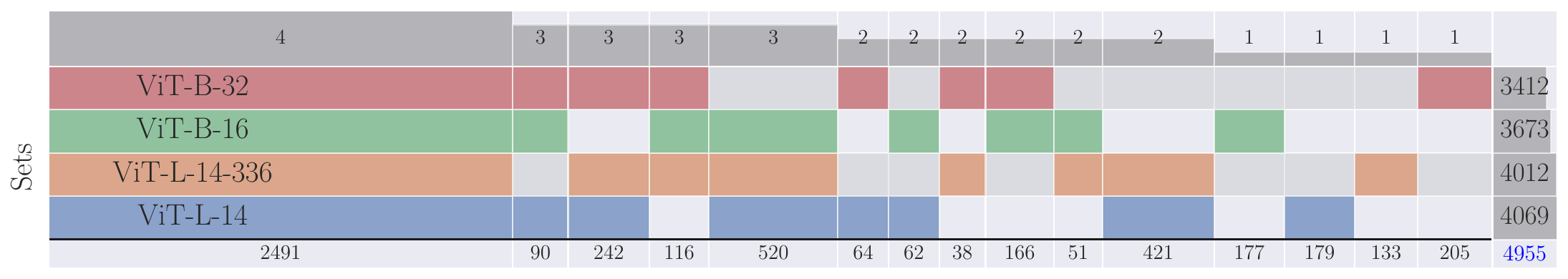}
    \includegraphics[width=\textwidth]{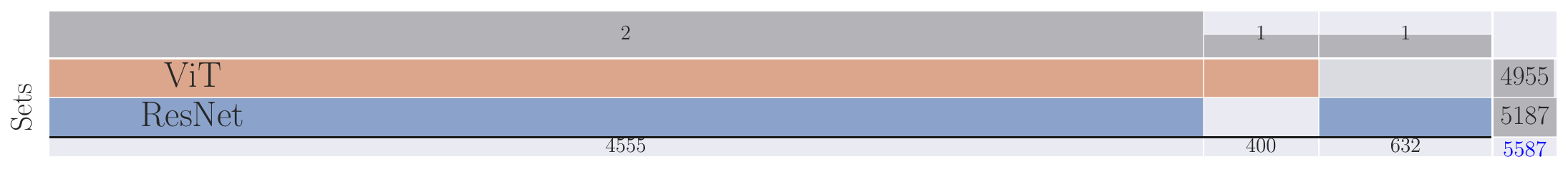}
    \caption{\Resisc{}45 Overlap diagrams with the correct prediction of each backbone. The Top part of the Overlap diagram shows the number of backbones that are predicting correctly a set of images. Each column represents a set of image instances that are predicted correctly by some group of backbones. Each row in the diagram shows in colour the backbone that correctly predicts a certain set of image instances, in grey when the backbone is not correctly predicting those instances. The bottom part of the Overlap diagram shows the number of images in a certain set. The right part is the total amount of correctly predicted images per backbone.}
    \label{fig:resisc45_diagram}
\end{figure}

\begin{figure}[bt]
    \centering
    \underline{\STL{}10}    
    \includegraphics[width=\textwidth]{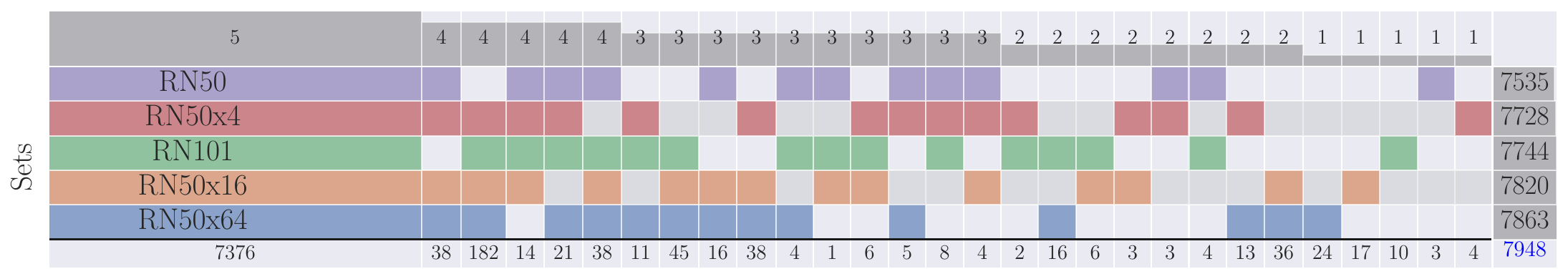}
    \includegraphics[width=\textwidth]{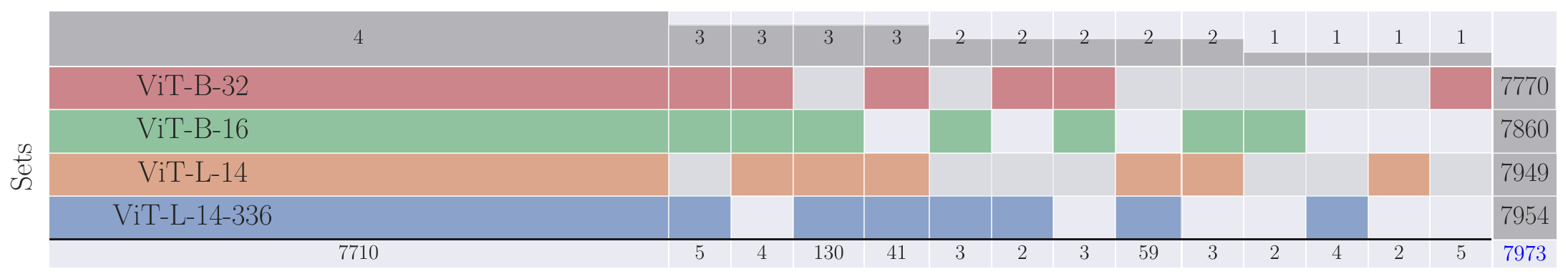}
    \includegraphics[width=\textwidth]{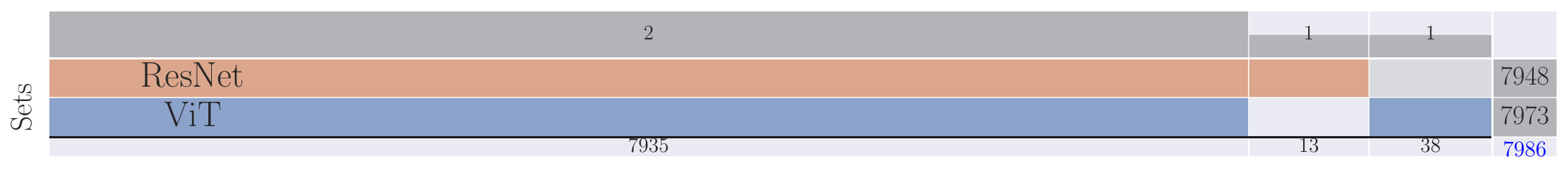}
    \caption{\STL{}10 Overlap diagrams with the correct prediction of each backbone. The Top part of the Overlap diagram shows the number of backbones that are predicting correctly a set of images. Each column represents a set of image instances that are predicted correctly by some group of backbones. Each row in the diagram shows in colour the backbone that correctly predicts a certain set of image instances, in grey when the backbone is not correctly predicting those instances. The bottom part of the Overlap diagram shows the number of images in a certain set. The right part is the total amount of correctly predicted images per backbone.}
    \label{fig:stl10_diagram}
\end{figure}

\begin{figure}[bt]
    \centering
    \underline{\SUN{}397}    
    \includegraphics[width=\textwidth]{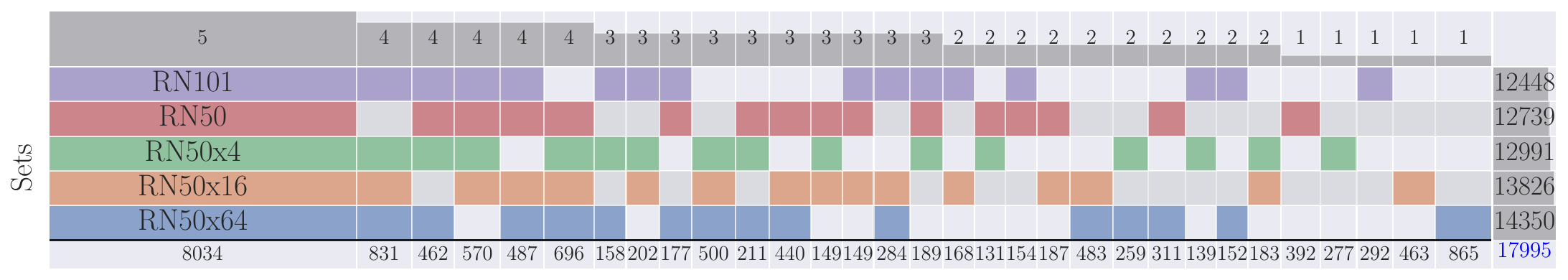}
    \includegraphics[width=\textwidth]{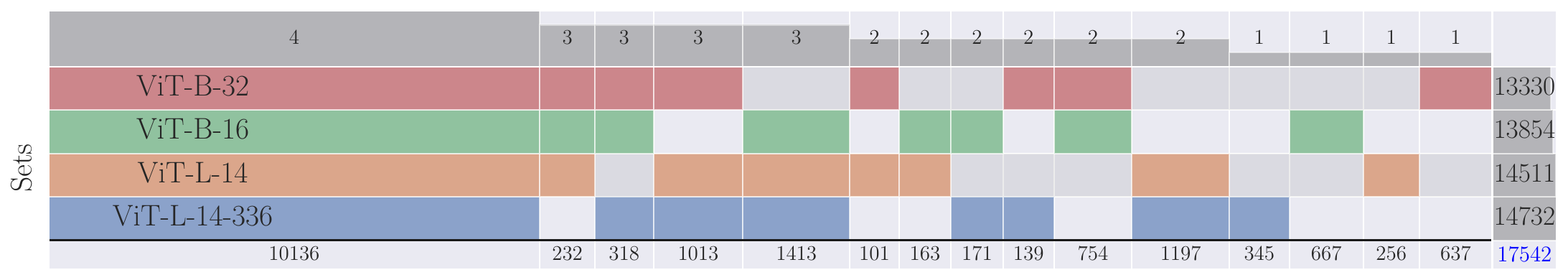}
    \includegraphics[width=\textwidth]{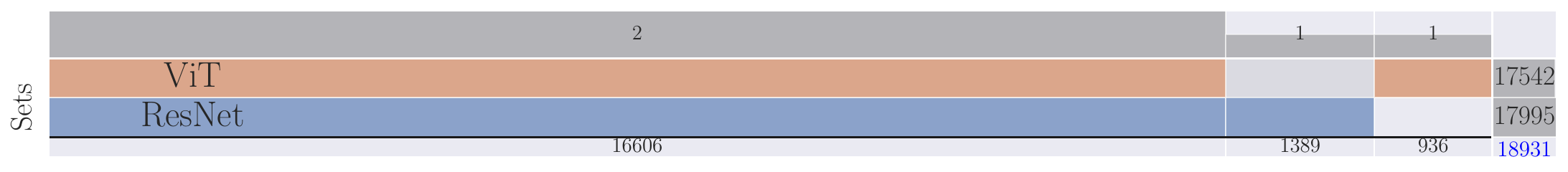}
    \caption{\SUN{}397 Overlap diagrams with the correct prediction of each backbone. The Top part of the Overlap diagram shows the number of backbones that are predicting correctly a set of images. Each column represents a set of image instances that are predicted correctly by some group of backbones. Each row in the diagram shows in colour the backbone that correctly predicts a certain set of image instances, in grey when the backbone is not correctly predicting those instances. The bottom part of the Overlap diagram shows the number of images in a certain set. The right part is the total amount of correctly predicted images per backbone.}
    \label{fig:sun397_diagram}
\end{figure}

In this section, we present the linear Venn Diagrams for each of the other datasets used in the experiment section \Caltech{} (Figure \ref{fig:caltech101_venn_diagram}), 
\Cars{} (Figure \ref{fig:cars_diagram}), 
\Cub{} (Figure \ref{fig:cub_diagram}), 
\Cifarten{} (Figure \ref{fig:cifar10_diagram}), 
\Cifaronehundred{} (Figure \ref{fig:cifar100_diagram}), 
\Clevr{} (Figure \ref{fig:clever_diagram}), 
\Country{}211 (Figure \ref{fig:country211_diagram}), 
\Cub{} (Figure \ref{fig:cub_diagram}), \Dtd{} (Figure \ref{fig:dtd_diagram}), 
\Eurosat (Figure \ref{fig:eurosat_diagram}), 
\FGVC{} (Figure \ref{fig:fgvc_diagram}), 
\Flowers{} (Figure \ref{fig:flowers_diagram}), 
\Food{} (Figure \ref{fig:food_diagram}), 
\Gtsrb{} (Figure \ref{fig:gtsrb_diagram}), 
\Mnist{} (Figure \ref{fig:mnist_diagram}),  
\Pcam{} (Figure \ref{fig:pcam_diagram}),  
\Pets{} (Figure \ref{fig:pets_diagram}),  
Rendered\Renderedsst{}2 (Figure \ref{fig:sst2_diagram}),  
\Resisc{}45 (Figure \ref{fig:resisc45_diagram}),  
\STL{}10 (Figure \ref{fig:stl10_diagram}), and  
\SUN{}397 (Figure \ref{fig:sun397_diagram}).  
We can see that in each dataset, the CLIP backbones present possible complementarities that could be exploited. 
\label{app:venndiagrams}


\section{Possible Combinations}
\label{app:combinations}
Tables \ref{tab:all_possible_combinations_pets}, \ref{tab:all_possible_combinations_cars}, \ref{tab:all_possible_combinations_cub}, \ref{tab:all_possible_combinations_dtd}, \ref{tab:all_possible_combinations_fgvc}, \ref{tab:all_possible_combinations_food}, \ref{tab:all_possible_combinations_flowers}, and \ref{tab:all_possible_combinations_inet} present the results of possible combinations of backbones using the non-parametric and parametric approaches proposed in the paper. Notably, the performance of \NNC{} consistently emerges as the best across various backbone combinations and datasets when compared to other methods.

Notably, instances exist where the combination of specific backbones yields a more substantial performance boost than utilizing all backbones together. For instance, in the \Pets{} dataset, combining ResNet 50, 101, and ViT-B-32 results in a delta improvement of 2.37\%, surpassing the 0.99\% improvement achieved by using the five backbones selected for this experiment. This phenomenon is consistent across datasets with different backbone combinations. In \Cars{}, there is a boost of 5.71\% when combining ResNet-101 and ViT-B-32, compared to the 2.55\% boost when using all five different backbones. While the best delta improvement among backbones may not necessarily come from combining all backbones, the best overall accuracy is consistently obtained when using the combination of all backbones.
\begin{table*}[b]
\centering
\caption{Our results on \Pets{} dataset for all the possible combinations of combining the zero-shot predictions of CLIP backbones, which we group intro non-parametric and parametric techniques. Also, the best-performing single backbone (\Best) and the \Oracle{} performance. We present, for each combination of backbones, the improvement \upg{}, constancy \samey{} and deterioration \downr{} of accuracy performance for each method when we compare it against the \Best{} backbone. Mean, Max, and Min $\Delta$ summarize the difference in performance across methods and backbone combinations.}
\Pets \\
\qquad
\resizebox{0.9\textwidth}{!}{
\begin{tabular}{ccccccccccccccc}
\toprule
\multicolumn{2}{c}{ResNet} & \multicolumn{3}{c}{ViT} & \multirow{2}{*}{\Best}  & \multicolumn{4}{c}{Non-Parametric} & \multicolumn{4}{c}{Parametric} & \multirow{2}{*}{\Oracle}\\
\cmidrule(lr){1-2} \cmidrule(lr){3-5} \cmidrule(lr){7-10} \cmidrule(lr){11-14}
50 & 101 & B-32 & B-16 & L-14 & & \VoteOne & \VoteThree & \Confidence & \LogitAvg & \CalibratedConfidence & \CLogitAvg & \GAC & \NNC &  \\
\midrule
\Checkmark &   &   &   &   &  \multicolumn{10}{c}{\xfill{.1em}  85.80 \samey{0.00} \xfill{.1em}} \\ 
  & \Checkmark &   &   &   & \multicolumn{10}{c}{\xfill{.1em}  86.86 \samey{0.00} \xfill{.1em}} \\ 
  &   & \Checkmark &   &   & \multicolumn{10}{c}{\xfill{.1em}  87.46 \samey{0.00} \xfill{.1em}} \\ 
  &   &   & \Checkmark &   & \multicolumn{10}{c}{\xfill{.1em}  89.07 \samey{0.00} \xfill{.1em}} \\ 
  &   &   &   & \Checkmark & \multicolumn{10}{c}{\xfill{.1em}  93.59 \samey{0.00} \xfill{.1em}} \\ \midrule
\Checkmark & \Checkmark &   &   &   & 86.86 & 87.84 \upg{0.98} & 87.90 \upg{1.04} & 88.20 \upg{1.34} & 88.03 \upg{1.17} & 87.93 \upg{1.06} & 87.84 \upg{0.98} & 87.14 \upg{0.27} & 88.09 \upg{1.23} & 91.88 \upg{5.01} \\ \midrule
\Checkmark &   & \Checkmark &   &   & 87.46 & 88.50 \upg{1.04} & 89.02 \upg{1.55} & 85.25 \downr{-2.21} & 89.40 \upg{1.94} & 87.98 \upg{0.52} & 89.21 \upg{1.74} & 89.48 \upg{2.02} & 89.45 \upg{1.99} & 92.64 \upg{5.18} \\ \midrule
\Checkmark &   &   & \Checkmark &   & 89.07 & 89.83 \upg{0.76} & 89.92 \upg{0.84} & 89.67 \upg{0.6} & 89.83 \upg{0.76} & 89.48 \upg{0.41} & 89.72 \upg{0.65} & 89.86 \upg{0.79} & 90.27 \upg{1.2} & 92.89 \upg{3.82} \\ \midrule
\Checkmark &   &   &   & \Checkmark & 93.59 & 94.03 \upg{0.44} & 94.00 \upg{0.41} & 94.00 \upg{0.41} & 93.81 \upg{0.22} & 93.73 \upg{0.14} & 93.98 \upg{0.38} & 93.73 \upg{0.14} & 94.19 \upg{0.6} & 96.18 \upg{2.59} \\ \midrule
  & \Checkmark & \Checkmark &   &   & 87.46 & 88.72 \upg{1.25} & 88.99 \upg{1.53} & 85.91 \downr{-1.55} & 89.07 \upg{1.61} & 88.55 \upg{1.09} & 89.07 \upg{1.61} & 89.10 \upg{1.64} & 89.04 \upg{1.58} & 93.00 \upg{5.53} \\ \midrule
  & \Checkmark &   & \Checkmark &   & 89.07 & 90.00 \upg{0.93} & 90.11 \upg{1.04} & 89.75 \upg{0.68} & 89.89 \upg{0.82} & 89.81 \upg{0.74} & 90.11 \upg{1.04} & 89.45 \upg{0.38} & 90.30 \upg{1.23} & 93.13 \upg{4.06} \\ \midrule
  & \Checkmark &   &   & \Checkmark & 93.59 & 93.32 \downr{-0.27} & 93.32 \downr{-0.27} & 93.13 \downr{-0.46} & 93.24 \downr{-0.35} & 93.27 \downr{-0.33} & 93.43 \downr{-0.16} & 93.70 \upg{0.11} & 93.70 \upg{0.11} & 96.51 \upg{2.92} \\ \midrule
  &   & \Checkmark & \Checkmark &   & 89.07 & 89.86 \upg{0.79} & 90.16 \upg{1.09} & 87.35 \downr{-1.72} & 90.41 \upg{1.34} & 89.45 \upg{0.38} & 90.38 \upg{1.31} & 90.27 \upg{1.2} & 90.30 \upg{1.23} & 93.35 \upg{4.28} \\ \midrule
  &   & \Checkmark &   & \Checkmark & 93.59 & 93.79 \upg{0.19} & 93.76 \upg{0.16} & 93.00 \downr{-0.6} & 93.38 \downr{-0.22} & 93.00 \downr{-0.6} & 93.73 \upg{0.14} & 93.89 \upg{0.3} & 93.87 \upg{0.27} & 95.99 \upg{2.4} \\ \midrule
  &   &   & \Checkmark & \Checkmark & 93.59 & 93.98 \upg{0.38} & 93.95 \upg{0.35} & 93.92 \upg{0.33} & 94.06 \upg{0.46} & 93.84 \upg{0.25} & 93.98 \upg{0.38} & 93.73 \upg{0.14} & 93.92 \upg{0.33} & 96.18 \upg{2.59} \\ \midrule
\Checkmark & \Checkmark & \Checkmark &   &   & 87.46 & 88.85 \upg{1.39} & 89.29 \upg{1.83} & 84.93 \downr{-2.53} & 89.72 \upg{2.26} & 88.42 \upg{0.95} & 89.45 \upg{1.99} & 89.48 \upg{2.02} & 89.83 \upg{2.37} & 94.69 \upg{7.22} \\ \midrule
\Checkmark & \Checkmark &   & \Checkmark &   & 89.07 & 89.29 \upg{0.22} & 89.81 \upg{0.74} & 90.02 \upg{0.95} & 90.11 \upg{1.04} & 89.86 \upg{0.79} & 90.08 \upg{1.01} & 90.11 \upg{1.04} & 90.38 \upg{1.31} & 94.74 \upg{5.67} \\ \midrule
\Checkmark & \Checkmark &   &   & \Checkmark & 93.59 & 91.58 \downr{-2.02} & 92.97 \downr{-0.63} & 93.43 \downr{-0.16} & 93.19 \downr{-0.41} & 93.30 \downr{-0.3} & 92.86 \downr{-0.74} & 93.70 \upg{0.11} & 94.06 \upg{0.46} & 97.36 \upg{3.76} \\ \midrule
\Checkmark &   & \Checkmark & \Checkmark &   & 89.07 & 90.16 \upg{1.09} & 90.46 \upg{1.39} & 86.73 \downr{-2.34} & 90.62 \upg{1.55} & 89.15 \upg{0.08} & 90.46 \upg{1.39} & 90.35 \upg{1.28} & 90.79 \upg{1.72} & 94.79 \upg{5.72} \\ \midrule
\Checkmark &   & \Checkmark &   & \Checkmark & 93.59 & 92.50 \downr{-1.09} & 93.08 \downr{-0.52} & 92.86 \downr{-0.74} & 93.02 \downr{-0.57} & 92.86 \downr{-0.74} & 93.19 \downr{-0.41} & 93.32 \downr{-0.27} & 94.25 \upg{0.65} & 97.03 \upg{3.43} \\ \midrule
\Checkmark &   &   & \Checkmark & \Checkmark & 93.59 & 92.97 \downr{-0.63} & 93.70 \upg{0.11} & 93.98 \upg{0.38} & 93.87 \upg{0.27} & 93.84 \upg{0.25} & 93.84 \upg{0.25} & 94.19 \upg{0.6} & 94.47 \upg{0.87} & 96.97 \upg{3.38} \\ \midrule
  & \Checkmark & \Checkmark & \Checkmark &   & 89.07 & 89.83 \upg{0.76} & 90.16 \upg{1.09} & 86.54 \downr{-2.53} & 90.65 \upg{1.58} & 89.83 \upg{0.76} & 90.41 \upg{1.34} & 90.35 \upg{1.28} & 90.71 \upg{1.64} & 95.12 \upg{6.05} \\ \midrule
  & \Checkmark & \Checkmark &   & \Checkmark & 93.59 & 92.20 \downr{-1.39} & 93.02 \downr{-0.57} & 92.56 \downr{-1.04} & 92.75 \downr{-0.84} & 92.70 \downr{-0.9} & 92.94 \downr{-0.65} & 94.00 \upg{0.41} & 93.98 \upg{0.38} & 97.27 \upg{3.68} \\ \midrule
  & \Checkmark &   & \Checkmark & \Checkmark & 93.59 & 92.91 \downr{-0.68} & 93.46 \downr{-0.14} & 93.46 \downr{-0.14} & 93.43 \downr{-0.16} & 93.57 \downr{-0.03} & 93.49 \downr{-0.11} & 93.84 \upg{0.25} & 93.92 \upg{0.33} & 97.30 \upg{3.71} \\ \midrule
  &   & \Checkmark & \Checkmark & \Checkmark & 93.59 & 92.97 \downr{-0.63} & 93.27 \downr{-0.33} & 92.89 \downr{-0.71} & 93.51 \downr{-0.08} & 93.13 \downr{-0.46} & 93.40 \downr{-0.19} & 94.03 \upg{0.44} & 94.17 \upg{0.57} & 97.03 \upg{3.43} \\ \midrule
\Checkmark & \Checkmark & \Checkmark & \Checkmark &   & 89.07 & 90.27 \upg{1.2} & 90.41 \upg{1.34} & 86.21 \downr{-2.86} & 90.57 \upg{1.5} & 89.48 \upg{0.41} & 90.54 \upg{1.47} & 90.35 \upg{1.28} & 90.90 \upg{1.83} & 95.88 \upg{6.81} \\ \midrule
\Checkmark & \Checkmark & \Checkmark &   & \Checkmark & 93.59 & 92.20 \downr{-1.39} & 92.80 \downr{-0.79} & 92.10 \downr{-1.5} & 92.91 \downr{-0.68} & 92.64 \downr{-0.95} & 92.80 \downr{-0.79} & 93.16 \downr{-0.44} & 94.17 \upg{0.57} & 97.77 \upg{4.17} \\ \midrule
\Checkmark & \Checkmark &   & \Checkmark & \Checkmark & 93.59 & 92.56 \downr{-1.04} & 92.94 \downr{-0.65} & 93.59 \samey{0.0} & 93.19 \downr{-0.41} & 93.54 \downr{-0.05} & 92.94 \downr{-0.65} & 93.81 \upg{0.22} & 94.52 \upg{0.93} & 97.77 \upg{4.17} \\ \midrule
\Checkmark &   & \Checkmark & \Checkmark & \Checkmark & 93.59 & 92.91 \downr{-0.68} & 93.27 \downr{-0.33} & 92.78 \downr{-0.82} & 93.19 \downr{-0.41} & 93.05 \downr{-0.55} & 93.27 \downr{-0.33} & 93.79 \upg{0.19} & 94.33 \upg{0.74} & 97.52 \upg{3.92} \\ \midrule
  & \Checkmark & \Checkmark & \Checkmark & \Checkmark & 93.59 & 92.61 \downr{-0.98} & 93.00 \downr{-0.6} & 92.12 \downr{-1.47} & 92.94 \downr{-0.65} & 92.91 \downr{-0.68} & 93.05 \downr{-0.55} & 93.68 \upg{0.08} & 94.03 \upg{0.44} & 97.79 \upg{4.2} \\ \midrule
\Checkmark & \Checkmark & \Checkmark & \Checkmark & \Checkmark & 93.59 & 91.63 \downr{-1.96} & 93.00 \downr{-0.59} & 92.26 \downr{-1.34} & 92.86 \downr{-0.74} & 92.89 \downr{-0.71} & 92.94 \downr{-0.65} & 93.30 \downr{-0.29} & 94.58 \upg{0.99} & 98.06 \upg{4.47} \\ \midrule

\multicolumn{5}{c}{Mean $\Delta$} & & -0.05 & 0.35 & -0.77 & 0.42 & 0.06 & 0.40 & 0.58 & 0.98 & 4.31 \\ 
\multicolumn{5}{c}{Max $\Delta$} & & 1.39 & 1.83 & 1.34 & 2.26 & 1.09 & 1.99 & 2.02 & 2.37 & 7.22 \\ 
\multicolumn{5}{c}{Min $\Delta$} & & -2.02 & -0.79 & -2.86 & -0.84 & -0.95 & -0.79 & -0.44 & 0.11 & 2.40 \\ 
\bottomrule
\end{tabular}
}
\label{tab:all_possible_combinations_pets}
\end{table*}

\begin{table*}[]
\centering
\caption{Our results on \Cars{} dataset for all the possible combinations of combining the zero-shot predictions of CLIP backbones, which we group intro non-parametric and parametric techniques. Also, the best-performing single backbone (\Best{}) and the \Oracle{} performance. We present, for each combination of backbones, the improvement \upg{}, constancy \samey{} and deterioration \downr{} of accuracy performance for each method when we compare it against the \Best{} backbone. Mean, Max, and Min $\Delta$ summarize the difference in performance across methods and backbone combinations.}
\qquad
\Cars \\
\resizebox{0.9\textwidth}{!}{

\begin{tabular}{ccccccccccccccc}
\toprule
\multicolumn{2}{c}{ResNet} & \multicolumn{3}{c}{ViT} & \multirow{2}{*}{\Best}  & \multicolumn{4}{c}{Non-Parametric} & \multicolumn{4}{c}{Parametric} & \multirow{2}{*}{\Oracle}\\
\cmidrule(lr){1-2} \cmidrule(lr){3-5} \cmidrule(lr){7-10} \cmidrule(lr){11-14}
50 & 101 & B-32 & B-16 & L-14 & & \VoteOne & \VoteThree & \Confidence & \LogitAvg & \CalibratedConfidence & \CLogitAvg & \GAC & \NNC &  \\
\midrule

\Checkmark &   &   &   &   &  \multicolumn{10}{c}{\xfill{.1em}  54.23 \samey{0.00} \xfill{.1em}} \\ 
  & \Checkmark &   &   &   & \multicolumn{10}{c}{\xfill{.1em}  61.12 \samey{0.00} \xfill{.1em}} \\ 
  &   & \Checkmark &   &   & \multicolumn{10}{c}{\xfill{.1em}  59.73 \samey{0.00} \xfill{.1em}} \\ 
  &   &   & \Checkmark &   & \multicolumn{10}{c}{\xfill{.1em}  64.61 \samey{0.00} \xfill{.1em}} \\ 
  &   &   &   & \Checkmark & \multicolumn{10}{c}{\xfill{.1em}  77.75 \samey{0.00} \xfill{.1em}} \\ \midrule
\Checkmark & \Checkmark &   &   &   & 61.12 & 62.41 \upg{1.28} & 62.67 \upg{1.54} & 53.41 \downr{-7.71} & 63.03 \upg{1.9} & 61.81 \upg{0.68} & 62.83 \upg{1.7} & 63.52 \upg{2.4} & 63.59 \upg{2.46} & 71.58 \upg{10.46} \\ \midrule
\Checkmark &   & \Checkmark &   &   & 59.73 & 61.30 \upg{1.57} & 62.26 \upg{2.52} & 60.84 \upg{1.11} & 63.39 \upg{3.66} & 61.14 \upg{1.41} & 62.52 \upg{2.79} & 63.41 \upg{3.68} & 63.40 \upg{3.67} & 71.50 \upg{11.76} \\ \midrule
\Checkmark &   &   & \Checkmark &   & 64.61 & 65.27 \upg{0.66} & 65.55 \upg{0.95} & 53.94 \downr{-10.67} & 66.01 \upg{1.41} & 64.97 \upg{0.36} & 65.89 \upg{1.28} & 66.55 \upg{1.94} & 66.80 \upg{2.19} & 73.95 \upg{9.34} \\ \midrule
\Checkmark &   &   &   & \Checkmark & 77.75 & 77.58 \downr{-0.17} & 77.76 \upg{0.01} & 77.69 \downr{-0.06} & 77.24 \downr{-0.51} & 77.47 \downr{-0.29} & 77.64 \downr{-0.11} & 78.51 \upg{0.76} & 78.57 \upg{0.82} & 83.96 \upg{6.21} \\ \midrule
  & \Checkmark & \Checkmark &   &   & 61.12 & 64.42 \upg{3.3} & 64.99 \upg{3.87} & 63.41 \upg{2.29} & 66.89 \upg{5.77} & 63.71 \upg{2.59} & 65.66 \upg{4.54} & 66.86 \upg{5.73} & 66.83 \upg{5.71} & 73.60 \upg{12.47} \\ \midrule
  & \Checkmark &   & \Checkmark &   & 64.61 & 67.16 \upg{2.55} & 67.74 \upg{3.13} & 59.28 \downr{-5.32} & 67.99 \upg{3.38} & 66.66 \upg{2.05} & 68.11 \upg{3.51} & 68.03 \upg{3.42} & 68.20 \upg{3.59} & 76.27 \upg{11.67} \\ \midrule
  & \Checkmark &   &   & \Checkmark & 77.75 & 77.47 \downr{-0.29} & 77.78 \upg{0.02} & 77.42 \downr{-0.34} & 77.73 \downr{-0.02} & 77.44 \downr{-0.31} & 78.10 \upg{0.35} & 78.92 \upg{1.17} & 79.11 \upg{1.36} & 84.32 \upg{6.57} \\ \midrule
  &   & \Checkmark & \Checkmark &   & 64.61 & 66.34 \upg{1.73} & 66.84 \upg{2.24} & 65.41 \upg{0.81} & 67.95 \upg{3.35} & 65.87 \upg{1.27} & 67.26 \upg{2.65} & 68.00 \upg{3.4} & 67.77 \upg{3.16} & 75.48 \upg{10.87} \\ \midrule
  &   & \Checkmark &   & \Checkmark & 77.75 & 77.55 \downr{-0.2} & 77.86 \upg{0.11} & 77.55 \downr{-0.2} & 77.42 \downr{-0.34} & 77.52 \downr{-0.24} & 77.80 \upg{0.05} & 78.68 \upg{0.93} & 78.86 \upg{1.11} & 83.88 \upg{6.13} \\ \midrule
  &   &   & \Checkmark & \Checkmark & 77.75 & 77.54 \downr{-0.21} & 77.45 \downr{-0.3} & 77.32 \downr{-0.44} & 77.79 \upg{0.04} & 77.18 \downr{-0.57} & 77.63 \downr{-0.12} & 78.88 \upg{1.13} & 78.91 \upg{1.16} & 84.26 \upg{6.5} \\ \midrule
\Checkmark & \Checkmark & \Checkmark &   &   & 61.12 & 64.63 \upg{3.51} & 65.23 \upg{4.1} & 57.36 \downr{-3.77} & 66.56 \upg{5.43} & 63.69 \upg{2.56} & 66.01 \upg{4.89} & 66.76 \upg{5.63} & 66.76 \upg{5.63} & 78.86 \upg{17.73} \\ \midrule
\Checkmark & \Checkmark &   & \Checkmark &   & 64.61 & 66.34 \upg{1.73} & 67.35 \upg{2.75} & 59.66 \downr{-4.95} & 67.60 \upg{3.0} & 66.65 \upg{2.04} & 67.77 \upg{3.16} & 67.95 \upg{3.35} & 68.20 \upg{3.59} & 80.66 \upg{16.06} \\ \midrule
\Checkmark & \Checkmark &   &   & \Checkmark & 77.75 & 73.80 \downr{-3.95} & 76.23 \downr{-1.52} & 76.10 \downr{-1.65} & 76.59 \downr{-1.16} & 77.25 \downr{-0.5} & 76.86 \downr{-0.9} & 79.06 \upg{1.31} & 78.88 \upg{1.13} & 87.49 \upg{9.74} \\ \midrule
\Checkmark &   & \Checkmark & \Checkmark &   & 64.61 & 66.24 \upg{1.63} & 67.06 \upg{2.45} & 56.24 \downr{-8.37} & 67.94 \upg{3.33} & 66.07 \upg{1.47} & 67.62 \upg{3.01} & 68.46 \upg{3.86} & 68.29 \upg{3.68} & 80.24 \upg{15.63} \\ \midrule
\Checkmark &   & \Checkmark &   & \Checkmark & 77.75 & 75.04 \downr{-2.71} & 77.17 \downr{-0.58} & 77.50 \downr{-0.25} & 76.71 \downr{-1.04} & 77.34 \downr{-0.41} & 77.24 \downr{-0.51} & 78.73 \upg{0.98} & 78.71 \upg{0.96} & 87.15 \upg{9.4} \\ \midrule
\Checkmark &   &   & \Checkmark & \Checkmark & 77.75 & 74.53 \downr{-3.22} & 76.28 \downr{-1.47} & 75.90 \downr{-1.85} & 76.76 \downr{-0.99} & 76.97 \downr{-0.78} & 76.94 \downr{-0.81} & 78.40 \upg{0.65} & 79.01 \upg{1.26} & 87.43 \upg{9.68} \\ \midrule
  & \Checkmark & \Checkmark & \Checkmark &   & 64.61 & 67.86 \upg{3.26} & 68.95 \upg{4.34} & 60.40 \downr{-4.2} & 69.36 \upg{4.75} & 67.04 \upg{2.44} & 69.12 \upg{4.51} & 69.16 \upg{4.55} & 69.34 \upg{4.74} & 81.37 \upg{16.76} \\ \midrule
  & \Checkmark & \Checkmark &   & \Checkmark & 77.75 & 75.04 \downr{-2.71} & 76.96 \downr{-0.8} & 77.17 \downr{-0.58} & 76.88 \downr{-0.87} & 77.18 \downr{-0.57} & 77.14 \downr{-0.61} & 79.24 \upg{1.49} & 79.26 \upg{1.5} & 87.32 \upg{9.56} \\ \midrule
  & \Checkmark &   & \Checkmark & \Checkmark & 77.75 & 75.60 \downr{-2.15} & 76.55 \downr{-1.21} & 75.70 \downr{-2.05} & 77.25 \downr{-0.5} & 77.15 \downr{-0.6} & 77.37 \downr{-0.39} & 79.02 \upg{1.27} & 79.06 \upg{1.31} & 87.63 \upg{9.87} \\ \midrule
  &   & \Checkmark & \Checkmark & \Checkmark & 77.75 & 75.23 \downr{-2.52} & 76.71 \downr{-1.04} & 77.28 \downr{-0.47} & 77.35 \downr{-0.4} & 77.19 \downr{-0.56} & 77.15 \downr{-0.6} & 78.42 \upg{0.67} & 79.04 \upg{1.29} & 87.39 \upg{9.64} \\ \midrule
\Checkmark & \Checkmark & \Checkmark & \Checkmark &   & 64.61 & 67.72 \upg{3.11} & 68.54 \upg{3.93} & 60.59 \downr{-4.02} & 68.77 \upg{4.17} & 66.98 \upg{2.38} & 68.56 \upg{3.95} & 69.06 \upg{4.45} & 69.16 \upg{4.55} & 84.18 \upg{19.57} \\ \midrule
\Checkmark & \Checkmark & \Checkmark &   & \Checkmark & 77.75 & 73.81 \downr{-3.94} & 75.87 \downr{-1.88} & 75.97 \downr{-1.78} & 76.04 \downr{-1.72} & 77.08 \downr{-0.67} & 76.43 \downr{-1.32} & 78.56 \upg{0.81} & 79.12 \upg{1.37} & 89.37 \upg{11.62} \\ \midrule
\Checkmark & \Checkmark &   & \Checkmark & \Checkmark & 77.75 & 74.69 \downr{-3.06} & 76.06 \downr{-1.69} & 75.66 \downr{-2.09} & 76.41 \downr{-1.34} & 76.93 \downr{-0.82} & 76.47 \downr{-1.28} & 79.07 \upg{1.32} & 79.22 \upg{1.47} & 89.60 \upg{11.85} \\ \midrule
\Checkmark &   & \Checkmark & \Checkmark & \Checkmark & 77.75 & 74.51 \downr{-3.25} & 75.95 \downr{-1.8} & 75.90 \downr{-1.85} & 76.57 \downr{-1.18} & 77.01 \downr{-0.75} & 76.05 \downr{-1.7} & 78.83 \upg{1.08} & 79.19 \upg{1.44} & 89.30 \upg{11.55} \\ \midrule
  & \Checkmark & \Checkmark & \Checkmark & \Checkmark & 77.75 & 75.28 \downr{-2.47} & 76.48 \downr{-1.27} & 75.54 \downr{-2.21} & 76.61 \downr{-1.14} & 76.99 \downr{-0.76} & 76.91 \downr{-0.85} & 79.07 \upg{1.32} & 79.33 \upg{1.58} & 89.49 \upg{11.74} \\ \midrule
\Checkmark & \Checkmark & \Checkmark & \Checkmark & \Checkmark & 77.75 & 73.67 \downr{-4.08} & 75.70 \downr{-2.05} & 75.56 \downr{-2.19} & 75.75 \downr{-2.0} & 76.78 \downr{-0.97} & 75.96 \downr{-1.79} & 78.96 \upg{1.21} & 80.30 \upg{2.55} & 90.85 \upg{13.1} \\ \midrule

\multicolumn{5}{c}{Mean $\Delta$} & & -0.41 & 0.63 & -2.42 & 1.04 & 0.40 & 0.98 & 2.25 & 2.43 & 11.36 \\ 
\multicolumn{5}{c}{Max $\Delta$} & & 3.51 & 4.34 & 2.29 & 5.77 & 2.59 & 4.89 & 5.73 & 5.71 & 19.57 \\ 
\multicolumn{5}{c}{Min $\Delta$} & & -4.08 & -2.05 & -10.67 & -2.00 & -0.97 & -1.79 & 0.65 & 0.82 & 6.13 \\ 
\bottomrule
\end{tabular}
}
\label{tab:all_possible_combinations_cars}
\end{table*}

\begin{table*}[]
\centering
\caption{Our results on \Cub{} dataset for all the possible combinations of combining the zero-shot predictions of CLIP backbones, which we group intro non-parametric and parametric techniques. Also, the best-performing single backbone (\Best{}) and the \Oracle{} performance. We present, for each combination of backbones, the improvement \upg{}, constancy \samey{} and deterioration \downr{} of accuracy performance for each method when we compare it against the \Best{} backbone. Mean, Max, and Min $\Delta$ summarize the difference in performance across methods and backbone combinations.}
\Cub \\
\qquad
\resizebox{0.9\textwidth}{!}{

\begin{tabular}{ccccccccccccccc}
\toprule
\multicolumn{2}{c}{ResNet} & \multicolumn{3}{c}{ViT} & \multirow{2}{*}{\Best}  & \multicolumn{4}{c}{Non-Parametric} & \multicolumn{4}{c}{Parametric} & \multirow{2}{*}{\Oracle}\\
\cmidrule(lr){1-2} \cmidrule(lr){3-5} \cmidrule(lr){7-10} \cmidrule(lr){11-14}
50 & 101 & B-32 & B-16 & L-14 & & \VoteOne & \VoteThree & \Confidence & \LogitAvg & \CalibratedConfidence & \CLogitAvg & \GAC & \NNC &  \\
\midrule

\Checkmark &   &   &   &   &  \multicolumn{10}{c}{\xfill{.1em} 46.57 \samey{0.00} \xfill{.1em}} \\ 
  & \Checkmark &   &   &   & \multicolumn{10}{c}{\xfill{.1em}  49.64 \samey{0.00} \xfill{.1em}} \\ 
  &   & \Checkmark &   &   & \multicolumn{10}{c}{\xfill{.1em}  52.99 \samey{0.00} \xfill{.1em}} \\ 
  &   &   & \Checkmark &   & \multicolumn{10}{c}{\xfill{.1em}  55.28 \samey{0.00} \xfill{.1em}} \\ 
  &   &   &   & \Checkmark & \multicolumn{10}{c}{\xfill{.1em}  62.06 \samey{0.00} \xfill{.1em}} \\ \midrule
\Checkmark & \Checkmark &   &   &   & 49.64 & 51.28 \upg{1.64} & 52.33 \upg{2.69} & 45.41 \downr{-4.23} & 55.06 \upg{5.42} & 50.72 \upg{1.09} & 53.56 \upg{3.92} & 55.06 \upg{5.42} & 55.13 \upg{5.49} & 61.13 \upg{11.49} \\ \midrule
\Checkmark &   & \Checkmark &   &   & 52.99 & 54.92 \upg{1.93} & 55.70 \upg{2.71} & 53.94 \upg{0.95} & 57.77 \upg{4.78} & 53.90 \upg{0.91} & 56.51 \upg{3.52} & 57.68 \upg{4.69} & 57.75 \upg{4.76} & 64.03 \upg{11.05} \\ \midrule
\Checkmark &   &   & \Checkmark &   & 55.28 & 56.61 \upg{1.33} & 57.51 \upg{2.23} & 55.82 \upg{0.54} & 59.39 \upg{4.11} & 55.49 \upg{0.21} & 58.44 \upg{3.16} & 59.73 \upg{4.45} & 59.68 \upg{4.4} & 65.52 \upg{10.23} \\ \midrule
\Checkmark &   &   &   & \Checkmark & 62.06 & 61.65 \downr{-0.41} & 62.32 \upg{0.26} & 61.72 \downr{-0.35} & 63.96 \upg{1.9} & 61.01 \downr{-1.05} & 62.77 \upg{0.71} & 64.12 \upg{2.05} & 64.26 \upg{2.19} & 70.49 \upg{8.42} \\ \midrule
  & \Checkmark & \Checkmark &   &   & 52.99 & 54.94 \upg{1.95} & 55.49 \upg{2.5} & 48.72 \downr{-4.26} & 57.61 \upg{4.63} & 54.04 \upg{1.05} & 56.51 \upg{3.52} & 57.42 \upg{4.44} & 57.58 \upg{4.59} & 63.82 \upg{10.84} \\ \midrule
  & \Checkmark &   & \Checkmark &   & 55.28 & 56.56 \upg{1.28} & 57.16 \upg{1.88} & 56.25 \upg{0.97} & 58.78 \upg{3.5} & 56.27 \upg{0.98} & 58.08 \upg{2.8} & 58.80 \upg{3.52} & 59.25 \upg{3.97} & 65.34 \upg{10.06} \\ \midrule
  & \Checkmark &   &   & \Checkmark & 62.06 & 62.01 \downr{-0.05} & 62.70 \upg{0.64} & 62.03 \downr{-0.03} & 63.58 \upg{1.52} & 61.67 \downr{-0.4} & 63.24 \upg{1.17} & 63.69 \upg{1.62} & 63.93 \upg{1.86} & 70.31 \upg{8.25} \\ \midrule
  &   & \Checkmark & \Checkmark &   & 55.28 & 57.87 \upg{2.59} & 58.58 \upg{3.3} & 51.05 \downr{-4.23} & 60.87 \upg{5.59} & 57.40 \upg{2.12} & 59.63 \upg{4.35} & 60.72 \upg{5.44} & 60.74 \upg{5.45} & 67.10 \upg{11.82} \\ \midrule
  &   & \Checkmark &   & \Checkmark & 62.06 & 62.58 \upg{0.52} & 63.29 \upg{1.23} & 62.62 \upg{0.55} & 64.34 \upg{2.28} & 62.32 \upg{0.26} & 64.07 \upg{2.0} & 64.58 \upg{2.52} & 64.50 \upg{2.43} & 71.33 \upg{9.27} \\ \midrule
  &   &   & \Checkmark & \Checkmark & 62.06 & 62.63 \upg{0.57} & 63.36 \upg{1.29} & 54.87 \downr{-7.2} & 64.98 \upg{2.92} & 62.29 \upg{0.22} & 63.91 \upg{1.85} & 65.07 \upg{3.0} & 65.00 \upg{2.93} & 71.18 \upg{9.11} \\ \midrule
\Checkmark & \Checkmark & \Checkmark &   &   & 52.99 & 56.25 \upg{3.26} & 57.21 \upg{4.23} & 47.93 \downr{-5.06} & 59.32 \upg{6.33} & 54.50 \upg{1.52} & 58.18 \upg{5.2} & 59.60 \upg{6.61} & 59.25 \upg{6.27} & 69.71 \upg{16.72} \\ \midrule
\Checkmark & \Checkmark &   & \Checkmark &   & 55.28 & 57.34 \upg{2.05} & 58.42 \upg{3.14} & 53.31 \downr{-1.97} & 60.20 \upg{4.92} & 56.08 \upg{0.79} & 59.42 \upg{4.14} & 60.58 \upg{5.3} & 60.68 \upg{5.4} & 70.90 \upg{15.62} \\ \midrule
\Checkmark & \Checkmark &   &   & \Checkmark & 62.06 & 61.25 \downr{-0.81} & 62.81 \upg{0.74} & 59.89 \downr{-2.17} & 64.01 \upg{1.95} & 60.80 \downr{-1.26} & 63.41 \upg{1.35} & 65.08 \upg{3.02} & 65.03 \upg{2.97} & 74.75 \upg{12.69} \\ \midrule
\Checkmark &   & \Checkmark & \Checkmark &   & 55.28 & 58.94 \upg{3.66} & 60.11 \upg{4.83} & 51.86 \downr{-3.42} & 61.48 \upg{6.2} & 57.46 \upg{2.17} & 60.99 \upg{5.71} & 61.22 \upg{5.94} & 62.01 \upg{6.73} & 72.64 \upg{17.36} \\ \midrule
\Checkmark &   & \Checkmark &   & \Checkmark & 62.06 & 61.96 \downr{-0.1} & 63.88 \upg{1.81} & 62.39 \upg{0.33} & 65.36 \upg{3.3} & 61.60 \downr{-0.47} & 64.36 \upg{2.3} & 65.67 \upg{3.61} & 65.64 \upg{3.57} & 75.85 \upg{13.79} \\ \midrule
\Checkmark &   &   & \Checkmark & \Checkmark & 62.06 & 62.67 \upg{0.6} & 64.19 \upg{2.12} & 55.44 \downr{-6.63} & 65.67 \upg{3.61} & 61.72 \downr{-0.35} & 64.96 \upg{2.9} & 66.02 \upg{3.95} & 66.34 \upg{4.28} & 76.04 \upg{13.98} \\ \midrule
  & \Checkmark & \Checkmark & \Checkmark &   & 55.28 & 58.97 \upg{3.69} & 59.92 \upg{4.64} & 49.83 \downr{-5.45} & 61.24 \upg{5.95} & 57.73 \upg{2.45} & 60.68 \upg{5.4} & 61.18 \upg{5.9} & 61.65 \upg{6.37} & 71.99 \upg{16.71} \\ \midrule
  & \Checkmark & \Checkmark &   & \Checkmark & 62.06 & 62.15 \upg{0.09} & 63.44 \upg{1.38} & 60.25 \downr{-1.81} & 65.00 \upg{2.93} & 61.82 \downr{-0.24} & 63.98 \upg{1.92} & 65.36 \upg{3.3} & 65.31 \upg{3.24} & 75.42 \upg{13.36} \\ \midrule
  & \Checkmark &   & \Checkmark & \Checkmark & 62.06 & 62.43 \upg{0.36} & 64.03 \upg{1.97} & 55.47 \downr{-6.59} & 65.33 \upg{3.26} & 62.12 \upg{0.05} & 64.91 \upg{2.85} & 66.24 \upg{4.18} & 66.05 \upg{3.99} & 75.54 \upg{13.48} \\ \midrule
  &   & \Checkmark & \Checkmark & \Checkmark & 62.06 & 63.12 \upg{1.05} & 64.46 \upg{2.4} & 54.45 \downr{-7.61} & 65.93 \upg{3.87} & 62.77 \upg{0.71} & 64.79 \upg{2.73} & 66.05 \upg{3.99} & 66.05 \upg{3.99} & 76.60 \upg{14.53} \\ \midrule
\Checkmark & \Checkmark & \Checkmark & \Checkmark &   & 55.28 & 59.44 \upg{4.16} & 60.61 \upg{5.33} & 49.57 \downr{-5.71} & 61.81 \upg{6.52} & 57.49 \upg{2.21} & 61.39 \upg{6.11} & 61.96 \upg{6.68} & 62.50 \upg{7.21} & 75.60 \upg{20.31} \\ \midrule
\Checkmark & \Checkmark & \Checkmark &   & \Checkmark & 62.06 & 62.00 \downr{-0.07} & 63.62 \upg{1.55} & 59.37 \downr{-2.69} & 64.96 \upg{2.9} & 61.22 \downr{-0.85} & 64.10 \upg{2.04} & 65.38 \upg{3.31} & 65.91 \upg{3.85} & 78.32 \upg{16.26} \\ \midrule
\Checkmark & \Checkmark &   & \Checkmark & \Checkmark & 62.06 & 62.98 \upg{0.91} & 64.41 \upg{2.35} & 54.45 \downr{-7.61} & 65.22 \upg{3.16} & 61.49 \downr{-0.57} & 64.65 \upg{2.59} & 66.66 \upg{4.59} & 66.59 \upg{4.52} & 78.58 \upg{16.52} \\ \midrule
\Checkmark &   & \Checkmark & \Checkmark & \Checkmark & 62.06 & 63.63 \upg{1.57} & 65.15 \upg{3.09} & 54.73 \downr{-7.34} & 65.96 \upg{3.9} & 62.20 \upg{0.14} & 65.59 \upg{3.52} & 66.21 \upg{4.14} & 66.90 \upg{4.83} & 79.67 \upg{17.6} \\ \midrule
  & \Checkmark & \Checkmark & \Checkmark & \Checkmark & 62.06 & 63.19 \upg{1.12} & 64.38 \upg{2.31} & 54.18 \downr{-7.89} & 65.69 \upg{3.62} & 62.29 \upg{0.22} & 65.21 \upg{3.14} & 66.28 \upg{4.21} & 66.57 \upg{4.5} & 79.01 \upg{16.95} \\ \midrule
\Checkmark & \Checkmark & \Checkmark & \Checkmark & \Checkmark & 62.06 & 63.32 \upg{1.26} & 64.38 \upg{2.31} & 53.71 \downr{-8.35} & 65.53 \upg{3.47} & 61.79 \downr{-0.28} & 65.01 \upg{2.95} & 66.29 \upg{4.23} & 68.40 \upg{6.34} & 81.20 \upg{19.14} \\ \midrule

\multicolumn{5}{c}{Mean $\Delta$} & & 1.31 & 2.42 & -3.74 & 3.94 & 0.45 & 3.15 & 4.24 & 4.47 & 13.68 \\ 
\multicolumn{5}{c}{Max $\Delta$} & & 4.16 & 5.33 & 0.97 & 6.52 & 2.45 & 6.11 & 6.68 & 7.21 & 20.31 \\ 
\multicolumn{5}{c}{Min $\Delta$} & & -0.81 & 0.26 & -8.35 & 1.52 & -1.26 & 0.71 & 1.62 & 1.86 & 8.25 \\ 
\bottomrule
\end{tabular}
}
\label{tab:all_possible_combinations_cub}
\end{table*}

\begin{table*}[t]
\centering
\caption{Our results on \Dtd{} dataset for all the possible combinations of combining the zero-shot predictions of CLIP backbones, which we group intro non-parametric and parametric techniques. Also, the best-performing single backbone (\Best{}) and the \Oracle{} performance. We present, for each combination of backbones, the improvement \upg{}, constancy \samey{} and deterioration \downr{} of accuracy performance for each method when we compare it against the \Best{} backbone. Mean, Max, and Min $\Delta$ summarize the difference in performance across methods and backbone combinations.}
\Dtd \\
\qquad
\resizebox{0.9\textwidth}{!}{

\begin{tabular}{ccccccccccccccc}
\toprule
\multicolumn{2}{c}{ResNet} & \multicolumn{3}{c}{ViT} & \multirow{2}{*}{\Best}  & \multicolumn{4}{c}{Non-Parametric} & \multicolumn{4}{c}{Parametric} & \multirow{2}{*}{\Oracle}\\
\cmidrule(lr){1-2} \cmidrule(lr){3-5} \cmidrule(lr){7-10} \cmidrule(lr){11-14}
50 & 101 & B-32 & B-16 & L-14 & & \VoteOne & \VoteThree & \Confidence & \LogitAvg & \CalibratedConfidence & \CLogitAvg & \GAC & \NNC &  \\
\midrule

\Checkmark &   &   &   &   &  \multicolumn{10}{c}{\xfill{.1em} 41.22 \samey{0.00} \xfill{.1em}} \\ 
  & \Checkmark &   &   &   & \multicolumn{10}{c}{\xfill{.1em}  43.67 \samey{0.00} \xfill{.1em}} \\ 
  &   & \Checkmark &   &   & \multicolumn{10}{c}{\xfill{.1em}  43.99 \samey{0.00} \xfill{.1em}} \\ 
  &   &   & \Checkmark &   & \multicolumn{10}{c}{\xfill{.1em}  45.11 \samey{0.00} \xfill{.1em}} \\ 
  &   &   &   & \Checkmark & \multicolumn{10}{c}{\xfill{.1em}  55.32 \samey{0.00} \xfill{.1em}} \\ \midrule
\Checkmark & \Checkmark &   &   &   & 43.67 & 47.13 \upg{3.46} & 47.29 \upg{3.62} & 46.44 \upg{2.77} & 47.71 \upg{4.04} & 45.96 \upg{2.29} & 47.34 \upg{3.67} & 47.61 \upg{3.94} & 47.71 \upg{4.04} & 54.36 \upg{10.69} \\ \midrule
\Checkmark &   & \Checkmark &   &   & 43.99 & 44.63 \upg{0.64} & 45.00 \upg{1.01} & 41.22 \downr{-2.77} & 46.01 \upg{2.02} & 43.78 \downr{-0.21} & 45.43 \upg{1.44} & 45.74 \upg{1.76} & 45.90 \upg{1.91} & 51.70 \upg{7.71} \\ \midrule
\Checkmark &   &   & \Checkmark &   & 45.11 & 46.86 \upg{1.76} & 47.13 \upg{2.02} & 40.37 \downr{-4.73} & 47.39 \upg{2.29} & 45.85 \upg{0.74} & 47.23 \upg{2.13} & 47.87 \upg{2.77} & 47.39 \upg{2.29} & 53.24 \upg{8.14} \\ \midrule
\Checkmark &   &   &   & \Checkmark & 55.32 & 55.64 \upg{0.32} & 56.12 \upg{0.8} & 41.33 \downr{-13.99} & 55.64 \upg{0.32} & 55.05 \downr{-0.27} & 56.22 \upg{0.9} & 55.43 \upg{0.11} & 56.44 \upg{1.12} & 62.23 \upg{6.91} \\ \midrule
  & \Checkmark & \Checkmark &   &   & 43.99 & 47.02 \upg{3.03} & 47.34 \upg{3.35} & 41.60 \downr{-2.39} & 48.46 \upg{4.47} & 47.13 \upg{3.14} & 47.93 \upg{3.94} & 47.87 \upg{3.88} & 47.98 \upg{3.99} & 55.11 \upg{11.12} \\ \midrule
  & \Checkmark &   & \Checkmark &   & 45.11 & 48.46 \upg{3.35} & 48.94 \upg{3.83} & 48.30 \upg{3.19} & 49.15 \upg{4.04} & 48.35 \upg{3.24} & 48.67 \upg{3.56} & 48.78 \upg{3.67} & 48.78 \upg{3.67} & 55.74 \upg{10.64} \\ \midrule
  & \Checkmark &   &   & \Checkmark & 55.32 & 55.21 \downr{-0.11} & 55.59 \upg{0.27} & 43.35 \downr{-11.97} & 55.69 \upg{0.37} & 55.11 \downr{-0.21} & 56.33 \upg{1.01} & 56.76 \upg{1.44} & 56.70 \upg{1.38} & 61.60 \upg{6.28} \\ \midrule
  &   & \Checkmark & \Checkmark &   & 45.11 & 46.65 \upg{1.54} & 46.86 \upg{1.76} & 43.24 \downr{-1.86} & 47.71 \upg{2.61} & 45.48 \upg{0.37} & 47.23 \upg{2.13} & 47.07 \upg{1.97} & 47.61 \upg{2.5} & 53.56 \upg{8.46} \\ \midrule
  &   & \Checkmark &   & \Checkmark & 55.32 & 55.74 \upg{0.43} & 56.33 \upg{1.01} & 43.56 \downr{-11.76} & 56.22 \upg{0.9} & 55.32 \samey{0.0} & 56.22 \upg{0.9} & 56.49 \upg{1.17} & 56.70 \upg{1.38} & 62.02 \upg{6.7} \\ \midrule
  &   &   & \Checkmark & \Checkmark & 55.32 & 57.02 \upg{1.7} & 57.39 \upg{2.07} & 43.88 \downr{-11.44} & 56.28 \upg{0.96} & 56.60 \upg{1.28} & 56.76 \upg{1.44} & 56.81 \upg{1.49} & 56.86 \upg{1.54} & 62.02 \upg{6.7} \\ \midrule
\Checkmark & \Checkmark & \Checkmark &   &   & 43.99 & 47.50 \upg{3.51} & 47.71 \upg{3.72} & 40.37 \downr{-3.62} & 48.99 \upg{5.0} & 46.97 \upg{2.98} & 47.82 \upg{3.83} & 49.26 \upg{5.27} & 49.10 \upg{5.11} & 59.26 \upg{15.27} \\ \midrule
\Checkmark & \Checkmark &   & \Checkmark &   & 45.11 & 48.35 \upg{3.24} & 49.41 \upg{4.31} & 44.68 \downr{-0.43} & 49.95 \upg{4.84} & 48.30 \upg{3.19} & 49.95 \upg{4.84} & 50.32 \upg{5.21} & 50.21 \upg{5.11} & 60.59 \upg{15.48} \\ \midrule
\Checkmark & \Checkmark &   &   & \Checkmark & 55.32 & 54.26 \downr{-1.06} & 56.22 \upg{0.9} & 41.70 \downr{-13.62} & 55.32 \samey{0.0} & 54.89 \downr{-0.43} & 56.76 \upg{1.44} & 57.02 \upg{1.7} & 57.18 \upg{1.86} & 65.80 \upg{10.48} \\ \midrule
\Checkmark &   & \Checkmark & \Checkmark &   & 45.11 & 46.86 \upg{1.76} & 46.97 \upg{1.86} & 44.95 \downr{-0.16} & 48.19 \upg{3.09} & 45.53 \upg{0.43} & 47.34 \upg{2.23} & 47.77 \upg{2.66} & 48.24 \upg{3.14} & 57.87 \upg{12.77} \\ \midrule
\Checkmark &   & \Checkmark &   & \Checkmark & 55.32 & 53.19 \downr{-2.13} & 55.43 \upg{0.11} & 41.86 \downr{-13.46} & 54.73 \downr{-0.59} & 55.11 \downr{-0.21} & 54.73 \downr{-0.59} & 56.06 \upg{0.74} & 56.49 \upg{1.17} & 65.59 \upg{10.27} \\ \midrule
\Checkmark &   &   & \Checkmark & \Checkmark & 55.32 & 53.99 \downr{-1.33} & 56.33 \upg{1.01} & 42.87 \downr{-12.45} & 56.01 \upg{0.69} & 55.80 \upg{0.48} & 56.60 \upg{1.28} & 57.34 \upg{2.02} & 57.18 \upg{1.86} & 65.69 \upg{10.37} \\ \midrule
  & \Checkmark & \Checkmark & \Checkmark &   & 45.11 & 48.40 \upg{3.3} & 49.68 \upg{4.57} & 41.76 \downr{-3.35} & 49.26 \upg{4.15} & 48.62 \upg{3.51} & 49.68 \upg{4.57} & 49.95 \upg{4.84} & 49.20 \upg{4.1} & 60.32 \upg{15.21} \\ \midrule
  & \Checkmark & \Checkmark &   & \Checkmark & 55.32 & 54.47 \downr{-0.85} & 55.69 \upg{0.37} & 44.57 \downr{-10.74} & 55.27 \downr{-0.05} & 55.32 \samey{0.0} & 56.01 \upg{0.69} & 56.86 \upg{1.54} & 57.18 \upg{1.86} & 65.32 \upg{10.0} \\ \midrule
  & \Checkmark &   & \Checkmark & \Checkmark & 55.32 & 54.89 \downr{-0.43} & 56.76 \upg{1.44} & 43.56 \downr{-11.76} & 55.74 \upg{0.43} & 56.22 \upg{0.9} & 56.81 \upg{1.49} & 56.70 \upg{1.38} & 57.77 \upg{2.45} & 65.37 \upg{10.05} \\ \midrule
  &   & \Checkmark & \Checkmark & \Checkmark & 55.32 & 54.63 \downr{-0.69} & 56.44 \upg{1.12} & 44.47 \downr{-10.85} & 55.74 \upg{0.43} & 55.74 \upg{0.43} & 56.49 \upg{1.17} & 57.23 \upg{1.91} & 57.07 \upg{1.76} & 65.32 \upg{10.0} \\ \midrule
\Checkmark & \Checkmark & \Checkmark & \Checkmark &   & 45.11 & 48.67 \upg{3.56} & 49.52 \upg{4.41} & 43.30 \downr{-1.81} & 49.63 \upg{4.52} & 48.46 \upg{3.35} & 49.79 \upg{4.68} & 50.27 \upg{5.16} & 49.41 \upg{4.31} & 63.19 \upg{18.09} \\ \midrule
\Checkmark & \Checkmark & \Checkmark &   & \Checkmark & 55.32 & 54.36 \downr{-0.96} & 55.16 \downr{-0.16} & 42.50 \downr{-12.82} & 54.73 \downr{-0.59} & 55.00 \downr{-0.32} & 54.95 \downr{-0.37} & 56.22 \upg{0.9} & 57.34 \upg{2.02} & 68.03 \upg{12.71} \\ \midrule
\Checkmark & \Checkmark &   & \Checkmark & \Checkmark & 55.32 & 54.89 \downr{-0.43} & 56.38 \upg{1.06} & 42.93 \downr{-12.39} & 55.00 \downr{-0.32} & 55.53 \upg{0.21} & 56.81 \upg{1.49} & 57.87 \upg{2.55} & 57.71 \upg{2.39} & 68.14 \upg{12.82} \\ \midrule
\Checkmark &   & \Checkmark & \Checkmark & \Checkmark & 55.32 & 53.72 \downr{-1.6} & 55.96 \upg{0.64} & 42.50 \downr{-12.82} & 54.57 \downr{-0.74} & 55.48 \upg{0.16} & 55.48 \upg{0.16} & 56.38 \upg{1.06} & 56.97 \upg{1.65} & 67.71 \upg{12.39} \\ \midrule
  & \Checkmark & \Checkmark & \Checkmark & \Checkmark & 55.32 & 54.68 \downr{-0.64} & 56.12 \upg{0.8} & 44.47 \downr{-10.85} & 55.37 \upg{0.05} & 55.96 \upg{0.64} & 56.06 \upg{0.74} & 56.81 \upg{1.49} & 57.50 \upg{2.18} & 67.66 \upg{12.34} \\ \midrule
\Checkmark & \Checkmark & \Checkmark & \Checkmark & \Checkmark & 55.32 & 53.72 \downr{-1.6} & 55.69 \upg{0.37} & 42.55 \downr{-12.77} & 54.36 \downr{-0.96} & 55.59 \upg{0.27} & 55.21 \downr{-0.11} & 56.12 \upg{0.8} & 58.94 \upg{3.62} & 69.63 \upg{14.31} \\ \midrule

\multicolumn{5}{c}{Mean $\Delta$} & & 0.76 & 1.78 & -7.65 & 1.61 & 1.00 & 1.87 & 2.36 & 2.63 & 11.00 \\ 
\multicolumn{5}{c}{Max $\Delta$} & & 3.56 & 4.57 & 3.19 & 5.00 & 3.51 & 4.84 & 5.27 & 5.11 & 18.09 \\ 
\multicolumn{5}{c}{Min $\Delta$} & & -2.13 & -0.16 & -13.99 & -0.96 & -0.43 & -0.59 & 0.11 & 1.12 & 6.28 \\ 
\bottomrule
\end{tabular}
}
\label{tab:all_possible_combinations_dtd}
\end{table*}

\begin{table*}[]
\centering
\caption{Our results on \FGVC{} dataset for all the possible combinations of combining the zero-shot predictions of CLIP backbones, which we group intro non-parametric and parametric techniques. Also, the best-performing single backbone (\Best{}) and the \Oracle{} performance. We present, for each combination of backbones, the improvement \upg{}, constancy \samey{} and deterioration \downr{} of accuracy performance for each method when we compare it against the \Best{} backbone. Mean, Max, and Min $\Delta$ summarize the difference in performance across methods and backbone combinations.}
\FGVC \\
\qquad
\resizebox{0.9\textwidth}{!}{

\begin{tabular}{ccccccccccccccc}
\toprule
\multicolumn{2}{c}{ResNet} & \multicolumn{3}{c}{ViT} & \multirow{2}{*}{\Best}  & \multicolumn{4}{c}{Non-Parametric} & \multicolumn{4}{c}{Parametric} & \multirow{2}{*}{\Oracle}\\
\cmidrule(lr){1-2} \cmidrule(lr){3-5} \cmidrule(lr){7-10} \cmidrule(lr){11-14}
50 & 101 & B-32 & B-16 & L-14 & & \VoteOne & \VoteThree & \Confidence & \LogitAvg & \CalibratedConfidence & \CLogitAvg & \GAC & \NNC &  \\
\midrule

\Checkmark &   &   &   &   &  \multicolumn{10}{c}{\xfill{.1em} 17.07 \samey{0.00} \xfill{.1em}} \\ 
  & \Checkmark &   &   &   & \multicolumn{10}{c}{\xfill{.1em}  18.63 \samey{0.00} \xfill{.1em}} \\ 
  &   & \Checkmark &   &   & \multicolumn{10}{c}{\xfill{.1em}  19.65 \samey{0.00} \xfill{.1em}} \\ 
  &   &   & \Checkmark &   & \multicolumn{10}{c}{\xfill{.1em}  24.39 \samey{0.00} \xfill{.1em}} \\ 
  &   &   &   & \Checkmark & \multicolumn{10}{c}{\xfill{.1em}  31.71 \samey{0.00} \xfill{.1em}} \\ \midrule
\Checkmark & \Checkmark &   &   &   & 18.63 & 18.87 \upg{0.24} & 19.02 \upg{0.39} & 18.60 \downr{-0.03} & 19.23 \upg{0.6} & 18.48 \downr{-0.15} & 19.05 \upg{0.42} & 19.59 \upg{0.96} & 19.89 \upg{1.26} & 26.19 \upg{7.56} \\ \midrule
\Checkmark &   & \Checkmark &   &   & 19.65 & 20.52 \upg{0.87} & 20.46 \upg{0.81} & 16.62 \downr{-3.03} & 21.36 \upg{1.71} & 19.80 \upg{0.15} & 21.18 \upg{1.53} & 21.00 \upg{1.35} & 22.17 \upg{2.52} & 27.69 \upg{8.04} \\ \midrule
\Checkmark &   &   & \Checkmark &   & 24.39 & 24.24 \downr{-0.15} & 24.33 \downr{-0.06} & 17.19 \downr{-7.2} & 24.78 \upg{0.39} & 23.97 \downr{-0.42} & 25.23 \upg{0.84} & 25.65 \upg{1.26} & 25.89 \upg{1.5} & 31.02 \upg{6.63} \\ \midrule
\Checkmark &   &   &   & \Checkmark & 31.71 & 31.47 \downr{-0.24} & 31.47 \downr{-0.24} & 31.47 \downr{-0.24} & 31.68 \downr{-0.03} & 31.59 \downr{-0.12} & 32.55 \upg{0.84} & 32.49 \upg{0.78} & 33.57 \upg{1.86} & 37.59 \upg{5.88} \\ \midrule
  & \Checkmark & \Checkmark &   &   & 19.65 & 21.00 \upg{1.35} & 21.51 \upg{1.86} & 20.70 \upg{1.05} & 21.69 \upg{2.04} & 20.73 \upg{1.08} & 22.02 \upg{2.37} & 21.21 \upg{1.56} & 22.41 \upg{2.76} & 28.20 \upg{8.55} \\ \midrule
  & \Checkmark &   & \Checkmark &   & 24.39 & 23.67 \downr{-0.72} & 23.97 \downr{-0.42} & 23.52 \downr{-0.87} & 23.82 \downr{-0.57} & 23.79 \downr{-0.6} & 24.36 \downr{-0.03} & 24.81 \upg{0.42} & 25.38 \upg{0.99} & 31.92 \upg{7.53} \\ \midrule
  & \Checkmark &   &   & \Checkmark & 31.71 & 30.81 \downr{-0.9} & 31.29 \downr{-0.42} & 30.93 \downr{-0.78} & 30.54 \downr{-1.17} & 31.20 \downr{-0.51} & 31.50 \downr{-0.21} & 31.35 \downr{-0.36} & 33.03 \upg{1.32} & 38.31 \upg{6.6} \\ \midrule
  &   & \Checkmark & \Checkmark &   & 24.39 & 24.18 \downr{-0.21} & 24.30 \downr{-0.09} & 20.22 \downr{-4.17} & 25.08 \upg{0.69} & 23.76 \downr{-0.63} & 24.84 \upg{0.45} & 25.02 \upg{0.63} & 25.95 \upg{1.56} & 32.73 \upg{8.34} \\ \midrule
  &   & \Checkmark &   & \Checkmark & 31.71 & 31.86 \upg{0.15} & 31.68 \downr{-0.03} & 31.59 \downr{-0.12} & 32.04 \upg{0.33} & 31.71 \samey{-0.0} & 32.79 \upg{1.08} & 32.61 \upg{0.9} & 33.33 \upg{1.62} & 39.57 \upg{7.86} \\ \midrule
  &   &   & \Checkmark & \Checkmark & 31.71 & 31.62 \downr{-0.09} & 32.19 \upg{0.48} & 31.95 \upg{0.24} & 32.76 \upg{1.05} & 31.92 \upg{0.21} & 32.88 \upg{1.17} & 33.30 \upg{1.59} & 33.69 \upg{1.98} & 41.25 \upg{9.54} \\ \midrule
\Checkmark & \Checkmark & \Checkmark &   &   & 19.65 & 20.79 \upg{1.14} & 21.51 \upg{1.86} & 18.12 \downr{-1.53} & 21.75 \upg{2.1} & 20.52 \upg{0.87} & 21.69 \upg{2.04} & 21.72 \upg{2.07} & 23.49 \upg{3.84} & 33.60 \upg{13.95} \\ \midrule
\Checkmark & \Checkmark &   & \Checkmark &   & 24.39 & 22.89 \downr{-1.5} & 23.28 \downr{-1.11} & 17.55 \downr{-6.84} & 23.67 \downr{-0.72} & 23.40 \downr{-0.99} & 24.03 \downr{-0.36} & 24.69 \upg{0.3} & 26.61 \upg{2.22} & 36.57 \upg{12.18} \\ \midrule
\Checkmark & \Checkmark &   &   & \Checkmark & 31.71 & 28.80 \downr{-2.91} & 30.00 \downr{-1.71} & 30.87 \downr{-0.84} & 30.21 \downr{-1.5} & 31.14 \downr{-0.57} & 31.53 \downr{-0.18} & 32.58 \upg{0.87} & 34.08 \upg{2.37} & 42.36 \upg{10.65} \\ \midrule
\Checkmark &   & \Checkmark & \Checkmark &   & 24.39 & 23.94 \downr{-0.45} & 24.57 \upg{0.18} & 22.53 \downr{-1.86} & 24.81 \upg{0.42} & 23.61 \downr{-0.78} & 24.99 \upg{0.6} & 25.71 \upg{1.32} & 26.76 \upg{2.37} & 37.68 \upg{13.29} \\ \midrule
\Checkmark &   & \Checkmark &   & \Checkmark & 31.71 & 30.36 \downr{-1.35} & 31.29 \downr{-0.42} & 30.93 \downr{-0.78} & 32.07 \upg{0.36} & 31.59 \downr{-0.12} & 32.73 \upg{1.02} & 32.25 \upg{0.54} & 34.74 \upg{3.03} & 43.53 \upg{11.82} \\ \midrule
\Checkmark &   &   & \Checkmark & \Checkmark & 31.71 & 30.63 \downr{-1.08} & 31.77 \upg{0.06} & 30.27 \downr{-1.44} & 32.40 \upg{0.69} & 31.80 \upg{0.09} & 32.97 \upg{1.26} & 33.06 \upg{1.35} & 34.80 \upg{3.09} & 45.06 \upg{13.35} \\ \midrule
  & \Checkmark & \Checkmark & \Checkmark &   & 24.39 & 23.79 \downr{-0.6} & 24.15 \downr{-0.24} & 20.55 \downr{-3.84} & 25.02 \upg{0.63} & 23.40 \downr{-0.99} & 24.72 \upg{0.33} & 25.50 \upg{1.11} & 26.31 \upg{1.92} & 38.10 \upg{13.71} \\ \midrule
  & \Checkmark & \Checkmark &   & \Checkmark & 31.71 & 30.24 \downr{-1.47} & 30.90 \downr{-0.81} & 30.90 \downr{-0.81} & 31.62 \downr{-0.09} & 31.20 \downr{-0.51} & 32.07 \upg{0.36} & 31.95 \upg{0.24} & 33.72 \upg{2.01} & 44.01 \upg{12.3} \\ \midrule
  & \Checkmark &   & \Checkmark & \Checkmark & 31.71 & 30.75 \downr{-0.96} & 31.62 \downr{-0.09} & 31.23 \downr{-0.48} & 31.77 \upg{0.06} & 31.44 \downr{-0.27} & 32.40 \upg{0.69} & 33.21 \upg{1.5} & 34.23 \upg{2.52} & 45.69 \upg{13.98} \\ \midrule
  &   & \Checkmark & \Checkmark & \Checkmark & 31.71 & 31.14 \downr{-0.57} & 31.59 \downr{-0.12} & 31.02 \downr{-0.69} & 32.91 \upg{1.2} & 31.77 \upg{0.06} & 33.39 \upg{1.68} & 33.18 \upg{1.47} & 34.50 \upg{2.79} & 46.56 \upg{14.85} \\ \midrule
\Checkmark & \Checkmark & \Checkmark & \Checkmark &   & 24.39 & 23.97 \downr{-0.42} & 23.91 \downr{-0.48} & 22.11 \downr{-2.28} & 24.72 \upg{0.33} & 23.31 \downr{-1.08} & 24.60 \upg{0.21} & 25.50 \upg{1.11} & 27.45 \upg{3.06} & 41.79 \upg{17.4} \\ \midrule
\Checkmark & \Checkmark & \Checkmark &   & \Checkmark & 31.71 & 28.65 \downr{-3.06} & 30.63 \downr{-1.08} & 30.45 \downr{-1.26} & 31.20 \downr{-0.51} & 31.14 \downr{-0.57} & 32.04 \upg{0.33} & 32.37 \upg{0.66} & 34.80 \upg{3.09} & 47.04 \upg{15.33} \\ \midrule
\Checkmark & \Checkmark &   & \Checkmark & \Checkmark & 31.71 & 29.46 \downr{-2.25} & 30.72 \downr{-0.99} & 29.85 \downr{-1.86} & 31.02 \downr{-0.69} & 31.38 \downr{-0.33} & 32.31 \upg{0.6} & 32.55 \upg{0.84} & 34.92 \upg{3.21} & 48.54 \upg{16.83} \\ \midrule
\Checkmark &   & \Checkmark & \Checkmark & \Checkmark & 31.71 & 30.06 \downr{-1.65} & 31.44 \downr{-0.27} & 31.02 \downr{-0.69} & 32.28 \upg{0.57} & 31.65 \downr{-0.06} & 33.21 \upg{1.5} & 33.63 \upg{1.92} & 35.37 \upg{3.66} & 49.38 \upg{17.67} \\ \midrule
  & \Checkmark & \Checkmark & \Checkmark & \Checkmark & 31.71 & 29.82 \downr{-1.89} & 31.20 \downr{-0.51} & 30.51 \downr{-1.2} & 32.58 \upg{0.87} & 31.35 \downr{-0.36} & 32.25 \upg{0.54} & 33.33 \upg{1.62} & 34.23 \upg{2.52} & 49.83 \upg{18.12} \\ \midrule
\Checkmark & \Checkmark & \Checkmark & \Checkmark & \Checkmark & 31.71 & 28.95 \downr{-2.76} & 30.96 \downr{-0.75} & 30.63 \downr{-1.08} & 31.53 \downr{-0.18} & 31.29 \downr{-0.42} & 31.89 \upg{0.18} & 33.18 \upg{1.47} & 35.88 \upg{4.17} & 52.09 \upg{20.37} \\ \midrule

\multicolumn{5}{c}{Mean $\Delta$} & & -0.83 & -0.16 & -1.64 & 0.33 & -0.27 & 0.74 & 1.06 & 2.43 & 12.01 \\ 
\multicolumn{5}{c}{Max $\Delta$} & & 1.35 & 1.86 & 1.05 & 2.10 & 1.08 & 2.37 & 2.07 & 4.17 & 20.37 \\ 
\multicolumn{5}{c}{Min $\Delta$} & & -3.06 & -1.71 & -7.20 & -1.50 & -1.08 & -0.36 & -0.36 & 0.99 & 5.88 \\ 
\bottomrule
\end{tabular}
}
\label{tab:all_possible_combinations_fgvc}
\end{table*}

\begin{table*}[]
\centering
\caption{Our results on \Food{} dataset for all the possible combinations of combining the zero-shot predictions of CLIP backbones, which we group intro non-parametric and parametric techniques. Also, the best-performing single backbone (\Best{}) and the \Oracle{} performance. We present, for each combination of backbones, the improvement \upg{}, constancy \samey{} and deterioration \downr{} of accuracy performance for each method when we compare it against the \Best{} backbone. Mean, Max, and Min $\Delta$ summarize the difference in performance across methods and backbone combinations.}
\Food \\
\qquad
\resizebox{0.9\textwidth}{!}{

\begin{tabular}{ccccccccccccccc}
\toprule
\multicolumn{2}{c}{ResNet} & \multicolumn{3}{c}{ViT} & \multirow{2}{*}{\Best}  & \multicolumn{4}{c}{Non-Parametric} & \multicolumn{4}{c}{Parametric} & \multirow{2}{*}{\Oracle}\\
\cmidrule(lr){1-2} \cmidrule(lr){3-5} \cmidrule(lr){7-10} \cmidrule(lr){11-14}
50 & 101 & B-32 & B-16 & L-14 & & \VoteOne & \VoteThree & \Confidence & \LogitAvg & \CalibratedConfidence & \CLogitAvg & \GAC & \NNC &  \\
\midrule

\Checkmark &   &   &   &   &  \multicolumn{10}{c}{\xfill{.1em} 77.91 \samey{0.00} \xfill{.1em}} \\ 
  & \Checkmark &   &   &   & \multicolumn{10}{c}{\xfill{.1em}  81.86 \samey{0.00} \xfill{.1em}} \\ 
  &   & \Checkmark &   &   & \multicolumn{10}{c}{\xfill{.1em}  82.58 \samey{0.00} \xfill{.1em}} \\ 
  &   &   & \Checkmark &   & \multicolumn{10}{c}{\xfill{.1em}  87.91 \samey{0.00} \xfill{.1em}} \\ 
  &   &   &   & \Checkmark & \multicolumn{10}{c}{\xfill{.1em}  92.32 \samey{0.00} \xfill{.1em}} \\ \midrule
\Checkmark & \Checkmark &   &   &   & 81.86 & 82.63 \upg{0.78} & 82.90 \upg{1.04} & 82.26 \upg{0.4} & 83.08 \upg{1.22} & 82.47 \upg{0.61} & 83.14 \upg{1.29} & 83.28 \upg{1.43} & 83.27 \upg{1.41} & 87.16 \upg{5.3} \\ \midrule
\Checkmark &   & \Checkmark &   &   & 82.58 & 83.83 \upg{1.25} & 84.26 \upg{1.68} & 76.96 \downr{-5.62} & 84.55 \upg{1.97} & 83.60 \upg{1.02} & 84.40 \upg{1.82} & 84.41 \upg{1.83} & 84.68 \upg{2.1} & 88.42 \upg{5.84} \\ \midrule
\Checkmark &   &   & \Checkmark &   & 87.91 & 87.56 \downr{-0.36} & 87.62 \downr{-0.3} & 87.49 \downr{-0.43} & 87.58 \downr{-0.33} & 87.52 \downr{-0.4} & 87.80 \downr{-0.11} & 88.13 \upg{0.22} & 88.55 \upg{0.64} & 91.31 \upg{3.39} \\ \midrule
\Checkmark &   &   &   & \Checkmark & 92.32 & 91.94 \downr{-0.38} & 92.04 \downr{-0.29} & 91.88 \downr{-0.44} & 91.75 \downr{-0.57} & 91.75 \downr{-0.57} & 91.93 \downr{-0.39} & 92.82 \upg{0.49} & 92.79 \upg{0.47} & 94.52 \upg{2.19} \\ \midrule
  & \Checkmark & \Checkmark &   &   & 82.58 & 84.97 \upg{2.39} & 85.31 \upg{2.73} & 84.48 \upg{1.9} & 85.70 \upg{3.11} & 84.62 \upg{2.04} & 85.54 \upg{2.96} & 85.55 \upg{2.97} & 85.69 \upg{3.11} & 89.28 \upg{6.69} \\ \midrule
  & \Checkmark &   & \Checkmark &   & 87.91 & 88.17 \upg{0.26} & 88.30 \upg{0.38} & 87.92 \upg{0.01} & 88.40 \upg{0.49} & 87.87 \downr{-0.05} & 88.34 \upg{0.43} & 88.72 \upg{0.81} & 88.68 \upg{0.76} & 91.81 \upg{3.9} \\ \midrule
  & \Checkmark &   &   & \Checkmark & 92.32 & 92.25 \downr{-0.07} & 92.34 \upg{0.02} & 92.04 \downr{-0.29} & 92.08 \downr{-0.24} & 91.85 \downr{-0.48} & 92.22 \downr{-0.11} & 92.78 \upg{0.45} & 92.80 \upg{0.48} & 94.81 \upg{2.49} \\ \midrule
  &   & \Checkmark & \Checkmark &   & 87.91 & 88.08 \upg{0.16} & 88.18 \upg{0.27} & 82.55 \downr{-5.36} & 88.25 \upg{0.33} & 88.10 \upg{0.19} & 88.31 \upg{0.4} & 88.79 \upg{0.88} & 88.74 \upg{0.83} & 91.54 \upg{3.63} \\ \midrule
  &   & \Checkmark &   & \Checkmark & 92.32 & 92.13 \downr{-0.2} & 92.21 \downr{-0.12} & 92.00 \downr{-0.33} & 92.04 \downr{-0.28} & 91.92 \downr{-0.4} & 92.14 \downr{-0.19} & 92.70 \upg{0.37} & 92.73 \upg{0.41} & 94.76 \upg{2.44} \\ \midrule
  &   &   & \Checkmark & \Checkmark & 92.32 & 92.58 \upg{0.25} & 92.67 \upg{0.34} & 92.45 \upg{0.12} & 92.57 \upg{0.24} & 92.31 \downr{-0.01} & 92.59 \upg{0.27} & 92.70 \upg{0.38} & 92.88 \upg{0.55} & 94.96 \upg{2.64} \\ \midrule
\Checkmark & \Checkmark & \Checkmark &   &   & 82.58 & 84.54 \upg{1.96} & 85.28 \upg{2.7} & 79.64 \downr{-2.95} & 85.59 \upg{3.01} & 84.62 \upg{2.04} & 85.38 \upg{2.8} & 85.82 \upg{3.24} & 85.84 \upg{3.26} & 91.16 \upg{8.58} \\ \midrule
\Checkmark & \Checkmark &   & \Checkmark &   & 87.91 & 86.54 \downr{-1.37} & 87.34 \downr{-0.57} & 87.57 \downr{-0.34} & 87.72 \downr{-0.19} & 87.51 \downr{-0.4} & 87.63 \downr{-0.28} & 88.62 \upg{0.71} & 88.73 \upg{0.82} & 92.99 \upg{5.07} \\ \midrule
\Checkmark & \Checkmark &   &   & \Checkmark & 92.32 & 88.86 \downr{-3.47} & 90.41 \downr{-1.91} & 91.66 \downr{-0.67} & 90.99 \downr{-1.34} & 91.45 \downr{-0.87} & 90.59 \downr{-1.74} & 92.79 \upg{0.47} & 92.84 \upg{0.52} & 95.47 \upg{3.15} \\ \midrule
\Checkmark &   & \Checkmark & \Checkmark &   & 87.91 & 87.45 \downr{-0.47} & 87.98 \upg{0.07} & 80.07 \downr{-7.85} & 87.97 \upg{0.06} & 87.77 \downr{-0.15} & 88.09 \upg{0.17} & 88.84 \upg{0.93} & 88.85 \upg{0.94} & 93.01 \upg{5.09} \\ \midrule
\Checkmark &   & \Checkmark &   & \Checkmark & 92.32 & 89.70 \downr{-2.63} & 90.90 \downr{-1.43} & 90.01 \downr{-2.32} & 91.18 \downr{-1.15} & 91.54 \downr{-0.79} & 90.93 \downr{-1.39} & 92.71 \upg{0.38} & 92.83 \upg{0.5} & 95.56 \upg{3.23} \\ \midrule
\Checkmark &   &   & \Checkmark & \Checkmark & 92.32 & 91.38 \downr{-0.94} & 92.00 \downr{-0.33} & 92.11 \downr{-0.21} & 91.98 \downr{-0.34} & 91.93 \downr{-0.4} & 92.00 \downr{-0.33} & 92.77 \upg{0.44} & 93.02 \upg{0.69} & 95.77 \upg{3.45} \\ \midrule
  & \Checkmark & \Checkmark & \Checkmark &   & 87.91 & 87.78 \downr{-0.13} & 88.18 \upg{0.27} & 83.45 \downr{-4.46} & 88.40 \upg{0.48} & 87.96 \upg{0.05} & 88.32 \upg{0.4} & 89.02 \upg{1.1} & 88.95 \upg{1.04} & 93.36 \upg{5.45} \\ \midrule
  & \Checkmark & \Checkmark &   & \Checkmark & 92.32 & 90.17 \downr{-2.15} & 91.20 \downr{-1.13} & 91.75 \downr{-0.57} & 91.50 \downr{-0.82} & 91.54 \downr{-0.78} & 91.28 \downr{-1.05} & 92.90 \upg{0.57} & 92.93 \upg{0.61} & 95.71 \upg{3.39} \\ \midrule
  & \Checkmark &   & \Checkmark & \Checkmark & 92.32 & 91.54 \downr{-0.78} & 91.98 \downr{-0.34} & 92.06 \downr{-0.26} & 92.07 \downr{-0.25} & 91.87 \downr{-0.45} & 92.02 \downr{-0.3} & 92.93 \upg{0.6} & 92.97 \upg{0.65} & 95.89 \upg{3.56} \\ \midrule
  &   & \Checkmark & \Checkmark & \Checkmark & 92.32 & 91.24 \downr{-1.08} & 91.85 \downr{-0.48} & 90.73 \downr{-1.59} & 92.02 \downr{-0.31} & 92.04 \downr{-0.28} & 91.92 \downr{-0.4} & 92.93 \upg{0.61} & 92.98 \upg{0.66} & 95.79 \upg{3.47} \\ \midrule
\Checkmark & \Checkmark & \Checkmark & \Checkmark &   & 87.91 & 87.65 \downr{-0.26} & 87.98 \upg{0.07} & 81.00 \downr{-6.91} & 88.11 \upg{0.2} & 87.66 \downr{-0.25} & 88.10 \upg{0.19} & 89.07 \upg{1.16} & 89.03 \upg{1.12} & 94.01 \upg{6.1} \\ \midrule
\Checkmark & \Checkmark & \Checkmark &   & \Checkmark & 92.32 & 89.57 \downr{-2.75} & 90.21 \downr{-2.12} & 90.17 \downr{-2.15} & 90.61 \downr{-1.71} & 91.24 \downr{-1.09} & 90.24 \downr{-2.09} & 92.49 \upg{0.17} & 92.90 \upg{0.58} & 96.10 \upg{3.77} \\ \midrule
\Checkmark & \Checkmark &   & \Checkmark & \Checkmark & 92.32 & 91.10 \downr{-1.22} & 91.29 \downr{-1.03} & 91.83 \downr{-0.5} & 91.38 \downr{-0.95} & 91.60 \downr{-0.73} & 91.26 \downr{-1.07} & 92.93 \upg{0.61} & 93.03 \upg{0.7} & 96.26 \upg{3.94} \\ \midrule
\Checkmark &   & \Checkmark & \Checkmark & \Checkmark & 92.32 & 90.99 \downr{-1.33} & 91.49 \downr{-0.84} & 89.53 \downr{-2.79} & 91.53 \downr{-0.8} & 91.73 \downr{-0.59} & 91.49 \downr{-0.84} & 92.93 \upg{0.6} & 93.06 \upg{0.74} & 96.25 \upg{3.93} \\ \midrule
  & \Checkmark & \Checkmark & \Checkmark & \Checkmark & 92.32 & 91.26 \downr{-1.06} & 91.56 \downr{-0.76} & 90.61 \downr{-1.72} & 91.62 \downr{-0.7} & 91.67 \downr{-0.65} & 91.45 \downr{-0.88} & 92.86 \upg{0.54} & 93.03 \upg{0.71} & 96.35 \upg{4.02} \\ \midrule
\Checkmark & \Checkmark & \Checkmark & \Checkmark & \Checkmark & 92.32 & 90.08 \downr{-2.25} & 90.88 \downr{-1.45} & 89.62 \downr{-2.7} & 91.09 \downr{-1.24} & 91.43 \downr{-0.9} & 90.91 \downr{-1.41} & 92.91 \upg{0.59} & 93.07 \upg{0.75} & 96.60 \upg{4.27} \\ \midrule

\multicolumn{5}{c}{Mean $\Delta$} & & -0.61 & -0.14 & -1.85 & 0.00 & -0.17 & -0.07 & 0.87 & 0.96 & 4.19 \\ 
\multicolumn{5}{c}{Max $\Delta$} & & 2.39 & 2.73 & 1.90 & 3.11 & 2.04 & 2.96 & 3.24 & 3.26 & 8.58 \\ 
\multicolumn{5}{c}{Min $\Delta$} & & -3.47 & -2.12 & -7.85 & -1.71 & -1.09 & -2.09 & 0.17 & 0.41 & 2.19 \\ 
\bottomrule
\end{tabular}
}

\label{tab:all_possible_combinations_food}
\end{table*}

\begin{table*}[]
\centering
\caption{Our results on \Flowers{} dataset for all the possible combinations of combining the zero-shot predictions of CLIP backbones, which we group intro non-parametric and parametric techniques. Also, the best-performing single backbone (\Best{}) and the \Oracle{} performance. We present, for each combination of backbones, the improvement \upg{}, constancy \samey{} and deterioration \downr{} of accuracy performance for each method when we compare it against the \Best{} backbone. Mean, Max, and Min $\Delta$ summarize the difference in performance across methods and backbone combinations.}
\Flowers \\
\qquad
\resizebox{0.9\textwidth}{!}{

\begin{tabular}{ccccccccccccccc}
\toprule
\multicolumn{2}{c}{ResNet} & \multicolumn{3}{c}{ViT} & \multirow{2}{*}{\Best}  & \multicolumn{4}{c}{Non-Parametric} & \multicolumn{4}{c}{Parametric} & \multirow{2}{*}{\Oracle}\\
\cmidrule(lr){1-2} \cmidrule(lr){3-5} \cmidrule(lr){7-10} \cmidrule(lr){11-14}
50 & 101 & B-32 & B-16 & L-14 & & \VoteOne & \VoteThree & \Confidence & \LogitAvg & \CalibratedConfidence & \CLogitAvg & \GAC & \NNC &  \\
\midrule

\Checkmark &   &   &   &   &  \multicolumn{10}{c}{\xfill{.1em} 66.12 \samey{0.00} \xfill{.1em}} \\ 
  & \Checkmark &   &   &   & \multicolumn{10}{c}{\xfill{.1em}  65.20 \samey{0.00} \xfill{.1em}} \\ 
  &   & \Checkmark &   &   & \multicolumn{10}{c}{\xfill{.1em}  66.48 \samey{0.00} \xfill{.1em}} \\ 
  &   &   & \Checkmark &   & \multicolumn{10}{c}{\xfill{.1em}  71.43 \samey{0.00} \xfill{.1em}} \\ 
  &   &   &   & \Checkmark & \multicolumn{10}{c}{\xfill{.1em}  79.05 \samey{0.00} \xfill{.1em}} \\ \midrule
 \Checkmark & \Checkmark &   &   &   & 66.12 & 68.01 \upg{1.89} & 67.95 \upg{1.82} & 67.36 \upg{1.24} & 68.63 \upg{2.5} & 67.25 \upg{1.12} & 68.43 \upg{2.31} & 68.40 \upg{2.28} & 68.84 \upg{2.72} & 73.83 \upg{7.71} \\ \midrule
\Checkmark &   & \Checkmark &   &   & 66.48 & 69.21 \upg{2.73} & 69.77 \upg{3.29} & 68.68 \upg{2.2} & 69.17 \upg{2.68} & 68.76 \upg{2.28} & 69.77 \upg{3.29} & 69.28 \upg{2.8} & 70.00 \upg{3.51} & 75.49 \upg{9.01} \\ \midrule
\Checkmark &   &   & \Checkmark &   & 71.43 & 72.45 \upg{1.02} & 72.68 \upg{1.25} & 72.03 \upg{0.6} & 72.48 \upg{1.06} & 71.87 \upg{0.44} & 72.69 \upg{1.27} & 72.42 \upg{0.99} & 73.12 \upg{1.69} & 76.63 \upg{5.2} \\ \midrule
\Checkmark &   &   &   & \Checkmark & 79.05 & 78.29 \downr{-0.76} & 78.22 \downr{-0.83} & 79.02 \downr{-0.03} & 77.33 \downr{-1.72} & 78.78 \downr{-0.28} & 78.19 \downr{-0.86} & 78.91 \downr{-0.15} & 79.41 \upg{0.36} & 83.27 \upg{4.21} \\ \midrule
  & \Checkmark & \Checkmark &   &   & 66.48 & 68.56 \upg{2.08} & 68.74 \upg{2.26} & 68.35 \upg{1.87} & 68.87 \upg{2.39} & 68.32 \upg{1.84} & 68.71 \upg{2.23} & 68.74 \upg{2.26} & 69.21 \upg{2.73} & 74.39 \upg{7.9} \\ \midrule
  & \Checkmark &   & \Checkmark &   & 71.43 & 71.98 \upg{0.55} & 72.32 \upg{0.89} & 71.70 \upg{0.28} & 72.35 \upg{0.93} & 71.41 \downr{-0.02} & 72.37 \upg{0.94} & 73.10 \upg{1.68} & 73.02 \upg{1.59} & 77.35 \upg{5.92} \\ \midrule
  & \Checkmark &   &   & \Checkmark & 79.05 & 77.87 \downr{-1.19} & 77.74 \downr{-1.32} & 78.34 \downr{-0.72} & 77.23 \downr{-1.82} & 77.87 \downr{-1.19} & 77.52 \downr{-1.53} & 77.83 \downr{-1.22} & 79.23 \upg{0.18} & 82.62 \upg{3.56} \\ \midrule
  &   & \Checkmark & \Checkmark &   & 71.43 & 72.91 \upg{1.48} & 72.69 \upg{1.27} & 65.15 \downr{-6.28} & 72.16 \upg{0.73} & 72.52 \upg{1.09} & 72.69 \upg{1.27} & 71.88 \upg{0.46} & 72.99 \upg{1.56} & 78.29 \upg{6.86} \\ \midrule
  &   & \Checkmark &   & \Checkmark & 79.05 & 78.50 \downr{-0.55} & 78.31 \downr{-0.75} & 78.45 \downr{-0.6} & 77.43 \downr{-1.63} & 78.35 \downr{-0.7} & 77.88 \downr{-1.17} & 78.94 \downr{-0.11} & 79.07 \upg{0.02} & 82.86 \upg{3.81} \\ \midrule
  &   &   & \Checkmark & \Checkmark & 79.05 & 78.01 \downr{-1.04} & 78.01 \downr{-1.04} & 78.45 \downr{-0.6} & 77.90 \downr{-1.15} & 78.37 \downr{-0.68} & 78.29 \downr{-0.76} & 78.14 \downr{-0.91} & 79.66 \upg{0.6} & 83.36 \upg{4.31} \\ \midrule
\Checkmark & \Checkmark & \Checkmark &   &   & 66.48 & 69.38 \upg{2.89} & 69.98 \upg{3.5} & 69.12 \upg{2.63} & 69.57 \upg{3.09} & 69.12 \upg{2.63} & 70.09 \upg{3.61} & 69.64 \upg{3.15} & 70.92 \upg{4.44} & 78.57 \upg{12.08} \\ \midrule
\Checkmark & \Checkmark &   & \Checkmark &   & 71.43 & 71.74 \upg{0.31} & 72.27 \upg{0.85} & 71.78 \upg{0.36} & 72.48 \upg{1.06} & 71.49 \upg{0.07} & 72.52 \upg{1.09} & 73.10 \upg{1.68} & 73.82 \upg{2.39} & 79.46 \upg{8.03} \\ \midrule
\Checkmark & \Checkmark &   &   & \Checkmark & 79.05 & 74.70 \downr{-4.36} & 76.03 \downr{-3.02} & 78.34 \downr{-0.72} & 75.59 \downr{-3.46} & 77.77 \downr{-1.28} & 75.95 \downr{-3.11} & 78.00 \downr{-1.06} & 79.90 \upg{0.85} & 84.65 \upg{5.59} \\ \midrule
\Checkmark &   & \Checkmark & \Checkmark &   & 71.43 & 72.69 \upg{1.27} & 73.20 \upg{1.77} & 66.29 \downr{-5.14} & 72.74 \upg{1.32} & 72.55 \upg{1.12} & 72.92 \upg{1.5} & 72.97 \upg{1.54} & 73.90 \upg{2.47} & 80.34 \upg{8.91} \\ \midrule
\Checkmark &   & \Checkmark &   & \Checkmark & 79.05 & 75.87 \downr{-3.19} & 76.89 \downr{-2.16} & 78.48 \downr{-0.57} & 76.34 \downr{-2.72} & 78.32 \downr{-0.73} & 76.74 \downr{-2.31} & 78.83 \downr{-0.23} & 80.08 \upg{1.02} & 84.91 \upg{5.85} \\ \midrule
\Checkmark &   &   & \Checkmark & \Checkmark & 79.05 & 76.84 \downr{-2.21} & 77.41 \downr{-1.64} & 78.48 \downr{-0.57} & 77.33 \downr{-1.72} & 78.35 \downr{-0.7} & 77.44 \downr{-1.61} & 78.09 \downr{-0.96} & 80.27 \upg{1.22} & 84.83 \upg{5.77} \\ \midrule
  & \Checkmark & \Checkmark & \Checkmark &   & 71.43 & 71.69 \upg{0.26} & 72.55 \upg{1.12} & 66.34 \downr{-5.09} & 72.09 \upg{0.67} & 72.43 \upg{1.01} & 72.82 \upg{1.4} & 71.74 \upg{0.31} & 73.83 \upg{2.41} & 80.94 \upg{9.51} \\ \midrule
  & \Checkmark & \Checkmark &   & \Checkmark & 79.05 & 75.15 \downr{-3.9} & 76.39 \downr{-2.67} & 77.92 \downr{-1.14} & 76.19 \downr{-2.86} & 77.67 \downr{-1.38} & 76.45 \downr{-2.6} & 78.53 \downr{-0.52} & 79.44 \upg{0.39} & 84.16 \upg{5.11} \\ \midrule
  & \Checkmark &   & \Checkmark & \Checkmark & 79.05 & 76.34 \downr{-2.72} & 77.13 \downr{-1.92} & 78.05 \downr{-1.01} & 77.20 \downr{-1.85} & 77.67 \downr{-1.38} & 77.36 \downr{-1.69} & 78.16 \downr{-0.89} & 80.16 \upg{1.11} & 84.88 \upg{5.82} \\ \midrule
  &   & \Checkmark & \Checkmark & \Checkmark & 79.05 & 77.04 \downr{-2.02} & 77.65 \downr{-1.4} & 76.11 \downr{-2.94} & 76.74 \downr{-2.31} & 78.11 \downr{-0.94} & 77.80 \downr{-1.25} & 78.57 \downr{-0.49} & 79.69 \upg{0.63} & 85.01 \upg{5.95} \\ \midrule
\Checkmark & \Checkmark & \Checkmark & \Checkmark &   & 71.43 & 72.52 \upg{1.09} & 72.95 \upg{1.53} & 67.02 \downr{-4.41} & 72.60 \upg{1.17} & 72.22 \upg{0.8} & 73.00 \upg{1.58} & 72.61 \upg{1.19} & 74.45 \upg{3.02} & 82.14 \upg{10.72} \\ \midrule
\Checkmark & \Checkmark & \Checkmark &   & \Checkmark & 79.05 & 75.22 \downr{-3.84} & 75.69 \downr{-3.37} & 77.92 \downr{-1.14} & 75.18 \downr{-3.87} & 77.59 \downr{-1.46} & 75.52 \downr{-3.53} & 78.63 \downr{-0.42} & 79.80 \upg{0.75} & 85.62 \upg{6.57} \\ \midrule
\Checkmark & \Checkmark &   & \Checkmark & \Checkmark & 79.05 & 76.01 \downr{-3.04} & 76.52 \downr{-2.54} & 78.00 \downr{-1.06} & 76.26 \downr{-2.8} & 77.57 \downr{-1.48} & 76.55 \downr{-2.5} & 78.24 \downr{-0.81} & 80.27 \upg{1.22} & 85.67 \upg{6.62} \\ \midrule
\Checkmark &   & \Checkmark & \Checkmark & \Checkmark & 79.05 & 76.71 \downr{-2.34} & 77.22 \downr{-1.84} & 76.21 \downr{-2.85} & 76.31 \downr{-2.75} & 78.14 \downr{-0.91} & 77.05 \downr{-2.0} & 78.18 \downr{-0.88} & 80.18 \upg{1.12} & 85.77 \upg{6.72} \\ \midrule
  & \Checkmark & \Checkmark & \Checkmark & \Checkmark & 79.05 & 76.81 \downr{-2.24} & 77.09 \downr{-1.97} & 75.80 \downr{-3.25} & 76.22 \downr{-2.83} & 77.69 \downr{-1.37} & 77.10 \downr{-1.95} & 78.19 \downr{-0.86} & 79.93 \upg{0.88} & 85.75 \upg{6.7} \\ \midrule
\Checkmark & \Checkmark & \Checkmark & \Checkmark & \Checkmark & 79.05 & 75.70 \downr{-3.35} & 76.28 \downr{-2.77} & 75.85 \downr{-3.20} & 75.54 \downr{-3.51} & 77.60 \downr{-1.45} & 76.25 \downr{-2.80} & 78.16 \downr{-0.89} & 81.10 \upg{2.05} & 86.32 \upg{7.27} \\ \midrule

\multicolumn{5}{c}{Mean $\Delta$} & & -0.81 & -0.37 & -1.24 & -0.75 & -0.14 & -0.35 & 0.30 & 1.57 & 6.76 \\ 
\multicolumn{5}{c}{Max $\Delta$} & & 2.89 & 3.50 & 2.63 & 3.09 & 2.63 & 3.61 & 3.15 & 4.44 & 12.08 \\ 
\multicolumn{5}{c}{Min $\Delta$} & & -4.36 & -3.37 & -6.28 & -3.87 & -1.48 & -3.53 & -1.22 & 0.02 & 3.56 \\ 
\bottomrule
\end{tabular}
}
\label{tab:all_possible_combinations_flowers}
\end{table*}

\begin{table*}[]
\centering
\caption{Our results on \Imagenet{} dataset for all the possible combinations of combining the zero-shot predictions of CLIP backbones, which we group intro non-parametric and parametric techniques. Also, the best-performing single backbone (\Best{}) and the \Oracle{} performance. We present, for each combination of backbones, the improvement \upg{}, constancy \samey{} and deterioration \downr{} of accuracy performance for each method when we compare it against the \Best{} backbone. Mean, Max, and Min $\Delta$ summarize the difference in performance across methods and backbone combinations.}
\Imagenet \\
\qquad
\resizebox{0.9\textwidth}{!}{

\begin{tabular}{ccccccccccccccc}
\toprule
\multicolumn{2}{c}{ResNet} & \multicolumn{3}{c}{ViT} & \multirow{2}{*}{\Best}  & \multicolumn{4}{c}{Non-Parametric} & \multicolumn{4}{c}{Parametric} & \multirow{2}{*}{\Oracle}\\
\cmidrule(lr){1-2} \cmidrule(lr){3-5} \cmidrule(lr){7-10} \cmidrule(lr){11-14}
50 & 101 & B-32 & B-16 & L-14 & & \VoteOne & \VoteThree & \Confidence & \LogitAvg & \CalibratedConfidence & \CLogitAvg & \GAC & \NNC &  \\
\midrule

\Checkmark &   &   &   &   &  \multicolumn{10}{c}{\xfill{.1em} 59.84 \samey{0.00} \xfill{.1em}} \\ 
  & \Checkmark &   &   &   & \multicolumn{10}{c}{\xfill{.1em}  62.28 \samey{0.00} \xfill{.1em}} \\ 
  &   & \Checkmark &   &   & \multicolumn{10}{c}{\xfill{.1em}  63.35 \samey{0.00} \xfill{.1em}} \\ 
  &   &   & \Checkmark &   & \multicolumn{10}{c}{\xfill{.1em}  68.34 \samey{0.00} \xfill{.1em}} \\ 
  &   &   &   & \Checkmark & \multicolumn{10}{c}{\xfill{.1em}  75.54 \samey{0.00} \xfill{.1em}} \\ \midrule
\Checkmark & \Checkmark &   &   &   & 62.30 & 63.76 \upg{1.46} & 64.16 \upg{1.86} & 59.05 \downr{-3.25} & 64.61 \upg{2.31} & 63.22 \upg{0.91} & 64.47 \upg{2.17} & 64.71 \upg{2.41} & 65.14 \upg{2.84} & 70.75 \upg{8.45} \\ \midrule
\Checkmark &   & \Checkmark &   &   & 63.36 & 65.00 \upg{1.65} & 65.37 \upg{2.01} & 64.40 \upg{1.04} & 65.78 \upg{2.42} & 64.39 \upg{1.03} & 65.54 \upg{2.19} & 65.86 \upg{2.51} & 66.07 \upg{2.71} & 71.76 \upg{8.41} \\ \midrule
\Checkmark &   &   & \Checkmark &   & 68.34 & 68.47 \upg{0.12} & 68.72 \upg{0.37} & 68.15 \downr{-0.19} & 68.89 \upg{0.55} & 68.09 \downr{-0.26} & 68.85 \upg{0.51} & 69.60 \upg{1.26} & 69.86 \upg{1.52} & 74.51 \upg{6.17} \\ \midrule
\Checkmark &   &   &   & \Checkmark & 75.53 & 75.13 \downr{-0.4} & 75.27 \downr{-0.27} & 60.23 \downr{-15.3} & 75.09 \downr{-0.45} & 74.51 \downr{-1.02} & 74.97 \downr{-0.56} & 75.97 \upg{0.44} & 76.17 \upg{0.64} & 80.13 \upg{4.6} \\ \midrule
  & \Checkmark & \Checkmark &   &   & 63.36 & 66.03 \upg{2.67} & 66.41 \upg{3.05} & 60.29 \downr{-3.07} & 66.99 \upg{3.63} & 65.45 \upg{2.09} & 66.70 \upg{3.35} & 67.00 \upg{3.65} & 67.35 \upg{3.99} & 72.96 \upg{9.61} \\ \midrule
  & \Checkmark &   & \Checkmark &   & 68.34 & 69.09 \upg{0.75} & 69.29 \upg{0.95} & 68.54 \upg{0.2} & 69.58 \upg{1.24} & 68.45 \upg{0.11} & 69.40 \upg{1.05} & 69.90 \upg{1.56} & 70.29 \upg{1.95} & 75.15 \upg{6.8} \\ \midrule
  & \Checkmark &   &   & \Checkmark & 75.53 & 75.18 \downr{-0.35} & 75.35 \downr{-0.18} & 62.53 \downr{-13.0} & 75.16 \downr{-0.38} & 74.56 \downr{-0.97} & 75.10 \downr{-0.44} & 76.01 \upg{0.48} & 76.39 \upg{0.86} & 80.28 \upg{4.75} \\ \midrule
  &   & \Checkmark & \Checkmark &   & 68.34 & 68.90 \upg{0.56} & 69.17 \upg{0.82} & 63.17 \downr{-5.18} & 69.41 \upg{1.07} & 68.48 \upg{0.13} & 69.23 \upg{0.89} & 69.76 \upg{1.42} & 70.04 \upg{1.7} & 74.97 \upg{6.63} \\ \midrule
  &   & \Checkmark &   & \Checkmark & 75.53 & 75.37 \downr{-0.16} & 75.40 \downr{-0.13} & 63.53 \downr{-12.01} & 75.19 \downr{-0.35} & 74.79 \downr{-0.74} & 75.20 \downr{-0.33} & 75.98 \upg{0.45} & 76.31 \upg{0.78} & 80.49 \upg{4.96} \\ \midrule
  &   &   & \Checkmark & \Checkmark & 75.53 & 75.47 \downr{-0.06} & 75.61 \upg{0.08} & 68.42 \downr{-7.12} & 75.63 \upg{0.1} & 75.04 \downr{-0.5} & 75.50 \downr{-0.03} & 75.98 \upg{0.44} & 76.26 \upg{0.73} & 80.60 \upg{5.06} \\ \midrule
\Checkmark & \Checkmark & \Checkmark &   &   & 63.36 & 66.15 \upg{2.79} & 66.76 \upg{3.4} & 60.27 \downr{-3.08} & 67.33 \upg{3.97} & 65.51 \upg{2.16} & 67.08 \upg{3.72} & 67.47 \upg{4.11} & 67.87 \upg{4.51} & 76.39 \upg{13.03} \\ \midrule
\Checkmark & \Checkmark &   & \Checkmark &   & 68.34 & 68.22 \downr{-0.12} & 68.94 \upg{0.6} & 66.82 \downr{-1.53} & 69.23 \upg{0.89} & 68.17 \downr{-0.18} & 69.18 \upg{0.83} & 70.11 \upg{1.77} & 70.71 \upg{2.37} & 77.95 \upg{9.61} \\ \midrule
\Checkmark & \Checkmark &   &   & \Checkmark & 75.53 & 72.15 \downr{-3.38} & 74.09 \downr{-1.45} & 62.70 \downr{-12.83} & 74.22 \downr{-1.31} & 74.03 \downr{-1.5} & 74.08 \downr{-1.45} & 76.07 \upg{0.54} & 76.42 \upg{0.88} & 82.41 \upg{6.87} \\ \midrule
\Checkmark &   & \Checkmark & \Checkmark &   & 68.34 & 68.58 \upg{0.24} & 69.20 \upg{0.85} & 63.72 \downr{-4.62} & 69.45 \upg{1.11} & 68.35 \samey{0.0} & 69.45 \upg{1.11} & 70.18 \upg{1.84} & 70.60 \upg{2.26} & 78.06 \upg{9.72} \\ \midrule
\Checkmark &   & \Checkmark &   & \Checkmark & 75.53 & 72.80 \downr{-2.73} & 74.35 \downr{-1.18} & 63.26 \downr{-12.28} & 74.36 \downr{-1.17} & 74.24 \downr{-1.3} & 74.32 \downr{-1.21} & 76.14 \upg{0.61} & 76.35 \upg{0.82} & 82.61 \upg{7.08} \\ \midrule
\Checkmark &   &   & \Checkmark & \Checkmark & 75.53 & 73.95 \downr{-1.58} & 74.93 \downr{-0.61} & 67.33 \downr{-8.21} & 74.98 \downr{-0.55} & 74.58 \downr{-0.95} & 74.90 \downr{-0.63} & 76.16 \upg{0.63} & 76.54 \upg{1.01} & 82.76 \upg{7.23} \\ \midrule
  & \Checkmark & \Checkmark & \Checkmark &   & 68.34 & 69.15 \upg{0.81} & 69.73 \upg{1.39} & 62.46 \downr{-5.88} & 70.02 \upg{1.68} & 68.66 \upg{0.32} & 69.95 \upg{1.61} & 70.43 \upg{2.09} & 70.87 \upg{2.53} & 78.58 \upg{10.24} \\ \midrule
  & \Checkmark & \Checkmark &   & \Checkmark & 75.53 & 73.32 \downr{-2.21} & 74.61 \downr{-0.92} & 63.05 \downr{-12.49} & 74.63 \downr{-0.9} & 74.27 \downr{-1.27} & 74.58 \downr{-0.95} & 76.10 \upg{0.57} & 76.60 \upg{1.07} & 82.82 \upg{7.29} \\ \midrule
  & \Checkmark &   & \Checkmark & \Checkmark & 75.53 & 74.13 \downr{-1.4} & 75.04 \downr{-0.49} & 67.39 \downr{-8.14} & 75.11 \downr{-0.42} & 74.59 \downr{-0.94} & 75.14 \downr{-0.39} & 76.09 \upg{0.56} & 76.73 \upg{1.2} & 82.89 \upg{7.35} \\ \midrule
  &   & \Checkmark & \Checkmark & \Checkmark & 75.53 & 74.01 \downr{-1.52} & 74.90 \downr{-0.63} & 68.56 \downr{-6.97} & 75.02 \downr{-0.51} & 74.64 \downr{-0.9} & 74.98 \downr{-0.55} & 76.07 \upg{0.54} & 76.55 \upg{1.01} & 82.91 \upg{7.38} \\ \midrule
\Checkmark & \Checkmark & \Checkmark & \Checkmark &   & 68.34 & 69.10 \upg{0.75} & 69.51 \upg{1.17} & 62.51 \downr{-5.83} & 69.66 \upg{1.32} & 68.43 \upg{0.08} & 69.74 \upg{1.4} & 70.52 \upg{2.18} & 71.09 \upg{2.75} & 80.31 \upg{11.97} \\ \midrule
\Checkmark & \Checkmark & \Checkmark &   & \Checkmark & 75.53 & 72.37 \downr{-3.16} & 73.71 \downr{-1.82} & 63.11 \downr{-12.42} & 73.67 \downr{-1.86} & 73.94 \downr{-1.6} & 73.62 \downr{-1.92} & 76.12 \upg{0.58} & 76.69 \upg{1.16} & 84.08 \upg{8.55} \\ \midrule
\Checkmark & \Checkmark &   & \Checkmark & \Checkmark & 75.53 & 73.47 \downr{-2.06} & 74.46 \downr{-1.07} & 67.47 \downr{-8.06} & 74.46 \downr{-1.07} & 74.21 \downr{-1.32} & 74.41 \downr{-1.12} & 76.10 \upg{0.57} & 76.63 \upg{1.09} & 84.17 \upg{8.64} \\ \midrule
\Checkmark &   & \Checkmark & \Checkmark & \Checkmark & 75.53 & 73.54 \downr{-1.99} & 74.40 \downr{-1.13} & 67.54 \downr{-7.99} & 74.39 \downr{-1.14} & 74.32 \downr{-1.21} & 74.39 \downr{-1.14} & 76.14 \upg{0.61} & 76.65 \upg{1.11} & 84.26 \upg{8.73} \\ \midrule
  & \Checkmark & \Checkmark & \Checkmark & \Checkmark & 75.53 & 73.78 \downr{-1.75} & 74.64 \downr{-0.89} & 67.46 \downr{-8.07} & 74.58 \downr{-0.95} & 74.32 \downr{-1.21} & 74.59 \downr{-0.94} & 76.19 \upg{0.66} & 76.77 \upg{1.23} & 84.42 \upg{8.89} \\ \midrule
\Checkmark & \Checkmark & \Checkmark & \Checkmark & \Checkmark & 75.53 & 72.67 \downr{-2.86} & 73.95 \downr{-1.58} & 67.46 \downr{-8.07} & 73.89 \downr{-1.64} & 74.06 \downr{-1.47} & 73.88 \downr{-1.65} & 76.22 \upg{0.69} & 76.59 \upg{1.06} & 85.29 \upg{9.76} \\ \midrule

\multicolumn{5}{c}{Mean $\Delta$} & & -0.54 & 0.16 & -7.09 & 0.29 & -0.40 & 0.21 & 1.28 & 1.68 & 7.99 \\ 
\multicolumn{5}{c}{Max $\Delta$} & & 2.79 & 3.40 & 1.04 & 3.97 & 2.16 & 3.72 & 4.11 & 4.51 & 13.03 \\ 
\multicolumn{5}{c}{Min $\Delta$} & & -3.38 & -1.82 & -15.30 & -1.86 & -1.60 & -1.92 & 0.44 & 0.64 & 4.60 \\ 
\bottomrule
\end{tabular}
}
\label{tab:all_possible_combinations_inet}
\end{table*}

\end{document}